\RequirePackage{fix-cm}
\documentclass[twocolumn]{svjour3}          % twocolumn
\smartqed  % flush right qed marks, e.g. at end of proof
\usepackage{subfig,graphicx,multirow}
\usepackage{amsmath,amssymb,bm}
\usepackage{color}
\usepackage{lipsum}
\usepackage{amsfonts}
\usepackage{graphicx}
\usepackage{epstopdf}
\usepackage{algorithmic}
\usepackage{floatrow}
\ifpdf
  \DeclareGraphicsExtensions{.eps,.pdf,.png,.jpg}
\else
  \DeclareGraphicsExtensions{.eps}
\fi

\usepackage{amsopn}

\usepackage{graphicx}
\usepackage{subfig,graphicx,multirow}
\usepackage{amsmath,amssymb,bm}
\usepackage{color}
\usepackage{algorithm}
\usepackage{float}
\floatstyle{plaintop}
\restylefloat{table}
\usepackage{mathptmx}      % use Times fonts if available on your TeX system

\begin{document}

\title{Chan-Vese Reformulation for Selective Image Segmentation}

\author{Michael Roberts  \and Jack Spencer}

\institute{M. Roberts \at
Early Clinical Development, Innovative Medicines and Early Development, AstraZeneca and Centre for Mathematical Imaging Techniques, Department of Mathematical Sciences, University of Liverpool \\
\email{michael.roberts2@astrazeneca.com} \\
\and
J. Spencer \at
Quantitative Biology and Medicine @ Exeter, Living Systems Institute, University of Exeter \\
\email{j.a.spencer@exeter.ac.uk}}
\date{Received: date / Accepted: date}
% The correct dates will be entered by the editor

\maketitle

\begin{abstract}
Selective segmentation involves incorporating user input to partition an image into foreground and background, by discriminating between objects of a similar type. Typically, such methods involve introducing additional constraints to generic segmentation approaches. However, we show that this is often inconsistent with respect to common assumptions about the image. The proposed method introduces a new fitting term that is more useful in practice than the Chan-Vese framework. In particular, the idea is to define a term that allows for the background to consist of multiple regions of inhomogeneity. We provide comparitive experimental results to alternative approaches to demonstrate the advantages of the proposed method, broadening the possible application of these methods.
\end{abstract}

\section{Introduction} \label{sec:intro}

Image segmentation is an important application of image processing techniques in which some, or all, objects in an image are isolated from the background. In other words, for an image $z(\bm{x})\in\mathbb{R}^{2}$, we find the partitioning of the image domain $\Omega\subset\mathbb{R}^{2}$ into subregions of interest. In the case of two-phase approaches this consists of the foreground domain $\Omega_{F}$ and background domain $\Omega_{B}$, such that $\Omega=\Omega_{F}\cup\Omega_{B}$. In this work we concentrate on approaching this problem with variational methods, particularly in cases where user input is incorporated. Specifically, we consider the convex relaxation approach of \cite{Chan:06,Bresson:07} and many others. This consists of a binary labelling problem where the aim is to compute a function $u(x)\in\{0,1\}$ indicating regions belonging to $\Omega_{F}$ and $\Omega_{B}$, respectively. This is obtained by imposing a relaxed constraint on the function, $u\in[0,1]$, and minimising a functional that fits the solution to the data with certain conditions on the regularity of the boundary of the foreground regions. 

\begin{figure*}
\centering
\floatbox[{\capbeside\thisfloatsetup{capbesideposition={left,top},capbesidewidth=1.5in}}]{figure}[\FBwidth]
{\captionsetup{position=bottom}\captionsetup[subfigure]{labelformat=empty,font=normalsize} 
\subfloat[(i) Image with ground truth]{\includegraphics[width=1.5in, height = 1.5in]{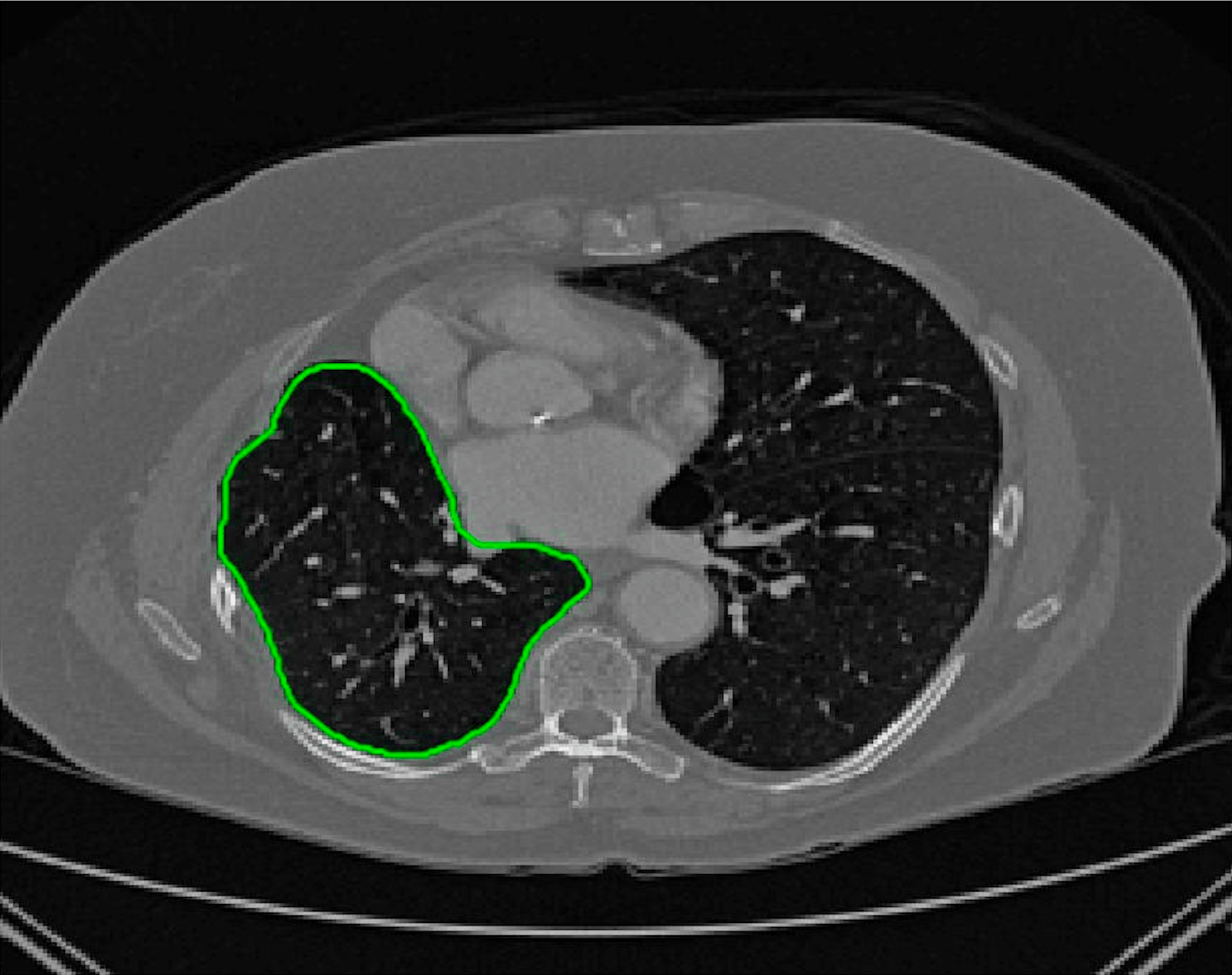}}\quad
\captionsetup{position=bottom}\captionsetup[subfigure]{labelformat=empty,font=normalsize} 
\subfloat[(ii) Foreground, $c_{1}=0.15$]{\includegraphics[width=1.5in, height = 1.5in]{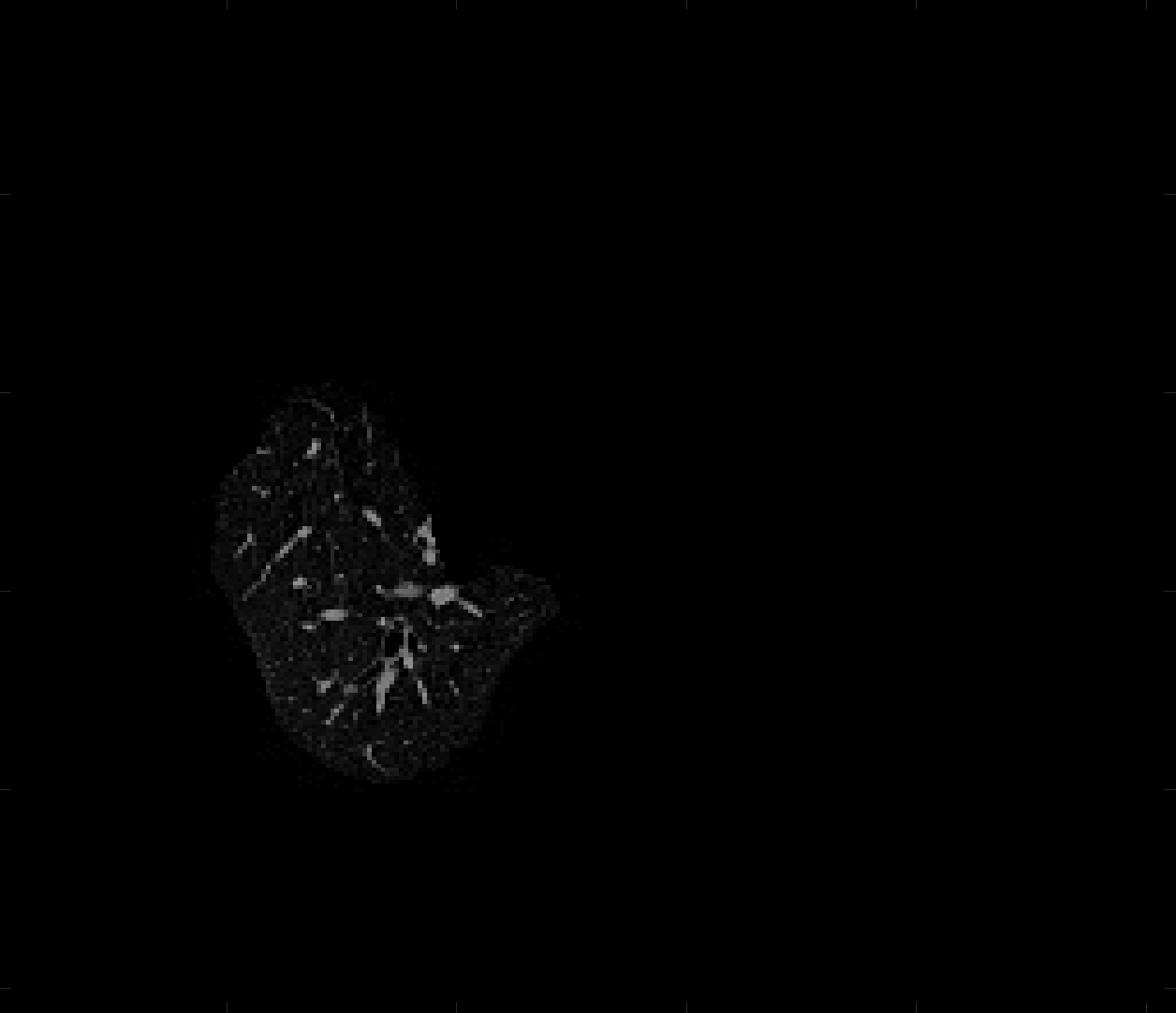}}\quad
\captionsetup{position=bottom}\captionsetup[subfigure]{labelformat=empty,font=normalsize} 
\subfloat[(iii) Background, $c_{2}=0.19$]{\includegraphics[width=1.5in, height = 1.5in]{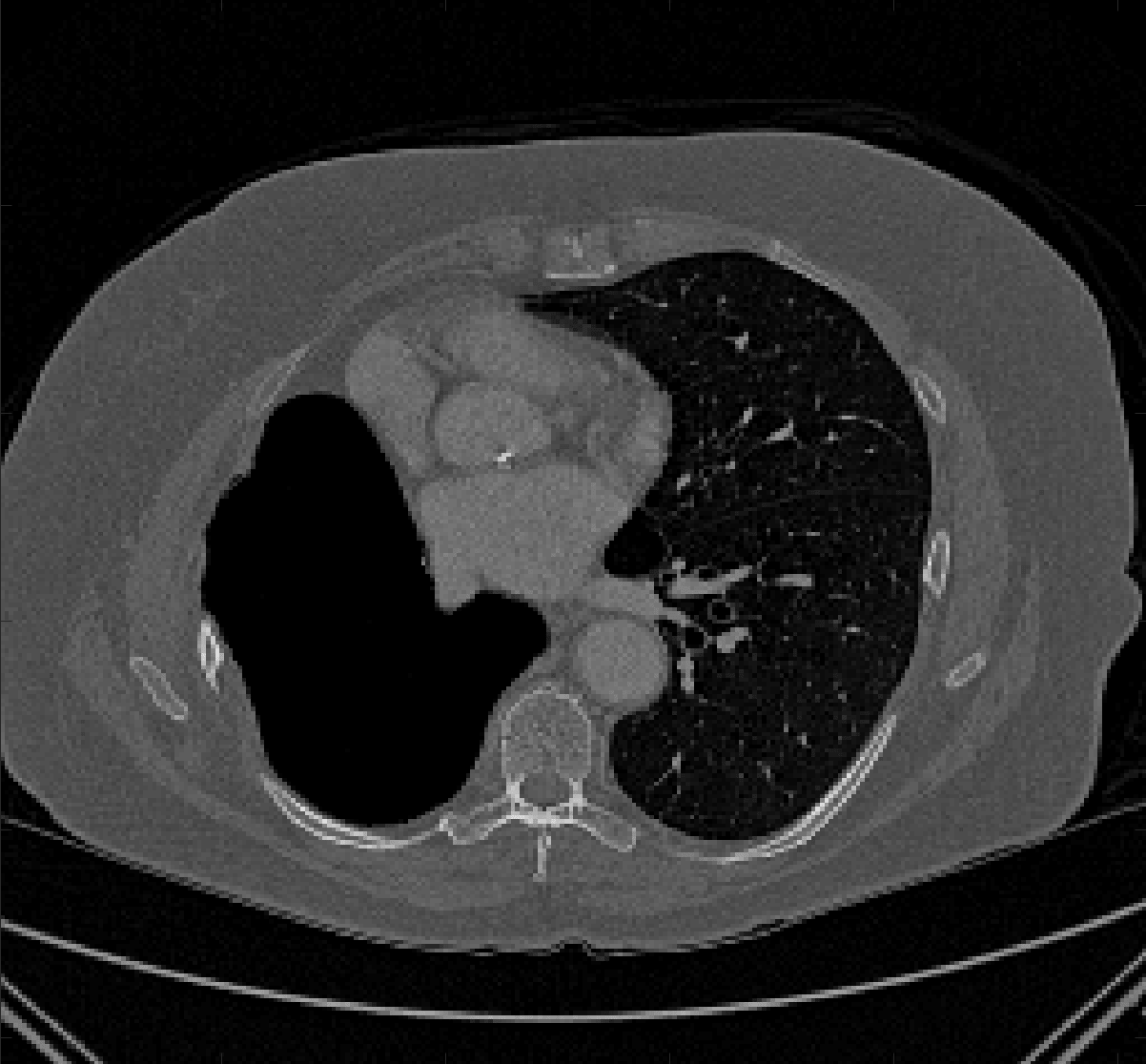}}}
{\caption{CT image with ground truth segmentation shown (green) and associated average intensity values ($c_{1}$ and $c_{2}$). \label{fig:CTscan}}}
\end{figure*}

We will first introduce the seminal work of Chan and Vese \cite{ACWE}, a segmentation model that uses the level set framework of Osher and Sethian \cite{Osher:88}. This approach assumes that the image $z$ is approximately piecewise-constant, but is dependent on the initialisation of the level set function as the minimisation problem is nonconvex. The Chan-Vese model was reformulated to avoid this by Chan et al. \cite{Chan:06}, using convex relaxation methods, that has the following data fitting functional
\begin{equation}\label{eqn:gensegfunc}
f_{CV}(u)=\int_{\Omega}\left(\lambda_{1}f_{1}(\bm{x})-\lambda_{2}f_{2}(\bm{x})\right)u(\bm{x})\ \mathrm{d}\Omega,
\end{equation}
where $f_{1}(\bm{x})$ and $f_{2}(\bm{x})$ are data fitting terms indicating the foreground and background regions, respectively. In particular, in \cite{ACWE} and \cite{Chan:06} these are given by
\begin{equation}\label{eqn:c1c2}
f_{1}(\bm{x})=|z-c_{1}|^{2},\ \ \ \ \ f_{2}(\bm{x})=|z-c_{2}|^{2}.
\end{equation}
It should be noted that it is common to fix $\lambda = \lambda_{1}=\lambda_{2}$. The introduction of binary labels to image segmentation was also proposed by Lie et al. \cite{LieLysakerTai}, with the connections between \cite{Chan:06} and \cite{LieLysakerTai} discussed in Wei et al. \cite{Wei:16}. The data fitting functional is balanced against a regularisation term. Typically, this penalises the length of the contour. This is represented by the total variation (TV) of the function \cite{ACWE,Rudin:92}, and is sometimes weighted by an edge detection function $g(s) = 1/(1+\beta s^{2})$ \cite{Bresson:07,Perona:90,Geo,CDSS}. Therefore, the regularisation term is given as
\begin{equation}
TV_{g}(u):=\int_{\Omega}g(|\nabla z(\bm{x})|)|\nabla u|\ \mathrm{d}\Omega.
\end{equation}
The convex segmentation problem, assuming fixed constants $c_{1}$ and $c_{2}$, is then defined by
\begin{equation}
\min_{u\in[0,1]}\left\lbrace F_{CV}(u,c_1,c_2)=TV_{g}(u)+f_{CV}(u,c_{1},c_{2})\right\rbrace.
\end{equation}
In the case where the intensity constants are unknown it is also possible to minimise $F_{CV}$ alternately with respect to $u, c_{1}$, and $c_{2}$, however, this would make the problem non-convex and hence dependent on the initialisation of $u$. Functionals of this type have been widely studied with respect to two-phase segmentation \cite{Bresson:07,Chan:06,ACWE}, which is our main interest. Alternative choices of data fitting terms can be used when different assumptions are made on the image, $z$. Examples include \cite{Ali:16,Ali:17,VMS,RSF,SBF,LCV}. We note that multiphase approaches \cite{Brox:06,VeseChan:02} are also closely related to this formulation although in this paper we focus on the two-phase problem due to associated applications of interest. It is also important to acknowledge analogous methods in the discrete setting such as \cite{Bai:07,Falcao:02,RW,Grabcut}. However, we do not go into detail about such methods here, although we introduce the work of \cite{SRW} in \S \ref{sec:selective} and compare corresponding results in \S\ref{sec:results}.

In selective segmentation the idea is to apply additional constraints such that user input is incorporated to isolate specific objects of interest. It is common for the user to input marker points to form a set $\mathcal{M}$, where
\(
\mathcal{M}=\{ (x_{i},y_{i})\in\Omega, 1\le i\le k\}
\)
and from this we can form a foreground region $\mathcal{P}$ whose interior points are inside the object to be segmented. In the case that $\mathcal{M}$ is provided $\mathcal{P}$ will be a polygon, but any user-defined region in the foreground is consistent with the proposed method. Some examples of selective or interactive methods include \cite{Cai:13,SRW,Gout:05,Liu:18,Nguyen:12,Geo,RW,LRW,Zhang:10,PFS}. A particular application of this in medical imaging is organ contouring in computed tomography (CT) images. This is often done manually which can be laborious and inefficient and it is often not possible to enhance existing methods with training data. In cases where learning based methods are applicable, the work of Xu et al. \cite{Xu:16} and Bernard and Gygli \cite{Benard:17} are state of the art approaches. At this stage we define the additional constraints in selective segmentation as follows:
\begin{equation}
f_{S}(u)=\theta\int_{\Omega}\mathcal{D}(\bm{x})u\ \mathrm{d}\Omega,
\end{equation}
where $\mathcal{D}(\bm{x})$ is some distance penalty term, such as \cite{Rada:13,Geo,CDSS}, and $\theta$ is a selection parameter. Essentially, the idea is that the selection term $\mathcal{D}(\bm{x})$ (based on the region $\mathcal{P}$ formed by the user input marker set) should penalise regions of the background (as defined by the data fitting term $f_{2}({\bm x})$) and also pixels far from $\mathcal{P}$. In this paper we choose $\mathcal{D}(\bm{x})$ to be the geodesic distance penalty proposed in \cite{Geo}. Explicitly, the geodesic distance from the region $\mathcal{P}$ formed from the marker set is given by:
\[
\mathcal{D}_{M}(\bm{x}) = 0 \text{ for } \bm{x}\in
\mathcal{P},
\]
\[
\mathcal{D}_{M}(\bm{x}) = \frac{\mathcal{D}_{M}^{0}(\bm{x})}{||\mathcal{D}_{M}^{0}(\bm{x})||_{L^{\infty}}} \text{ for } \bm{x}\not\in\mathcal{P},
\]
where $\mathcal{D}_{M}^{0}(\bm{x})$ is the solution of the following PDE:
\begin{equation}\label{eqn:DG0}
|\nabla\mathcal{D}_{M}^{0}(\bm{x})| = q(\bm{x}), \qquad \mathcal{D}^{0}_{M}(\bm{x}_{0}) = 0, \,(\bm{x}_{0})\in\mathcal{P}.
\end{equation}
The function $q(\bm{x})$ is image dependent and controls the rate of increase in the distance. It is defined as a function similar to
\begin{equation}\label{eqn:edgemap}
q({\bm x}) = \varepsilon_{\mathcal{D}} + \beta_{G}|\nabla z(\bm{x})|^{2},
\end{equation}
where $\varepsilon_{\mathcal{D}}$ is a small non-zero parameter and $\beta_{G}$ is a non-negative tuning parameter. We set the value of $\beta_{G} =1000$ and $\varepsilon_{\mathcal{D}} = 10^{-3}$ throughout. Note that
if $q(\bm{x})\equiv 1$ then the distance penalty $\mathcal{D}_{M}(\bm{x})$ is simply the normalised Euclidean distance, as used in \cite{CDSS}. 

A general selective segmentation functional, assuming homogeneous target regions, is therefore given by:
\begin{equation}\label{eqn:selectCV}
F_{S}(u,c_1,c_2)=TV_{g}(u)+f_{CV}(u,c_{1},c_{2})+f_{S}(u).
\end{equation}
Assuming that the optimal intensity constants $c_{1}$ and $c_{2}$ are fixed, the minimisation problem is then:
\begin{equation}\label{eqn:selectCV2}
\min_{u\in[0,1]}F_{S}(u,c_{1},c_{2}).
\end{equation}
Again, it is possible to alternately minimise $F_{S}(u,c_1,c_2)$ with respect to the constants $c_{1}$ and $c_{2}$ to obtain the average intensity in $\Omega_{F}$ and $\Omega_{B}$, respectively. However, in selective segmentation it is often sufficient to fix these according to the user input.  In the framework of \eqref{eqn:selectCV2} the Chan-Vese terms \cite{Chan:06,ACWE,MumfordShah} have limitations due to the dependence on $c_{2}$. In conventional two-phase segmentation problems it makes sense to penalise deviances from $c_{2}$ outside the contour, however for selective segmentation we need not consider the intensities outside of the object we have segmented. Regardless of whether the intensity of regions outside the object is above or below $c_{1}$, it should be penalised positively. The Chan-Vese terms cannot ensure this as they work based on a fixed "exterior'' intensity $c_{2}$ and can lead to negative penalties on regions which are outside the object of interest. It is our aim in this paper to address this problem.

The motivation for this work comes from observing contradictions in using piecewise-constant intensity fitting terms in selective segmentation. Whilst good results are possible with this approach, the exceptional cases lead to severe limitations in practice. This is quite common in medical imaging as demonstrated in Fig. \ref{fig:CTscan}, where the target foreground has a low intensity. Given that the corresponding background includes large regions of low intensity, the optimal average intensities for this segmentation problem are $c_{1}=0.1534$ and $c_{2}=0.1878$. For cases where $c_{1}\approx c_{2}$, we see that by (\ref{eqn:gensegfunc}), $f_{1}-f_{2}\approx0$ almost everywhere in the domain $\Omega$. This means that it is very difficult to achieve an adequate result, without an over-reliance on the user input or parameter selection.

The central premise for applying Chan-Vese type methods is the assumption that the image approximately consists of
\begin{equation}
z(\bm{x})=c_{1}\chi_{F}+c_{2}\chi_{B}+\eta,
\end{equation}
where $\eta$ is noise, $\chi_{i}$ is the characteristic function of the region $\Omega_{i}$, for $i=F,B$ respectively. The idea of selective segmentation is to incorporate user input to apply constraints that exclude regions classified as foreground, based on their location in the image. We use a distance constraint which penalises the distance from the user input markers. However, a key problem for selective segmentation is that for cases where the optimal intensity values $c_{1}$ and $c_{2}$ are similar, the intensity fitting term will become obsolete as the contour evolves. This is illustrated in Fig. \ref{fig:newfittingterm}. The purpose of our approach is to construct a model that is based on assumptions that are consistent with the observed image and any homogeneous target region of interest. A common approach in selective segmentation is to discriminate between objects of a similar intensity \cite{Rada:13,Geo,CDSS}. However, the fitting terms in previous formulations \cite{Klodt:13,Rada:13,Geo,CDSS} aren't applicable in many cases as there are contradictions in the formulation in this context. We will address this in detail in the following section.

In this paper our main contribution is to highlight a crucial flaw in the assumptions behind many current selective segmentation approaches and propose a new fitting term in relation to such methods. We demonstrate how our reformulation is capable of achieving superior results and is more robust to parameter choices than existing approaches, allowing for more consistency in practice. In \S\ref{sec:related} we give a brief review of alternative intensity fitting terms proposed in the literature, and detail them in relation to selective segmentation. We then briefly detail alternative selective segmentation approaches to compare our method against in \S\ref{sec:selective}. In \S\ref{sec:proposed} we introduce the proposed model, focussing on a fitting term that allows for significant intensity variation in the background domain. In \S\ref{sec:numerics} we discuss the implementation of each approach in a convex relaxation framework, provide the algorithm in \S\ref{sec:alg}, and detail some experimental results in \S\ref{sec:results}. Finally, in \S\ref{sec:conclude} we give some concluding remarks.

\section{Related Approaches}
\label{sec:related}

Here, we introduce and discuss work that has introduced alternative data fitting terms closely related to Chan-Vese \cite{ACWE}.  In order to make direct comparisons, we convert each approach to the unified framework of convex relaxation \cite{Chan:06}. It is worth noting that this alternative implementation is equivalent in some respects, but that the results might differ slightly if using the original methods. We are considering these models in the terms of selective segmentation, so all formulations have the following structure:
\begin{equation} \label{eqn:Fu}
\min_{u\in[0,1]}\left\lbrace F(u)=TV_{g}(u)+f_{S}(u)+f(u)\right\rbrace.
\end{equation}
We are interested in the effectiveness of $f(u)$ in this context, which we will focus on next. In particular, we detail various choices of $f(u)$ from the literature that are generalisations of the Chan-Vese approach. In the following we refer to minimisers of convex formulations, such as \eqref{eqn:Fu}, by $u_{\gamma}$. Here, the minimiser of $F(u)$ is thresholded for $\gamma\in(0,1)$ in a conventional way \cite{Chan:06}. 

\subsection{Region-Scalable Fitting (RSF) \cite{RSF}} 

The data fitting term from the work of Li et al. \cite{RSF}, known as Region-Scalable Fitting (RSF), consistent with the convex relaxation technique of \cite{Chan:06} is given by
\begin{equation}
f_{RSF}(u)=\int_{\Omega} \left(\lambda_{1}f_{1}(\bm{x}) - \lambda_{2}f_{2}(\bm{x})\right)u \ \mathrm{d}\Omega,
\end{equation}
where
\begin{align}
f_{1}(\bm{x})&=\int_{\Omega}K_{\sigma}(\bm{x}-\bm{y})\left|z-h_{1}(\bm{x})\right|^{2}\, \mathrm{d}\Omega, \notag \\
f_{2}(\bm{x}) &= \int_{\Omega} K_{\sigma}(\bm{x}-\bm{y})\left|z - h_{2}(\bm{x})\right|^{2} \, \mathrm{d}\Omega,
\end{align}
and $K_{\sigma}(\bm{x})$ is chosen as a Gaussian kernel with scale parameter $\sigma>0$. The RSF selective formulation is then given as follows:
\begin{equation}
F_{RSF}(u)=TV_{g}(u)+f_{S}(u)+f_{RSF}(u).
\end{equation}
The functions $h_{1}(\bm{x})$ and $h_{2}(\bm{x})$, which are generalisations of $c_{1}$ and $c_{2}$ from Chan-Vese, are updated iteratively by
\begin{align}
h_{1}(\bm{x})&= \frac{K_{\sigma}(\bm{x})*\left( u_{\gamma}\,z \right)}{K_{\sigma}(\bm{x})*u_{\gamma}}, \notag \\
h_{2}(\bm{x})&= \frac{K_{\sigma}(\bm{x})*\left( \left(1 - u_{\gamma}\right)z \right)}{K_{\sigma}(\bm{x})*\left(1 - u_{\gamma}\right)}.
\end{align}
Using the RSF fitting term, any deviations of $z$ from $h_{1}$ and $h_{2}$ are smoothed by the convolution operator, $K_{\sigma}$. This allows for intensity inhomogeneity in the foreground and background of target objects.

\subsection{Local Chan-Vese (LCV) Fitting \cite{LCV}} 

Wang et al. \cite{LCV} proposed the Local Chan-Vese (LCV) model. In terms of the equivalent convex formulation, the data fitting term is given by
\begin{equation}
f_{LCV}(u)= \int_{\Omega} \left(f_{1}(\bm{x}) - f_{2}(\bm{x})\right)u \ \mathrm{d}\Omega
\end{equation}
where
\begin{align}
f_{1}(\bm{x}) &=  \alpha\left|z - c_{1}\right|^{2}  + \beta\left|z^{*} - z- d_{1}\right|^{2}, \notag \\
f_{2}(\bm{x}) &= \alpha\left|z - c_{2}\right|^{2}  + \beta\left|z^{*} - z - d_{2}\right|^{2},
\end{align}
and $z^{*} = M_{k}*z$. Here, $M_{k}$ is an averaging convolution with $k \times k$ window. The LCV selective formulation is then given as
\begin{equation}
F_{LCV}(u)=\ TV_{g}(u)+f_{S}(u)+f_{LCV}(u).
\end{equation}
The values $c_{1},c_{2},d_{1},d_{2}$ which minimise this functional for $u_{\gamma}$ are given by
\begin{equation}
\begin{gathered}
\begin{aligned}
c_{1} &= \frac{\int_{\Omega}z u_{\gamma}\,\mathrm{d}\Omega}{\int_{\Omega}u_{\gamma}\,\mathrm{d}\Omega},\quad 
c_{2} = \frac{\int_{\Omega}z(1-u_{\gamma})\,\mathrm{d}\Omega}{\int_{\Omega}(1-u_{\gamma})\,\mathrm{d}\Omega},\\
d_{1} &= \frac{\int_{\Omega}\left(
z^{*}- z
\right)u_{\gamma}\,\mathrm{d}\Omega}{\int_{\Omega}u_{\gamma} \,\mathrm{d}\Omega},\quad
d_{2} = \frac{\int_{\Omega}\left(
z^{*}- z
\right)(1-u_{\gamma})\,\mathrm{d}\Omega}{\int_{\Omega}(1-u_{\gamma}) \,\mathrm{d}\Omega}.
\end{aligned}
\end{gathered}
\end{equation}
The formulation is minimised iteratively. The LCV fitting term that $f_{1}(\bm{x})$ and $f_{2}(\bm{x})$ includes an additional term weighted by the parameters $\alpha$ and $\beta$. The principle for the LCV model is that the difference image $z^{*}-z$ is a higher contrast image than $z$ and a two-phase segmentation on this image can be computed.

\subsection{Hybrid (HYB) Fitting \cite{Ali:16}} 

Based on extending the LCV model, Ali et al. \cite{Ali:16} proposed the following data fitting term, 
\begin{equation}
f_{HYB}(u,c_{1},c_{2},d_{1},d_{2})= \int_{\Omega} \left(f_{1}(\bm{x})-f_{2}(\bm{x})\right)u \ \mathrm{d}\Omega
\end{equation}
where
\begin{align}
f_{1}(\bm{x}) &=  \alpha\left|w - c_{1}\right|^{2}  + \beta\left|w^{*} - w - d_{1}\right|^{2}, \notag \\
f_{2}(\bm{x}) &= \alpha\left|w - c_{2}\right|^{2}  + \beta\left|w^{*} - w - d_{2}\right|^{2}.
\end{align}
Here, $z^{*} = M_{k}*z$, $w = z^{*}z$, and $w^{*} = M_{k}*w$, with $M_{k}$ the averaging convolution as used in the LCV model. The values $c_{1},c_{2},d_{1},d_{2}$ are updated in a similar way to \cite{LCV}, with further details found in \cite{Ali:16}. The authors refer to this approach as the Hybrid (HYB) Model. The HYB selective formulation is then given as
\begin{align}
F_{HYB}(u)=\ &TV_{g}(u)+f_{S}(u)+f_{HYB}(u).
\end{align}
The key aim of the HYB model is to account for intensity inhomogeneity in the foreground and background of the image through the product image $w$. In LCV, the presence of the blurred image $z^{*}$ in the data fitting term deals with intensity inhomogeneity, whilst including $z$ helps identify contrast between regions. The authors found that the product image $w=z^{*}z$ can improve the data fitting in both respects. Therefore they construct a LCV-type function with $w$ rather than the original $z$. Their results suggest that this approach is more robust.

\subsection{Generalised Averages (GAV) Fitting \cite{Ali:17}} 
\label{sec:GAV}

Recently, Ali et al. \cite{Ali:17} proposed using the data fitting terms of Chan-Vese in a signed pressure force function framework \cite{Zhang:10}. They refer to this approach as Generalised Averages (GAV) as they update the intensity constants in an alternative way, detailed below. In the convex framework, we consider the selective GAV functional:
\begin{equation}
F_{GAV}(u)=TV_{g}(u)+f_{S}(u)+f_{GAV}(u),
\end{equation}
where $f_{GAV}(u)=f_{CV}(u)$. This is identical to the CV selective formulation (\ref{eqn:selectCV}). However, the authors propose an alternative update for the fitting constants $c_{1}$ and $c_{2}$, given as follows:
\begin{equation}
c_{1} = \frac{\int_{\Omega}z^{\beta}u_{\gamma}\,\mathrm{d}\Omega}{\int_{\Omega}z^{\beta-1}u_{\gamma}\,\mathrm{d}\Omega},\qquad c_{2} = \frac{\int_{\Omega}z^{\beta}(1-u_{\gamma})\,\mathrm{d}\Omega}{\int_{\Omega}z^{\beta-1}(1-u_{\gamma})\,\mathrm{d}\Omega},
\end{equation}
with $\beta\in\mathbb{R}$. If $\beta = 1$, the approach is identical to CV. In \cite{Ali:17} the authors assert that the proposed adjustments have the following properties. As $\beta\rightarrow\infty$, $c_{1}$ and $c_{2}$ approach the maximum and minimum intensity in the foreground and background of the image, respectively. Also, as $\beta\rightarrow -\infty$, $c_{1}$ and $c_{2}$ approach the minimum intensity in the foreground and background of the image, respectively. For example, if a high value of $\beta$ is set, $c_{1}$ will take a larger value than in CV which can be useful for selective segmentation. For example, if we consider the image in Fig.~\ref{fig:CTscan} we can achieve a larger $c_{2}$ value by setting $\beta >1$ and a smaller value by setting $\beta <1$. Therefore, there is more flexibility when using this data fitting term in selective formulations. However, it should be noted that it involves the selection of the parameter $\beta$, which can be difficult to optimise.

\section{Alternative Selective Segmentation Models}
\label{sec:selective}

We now introduce two recent methods that incorporate user input to perform selective segmentation. Each involves input in the form of foreground/background regions to indicate relevant structures of interest. An example of this can be seen in Fig. \ref{fig:inputimages}, where red regions indicate foreground and blue regions indicate background. We compare against the work of Nguyen et al. \cite{Nguyen:12}, which uses a similar convex relaxation framework to the proposed approach, and Dong et al. \cite{SRW}, which uses a variation of the random walk approach. We summarise the essential aspects of each approach in the following.

\subsection{Constrained Active Contours (CAC) \cite{Nguyen:12}}
\label{sec:CAC}

The authors use a probability map, $P({\bm x})$, from Bai and Sapiro \cite{Bai:07} where the geodesic distances to the foreground/background regions are denoted by $D_{F}({\bm x})$ and $D_{B}({\bm x})$, respectively. An approximation of the probability that a point ${\bm x}$ belongs to the foreground is then given by 
\begin{equation} \label{eqn:P1}
P({\bm x})=\frac{D_{B}({\bm x})}{D_{F}({\bm x})+D_{B}({\bm x})}.
\end{equation}
Foreground/background Gaussian mixture models (GMM) are estimated from the user input. The terms $Pr({\bm x}|F)$ and $Pr({\bm x}|B)$ denote the probability that a point, ${\bm x}$, belongs to the the foreground and background, respectively. The normalised log likelihood for each is then given by 
\begin{align}
P_{F}({\bm x})&=-\log Pr({\bm x}|F)/(-\log Pr({\bm x}|F)-\log Pr({\bm x}|B)), \notag \\ 
P_{B}({\bm x})&=-\log Pr({\bm x}|B)/(-\log Pr({\bm x}|F)-\log Pr({\bm x}|B)).
\end{align}
GMMs are widely used in selective segmentation \cite{Falcao:02,Grabcut,Bai:07,RW,SRW} and the authors in \cite{Nguyen:12} incorporate this idea into the framework we consider with the following data fitting term:
\begin{equation}
h_{c}({\bm x})=\alpha_{0}\left(P_{B}({\bm x})-P_{F}({\bm x})\right)+(1-\alpha_{0})\left(1-2P({\bm x})\right),
\end{equation}
for a weighting parameter $\alpha_{0}\in[0,1]$. It is proposed that $\alpha_{0}$ is selected automatically as follows:
\begin{equation} \label{eqn:alpha}
\alpha_{0}=\frac{1}{N}\sum_{i=1}^{N}\left\lvert\frac{\log Pr(x_{i}|F)-\log Pr(x_{i}|B)}{\log Pr(x_{i}|F)+\log Pr(x_{i}|B)}\right\rvert,
\end{equation}
where $N$ is the total number of pixels in the image. Defining $g_{0}$ as the function $g(s)$ applied to the image $z({\bm x})$ and $g_{p}$ applied to the GMM probability map $P_{F}({\bm x})$, an enhanced edge function is defined as
\begin{equation}
g_{c}({\bm x})=\beta_{0} g_{p}+(1-\beta_{0})g_{0},
\end{equation}
for a weighting parameter $\beta_{0}\in[0,1]$, which can be set automatically in a similar way to \eqref{eqn:alpha}. Thus, Nguyen et al. \cite{Nguyen:12} define the Constrained Active Contours (CAC) Model as
\begin{equation}
\min_{u\in[0,1]}\left\lbrace \int_{\Omega}g_{c}({\bm x})|\nabla u({\bm x})|\ \mathrm{d}\Omega+\lambda\int_{\Omega}h_{c}({\bm x})u({\bm x})\ \mathrm{d}\Omega \right\rbrace.
\end{equation}
They obtain a solution using the split Bregman method of Goldstein et al. \cite{Goldstein:10}, although other methods are applicable and will yield similar results. However, that is not the focus of this paper so we omit the details here. In the results section, \S\ref{sec:results}, we will compare our method against CAC to see how our data fitting term compares against a GMM-based approach. 

\subsection{Submarkov Random Walks (SRW) \cite{SRW}}
\label{sec:SRW}

We now introduce a recent selective segmentation method by Dong et al. \cite{SRW} known as Submarkov Random Walks (SRW). Rather than using the continuous framework of \cite{Chan:06}, this approach is based in the discrete setting where each pixel in the image is treated as a node in a weighted graph. Random walks (RW) have been widely used for segmentation since the work of Grady \cite{RW}. SRW is capable of achieving impressive results with user-defined foreground and background regions. The selective segmentation result can be obtained by assigning a label to each pixel based on the computed probabilities of the random walk approach. For brevity, we do not provide the full details of the method here, however, further details can be found in \cite{SRW}. We compare SRW to our proposed approach on a CT data set in \S\ref{sec:r4}.

We now introduce essential notation to understand the approach of \cite{SRW}. In RW an image is formulated as a weighted undirected graph $G=(V,E)$ with nodes $v\in V$ and edges $e\in E\subseteq V\times V$. Each node $v_{i}$ represents an image pixel $x_{i}$. An edge $e_{ij}$ connects two nodes $v_{i}$ and $v_{j}$ and a weight $w_{ij}\in W$ of edge $e_{ij}$ measures the likelihood that a random walker will cross this edge:
\begin{equation} \label{eqn:wij}
w_{ij}=\exp\left(-\frac{||I_{i}-I_{j}||^{2}}{\sigma_{0}}\right)+\epsilon_{0},
\end{equation}
where $I_{i}$ and $I_{j}$ are pixel intensities, with $\sigma_{0},\epsilon_{0}\in\mathbb{R}$. In SRW a user indicates foreground/background regions in a similar way to CAC, as shown in Fig. \ref{fig:inputimages}, and can be viewed as a traditional random walker with added auxiliary nodes. In \cite{SRW}, these are defined as a set of labelled nodes $V_{M}=\{V^{l_{1}},V^{l_{2}},...,V^{l_{K}}\}$. A set of labels is defined, $LS=\{l_{1},l_{2},...,l_{K}\}$, with $K$ the number of labels $V^{l_{k}}=\{V_{1}^{l_{1}},V_{2}^{l_{1}},...,V_{M_{K}}^{l_{K}}\}$, and $M_{k}$ the number of seeds labelled $l_{k}$. The prior is then constructed from the seeded nodes (defined by the user). Assuming a label $l_{k}$ has an intensity distribution $H_{k}$ (based on GMM learning), a set of auxiliary nodes $H_{k}=\{h_{1},h_{2},\cdots,h_{K}\}$ is added into an expanded graph $G_{e}$ to define a graph with prior $\bar{G}$. Each prior node is connected with all nodes in $V$ and the weight, $w_{ih_{k}}$, of an edge between a prior node $h_{k}$ and a node $v_{i}\in V$ is proportional to $u^{k}_{i}$, the probability density belonging to $H_{k}$ at $v_{i}$ .

The authors define the probabilities of each node $v_{i}\in V$ belonging to label $l_{k}$ as the average reaching probability, denoted $\bar{r}_{i}^{l_{k}}$. This term incorporates the auxillary nodes introduced above and is dependent on multiple variables and parameters, including $w_{ij}$ \eqref{eqn:wij}. Further details can be found in \cite{SRW}. The segmentation result is then found by solving the following discrete optimisation problem:
\begin{equation}
\bar{R}_{i}=\arg \max_{l_{k}}\bar{r}_{i}^{l_{k}},
\end{equation}
where $\bar{R}_{i}$ represents the final label for each node. In other words, for a two-phase segmentation problem, $\bar{R}_{i}$ is analogous to the discretised solution of a convex relaxation problem in the continuous setting. Comparisons in terms of accuracy can therefore be made directly, which we elaborate on further in \S\ref{sec:results}. The authors also detail the optimisation procedure and aspects of dealing with noise reduction.

\section{Proposed Model}
\label{sec:proposed}

\begin{figure*}
\centering
\floatbox[{\capbeside\thisfloatsetup{capbesideposition={left,top},capbesidewidth=1.5in}}]{figure}[\FBwidth]
{
\captionsetup{position=bottom}\captionsetup[subfigure]{labelformat=empty,font=normalsize} 
\subfloat[{(i) $\gamma_{1} = 0.1 , \gamma_{2} = 0.2$}]{\includegraphics[width= 1.65in,height=3.6cm]{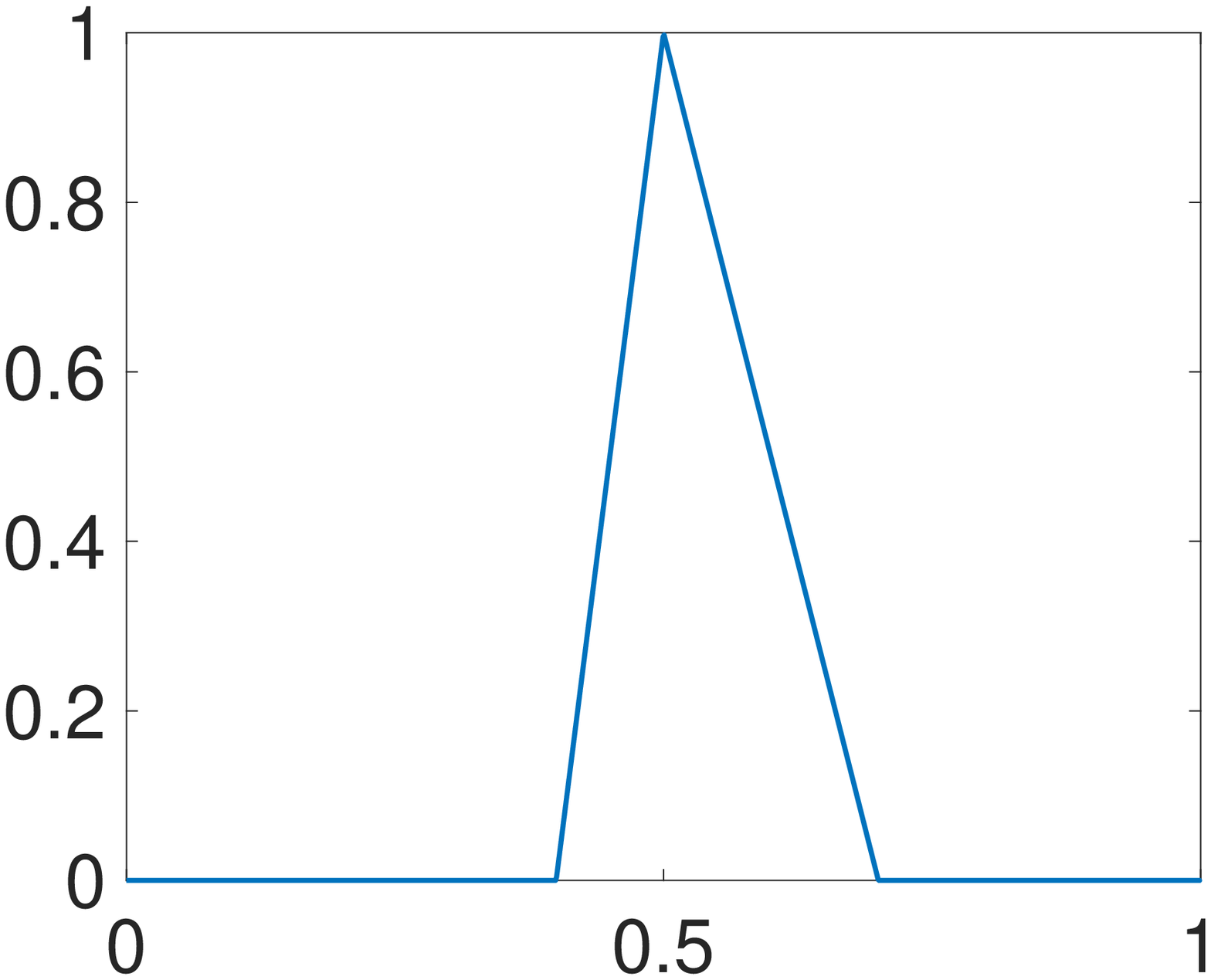}}
\captionsetup{position=bottom}\captionsetup[subfigure]{labelformat=empty,font=normalsize} 
\subfloat[{(ii) $\gamma_{1} = 0.3, \gamma_{2} = 0.2$}]{\includegraphics[width= 1.65in,height=3.6cm]{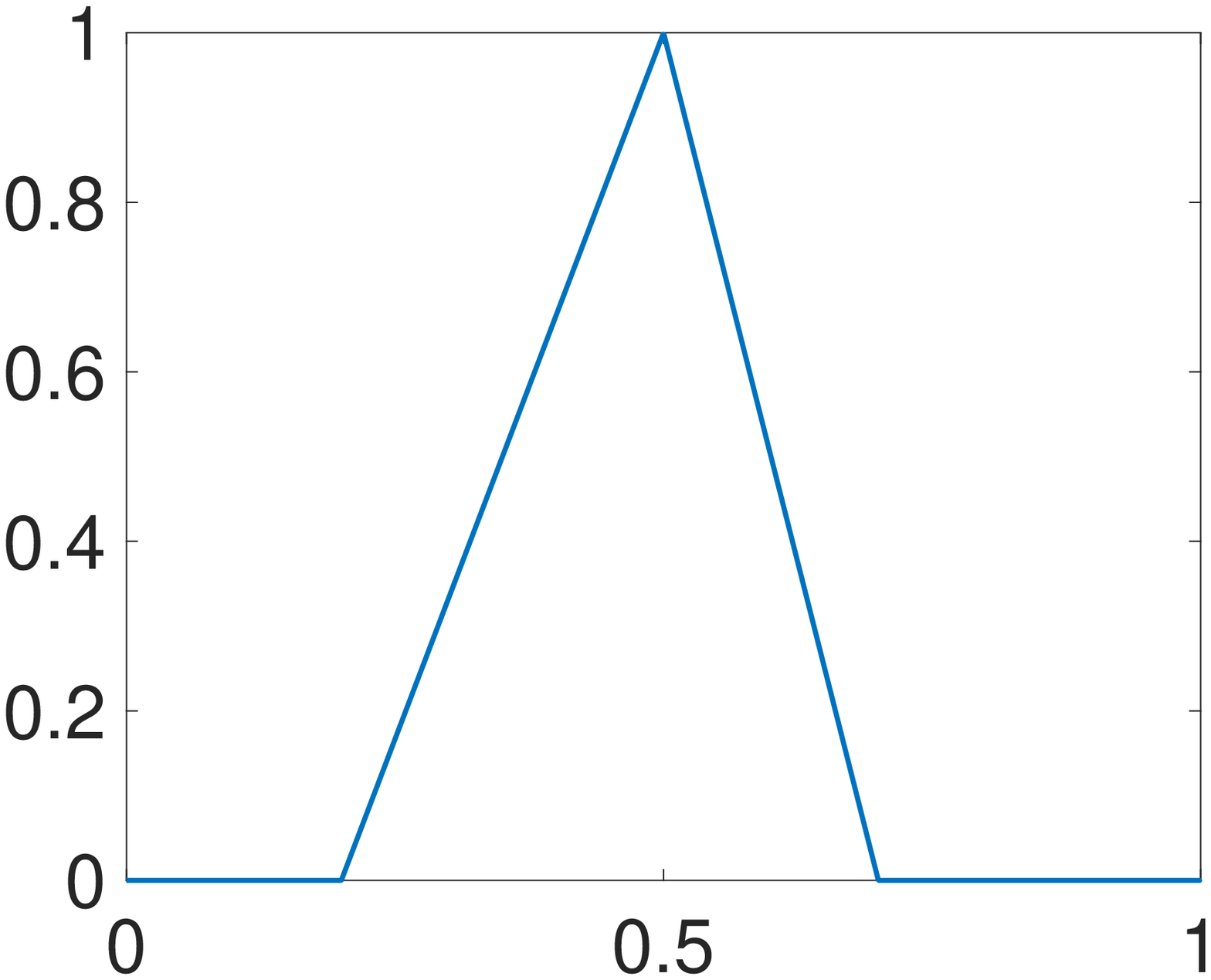}}
\captionsetup{position=bottom}\captionsetup[subfigure]{labelformat=empty,font=normalsize} 
\subfloat[{(iii) $\gamma_{1} = 0.3, \gamma_{2} = 0.4$}]{\includegraphics[width= 1.65in, height = 3.6cm]{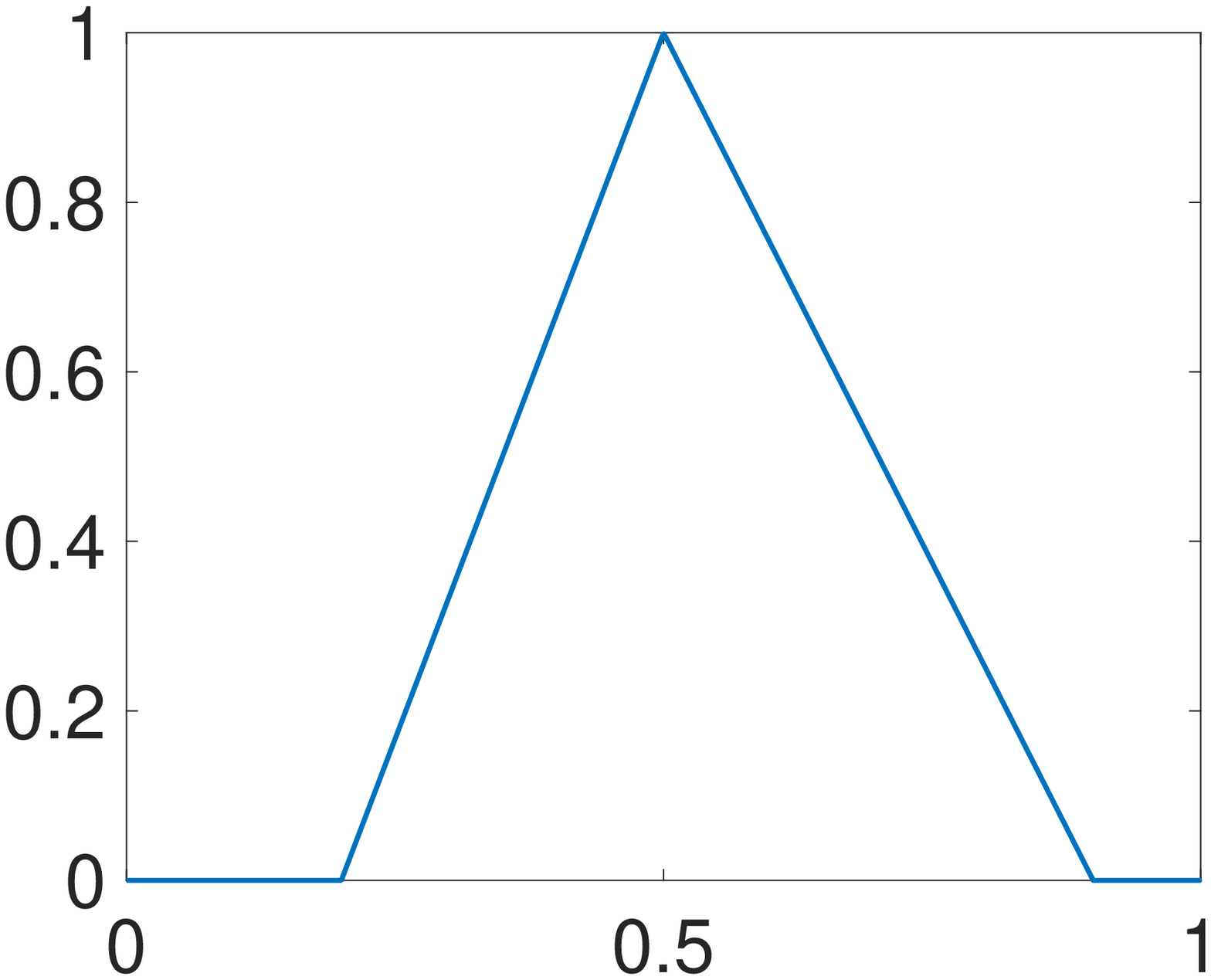}}
}
{ \caption{Three 1D plots of $\tilde{f}_{2}(\bm{x})$ whilst varying $\gamma_{1}$ and $\gamma_{2}$ (with $c_{1} = 0.5$). \label{fig:g1g2plots}}}
\end{figure*}

In this section we introduce the proposed data fitting term for selective segmentation. We consider objects that are approximately homogeneous in the target region. Intrinsically, it is then assumed that the region $\mathcal{P}$, provided by the user, is likely to provide a reasonable approximation of the optimal $c_{1}$ value and therefore an appropriate foreground fitting function, $f_{1}$, is given by CV (\ref{eqn:c1c2}). For this reason, it makes sense to retain this term in the proposed approach. The contradiction is in how the background fitting function $f_{2}$ is defined. Considering piecewise-constant assumptions of the image, and many of the related approaches, the background is expected to be defined by a single constant value, $c_{2}$. If $c_{1}\approx c_{2}$ then $f_{2}\approx f_{1}$ everywhere, and therefore the fitting term can't accurately separate background regions from the foreground. It is not practical to rely on $f_{S}(u)$ to overcome this difficulty as it will produce an over-dependence on the choice of $\mathcal{M}$ and $\mathcal{P}$. This is prohibitive in practice. An alternative function $f_{2}$ must therefore be defined which is compatible with $f_{1}$ and $f_{S}(u)$. Here, we define a new data fitting term that penalises background objects in such a way that avoids these problems by allowing intensity variation above and below the value $c_{1}$. In order to design a new functional, we first look at the original CV background fitting function
\[
f_{2}=(z(\bm{x})-c_{2})^{2}.
\]
It is clear that in an approximately piecewise-constant image this function will be small outside the target region (i.e. where the image takes values near $c_{2}$) and larger inside the target region. Our aim in a new fitting term is to mimic this in such a way that is consistent with selective segmentation, where regions with a `foreground intensity' are forced to be in the background. It is beneficial to introduce two parameters, $\gamma_{1}$ and $\gamma_{2}$, to enforce the penalty on regions of intensity in the range $[c_{1}-\gamma_{1},c_{1}+\gamma_{2}]$, i.e. enforce the penalty asymmetrically around $c_{1}$. We propose the following function to achieve this:
\begin{equation}\label{eqn:newfuncdef}
\tilde{f}_{2}(\bm{x})= 
\begin{cases}
1+\frac{z(\bm{x})-c_{1}}{\gamma_{1}}, & c_{1} - \gamma_{1} \le z(\bm{x}) \le c_{1} \\
1-\frac{z(\bm{x})-c_{1}}{\gamma_{2}}, & c_{1} < z(\bm{x})\le c_{1} + \gamma_{2}\\
0, & else.
\end{cases}
\end{equation}
This function takes its maximum value where $z(\bm{x})=c_{1}$ and is $0$ for $z(\bm{x})>c_{1}-\gamma_{1}$ and $z(\bm{x}) < c_{1}+\gamma_{2}$. In Fig.~\ref{fig:g1g2plots} we provide a 1D representation of $\tilde{f}_{2}(\bm{x})$ for various choices of $\gamma_{1}$ and $\gamma_{2}$, with $z(\bm{x})\in[0,1]$ and $c_{1} = 0.5$. Here, it can be seen how the proposed data fitting term acts as a penalty in relation to a fixed constant $c_{1}$. It is analogous to CV, whilst accounting for the idea of selective segmentation with a data fitting term. The main advantage of this term is that it replaces the dependence on $c_{2}$ in the formulation, which has no meaningful relation to the solution of a selective segmentation problem. Even when the foreground is relatively homogeneous, the background may have intensities of a similar value to $c_{1}$ which will cause difficulties in obtaining an accurate solution. We detail the proposed fitting term in the following section.

\subsection{New Fitting Term}

\begin{figure*}
\centering
\floatbox[{\capbeside\thisfloatsetup{capbesideposition={left,top},capbesidewidth=1.5in}}]{figure}[\FBwidth]
{
\captionsetup{position=bottom}\captionsetup[subfigure]{labelformat=empty,font=normalsize} 
\subfloat[{(i) Image and $\mathcal{P}$}]{\includegraphics[width= 1.65in,height=3.6cm]{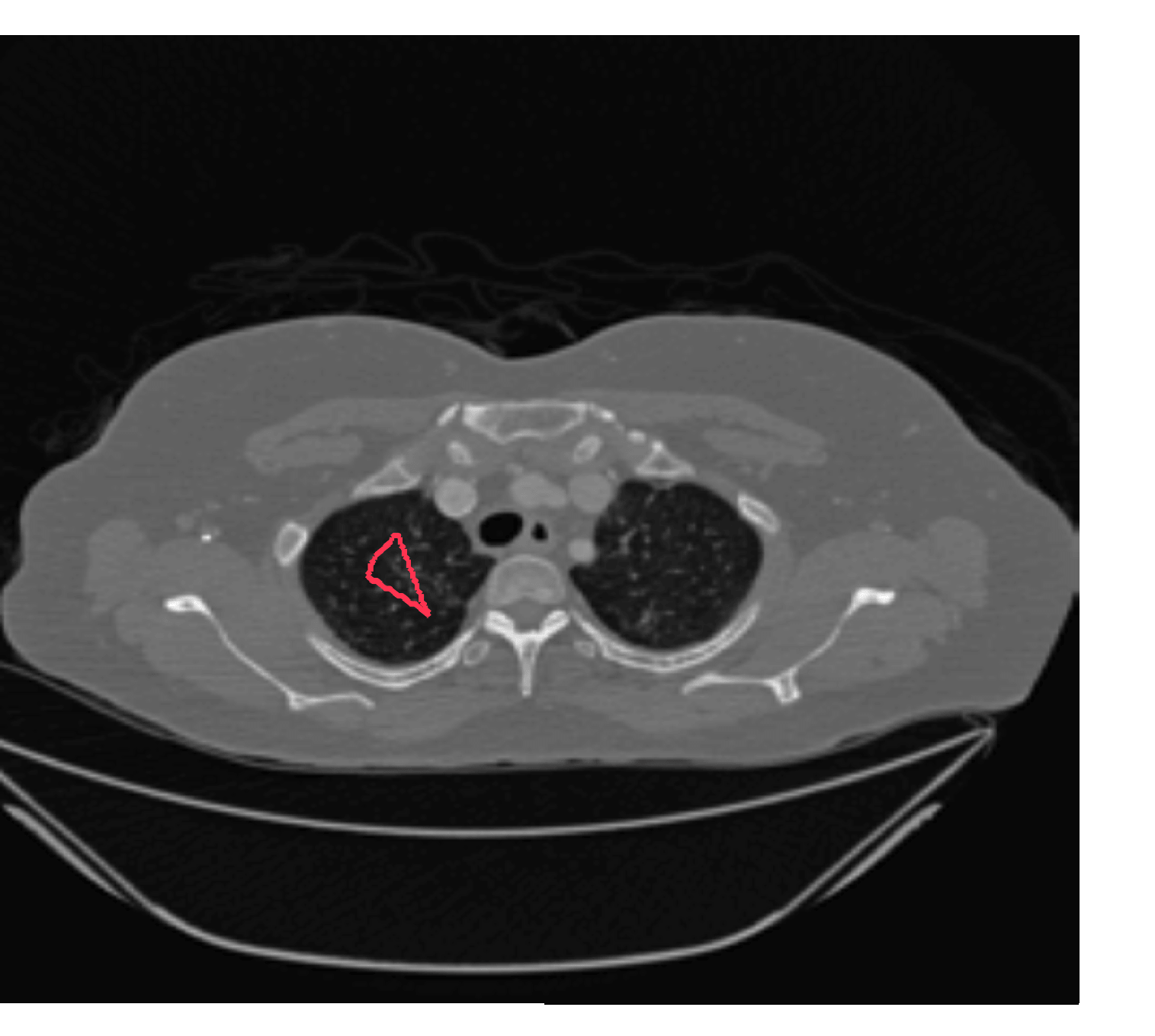}}\quad
\captionsetup{position=bottom}\captionsetup[subfigure]{labelformat=empty,font=normalsize} 
\subfloat[\kern-3em{(ii) CV fitting}]{\includegraphics[width= 1.65in,height=3.6cm]{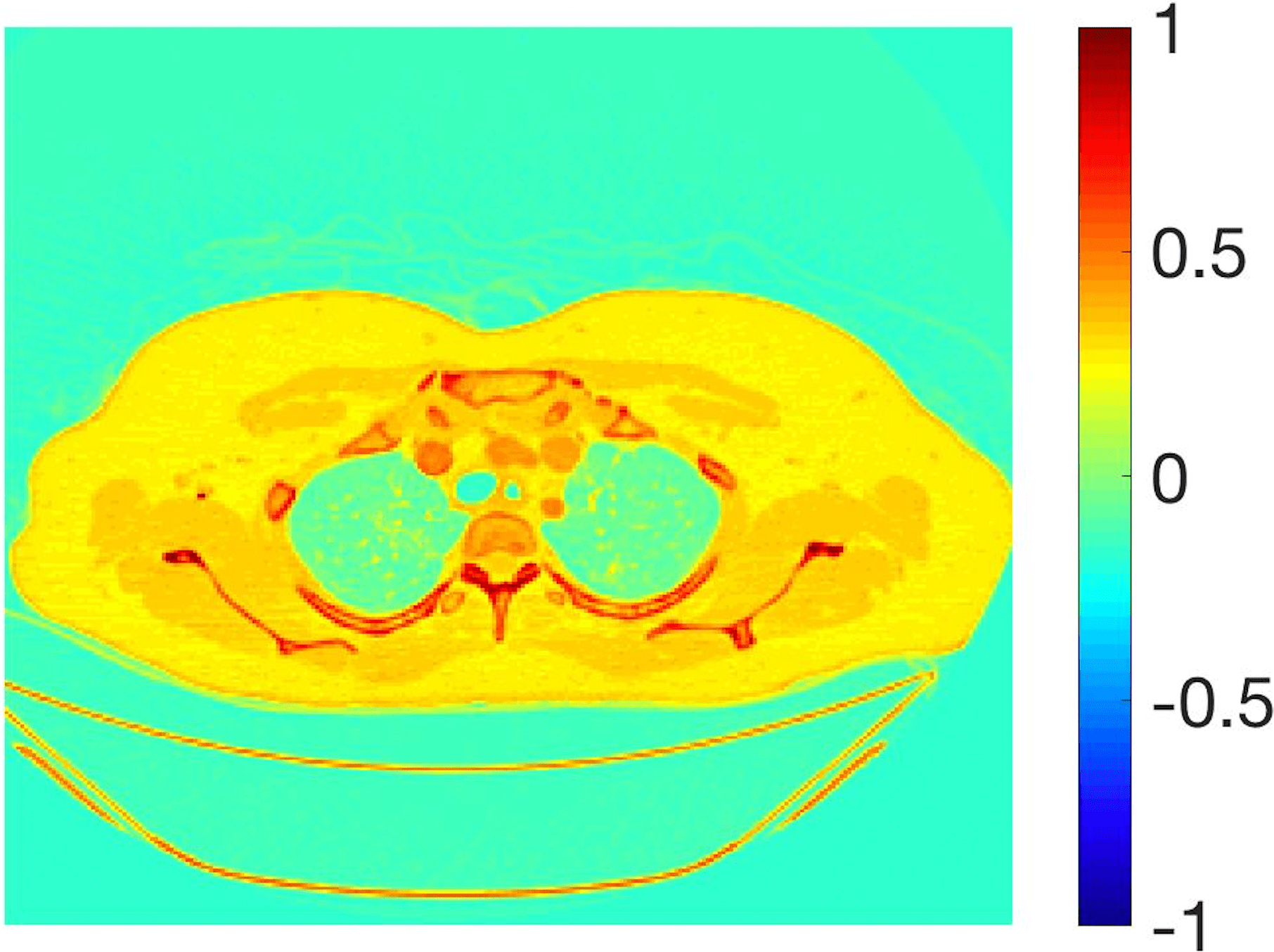}}\quad
\captionsetup{position=bottom}\captionsetup[subfigure]{labelformat=empty,font=normalsize} 
\subfloat[\kern-2em{(iii) New fitting}]{\includegraphics[width= 1.65in,height=3.6cm]{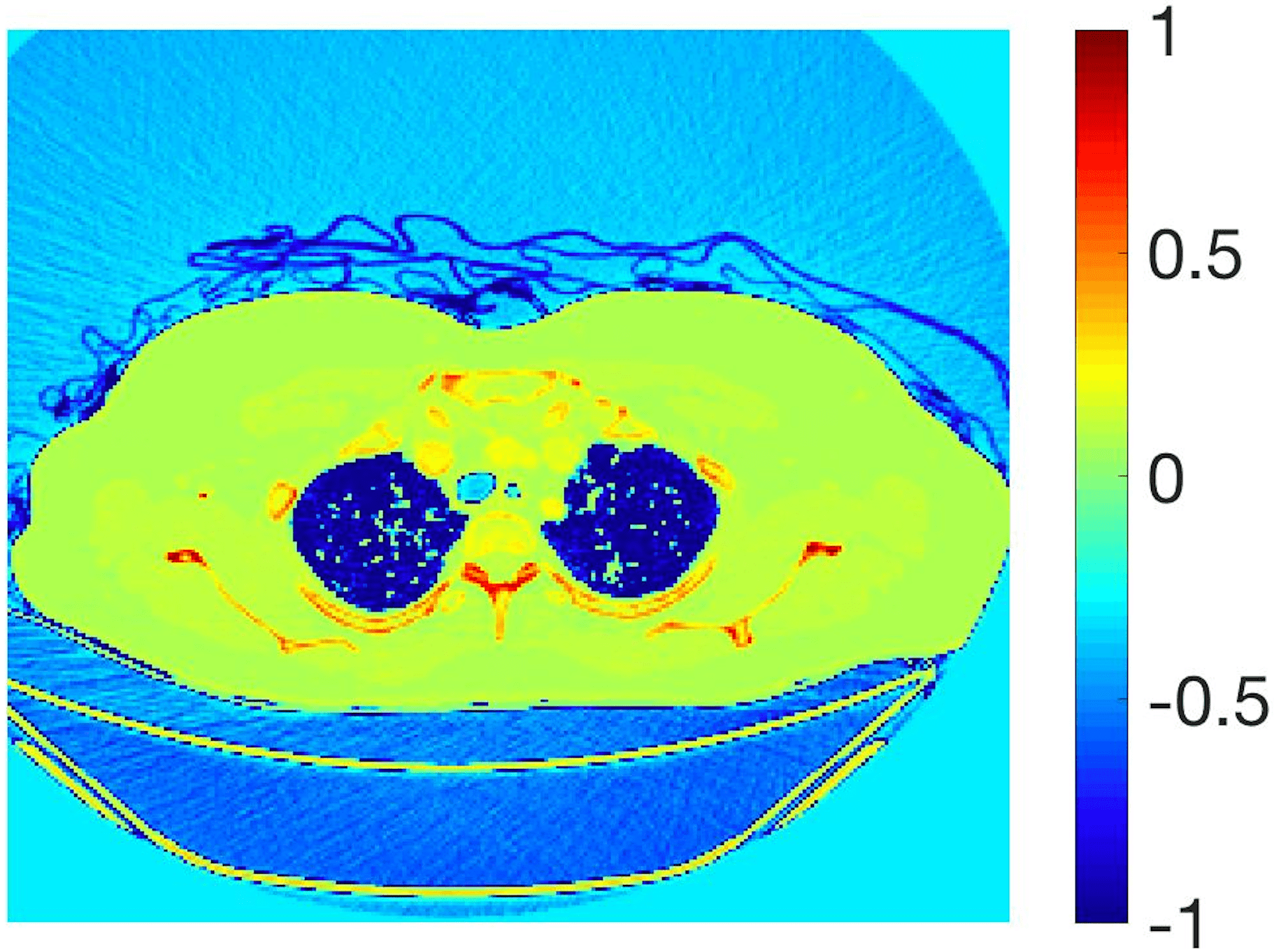}}
}
{\caption{An image with user input $\mathcal{P}$ shown in red ($c_{1} = 0.152, c_{2} = 0.188$). Here, we show the difference between the CV fitting function and the proposed approach. The target region is clearly defined by negative values in (iii). \label{fig:newfittingterm}}}
\end{figure*}

We define the proposed data fitting functional as follows:
\begin{equation}\label{eqn:newfitting}
f_{PM}(u):=\int_{\Omega}(\lambda_{1}f_{1}(\bm{x})-\lambda_{2}\tilde{f_{2}}(\bm{x}))u\ \ \mathrm{d}\Omega,
\end{equation}
for $f_{1}(\bm{x})=(z-c_{1})^{2}$ and $\tilde{f}_{2}(\bm{x})$ as defined in (\ref{eqn:newfuncdef}). This is consistent with respect to the intensities of the observed object and the concept of selective segmentation. In Fig.~\ref{fig:newfittingterm} we see the difference between CV and the proposed fitting terms for given user input on a CT image. For the CT image, the CV fitting terms are near 0 within the target region. This is despite there being a distinct homogeneous area with good contrast on the boundary. This illustrates the problem we are aiming to overcome. With the proposed fitting term this phenomenon should be avoided in cases like this. By defining $\tilde{f}_{2}$ as in \eqref{eqn:newfuncdef} there is no contradiction if the foreground and background intensities of the target region are similar.

For images where we assume that the target foreground is approximately homogeneous, we have generally found that fixing $c_{1}$ according to the user input is preferable. We compute $c_{1}$ as the average intensity inside the region $\mathcal{P}$ formed from the user input marker point set.
We therefore propose to minimise the following functional with respect to $u\in[0,1]$, given a fixed $c_{1}$ :
\begin{equation}\label{eqn:gensegfunc2}
F_{PM}(u)=TV_{g}(u)+f_{PM}(u)+f_{S}(u).
\end{equation}
where $f_{S}$ is the geodesic distance computed as described earlier using (\ref{eqn:DG0}). The minimisation problem is given as
\begin{equation}
\min_{u\in[0,1]}F_{PM}(u)
\end{equation}
The model consists of weighted TV regularisation with a geodesic distance constraint as in \cite{Geo}. However, alternative constraints are possible, such as Euclidean \cite{CDSS}, or moments \cite{Klodt:13}. It is important to note that we have defined the model in a similar framework to the related approaches discussed previously. The main idea is to establish how the proposed fitting term, $f_{PM}(u)$, performs compared to alternative methods. Next we describe how we determine the values of $\gamma_{1}$ and $\gamma_{2}$ in the function $\tilde{f}_{2}(\bm{x})$ automatically. This is important in practice as it avoids any additional user input or parameter dependence to achieve an accurate result. In subsequent sections we provide details of how we obtain a solution for the proposed model.

\subsection{Parameter Selection}

\begin{figure*}
\centering
\floatbox[{\capbeside\thisfloatsetup{capbesideposition={left,top},capbesidewidth=1.5in}}]{figure}[\FBwidth]
{
\captionsetup{position=bottom}\captionsetup[subfigure]{labelformat=empty,font=normalsize} 
\subfloat[(i) Test Image 1]{\includegraphics[width= 1.65in,height=3.6cm]{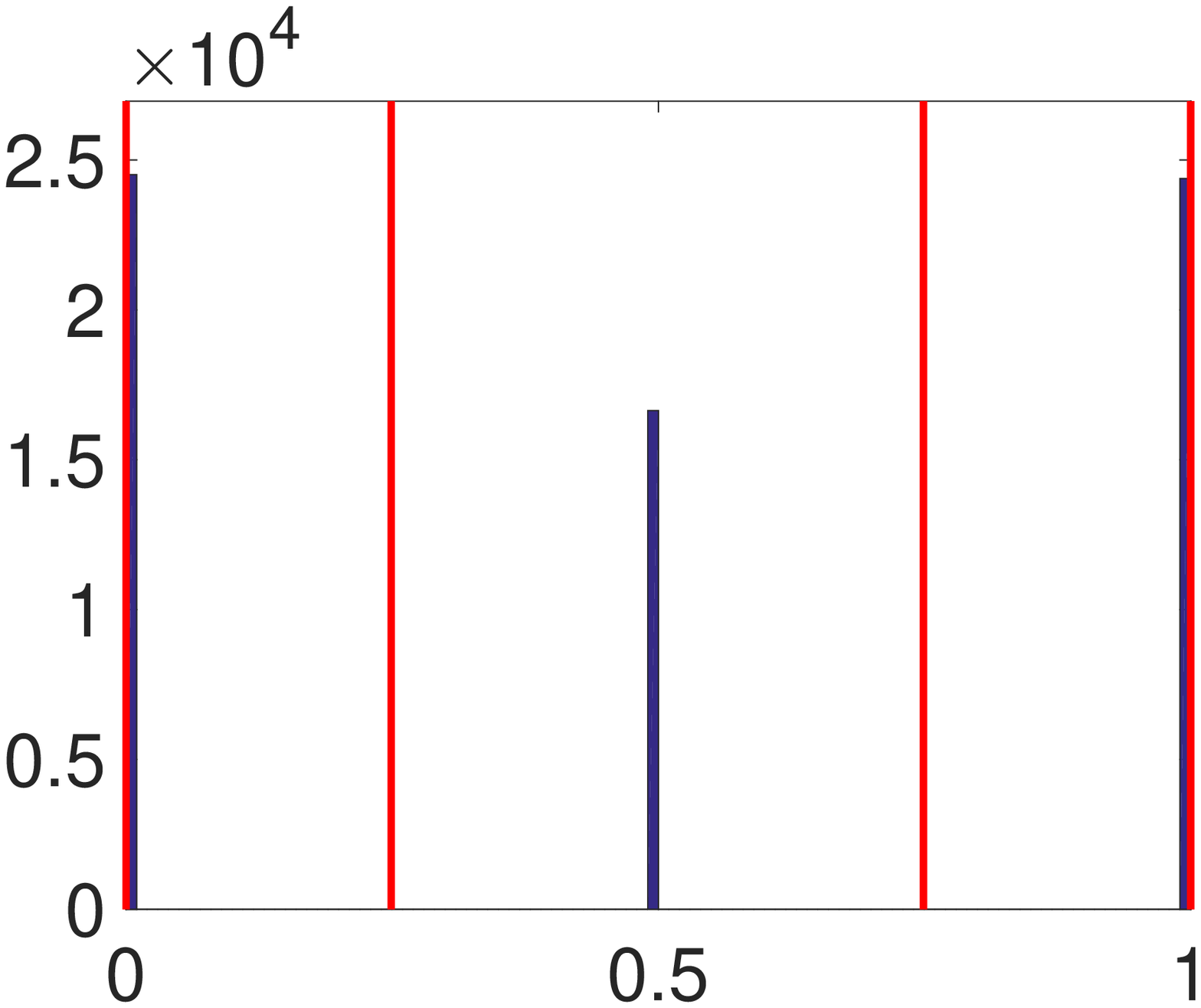}}
\captionsetup{position=bottom}\captionsetup[subfigure]{labelformat=empty,font=normalsize} 
\subfloat[(ii) Test Image 7]{\includegraphics[width= 1.65in,height=3.6cm]{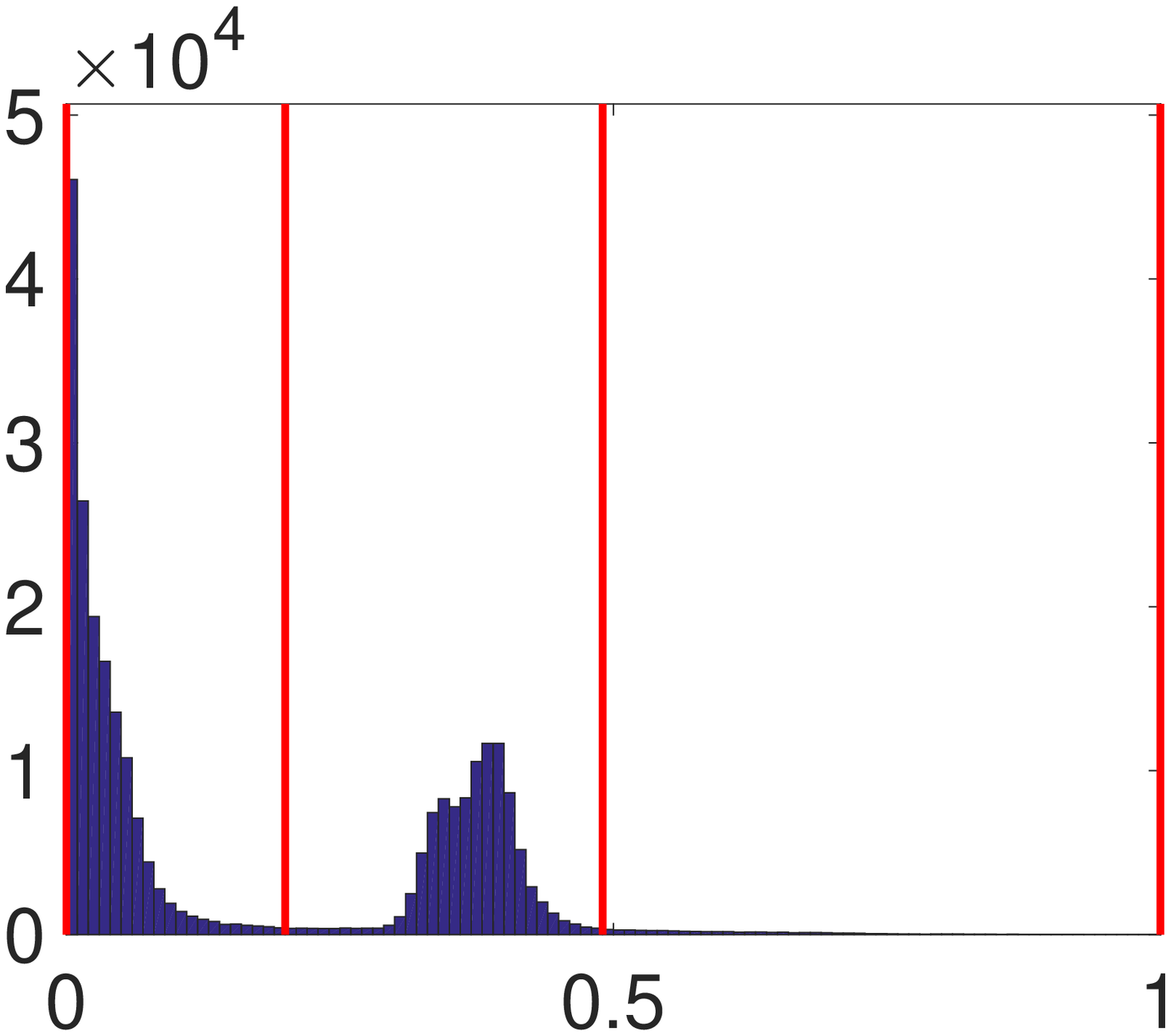}}
\captionsetup{position=bottom}\captionsetup[subfigure]{labelformat=empty,font=normalsize} 
\subfloat[(iii) Test Image 9]{\includegraphics[width= 1.65in, height = 3.6cm]{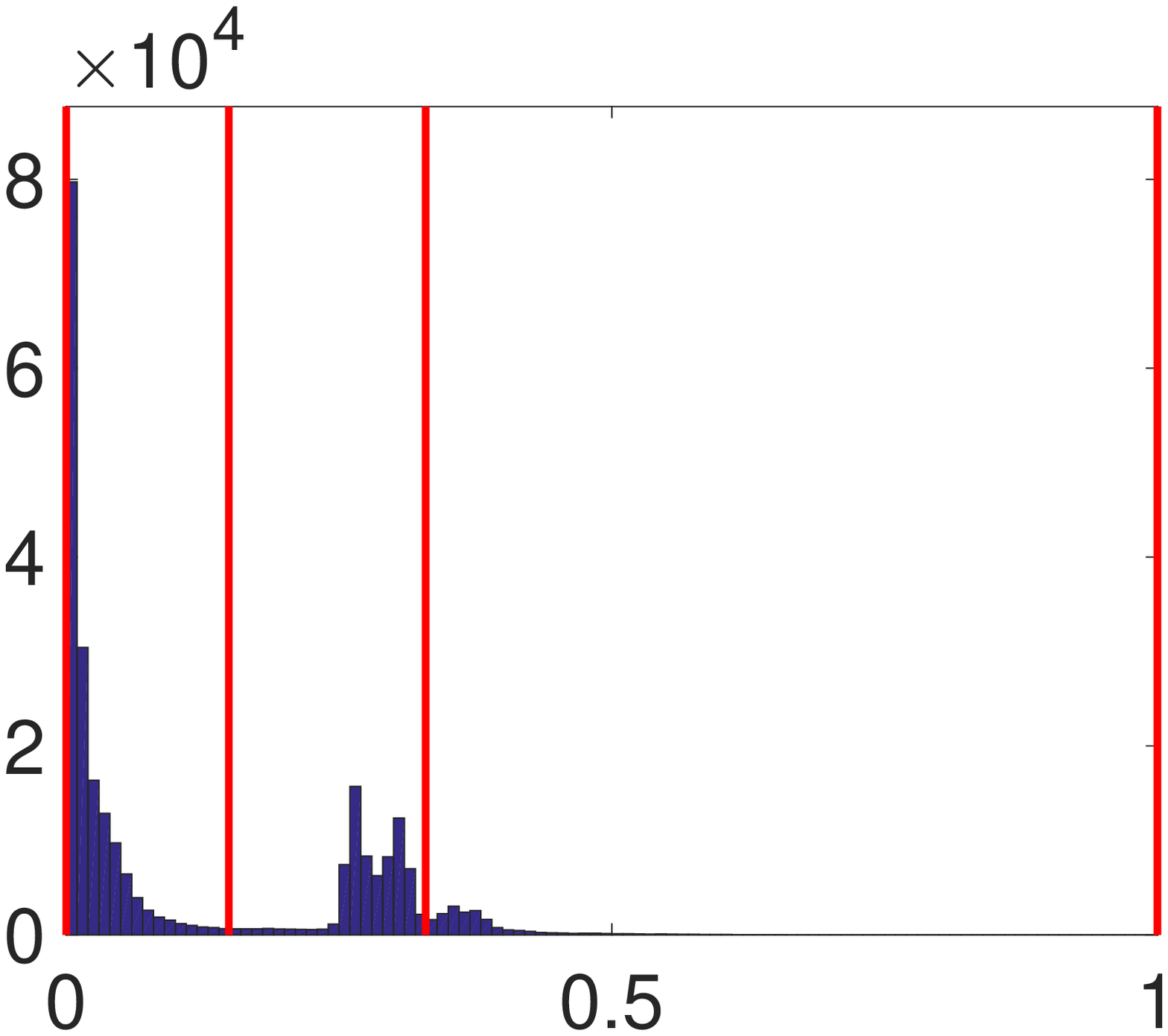}}
}
{\caption{The histograms of intensities for some example images. The red lines are the automatic thresholds $T_{i}$ obtained by Otsu's thresholding with $N=3$. \label{fig:otsu}}}
\end{figure*}

For a particular problem it is quite straightforward to optimise the choice of $\gamma_{1}$ and $\gamma_{2}$ experimentally, but we would like a method which is not sensitive to the choice of $\gamma_{1}$ and $\gamma_{2}$ and would also prefer that the user need not choose these values manually. Therefore, in this section we explain how to choose these values automatically based on justifiable assumptions about general selective segmentation problems. To select the parameters $\gamma_{1}$ and $\gamma_{2}$ we use Otsu's method \cite{Otsu:79} to divide the histogram of image intensities into $N$ partitions. Otsu's thresholding is an automatic clustering method which chooses optimal threshold values to minimise the intra-class variance. This has been implemented very efficiently in MATLAB in the function $\texttt{multithresh}$ for dividing a histogram such that there are $N-1$ thresholds $T_{i}$. 

We use the thresholds from Otsu's method to find $\gamma_{1}$ and $\gamma_{2}$ as follows. There are three cases to consider, based on the value of $c_{1}$ computed from the user input: i) $T_{i-1}\le c_{1}\le T_{i}$ for some $i>1$, ii) $ 0 \le c_{1}\le T_{1}$, iii) $ T_{N-1} \le c_{1}\le 1$. For each case we set the parameters as follows:

\begin{enumerate}
\item[(i)] $\gamma_{1} = c_{1} - T_{i-1},\ \ \ \gamma_{2} = T_{i} - c_{1}$
\item[(ii)] $\gamma_{1}=c_{1},\ \ \ \gamma_{2} = T_{1} - c_{1}$
\item[(iii)] $\gamma_{1} = c_{1}-T_{N-1},\ \ \ \gamma_{2} = 1-c_{1}$
\end{enumerate}
Choosing $N$ too large could mean $\gamma_{1}$ and $\gamma_{2}$ are too small as the histogram would be partitioned too precisely. Generally we only ever need to consider a maximum of 3 phases for selective segmentation. If there is a large number of pixels in the image with intensity above or below $c_{1}$ the image can be considered two-phase in practice. Conversely, if a large number of pixels in the image have intensity above and below $c_{1}$ the image can essentially be considered three-phase in the context of selective segmentation. This is due to the way $\tilde{f_{2}}$ has been defined. Therefore, we set $N=3$ for all tests. In Fig.~\ref{fig:otsu} we can see the Otsu thresholds chosen for various images given in this paper. They divide the peaks in the histogram well and once we know the value of $c_{1}$ (the approximation of the intensity of the object we would like to segment) we can automatically choose $\gamma_{1}$ and $\gamma_{2}$ according to this criteria.

\section{Numerical Implementation}
\label{sec:numerics}

We now introduce the framework in which we compute a solution to the minimisation of the proposed model, as well the related models introduced in \S\ref{sec:intro} and \S\ref{sec:related}. All consist of the minimisation problem
\begin{equation}\label{eqn:genmin}
\min_{u\in[0,1]}\left\lbrace F_{X}(u)=TV_{g}(u)+f_{X}(u)+f_{S}(u)\right\rbrace,
\end{equation}
for $X=\text{CV, RSF, LCV, HYB, GAV, PM}$ respectively. Minimisation problems of this type \eqref{eqn:genmin} have been widely studied in terms of continuous optimisation in imaging, including two-phase segmentation.  A summary of such methods in recent years is given by Chambolle and Pock \cite{CPintro}. Details of the introduction of binary labels to image segmentation can be found in Lie et al. \cite{LieLysakerTai} and Chan et al. \cite{Chan:06}, and our numerical scheme follows the approach in \cite{Chan:06}: enforcing the constraint in \eqref{eqn:genmin} with a penalty function, and deriving the Euler-Lagrange of the regularised functional. We then solve the corresponding PDE by following a splitting scheme first applied to this kind of problem by Spencer and Chen \cite{CDSS}. Whilst the numerical details are not the focus of the work, it is important to note widely used alternatives. A summary of such approaches, describing major developments in this area and the connections between each method is given in a review by Wei et al. \cite{Wei:16}.

It has proved very effective to exploit the duality in the functional and avoid smoothing the TV term. A prominent example is the split Bregman approach for segmentation by Goldstein et al. \cite{Goldstein:10}. This is closely related to augmented lagrangian methods, a matter further discussed by Boyd et al. \cite{Boyd:11}. Analogous approaches also consist of the first-order primal dual algorithm of Chambolle and Pock \cite{ChambollePock} and the max-flow/min-cut framework detailed by Yuan et al. \cite{Yuan:13}. There are practical advantages in implementing such a numerical scheme for our problem, primarily in terms of computational speed. However, in the numerical tests we include we're mainly interested in accuracy comparisons. For this purpose the convex splitting algorithm of \cite{CDSS} is sufficient, and the extension of splitting schemes for convex segmentation problems may be of interest. Further details can be found in \cite{CDSS} and \cite{Geo}. In the following, we first discuss the minimisation of \eqref{eqn:genmin} in a general sense and then mention some important aspects in relation to the alternative fitting terms discussed in \S\ref{sec:related}. 

\subsection{Finding the Global Minimiser}

To solve this constrained convex minimisation problem (\ref{eqn:unconmin}) we use the Additive Operator Splitting (AOS) scheme from Gordeziani et al. \cite{Gordeziani:74}, 
Lu et al. \cite{Tai:91} and Weickert et al. \cite{Weickert:98}. This is used extensively for image segmentation models \cite{Rada:13,Geo,CDSS}. It allows the 2D problem to be split into two 1D problems, each solved separately, with the results combined in an efficient manner. We address some aspects of AOS in \S\ref{sec:alg}, with further details provided in \cite{Geo,CDSS}.

A challenge with the functional (\ref{eqn:gensegfunc2}), particularly with respect to AOS, is that this is a constrained minimisation problem. Consequently, it is reformulated by introducing an exact penalty function, $\nu(u)$, given in \cite{Chan:06}. To simplify the formulation we define
\[
r(\bm{x})=\theta\mathcal{D}(\bm{x})+f(\bm{x}),
\]
$f(\bm{x})$ is the function associated with $f_{X}(u)$. We introduce a new parameter, $\tilde{\lambda}$, which  allows us to balance the data fitting terms to the regularisation term more reliably. To be clear, we still only have two main tuning parameters ($\theta$ and $\tilde{\lambda}$) as we fix any variable parameters in $f({\bm x})$ according to the choices in the corresponding papers. The unconstrained minimisation problem is then given as:
\begin{equation}\label{eqn:unconmin}
\min_{u}\bigg\{TV_{g}(u)+\tilde{\lambda}\int_{\Omega}r(\bm{x})u\ \mathrm{d}\Omega+\alpha\int_{\Omega}\nu(u)\ \mathrm{d}\Omega\bigg\}.
\end{equation}
We rescale the data term with $\mathcal{F}(\bm{x}) = r(\bm{x}) / ||r(\bm{x})||_{\infty}$. In effect this change is simply a rescaling of the parameters. This allows for the parameter choices between different models to be more consistent, as the fitting terms are similar in value. The problem (\ref{eqn:unconmin}) has the corresponding Euler-Lagrange equation (for fixed $c_{1}$):
\begin{equation}\label{eqn:segpde}
\nabla\cdot\left( g(|\nabla z|)\frac{\nabla u}{|\nabla u|_{\varepsilon_{1}}}\right) - \tilde{\lambda}\mathcal{F}(\bm{x}) -\alpha\nu'_{\varepsilon_{2}}(u)= 0.
\end{equation}
in $\Omega$ and $\frac{\partial u}{\partial \bm{n}} = 0$ where $\bm{n}$ is the outward unit normal. The constraint is enforced for $\alpha>\frac{\tilde{\lambda}}{2}||r(\bm{x})||$ by \cite{Chan:06}. Two parameters, $\varepsilon_{1}$ and $\varepsilon_{2}$, are introduced here. The former is to avoid singularities in the TV term and the latter is associated with the regularised penalty function $\nu_{\varepsilon_{2}}(u)$ from \cite{CDSS}:
\begin{equation}
\nu_{\varepsilon_{2}}(u) = H_{\varepsilon_{2}}\left(b_{\varepsilon_{2}}(u)\right)\left[b_{\varepsilon_{2}}(u)\right],
\end{equation}
with $b_{\varepsilon_{2}}(u)=\sqrt{(2u-1)^{2}+\varepsilon_{2}}-1$ and regularised Heaviside function
\begin{equation}
H_{\varepsilon_{2}}(u) = \frac{1}{2}\left( 1 + \frac{2}{\pi}\arctan\left( \frac{u}{\varepsilon_{2}} \right)  \right).
\end{equation}
The viscosity solution of the parabolic formulation of \eqref{eqn:segpde}, obtained by multiplying the PDE by $|\nabla u|$, exists and is unique. The general proof for a class of PDEs to which (\ref{eqn:segpde}) belongs, is included in \cite{Geo} and we refer the reader there for the details. Once the solution to \eqref{eqn:segpde} is found, denoted $u^{*}$, we define the computed foreground region as follows:
\begin{equation}
u_{\gamma}=\{x\in\Omega|\ u^{*}(x)>\gamma\}.
\end{equation}
We select $\gamma=0.5$ (although other values $\gamma\in(0,1)$ would yield a similar result according to Chan et al. \cite{Chan:06}). In the following we use the binary form of the solution, $u^{*}$, denoted $u_{\gamma}$. This partitions the domain into $\Omega_{F}$ and $\Omega_{B}$ according to the labelling function $u_{\gamma}$.

\subsection{Implementation for Related Models}

The discussion in this section so far has used the function $f(\bm{x})$ associated with the data fitting functional $f_{X}(u)$. This corresponding equations for the RSF, LCV, HYB and GAV models are detailed in \S\ref{sec:related},  CV is discussed in \S\ref{sec:intro}, and our approach is given by eqn. \eqref{eqn:newfitting}. We use this implementation to obtain selective segmentation versions of each of those models, given by \eqref{eqn:genmin}. When these terms contain parameter choices we follow the advice in the corresponding papers as far as possible, unless we have found that alternatives will improve results. In the next section we will give the results of these models and compare them to our proposed approach. 

{\bf Note.} We now discuss details behind tuning parameters for the GAV model. It is noted in \S\ref{sec:related} that the GAV model requires a parameter $\beta$ to adapt the $c_{1}$ and $c_{2}$ calculation. We find that it is actually better to consider $c_{1}$ and $c_{2}$ separately to achieve improved results, as sometimes we wish to tune the values to have a higher $c_{1}$ and lower $c_{2}$ (or vice-versa) simultaneously. Therefore we introduce parameters $\beta_{1}$ and $\beta_{2}$ to tune $c_{1}$ and $c_{2}$ as follows:
\begin{equation}
c_{1} = \frac{\int_{\Omega}z^{\beta_{1}}u}{\int_{\Omega}z^{\beta_{1}-1}u}\ \mathrm{d}\Omega,\qquad c_{2} = \frac{\int_{\Omega}z^{\beta_{2}}(1-u)}{\int_{\Omega}z^{\beta_{2}-1}(1-u)}\ \mathrm{d}\Omega,
\end{equation}
In all experiments, we tested the following combinations of $(\beta_{1},\beta_{2})$: $(1.5,0.5)$, $(2,0)$, $(3,-1)$, $(4,-2)$, $(0.5,1.5)$, $(0,2)$, $(-1,3)$ and $(-2,4)$. For each choice, we optimised the values of $\tilde{\lambda}$ and $\theta$ according to the procedure described in \S\ref{sec:r2}. This allowed us to select the optimal combination of $(\beta_{1},\beta_{2})$ for each image.

\section{Algorithm}
\label{sec:alg}

Here, we will discuss the algorithm that we use to minimise the selective segmentation model \eqref{eqn:genmin}. We utilise additive operator splitting techniques to solve the minimisation problem efficiently.

\subsection{An Additive Operator Splitting (AOS) Scheme}

\begin{figure*}
\centering
\floatbox[{\capbeside\thisfloatsetup{capbesideposition={left,top},capbesidewidth=1.5in}}]{figure}[\FBwidth]
{\captionsetup{position=top}\captionsetup[subfigure]{labelformat=empty,font=normalsize} 
\subfloat[Test Image 1]{\includegraphics[width=1.6in,height = 1.4in]{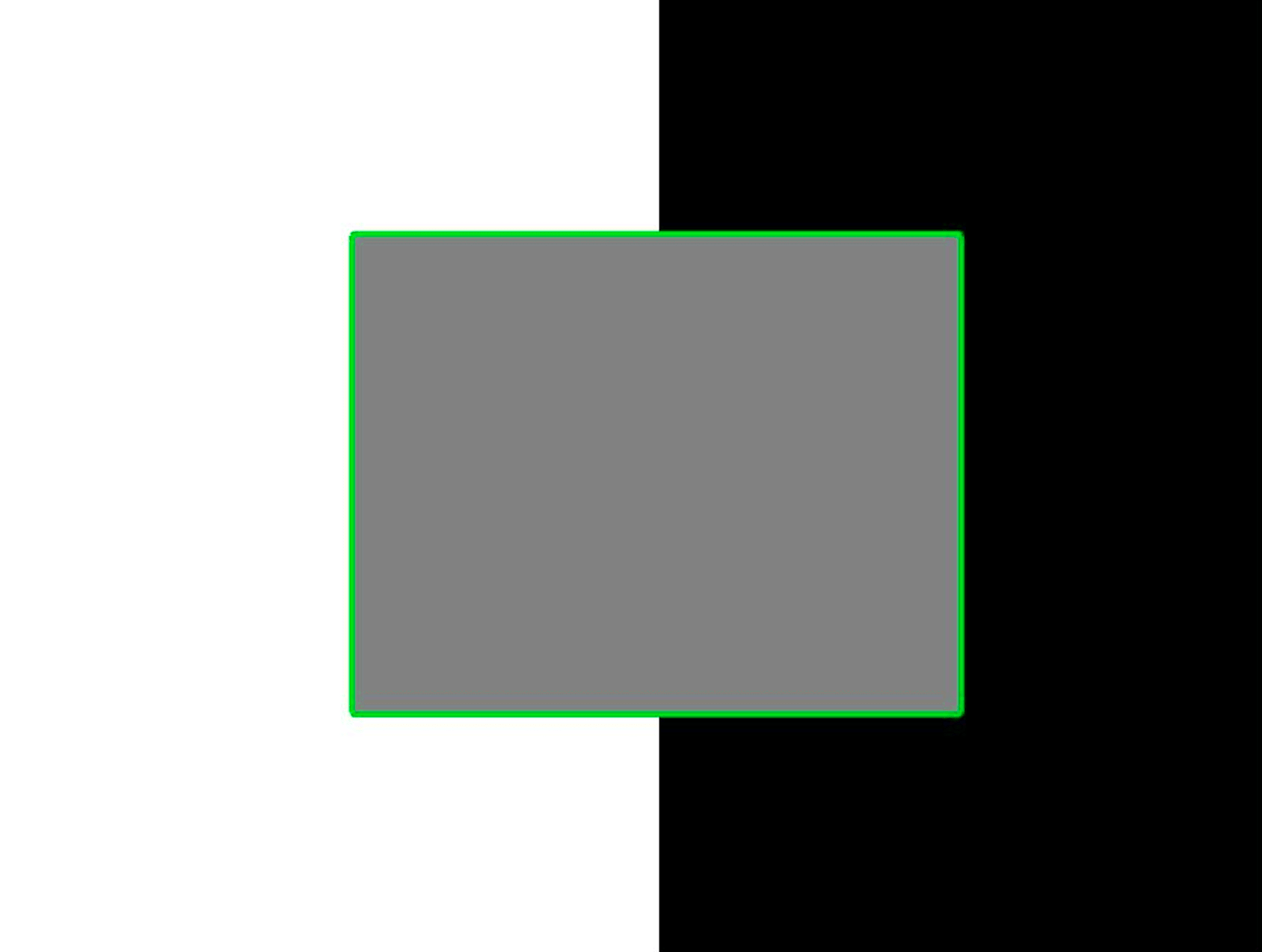}}\quad
\captionsetup{position=top}\captionsetup[subfigure]{labelformat=empty,font=normalsize} 
\subfloat[Test Image 2]{\includegraphics[width=1.6in,height = 1.4in]{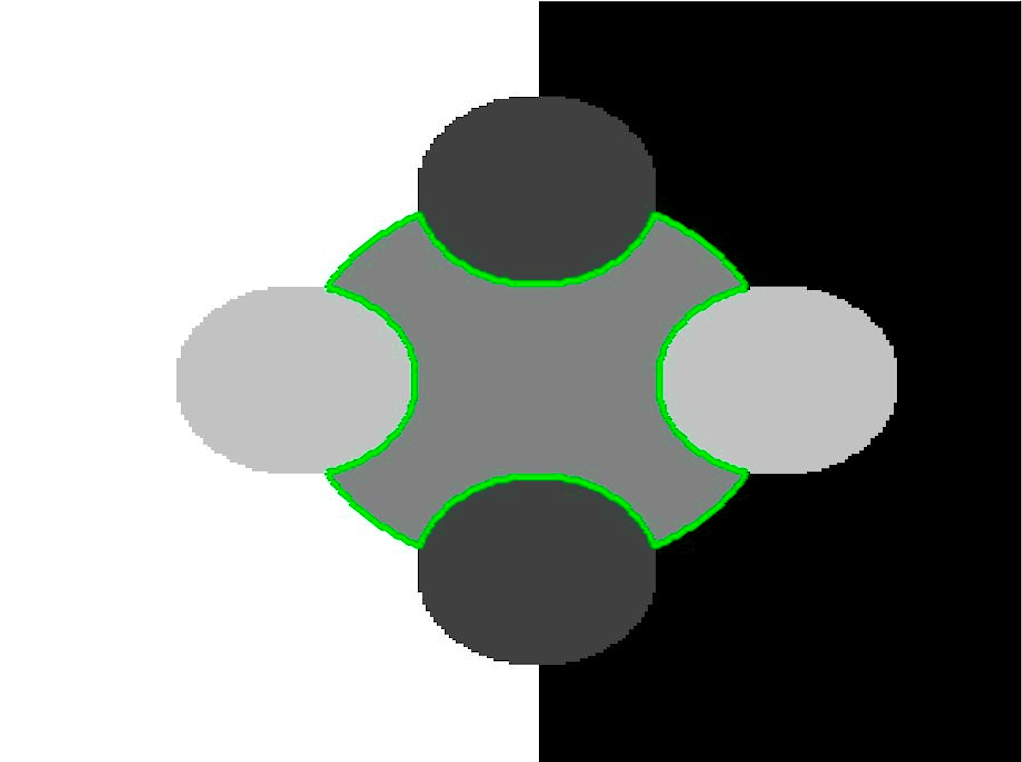}}\quad
\captionsetup{position=top}\captionsetup[subfigure]{labelformat=empty,font=normalsize} 
\subfloat[Test Image 3]{\includegraphics[width=1.6in,height = 1.4in]{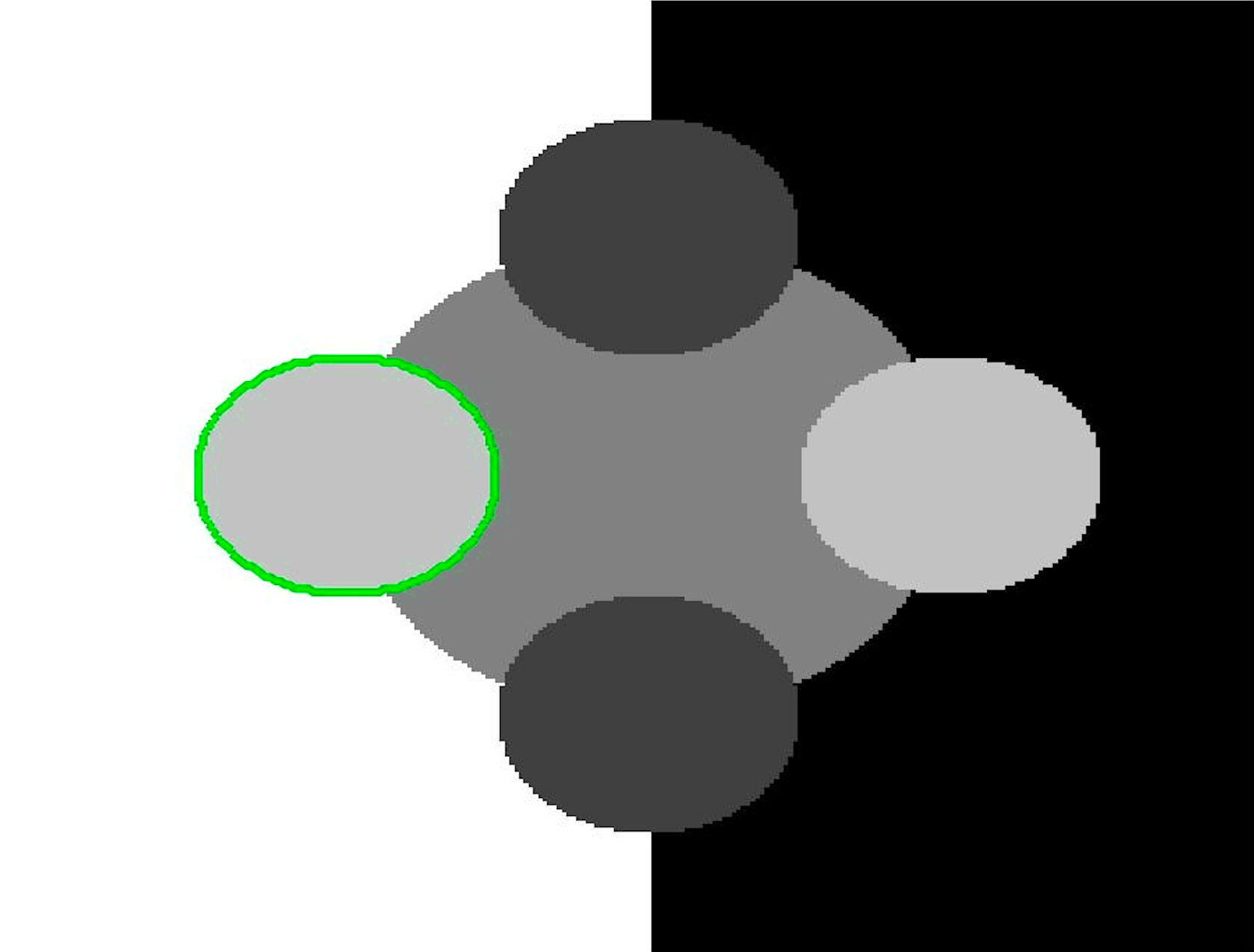}}} 
{\caption{Test Images 1--3; the ground truth contours are defined in the first row and the corresponding user input marker set is shown in the second row. These are synthetic images with homogeneous foregrounds selected to highlight the benefits of the proposed model. \label{fig:testimages1}}}\vspace{-0.3in}
\centering
\floatbox[{\capbeside\thisfloatsetup{capbesideposition={left,top},capbesidewidth=1.5in}}]{figure}[\FBwidth]
{\caption*{ }}
{\subfloat{\includegraphics[width=1.6in,height = 1.4in]{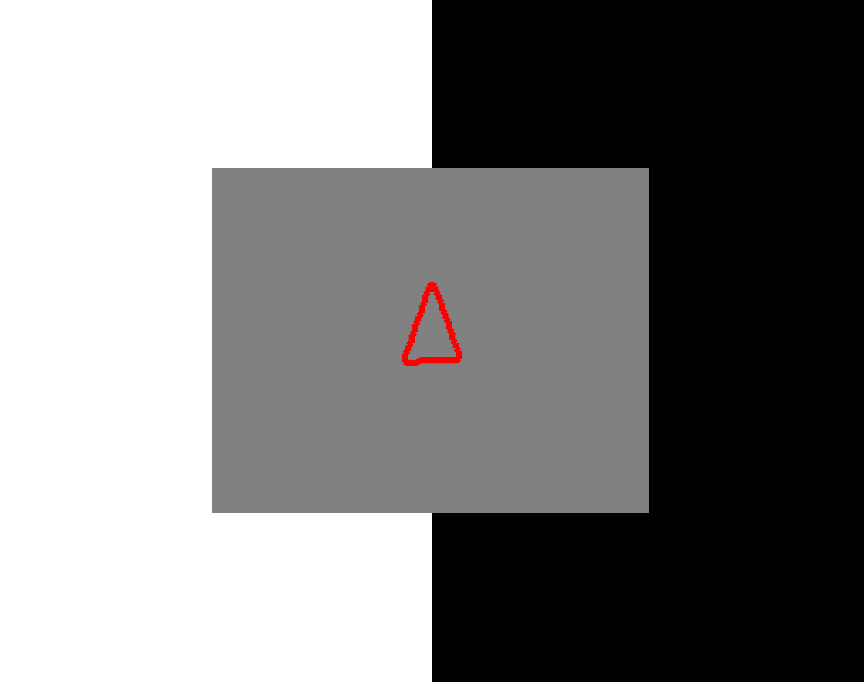}}\quad
\subfloat{\includegraphics[width=1.6in,height = 1.4in]{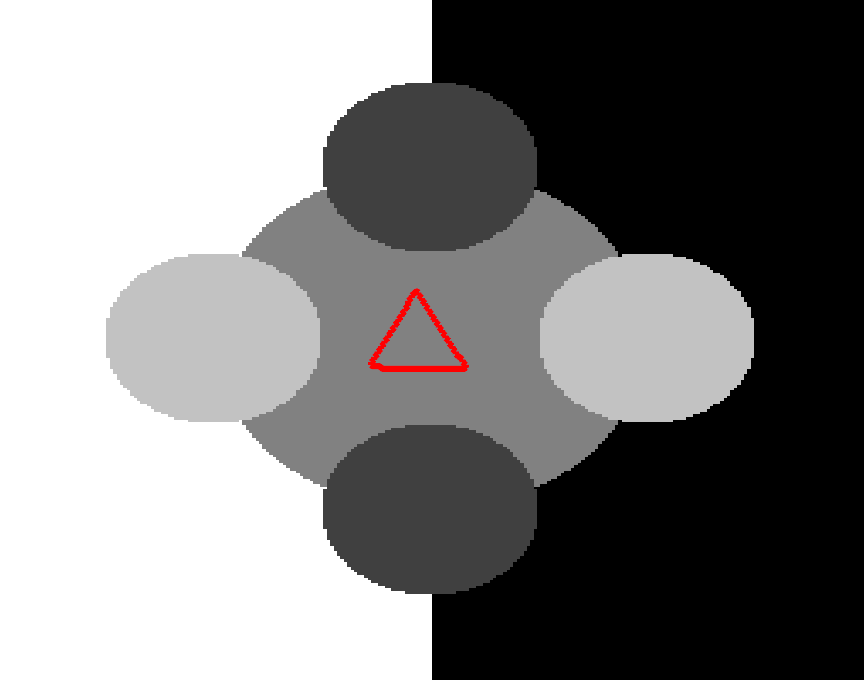}}\quad
\subfloat{\includegraphics[width=1.6in,height = 1.4in]{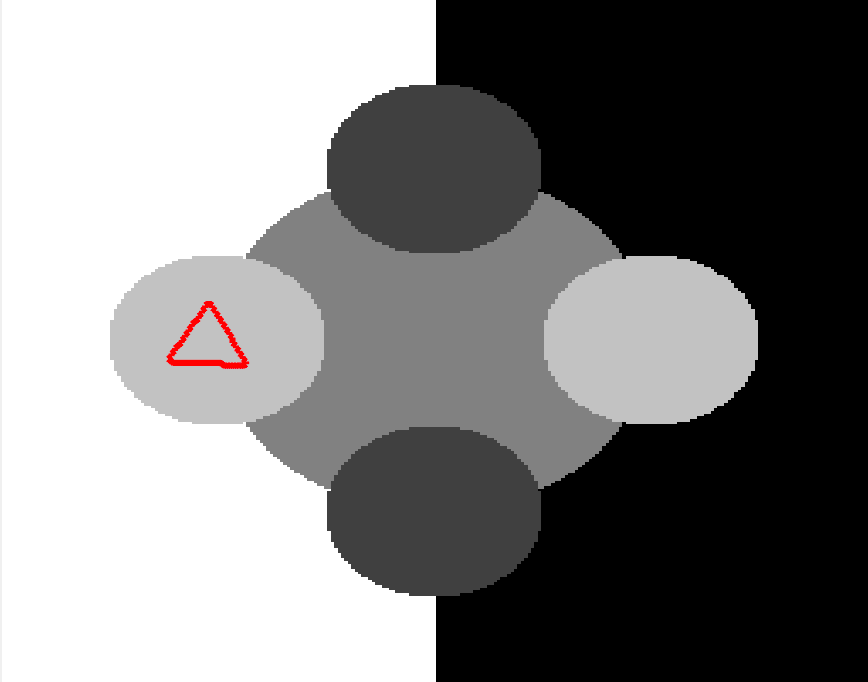}}}
\end{figure*}

Additive Operator Splitting (AOS) \cite{Gordeziani:74,Tai:91,Weickert:98} is a widely used method for solving PDEs with linear and non-linear diffusion terms
\cite{Rada:13,Geo,CDSS} such as
\begin{equation}\label{eqn:AOSeqn}
\frac{\partial u}{\partial t} = \mu\nabla\cdot (G(u)\nabla u) - f_{0}.
\end{equation}
AOS allows us to split the two-dimensional problem into two one-dimensional problems, which we solve separately and then combine. Each one-dimensional problem gives rise to a tridiagonal system of equations which can be solved efficiently by Thomas' algorithm, hence AOS is a very efficient method for solving PDEs of this type.
AOS is a semi-implicit method and permits far larger time-steps than the corresponding explicit schemes would. Hence AOS is more stable than an explicit method \cite{Weickert:98}. Note here that
\begin{equation}
G(u)=\frac{g(|\nabla z|)}{|\nabla u|_{\varepsilon_{1}}},\ \ \ \ \ f_{0}=\tilde{\lambda}\mathcal{F}(\bm{x})+\alpha\nu'_{\varepsilon_{2}}(u),
\end{equation}
and $\mu=1$. The standard AOS scheme assumes $f_{0}$ does not depend on $u$, however in this instance that is not the case. This requires a modification to be used for convex segmentation problems, first introduced by \cite{CDSS}. This non-standard formulation incorporates the regularised penalty term, $\nu_{\varepsilon_{2}}(u)$, into the AOS scheme which we briefly detail next.

The authors consider the Taylor expansions of $\nu'_{\varepsilon_{2}}(u)$ around $u=0$ and $u=1$. They find that the coefficient $b$ of the linear term in $u$ is the same for both expansions. Therefore, for a change in $u$ of $\delta u$ around $u=0$ and $u=1$ the change in $\nu'_{\varepsilon_{2}}(u)$ can be approximated by $b\cdot\delta(u)$. To address this, the relevant interval is defined as
\[
I_{\zeta} := [0-\zeta,0+\zeta]\cup [1-\zeta,1+\zeta]
\]
and a corresponding update function is given as
\[
\tilde{b}({\bm x}) =
\begin{cases}
b, & x\in\Omega,\ \ u({\bm x}) \in I_{\zeta}\\
0, & else.
\end{cases}
\]
The solution for \eqref{eqn:AOSeqn} is then obtained by discretising the equation as follows:
\begin{equation*}
\resizebox{\columnwidth}{!}{$
\begin{gathered}
\begin{aligned}
\frac{u^{(k+1)} -u^{(k)}}{\tau} = \mu\sum_{\ell = 1,2}A_{\ell}(u^{(k)}) u^{(k+1)} + \alpha\tilde{b}^{(k)}(u^{(k)}-u^{(k+1)})- f_{0}^{(k)}.
\end{aligned}
\end{gathered}
$}
\end{equation*}
where $A_{1}$ and $A_{2}$ are discrete forms of $\partial_{x}(G(u)\partial_{x})$ and \\ $\partial_{y}(G(u)\partial_{y})$, respectively (given in \cite{CDSS,Geo}). The modified AOS update is then given by
\begin{equation}\label{eqn:AOSgeo}
\begin{gathered}
\begin{aligned}
u^{(k+1)} =\frac{1}{2}\sum_{\ell=1}^{2}\bigg(I-2\tau\mu (I+\tilde{B}^{(k)})^{-1}A_{\ell}(u^{(k)})\bigg)^{-1}\tilde{u}^{(k)},
\end{aligned}
\end{gathered}
\end{equation}
where $\tilde{B}^{(k)} = \text{diag}(\tau\alpha\tilde{b}^{(k)})$ and $\tilde{u}^{(k)} = u^{(k)}+\tau (I+\tilde{B}^{(k)})^{-1}f_{0}^{(k)}$. This scheme allows for more control on the changes in $f_{0}$ between iterations due to the function $\tilde{b}$ and parameter $\zeta$, and therefore leads to a more stable convergence. We refer the reader to \cite{CDSS} for full details of the numerical method.

\subsection{The Proposed Algorithm}

In Algorithm~\ref{alg:1} we provide details of how we find the minimiser of the various selective segmentation models detailed above, defined by \eqref{eqn:genmin}. The algorithm is in a general form to be applied to any of the approaches discussed so far. It is important to reiterate that alternative solvers to AOS are available, such as the dual formulation \cite{Aujol:06,Bresson:07,Chambolle:04}, split-Bregman \cite{Goldstein:10}, augmented Lagrange \cite{Bertsekas:14}, primal dual \cite{ChambollePock}, and max-flow/min-cut \cite{Yuan:13}. In all experiments we use the tolerance of $10^{-4}$ for the stopping criteria and set $\varepsilon_{1} = 10^{-4}$, $\varepsilon_{2} = 10^{-1}$ and $\tau = 10^{-2}$.
\begin{algorithm}
\caption{Selective Segmentation Algorithm}
\label{alg:1}
\begin{algorithmic}
\STATE{Provide user input region $\mathcal{P}$ and compute $\mathcal{D}$, according to (\ref{eqn:DG0}).}
\STATE{Define $f(\bm{x})$ appropriately for the model (CV, RSF, LCV, HYB, GAV, or the proposed approach).}
\STATE{Compute $r(\bm{x})=\theta\mathcal{D}(\bm{x})+f(\bm{x}).$ and $\mathcal{F}(\bm{x}) = r(\bm{x}) / ||r(\bm{x})||_{\infty}$.}
\STATE{Initialise $u$ (arbitrary for $u\in[0,1]$).}
\WHILE{$\delta > tolerance $}
\STATE{$u_{old} := u$.}
\STATE{Update $u$ according to the AOS iteration \eqref{eqn:AOSgeo}.}
\STATE{$\delta = || u - u_{old} || / || u_{old} ||$.}
\ENDWHILE
\RETURN $u^{*}=u$ and binary labelling function, $u_{\gamma}$.
\end{algorithmic}
\end{algorithm}

\section{Results}
\label{sec:results}

\begin{figure*}
\centering
\floatbox[{\capbeside\thisfloatsetup{capbesideposition={left,top},capbesidewidth=1.5in}}]{figure}[\FBwidth]
{\captionsetup{position=top}\captionsetup[subfigure]{labelformat=empty,font=normalsize} 
\subfloat[Test Image 4]{\includegraphics[width=1.6in,height = 1.4in]{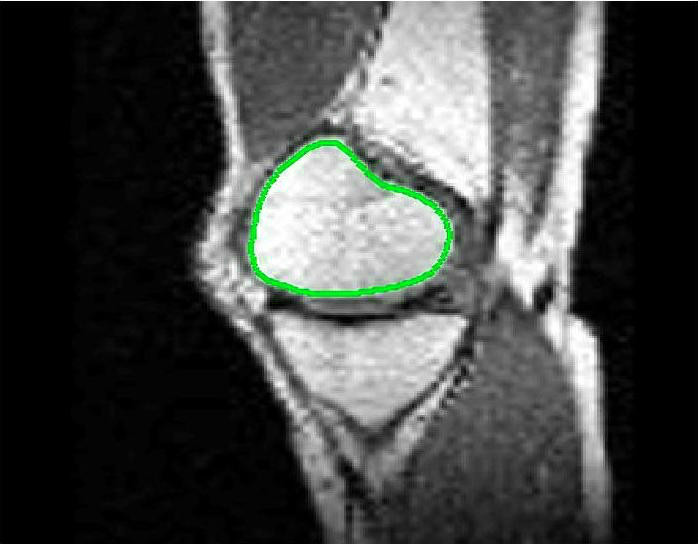}}\quad
\captionsetup{position=top}\captionsetup[subfigure]{labelformat=empty,font=normalsize} 
\subfloat[Test Image 5]{\includegraphics[width=1.6in,height = 1.4in]{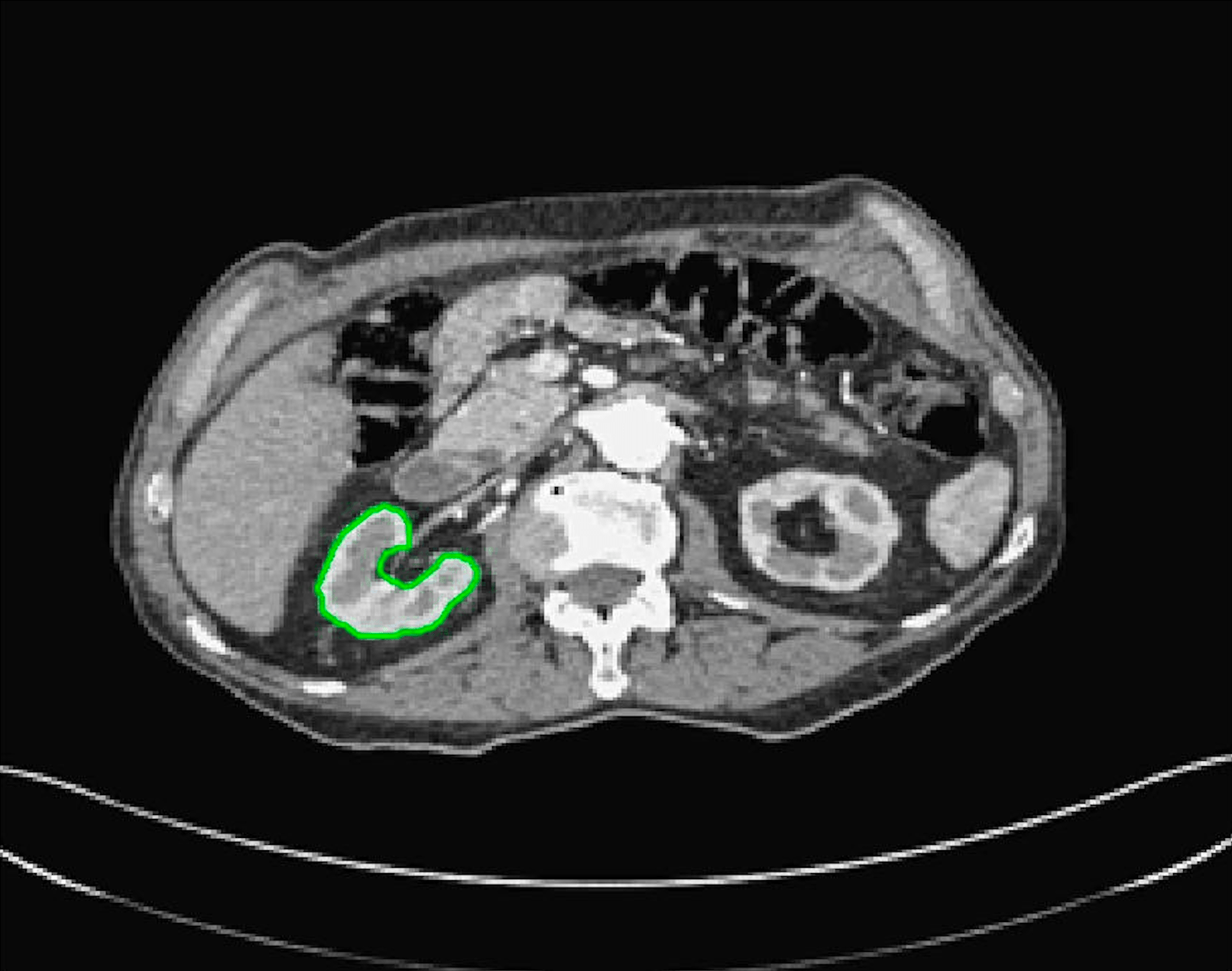}}\quad
\captionsetup{position=top}\captionsetup[subfigure]{labelformat=empty,font=normalsize} 
\subfloat[Test Image 6]{\includegraphics[width=1.6in,height = 1.4in]{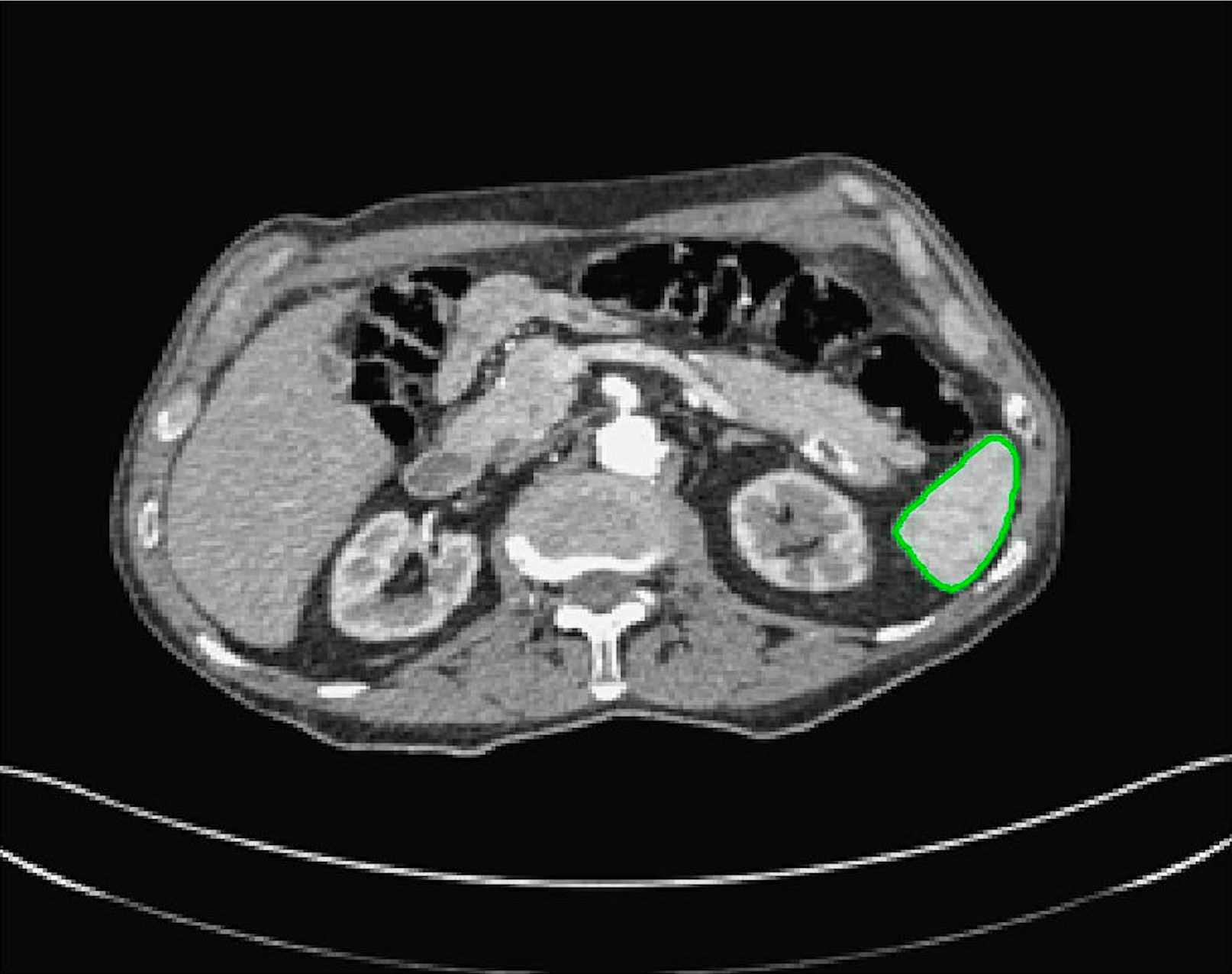}}}
{\caption{Test Images 4--6; the ground truth contours are defined in the first row and the corresponding user input marker set is shown in the second row. These are real images with some degree of intensity inhomogeneity in the foreground, with potential medical applications in mind. \label{fig:testimages2}}}\vspace{-0.3in}
\centering
\vspace{-0.175in}
\floatbox[{\capbeside\thisfloatsetup{capbesideposition={left,top},capbesidewidth=1.5in}}]{figure}[\FBwidth]
{\caption*{ }}
{\subfloat{\includegraphics[width=1.6in,height = 1.4in]{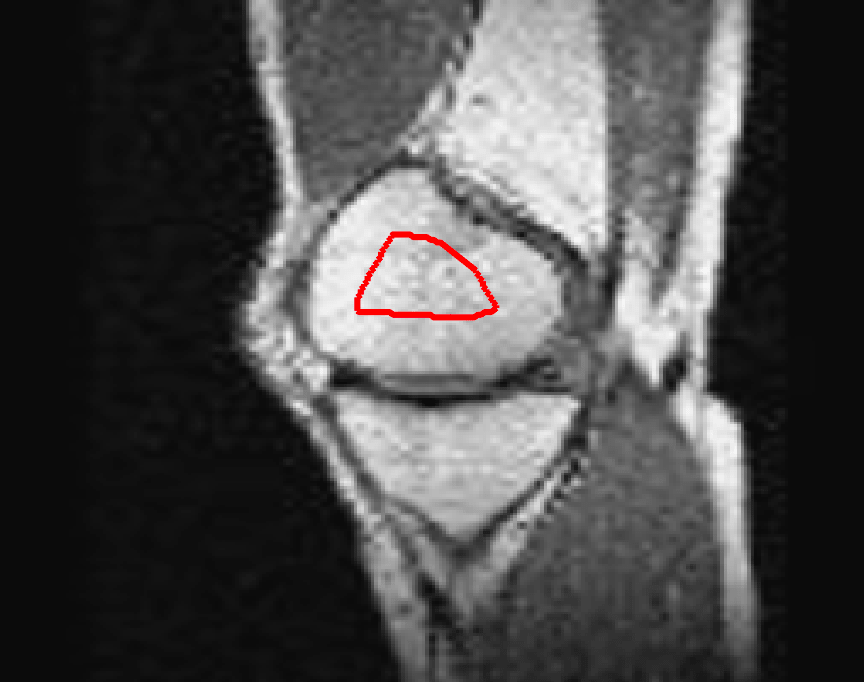}}\quad
\subfloat{\includegraphics[width=1.6in,height = 1.4in]{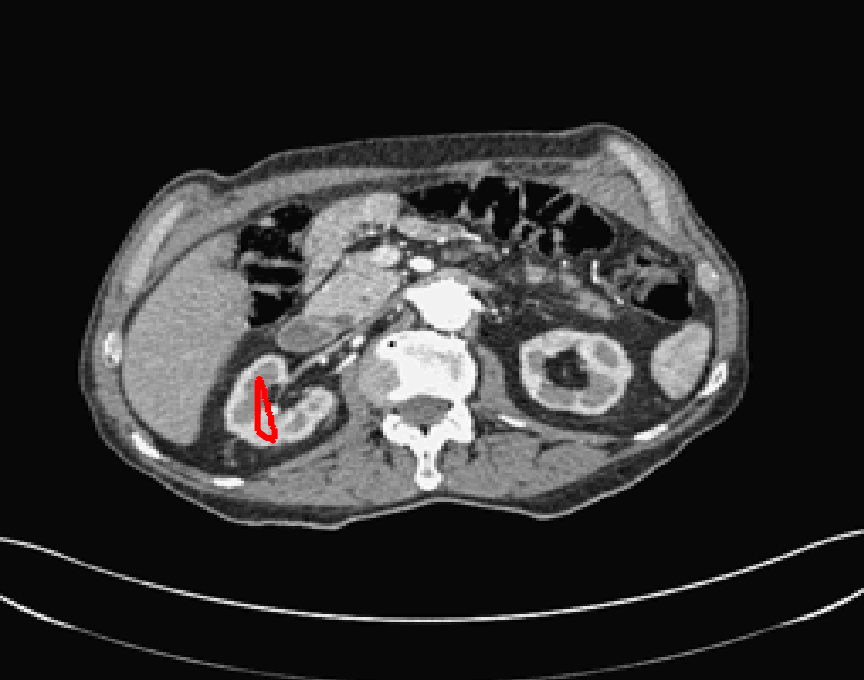}}\quad
\subfloat{\includegraphics[width=1.6in,height = 1.4in]{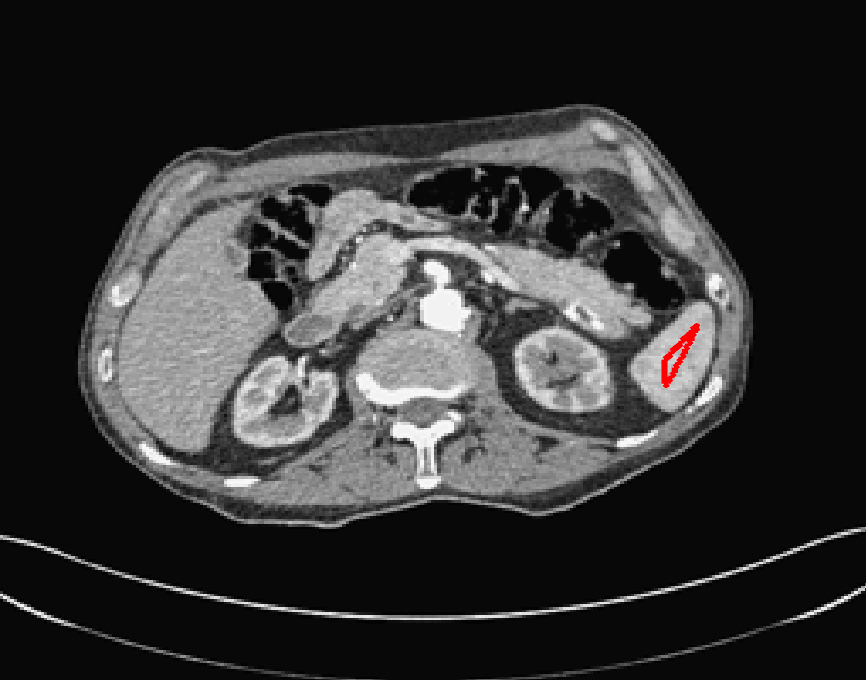}}}
\end{figure*}

In this section we will present results obtained using the proposed model and compare them to using fitting terms from similar models (CV \cite{ACWE}, RSF \cite{RSF}, LCV \cite{LCV}, HYB \cite{Ali:16}, GAV \cite{Ali:17}), detailed in \S\ref{sec:related}, and additional comparisons to alternative selective models. Specifically, we compare against the work of Nguyen et al. \cite{Nguyen:12} and Dong et al. \cite{SRW}, referred to as CAC and SRW respectively and detailed in \S\ref{sec:selective}. We intend to provide an overview of how effective each approach is in a number of key respects and analyse their potential for practical use in a reliable and consistent manner. Our focus is on how each fitting term can be applied to a consistent selective segmentation framework, and how robust the proposed model is overall. The key questions we consider are:
\begin{enumerate}
\item[(i)] How sensitive are the results to variations of the parameters $\tilde{\lambda}$ and $\theta$?
\item[(ii)] Is the model capable of achieving accurate results? 
\item[(iii)] To what extent is the proposed model dependent on the user input?
\item[(iv)] Does the model compare favourably against alternative selective methods?
\end{enumerate}

\begin{figure*}[t!]
\centering
\floatbox[{\capbeside\thisfloatsetup{capbesideposition={left,top},capbesidewidth=1.5in}}]{figure}[\FBwidth]
{\captionsetup{position=top}\captionsetup[subfigure]{labelformat=empty,font=normalsize} 
\subfloat[Test Image 7]{\includegraphics[width=1.6in,height = 1.4in]{figs/GT_Lung-min}}\quad
\captionsetup{position=top}\captionsetup[subfigure]{labelformat=empty,font=normalsize} 
\subfloat[Test Image 8]{\includegraphics[width=1.6in,height = 1.4in]{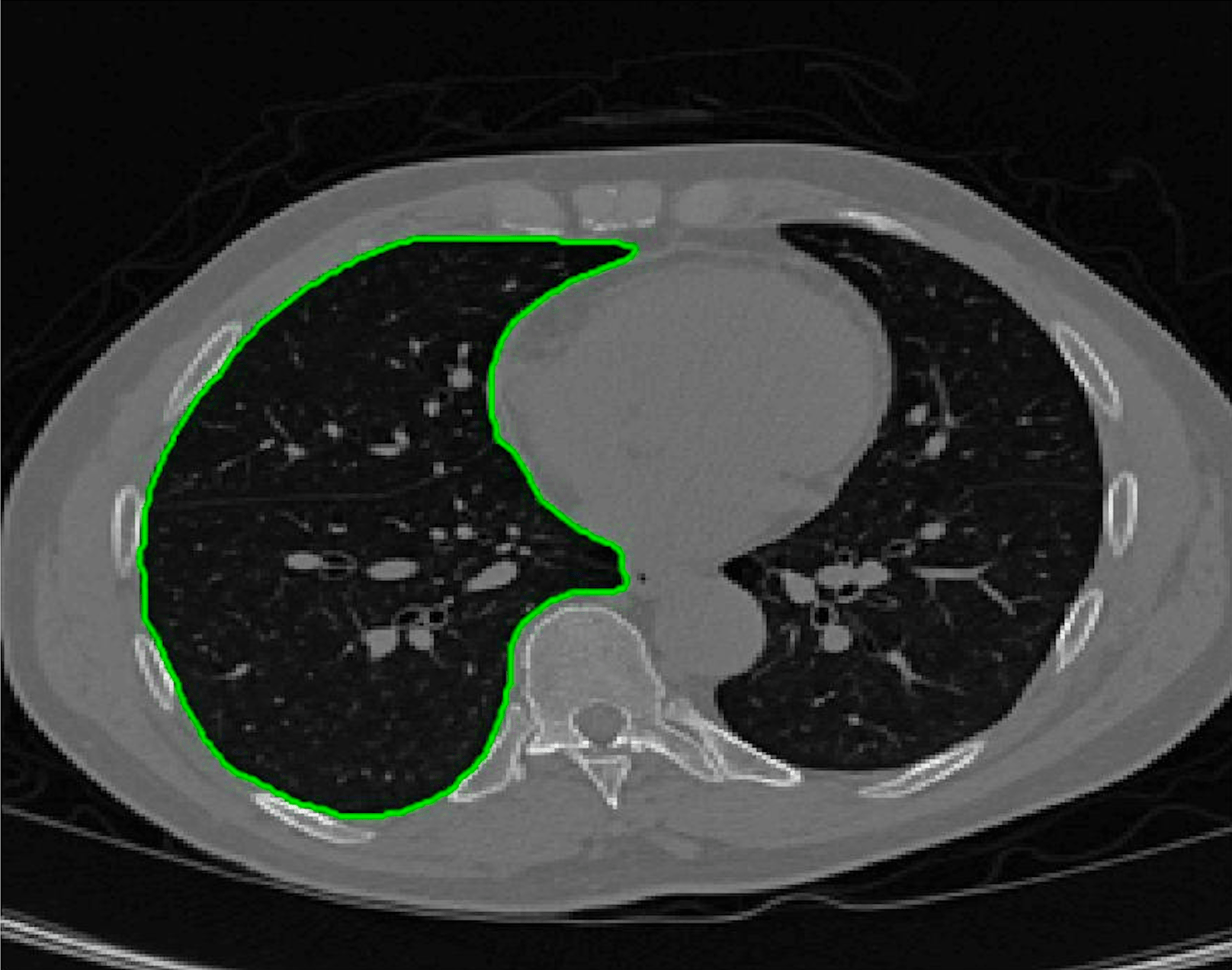}}\quad
\captionsetup{position=top}\captionsetup[subfigure]{labelformat=empty,font=normalsize} 
\subfloat[Test Image 9]{\includegraphics[width=1.6in,height = 1.4in]{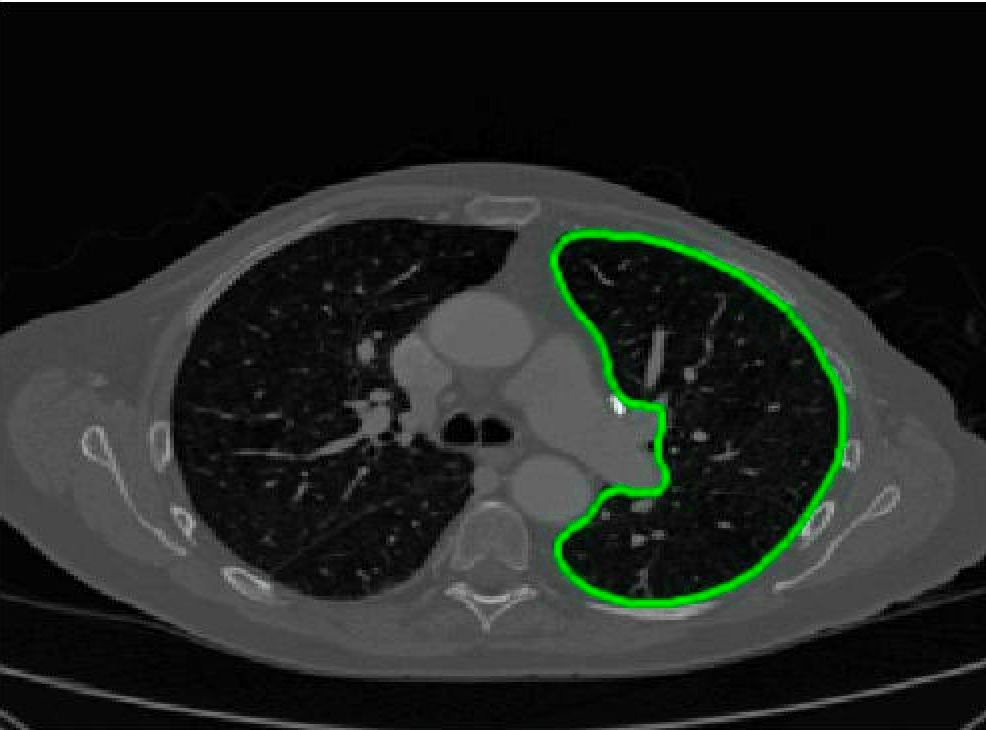}}}
{\caption{Test Images 7--9; the ground truth contours are defined in the first row and the corresponding user input marker set is shown in the second row. These are real images with approximately homogeneous foregrounds. The challenge is that the background contains substantial regions of a similar intensity. \label{fig:testimages3}}}\vspace{-0.45in}
\centering
\floatbox[{\capbeside\thisfloatsetup{capbesideposition={left,top},capbesidewidth=1.5in}}]{figure}[\FBwidth]
{\caption*{ }}
{\subfloat{\includegraphics[width=1.6in,height = 1.4in]{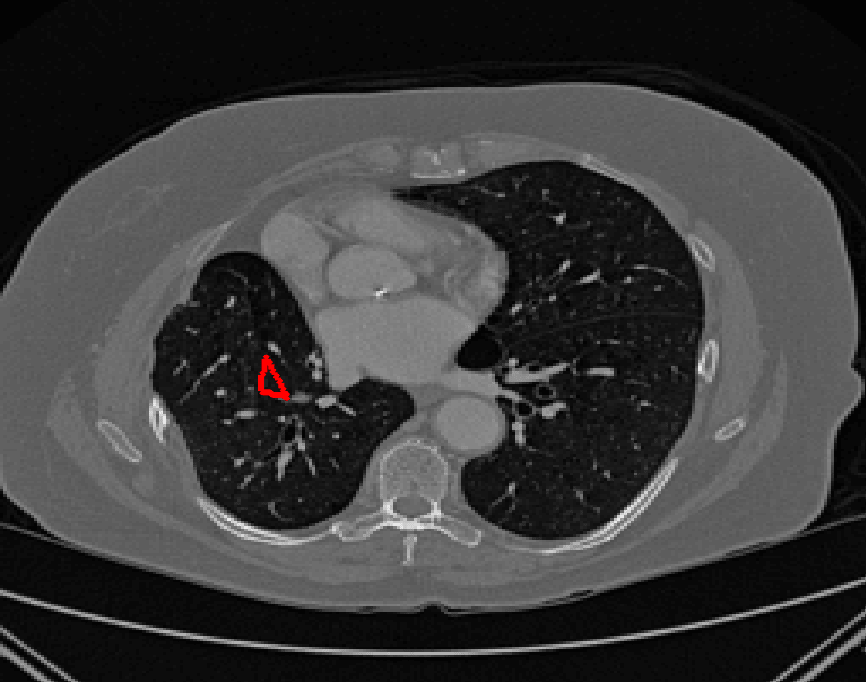}}\quad
\subfloat{\includegraphics[width=1.6in,height = 1.4in]{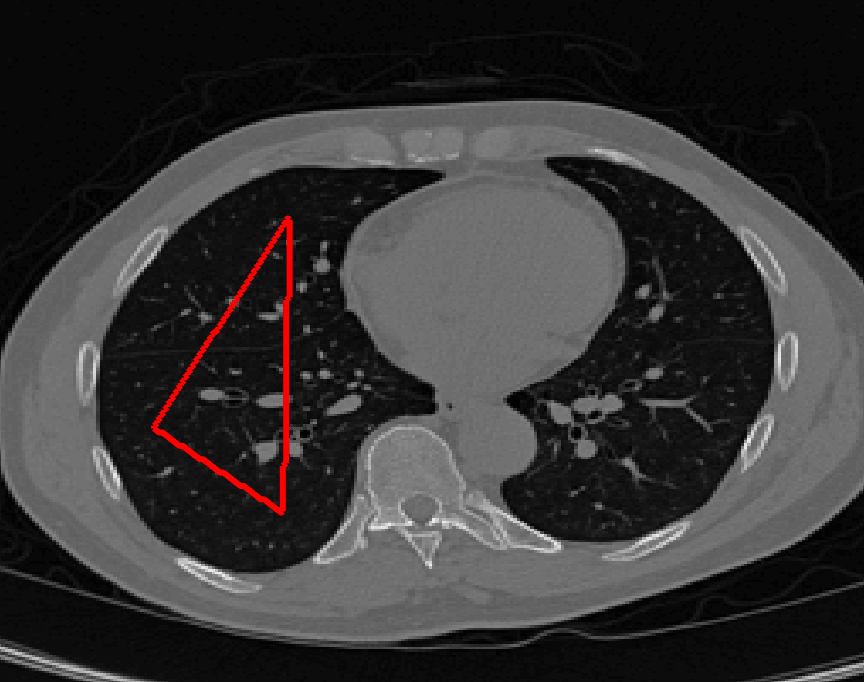}}\quad
\subfloat{\includegraphics[width=1.6in,height = 1.4in]{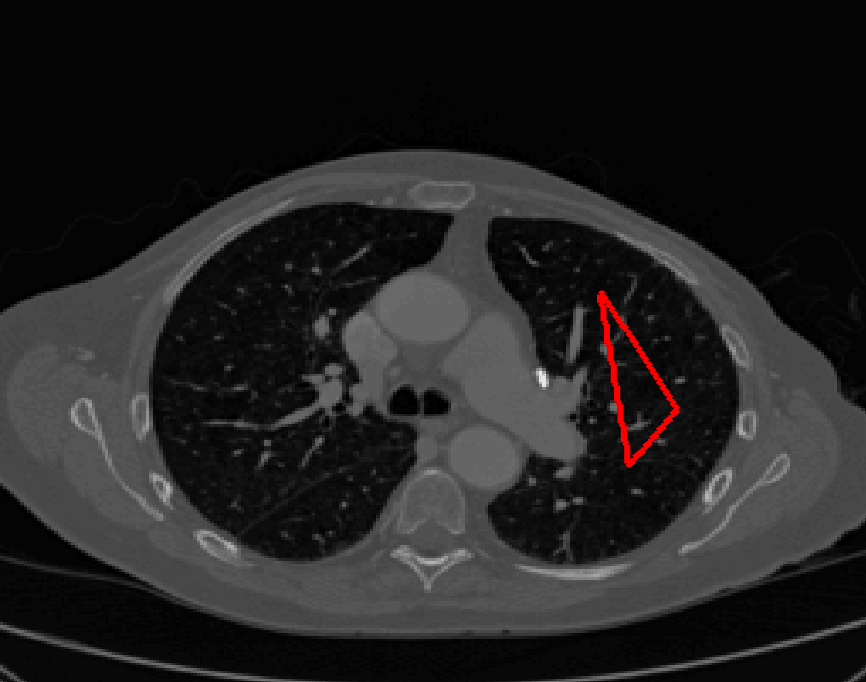}}}
\end{figure*}

{\bf Test Images.} We will perform initial tests on the images shown in Figs.~\ref{fig:testimages1}--\ref{fig:testimages3}. We have provided the ground truth and initialisation used for each image. Test Images 1--3 are synthetic, Test Image 4 is an MRI scan of a knee, Test Images 5--6 are abdominal CT scans, and Test Images 7--9 are lung CT scans. They have been selected to present challenges relevant to the discussion in \S\ref{sec:related}. We focus on medical images as this is the application of most interest to our work. In the following we will discuss the results in terms of synthetic images (1--3) and real images (4--9). We also test the proposed approach on a larger data set of 30 CT images (a sample of which is presented in Fig. \ref{fig:inputimages}), comparing against existing selective methods detailed in \S\ref{sec:selective}.

{\bf Measuring Segmentation Accuracy.} In our tests we use the Jaccard Coefficient \cite{Jaccard:12}, often referred to as the Tanimoto Coefficient (TC), to measure the quality of the segmentation. We define accuracy with respect to a ground truth, $GT$, given by a manual segmentation:
\begin{equation*}
GT=\{x\in\Omega|\ x\in \text{foreground}\}.
\end{equation*}
The Tanimoto Coefficient is then calculated as
\[
\text{TC} = \frac{|N(u_{\gamma} \cap GT)|}{|N(u_{\gamma} \cup GT)|},
\]
where $N(\cdot)$ refers to the number of points in the enclosed region. This takes values in the range $[0,1]$, with higher TC values indicating a more accurate segmentation. In the following we will represent accuracy visually from red ($\text{TC}=0$) to green ($\text{TC}=1$), with the intermediate scaling of colours used shown in Fig.~\ref{fig:colorbar}. This will be particularly relevant in \S\ref{sec:r1}. 

{\bf Note.} In \S\ref{sec:GAV} we mentioned the tuning of parameters in the GAV model. To be explicit the optimal $(\beta_{1},\beta_{2})$ pairs used in the following tests were (4,-2) for Test Images 1 and 2, (1.5,0.5) for Test Images 3,4, and 6, (2,0) for Test Image 5, and (-2,4) for Test Images 7,8, and 9. Results vary significantly as $(\beta_{1},\beta_{2})$ are varied, but we found these to be the best choices for each image.

The discussion of results is split into four sections, addressing the questions introduced above. First, in \S\ref{sec:r2}, we will examine the robustness to the parameters $\tilde{\lambda}$ and $\theta$ for each model. Then, in \S\ref{sec:r1}, we will compare the optimal accuracy achieved by each method to determine what they are capable of in the context of selective segmentation for these examples. In \S\ref{sec:r3}, we will test the proposed model with respect to the user input. By randomising the input we will determine to what extent the proposed model is suitable for use in practice. Finally, in \S\ref{sec:r4} we will compare the proposed approach to the methods introduced in \S\ref{sec:selective} on an additional CT data set. This will help further establish how the algorithm performs against competitive approaches in the literature.

\subsection{Parameter Robustness}
\label{sec:r2}

\begin{figure}[h]
\includegraphics[width=3.0in,height = 0.4in]{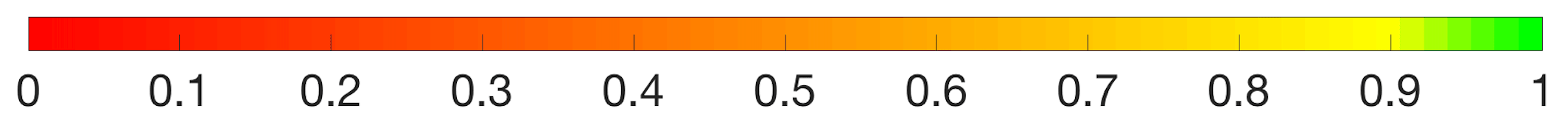}
\caption{Colour scaling corresponding to TC values, representing the accuracy of the result. This scale is used in subsequent figures. \label{fig:colorbar}}
\end{figure}

\begin{figure}[h]
\includegraphics[width=3.4in,height = 2.42in]{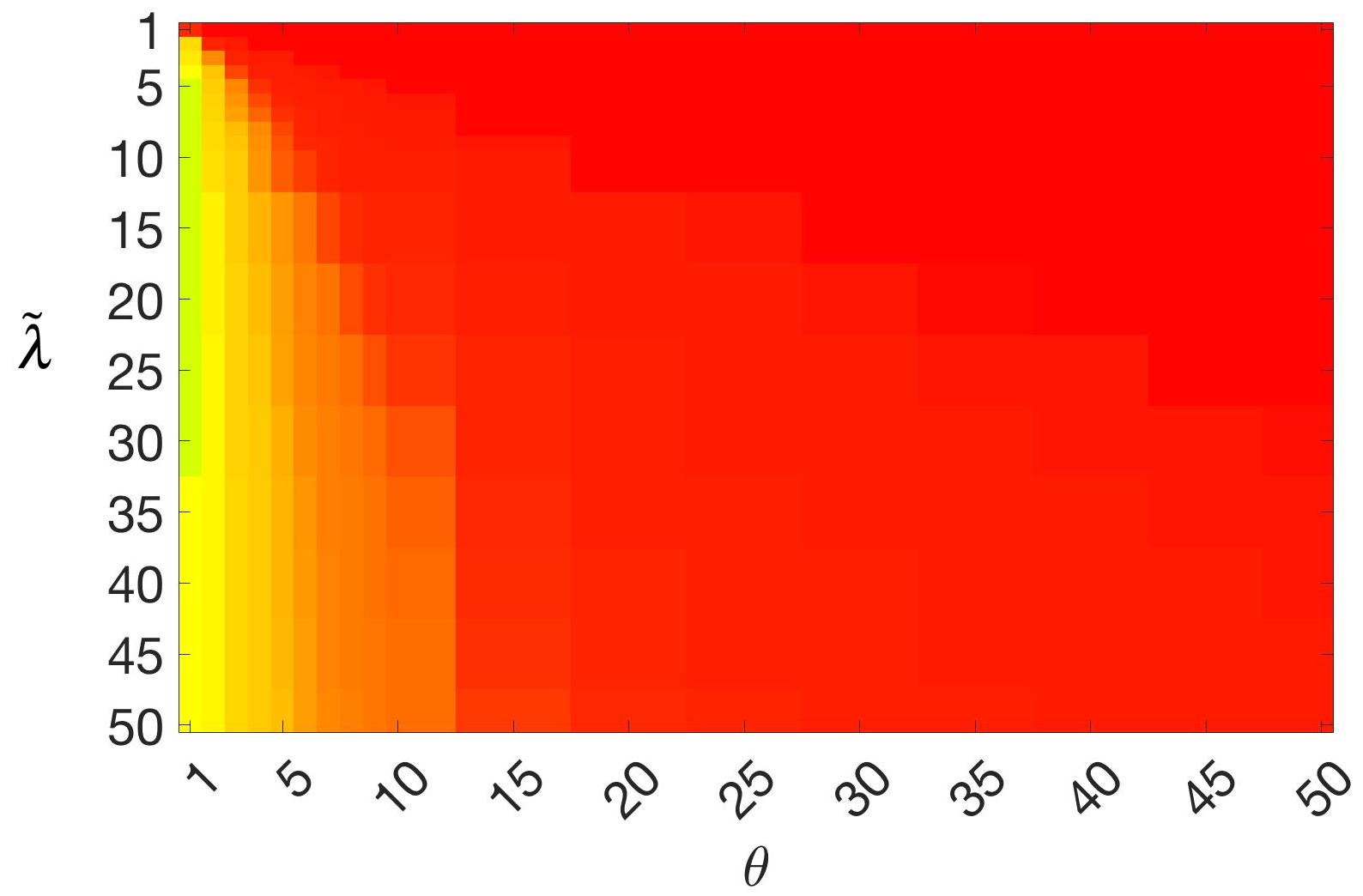}
\caption{Example heatmap of TC values to display segmentation accuracy for parameters $(\tilde{\lambda},\theta)$. \label{fig:exampleheat}}
\end{figure}

\begin{figure*}
\centering
\floatbox{figure}[\FBwidth]
{\captionsetup[subfigure]{labelformat=empty,font=normalsize}
\subfloat[{(i) $\tilde{\lambda}$ = 1, TC = 0.00}]{\includegraphics[width=1.5in,height = 1.5in]{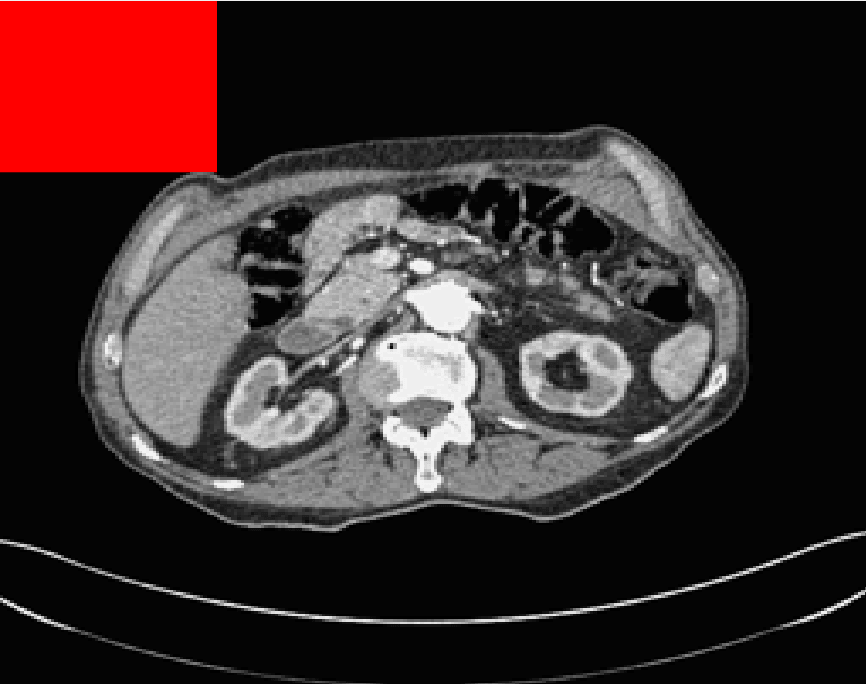}}
\subfloat[{(ii) $\tilde{\lambda}$ = 2, TC = 0.79}]{\includegraphics[width=1.5in,height = 1.5in]{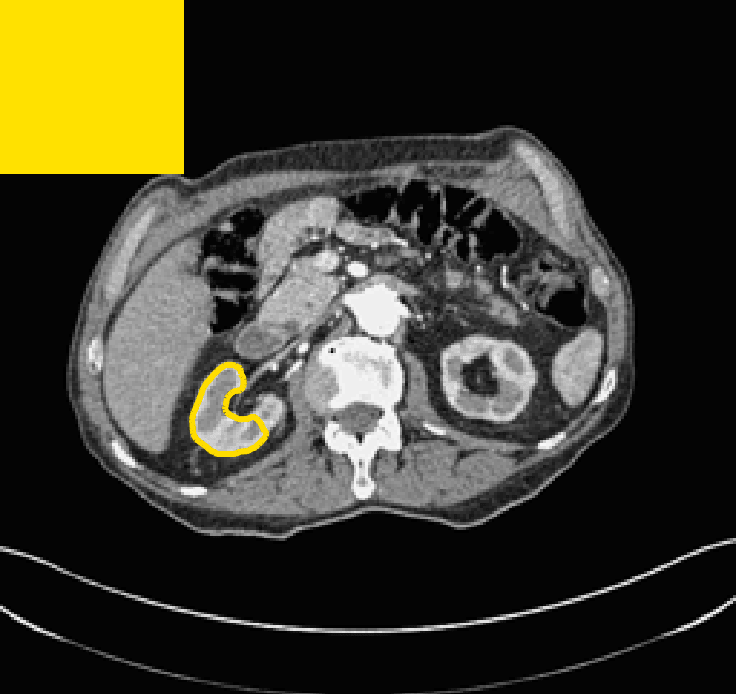}}
\subfloat[{(iii) $\tilde{\lambda}$ = 3, TC = 0.91}]{\includegraphics[width=1.5in,height = 1.5in]{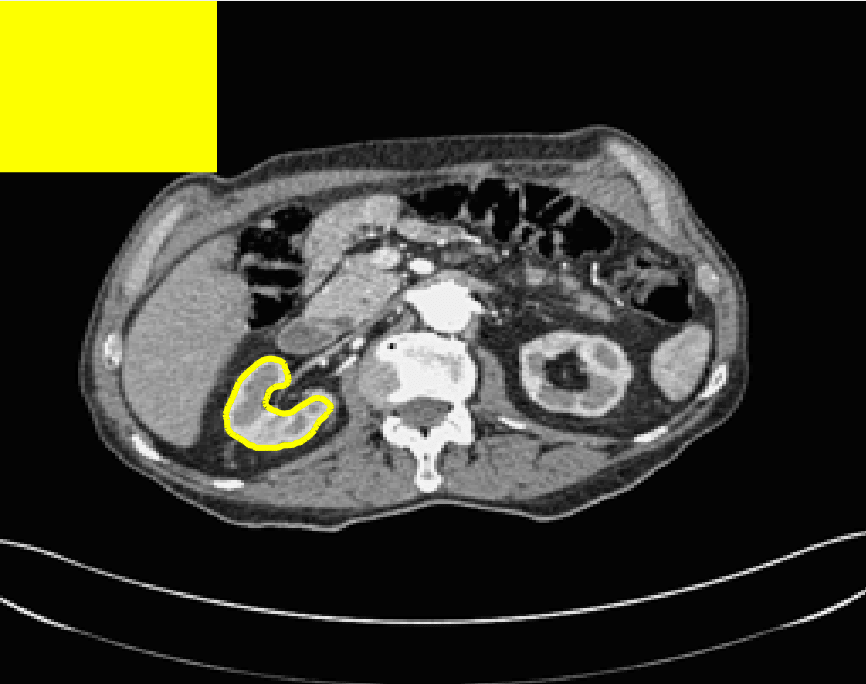}}
\subfloat[{(iv) $\tilde{\lambda}$ = 4, TC = 0.95}]{\includegraphics[width=1.5in,height = 1.5in]{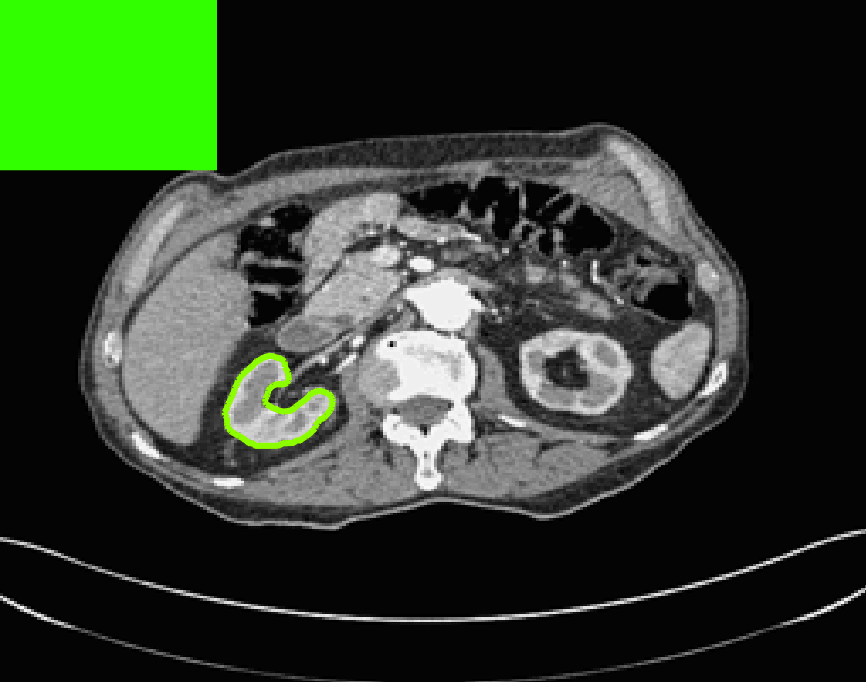}}}
\centering
\floatbox{figure}[\FBwidth]
{\captionsetup[subfigure]{labelformat=empty,font=normalsize}
\subfloat[{(v) $\tilde{\lambda}$ = 5, TC = 0.95}]{\includegraphics[width=1.5in,height = 1.5in]{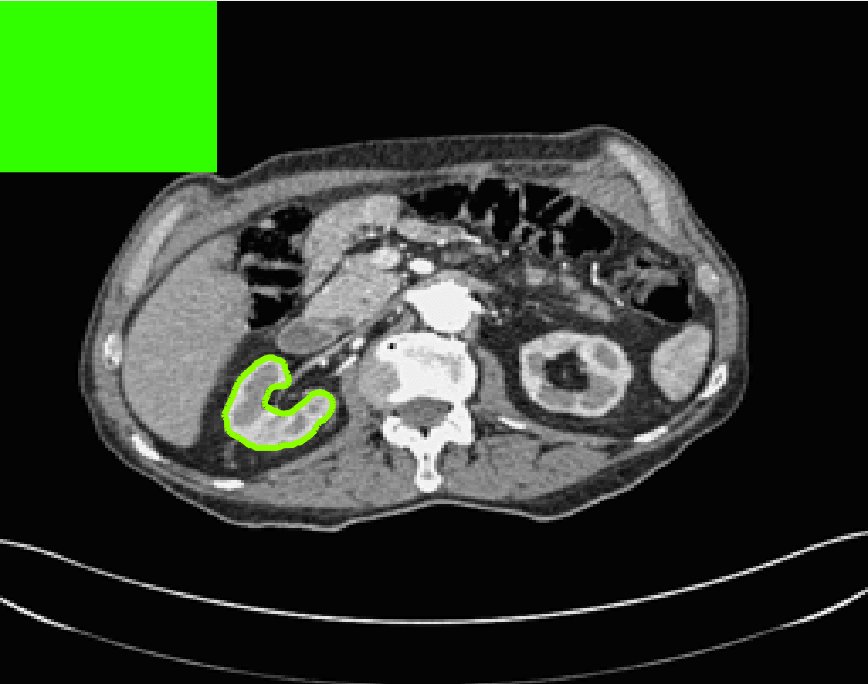}}
\subfloat[{(vi) $\tilde{\lambda}$ = 6, TC = 0.95}]{\includegraphics[width=1.5in,height = 1.5in]{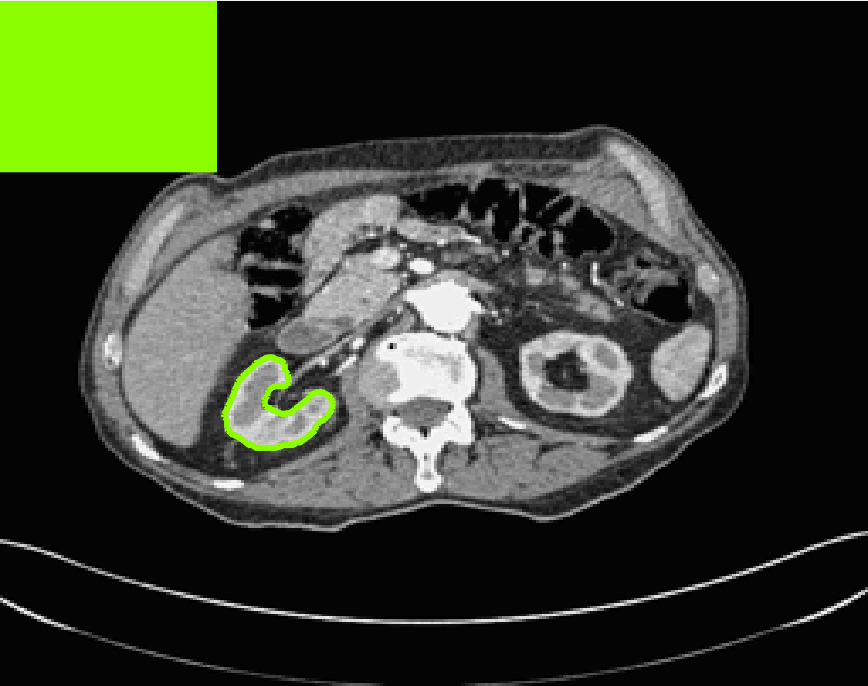}}
\subfloat[{(vii) $\tilde{\lambda}$ = 7, TC = 0.94}]{\includegraphics[width=1.5in,height = 1.5in]{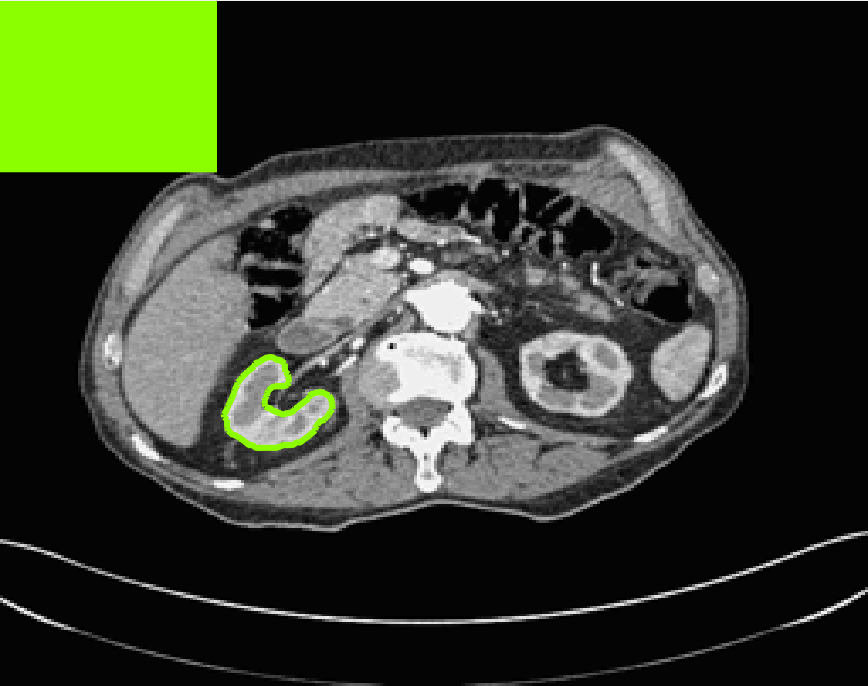}}
\subfloat[{(viii) $\tilde{\lambda}$ = 8, TC = 0.94}]{\includegraphics[width=1.5in,height = 1.5in]{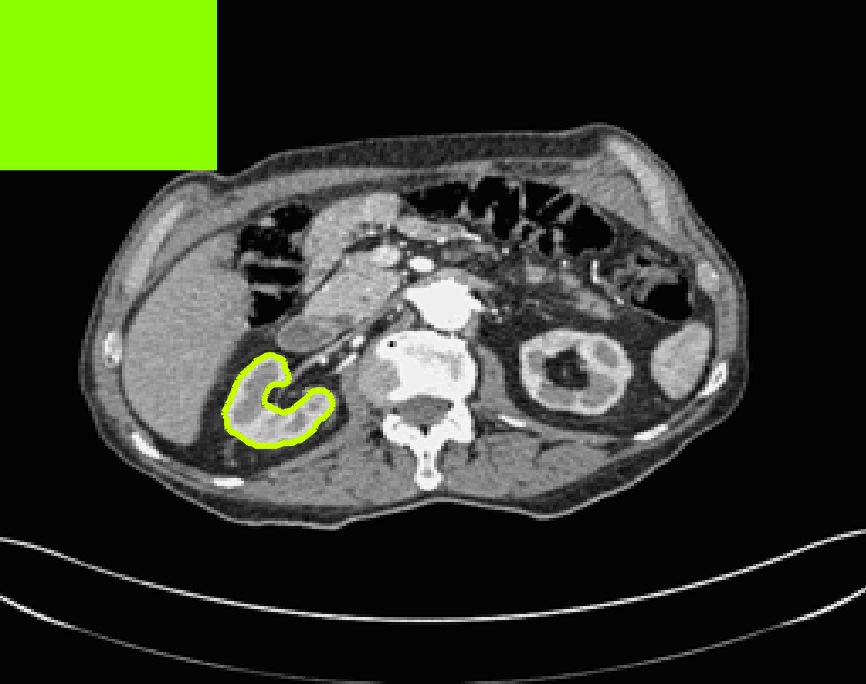}}}
\centering
\floatbox{figure}[\FBwidth]
{\captionsetup[subfigure]{labelformat=empty,font=normalsize}
\subfloat[{(ix) $\tilde{\lambda}$ = 9, TC = 0.93}]{\includegraphics[width=1.5in,height = 1.5in]{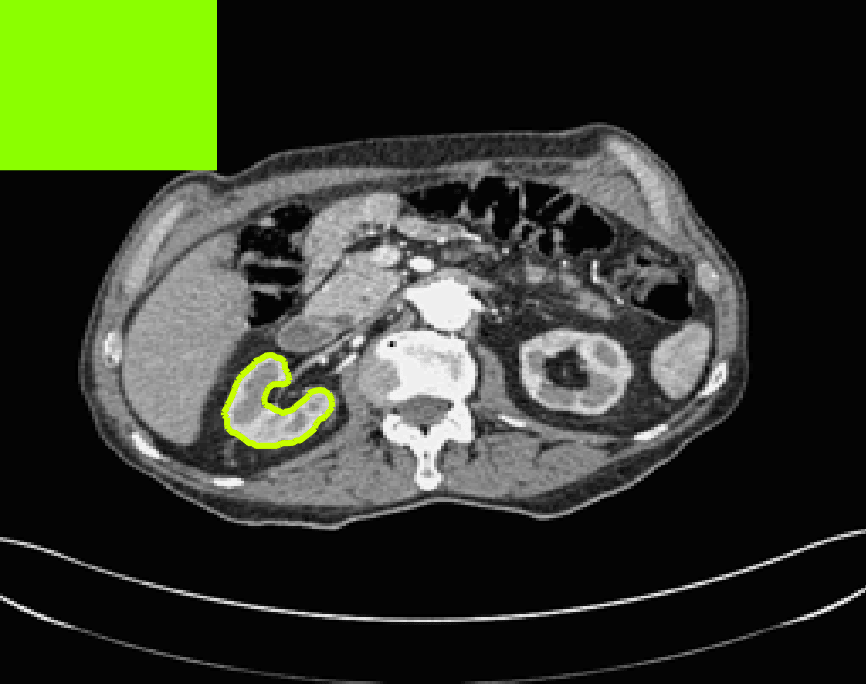}}
\subfloat[{(x) $\tilde{\lambda}$ = 10, TC = 0.93}]{\includegraphics[width=1.5in,height = 1.5in]{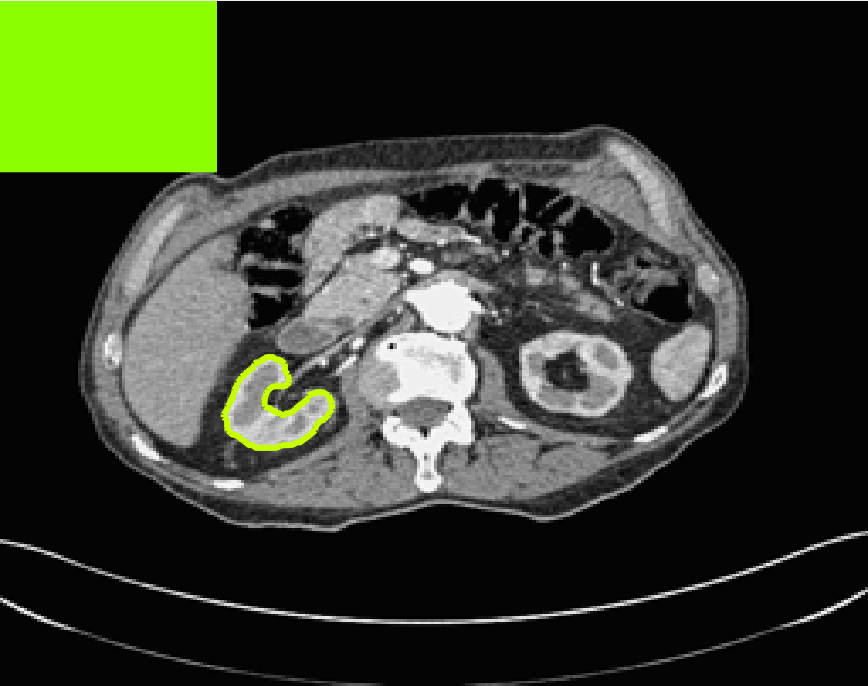}}
\subfloat[{(xi) $\tilde{\lambda}$ = 15, TC = 0.93}]{\includegraphics[width=1.5in,height = 1.5in]{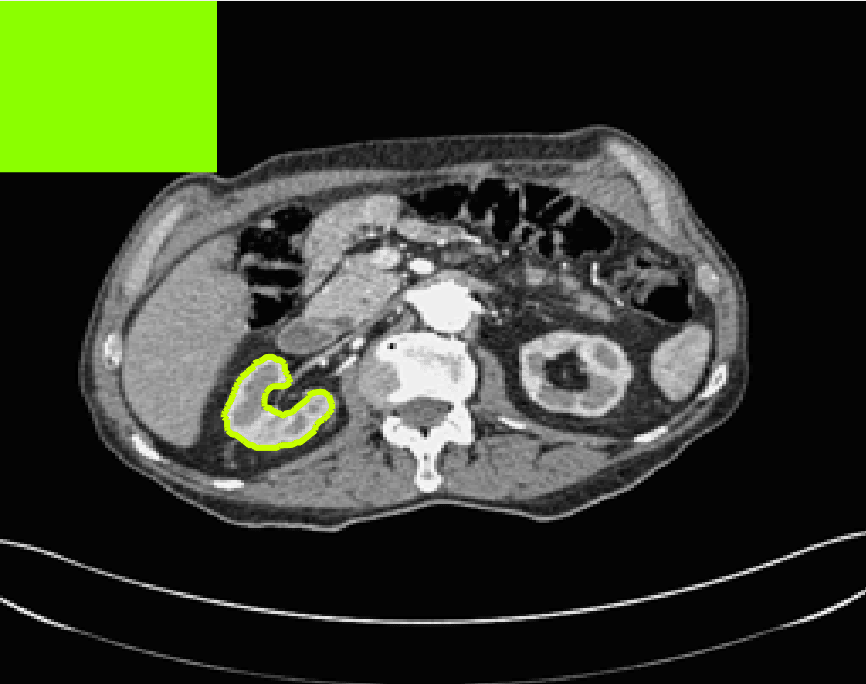}}
\subfloat[{(xii) $\tilde{\lambda}$ = 20, TC = 0.85}]{\includegraphics[width=1.5in,height = 1.5in]{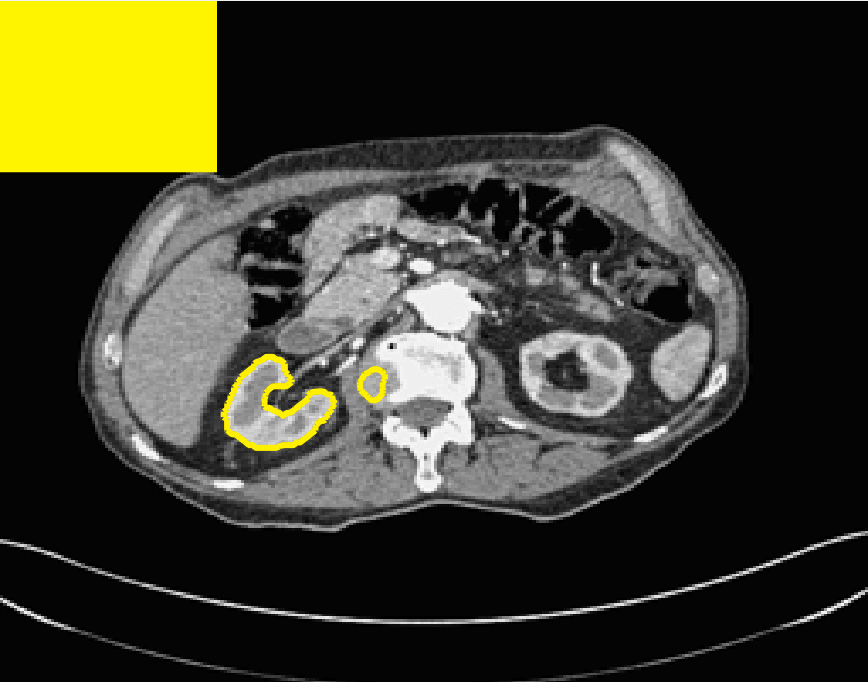}}}
{\caption{Segmentation results and TC values for the proposed model whilst varying $\tilde{\lambda}$ (with $\theta = 4$). The colours correspond to the TC value (green is TC = 1, red is TC = 0), consistent with the scale in Fig. \ref{fig:colorbar}. This is for Test Image 5, with the corresponding heatmap provided in Fig.~\ref{fig:heat2}.\label{fig:kidneyres}}}
\end{figure*}

\begin{figure*}
\centering
\floatbox[{\capbeside\thisfloatsetup{capbesideposition={left,top},capbesidewidth=1.5in}}]{figure}[\FBwidth]
{\captionsetup{position=top}\captionsetup[subfigure]{labelformat=empty,font=normalsize} 
\subfloat[Test Image 1]{\includegraphics[width=1.6in,height = 1.4in]{figs/GT_Squares-min}}\quad
\captionsetup{position=top}\captionsetup[subfigure]{labelformat=empty,font=normalsize} 
\subfloat[Test Image 2]{\includegraphics[width=1.6in,height = 1.4in]{figs/GT_Circles1-min}}\quad
\captionsetup{position=top}\captionsetup[subfigure]{labelformat=empty,font=normalsize} 
\subfloat[Test Image 3]{\includegraphics[width=1.6in,height = 1.4in]{figs/GT_Circles2-min}}} 
{\caption{Heatmaps of TC values for permutations of $\tilde{\lambda}$ and $\theta$. Each row and column is labelled according to the model used and the image tested. The colour is consistent with the scale in Fig.~\ref{fig:colorbar}. Here, we present Test Images 1 -- 3.\label{fig:heat1}}}\vspace{-0.1in}
\centering
\floatbox[{\capbeside\thisfloatsetup{capbesideposition={left,center},capbesidewidth=1.5in,font =normalsize}}]{figure}[\FBwidth]
{\caption*{CV \cite{ACWE}}}
{\subfloat{\includegraphics[width=1.6in,height=1.1in]{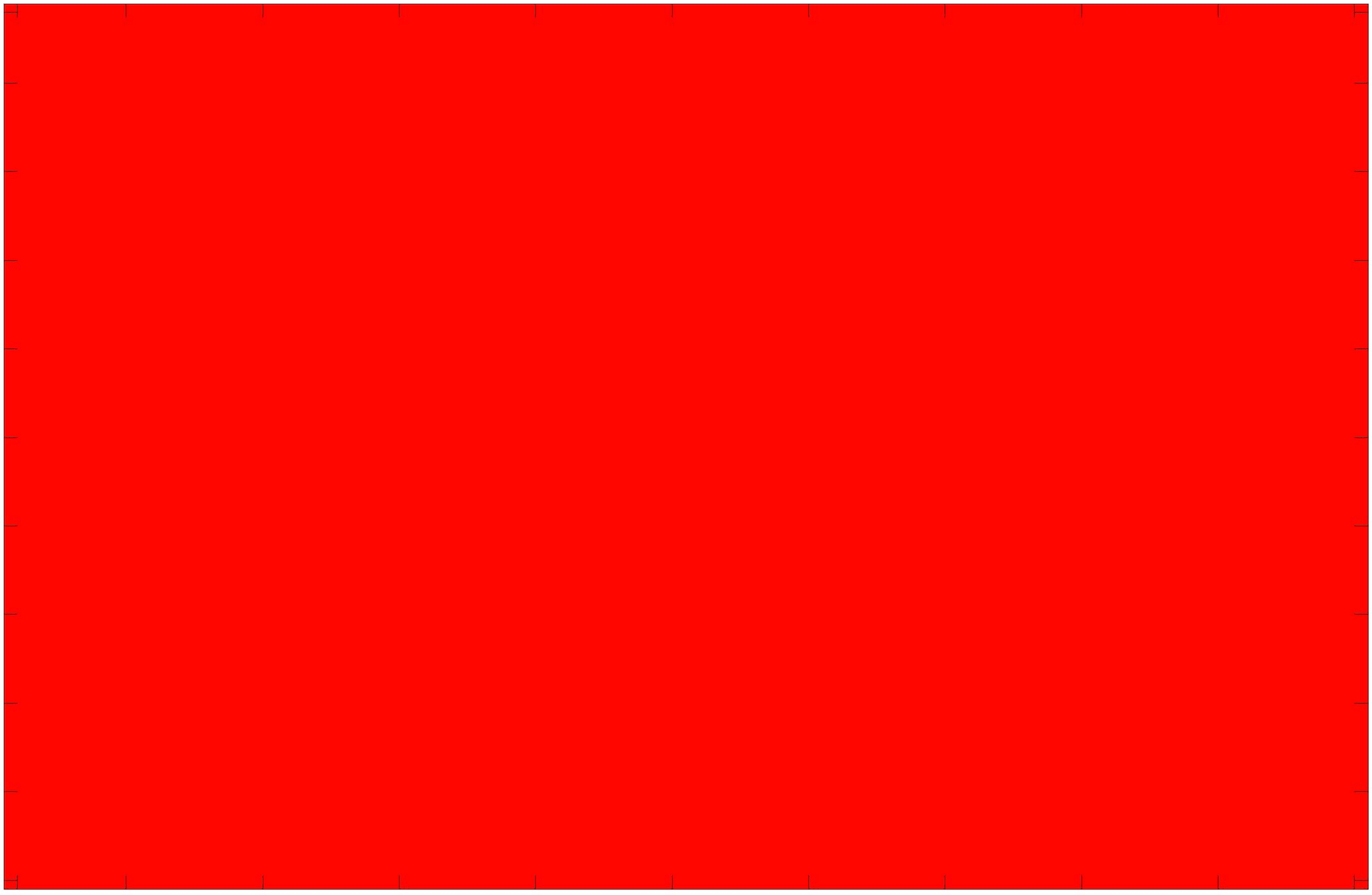}}\quad
\subfloat{\includegraphics[width=1.6in,height=1.1in]{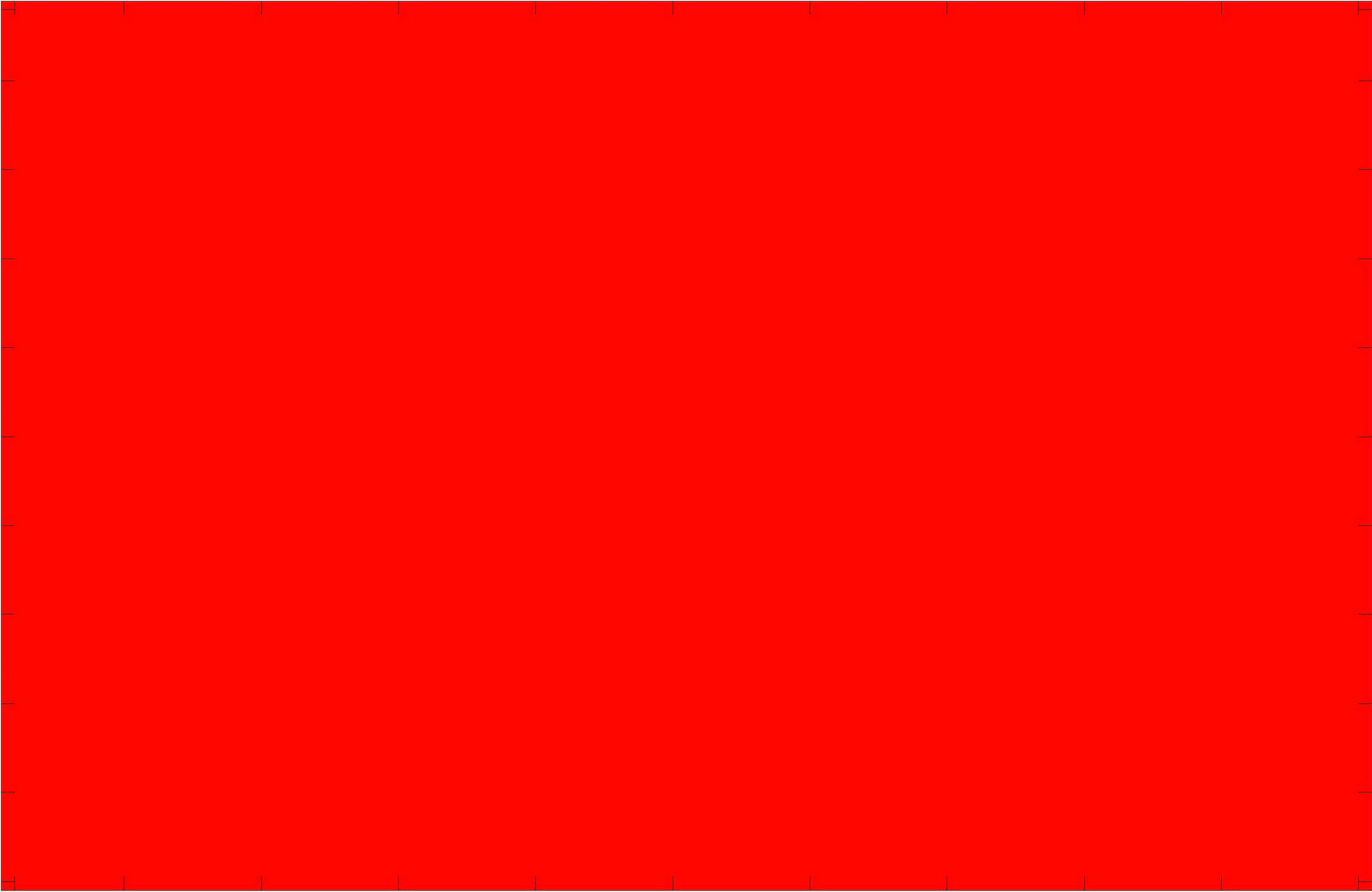}}\quad
\subfloat{\includegraphics[width=1.6in,height=1.1in]{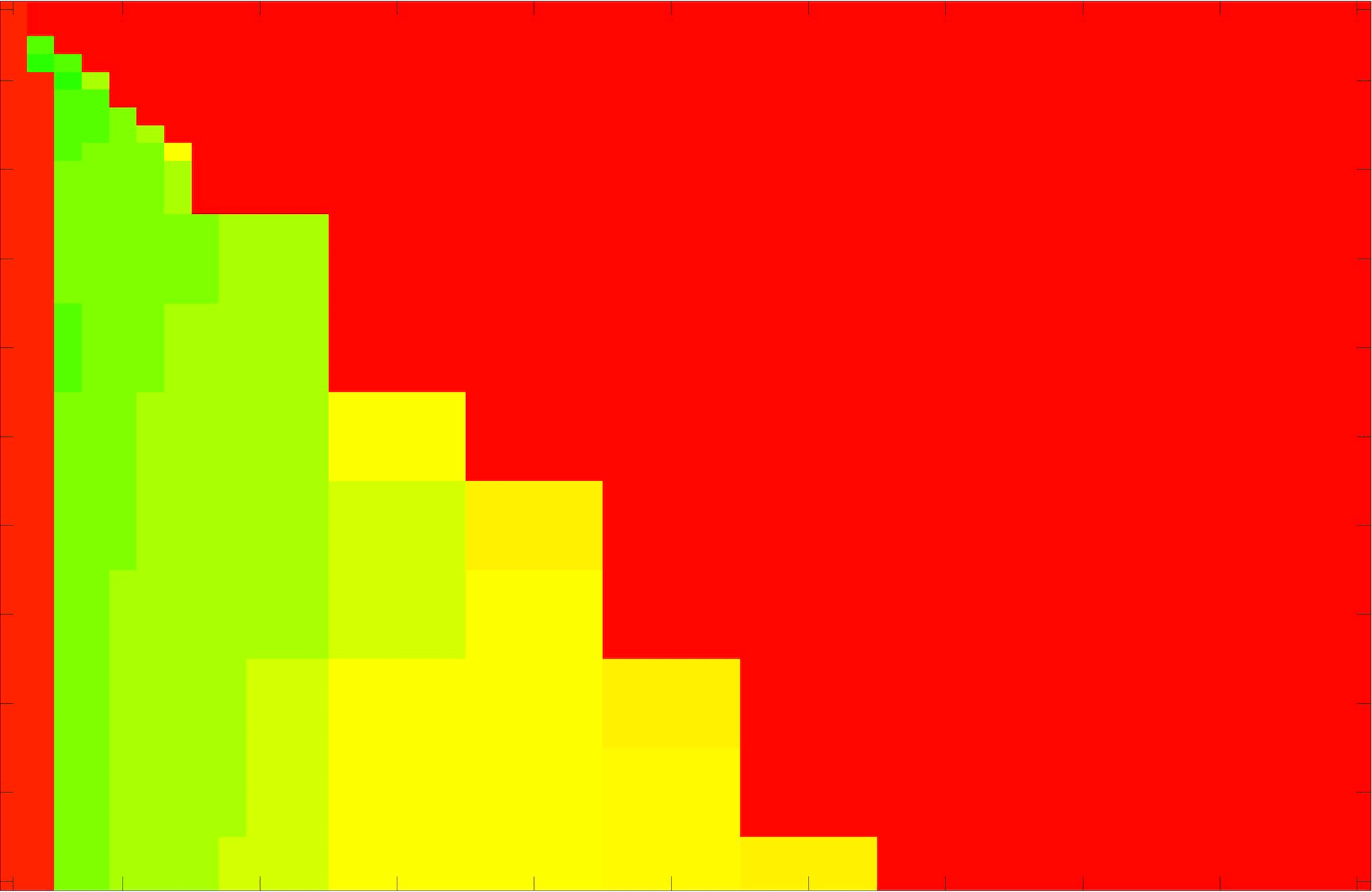}}}
\centering
\floatbox[{\capbeside\thisfloatsetup{capbesideposition={left,center},capbesidewidth=1.5in,font =normalsize}}]{figure}[\FBwidth]
{\caption*{RSF \cite{RSF}}}
{\subfloat{\includegraphics[width=1.6in,height=1.1in]{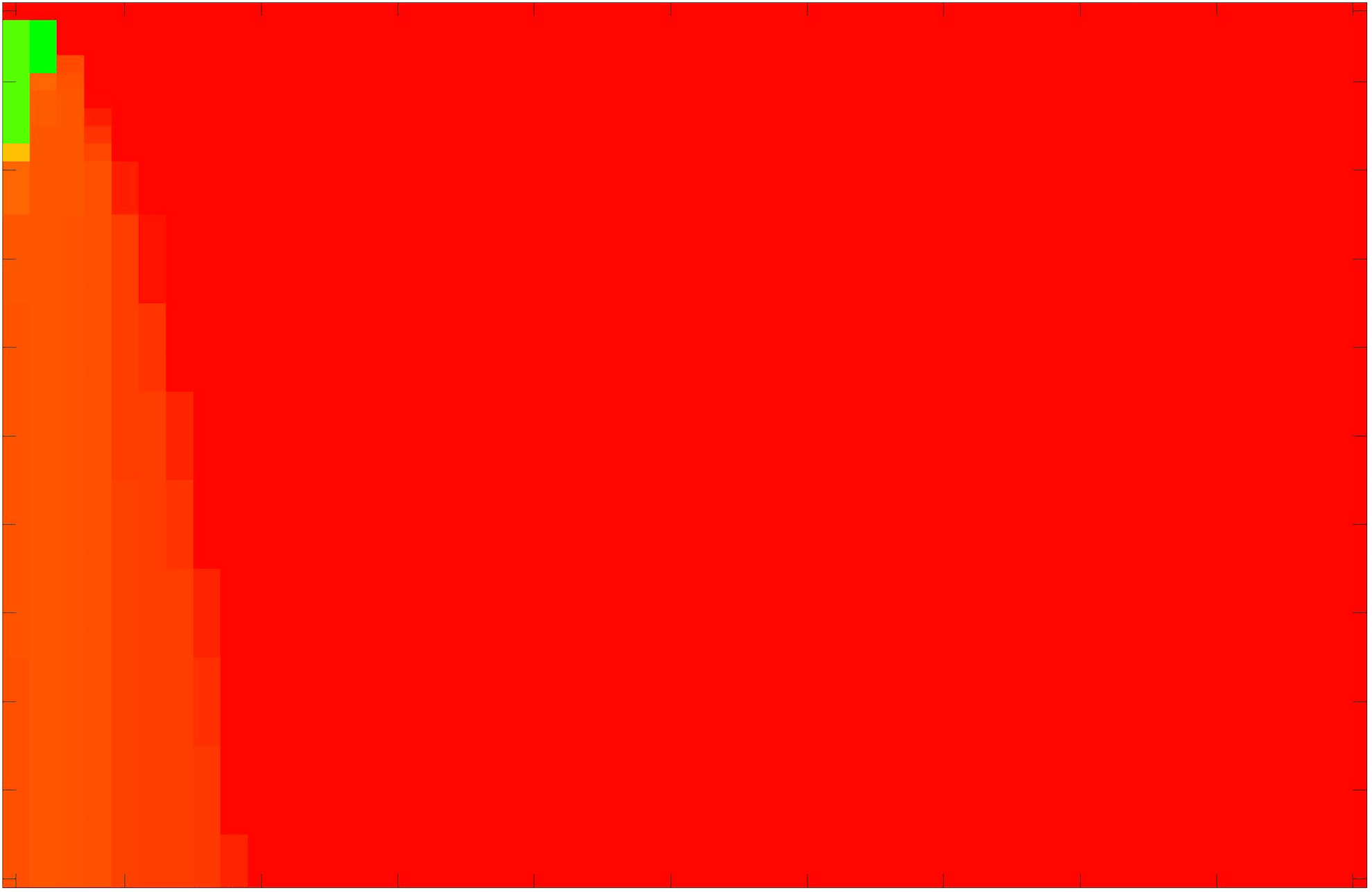}}\quad
\subfloat{\includegraphics[width=1.6in,height=1.1in]{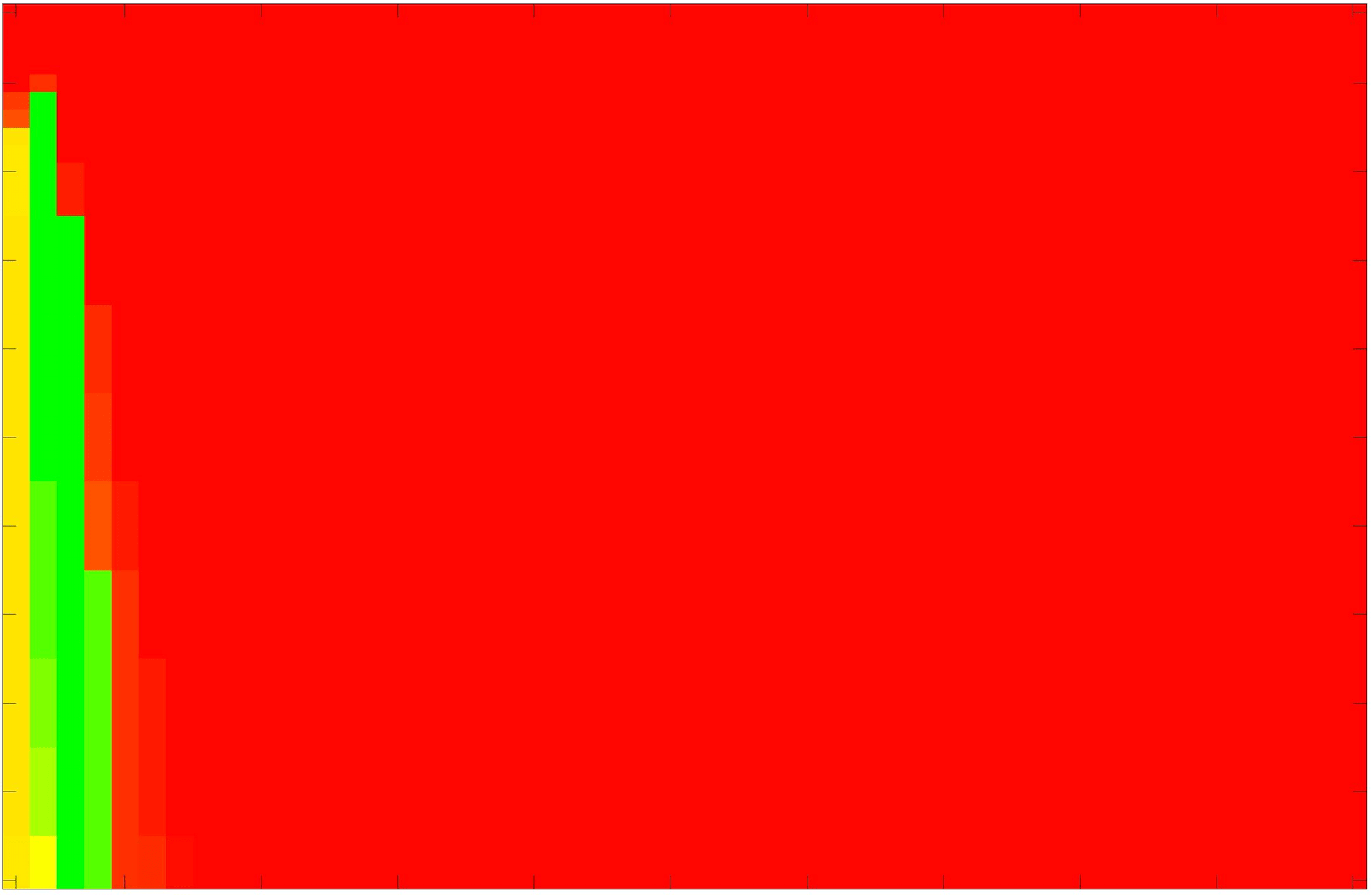}}\quad
\subfloat{\includegraphics[width=1.6in,height=1.1in]{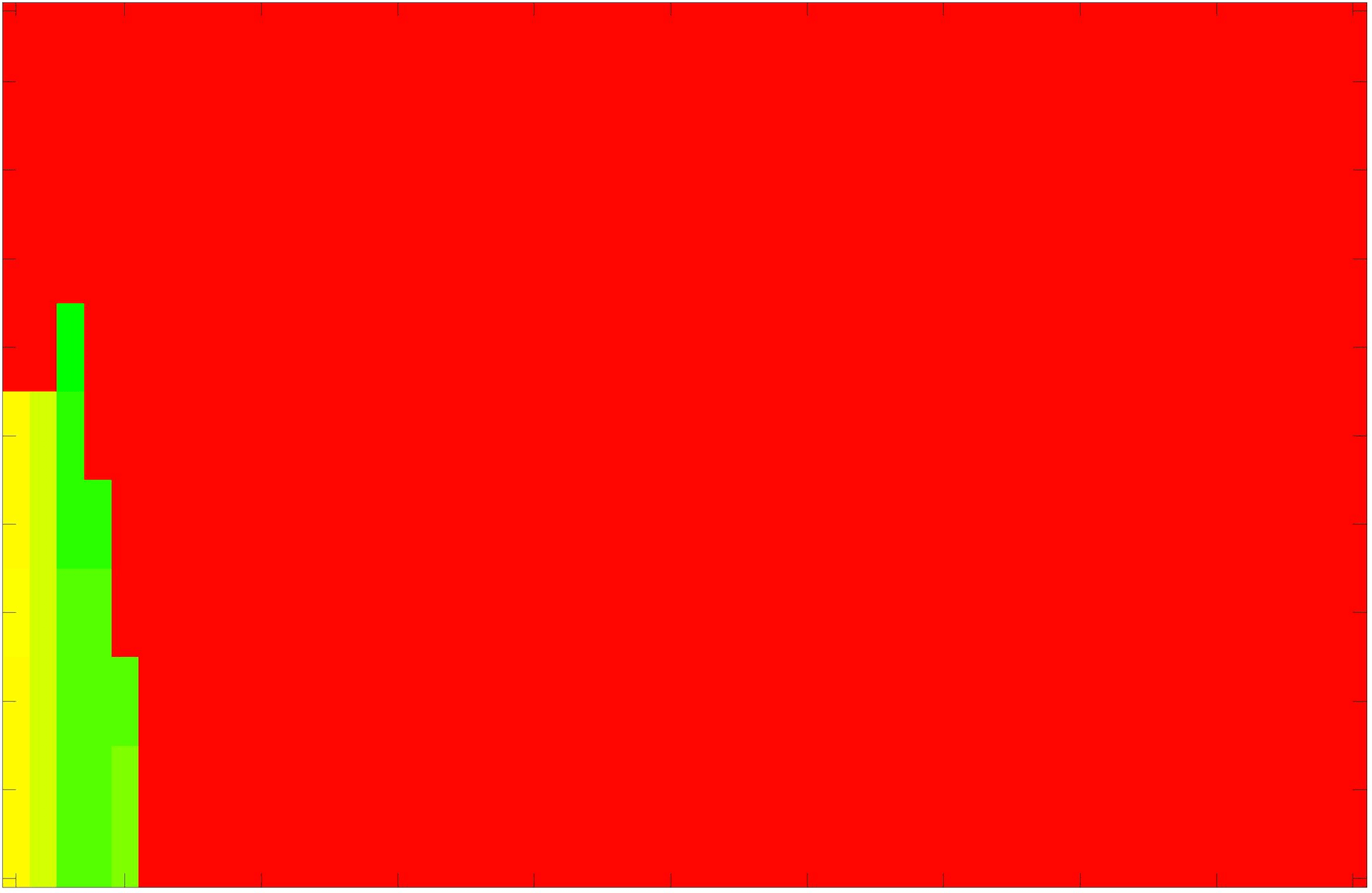}}}
\centering
\floatbox[{\capbeside\thisfloatsetup{capbesideposition={left,center},capbesidewidth=1.5in,font =normalsize}}]{figure}[\FBwidth]
{\caption*{LCV \cite{LCV}}}
{\subfloat{\includegraphics[width=1.6in,height=1.1in]{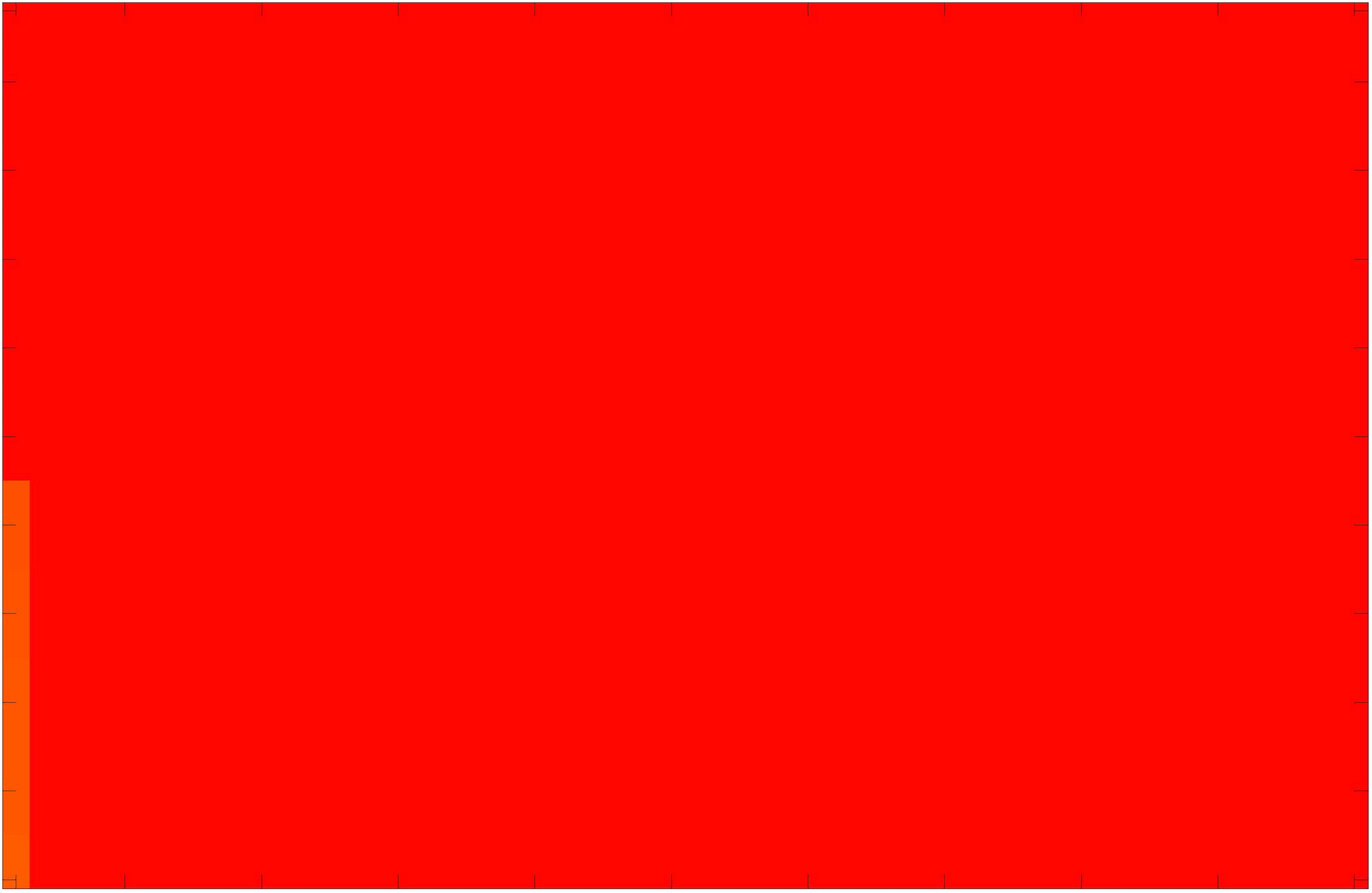}}\quad
\subfloat{\includegraphics[width=1.6in,height=1.1in]{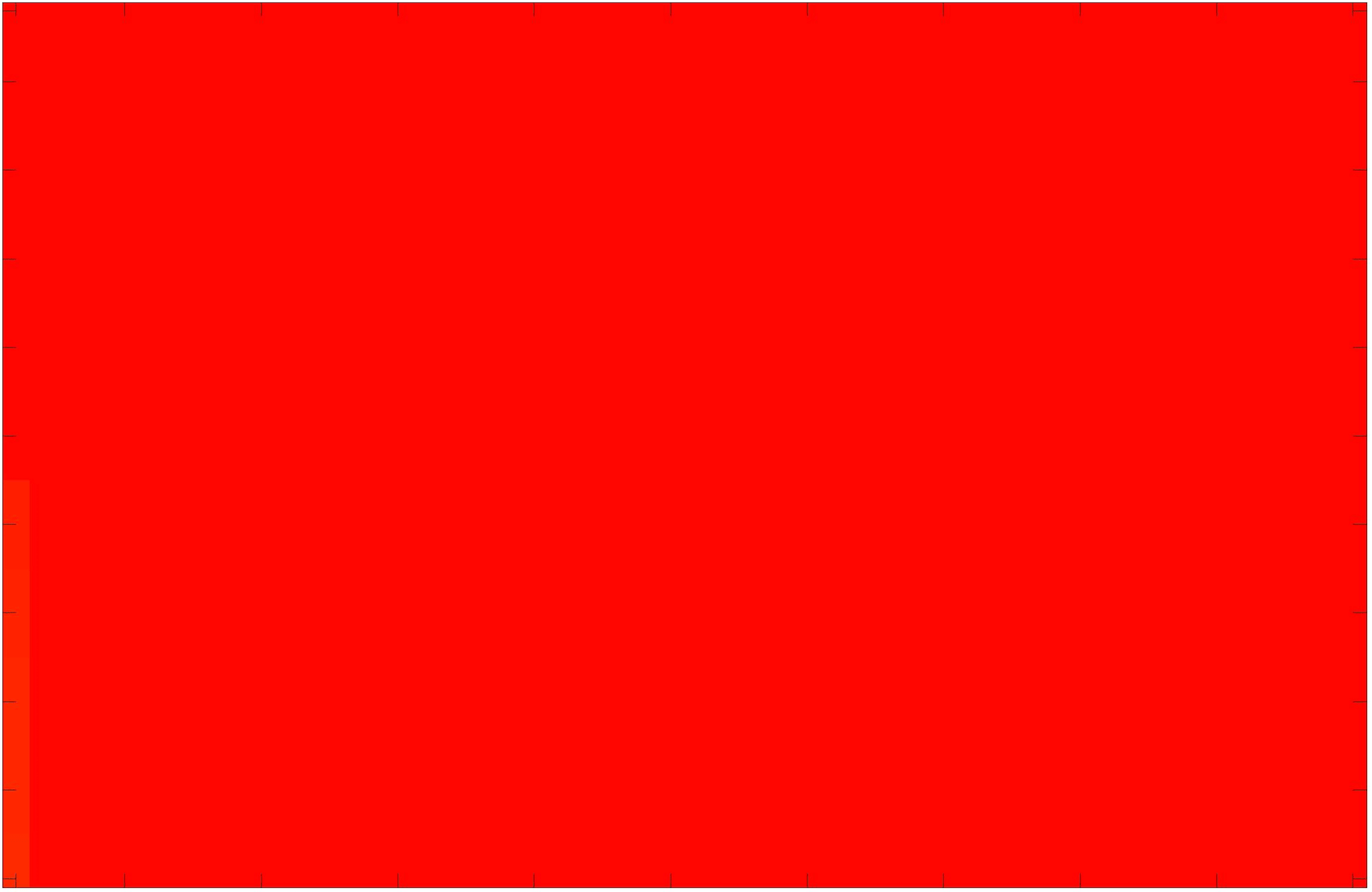}}\quad
\subfloat{\includegraphics[width=1.6in,height=1.1in]{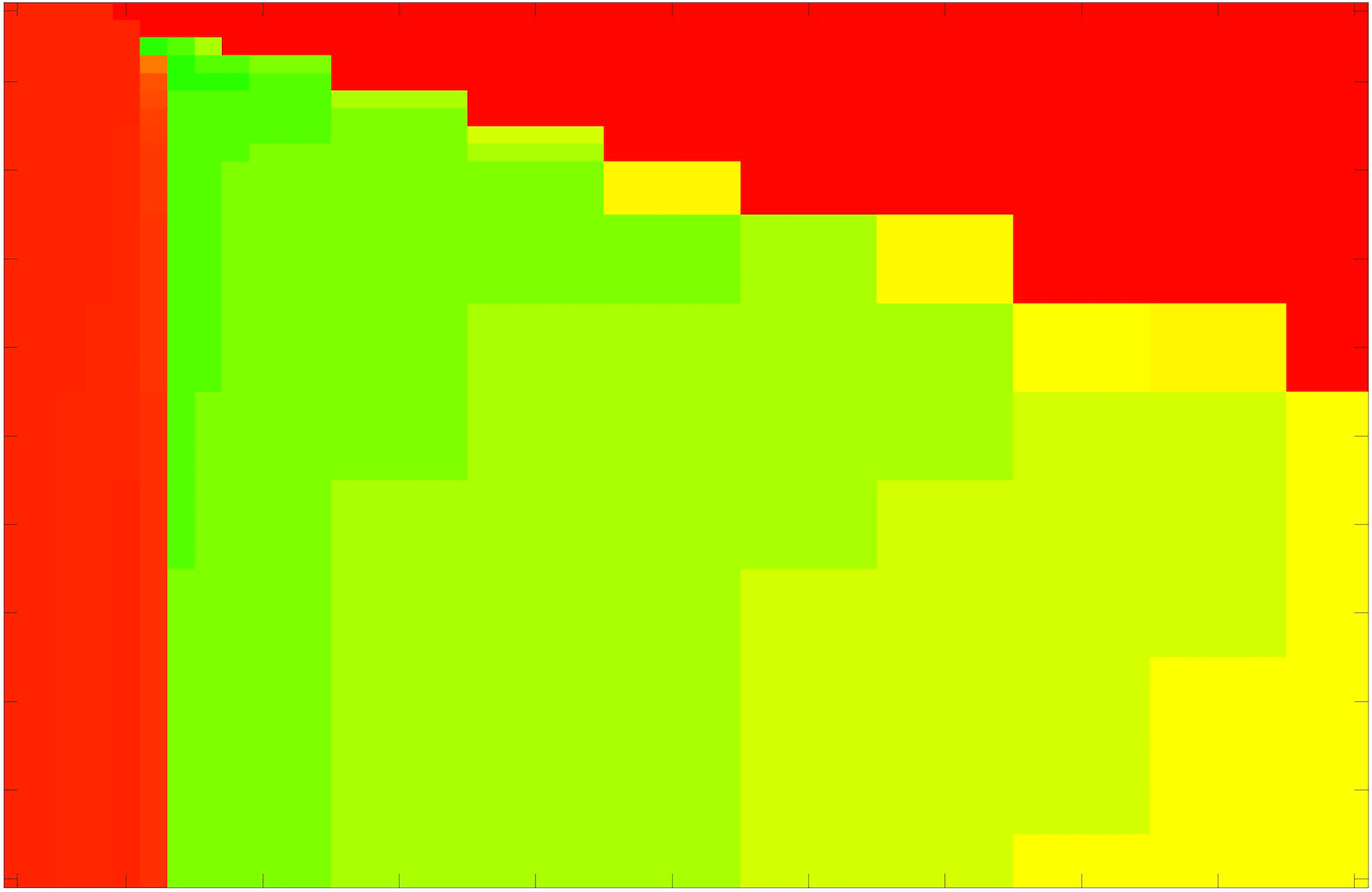}}} 
\centering
\floatbox[{\capbeside\thisfloatsetup{capbesideposition={left,center},capbesidewidth=1.5in,font =normalsize}}]{figure}[\FBwidth]
{\caption*{HYB \cite{Ali:16}}}
{\subfloat{\includegraphics[width=1.6in,height=1.1in]{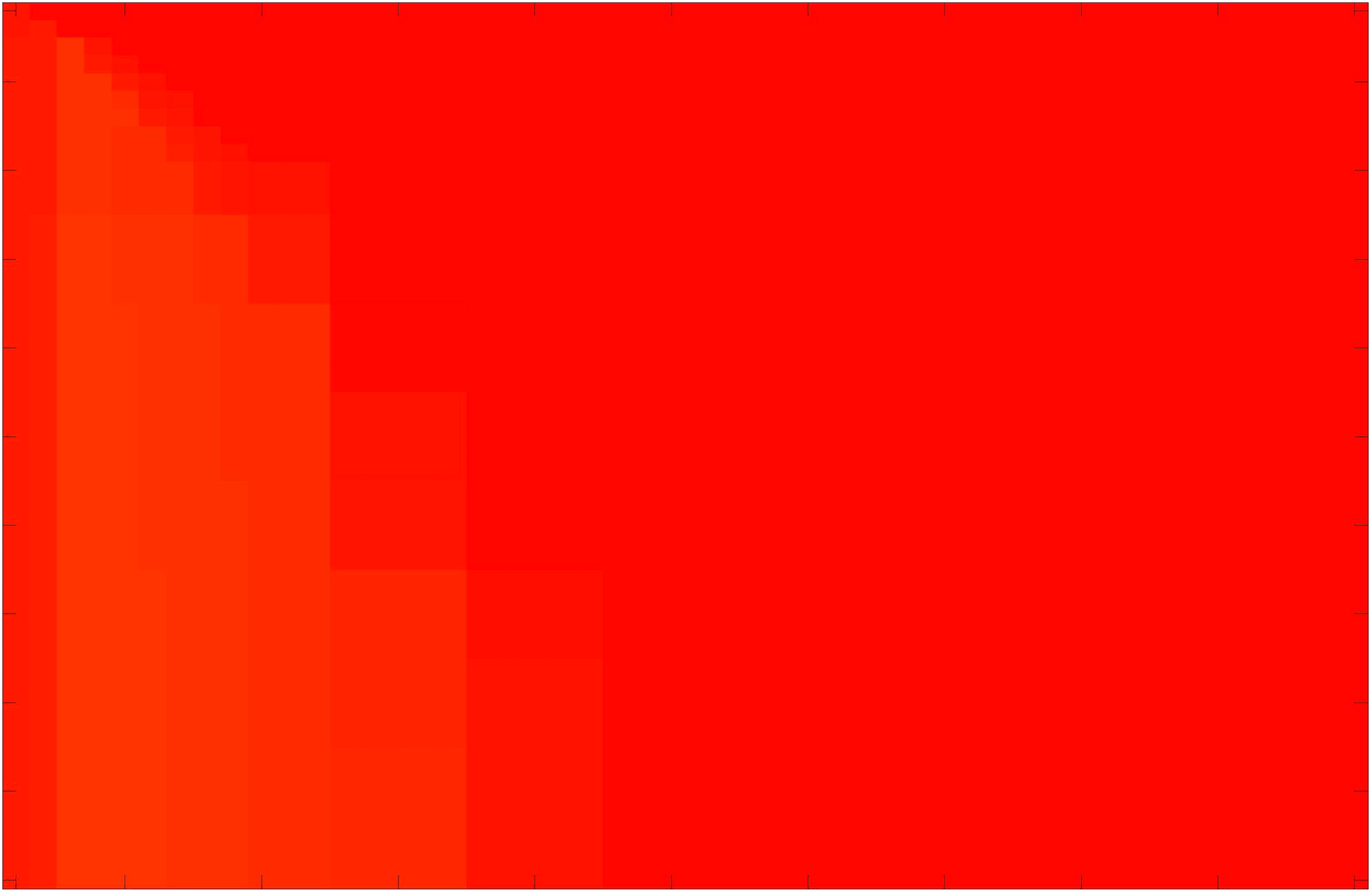}}\quad
\subfloat{\includegraphics[width=1.6in,height=1.1in]{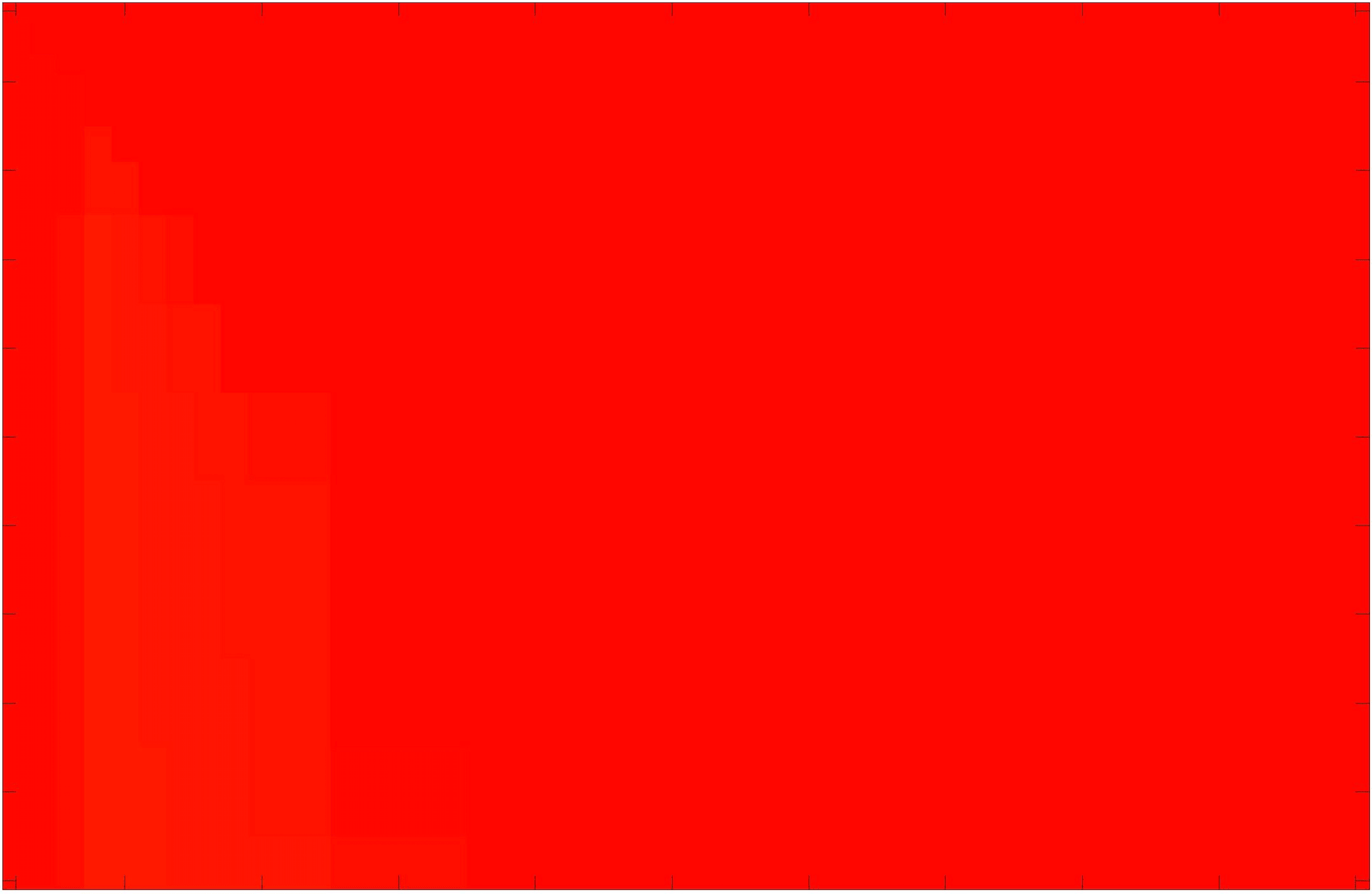}}\quad
\subfloat{\includegraphics[width=1.6in,height=1.1in]{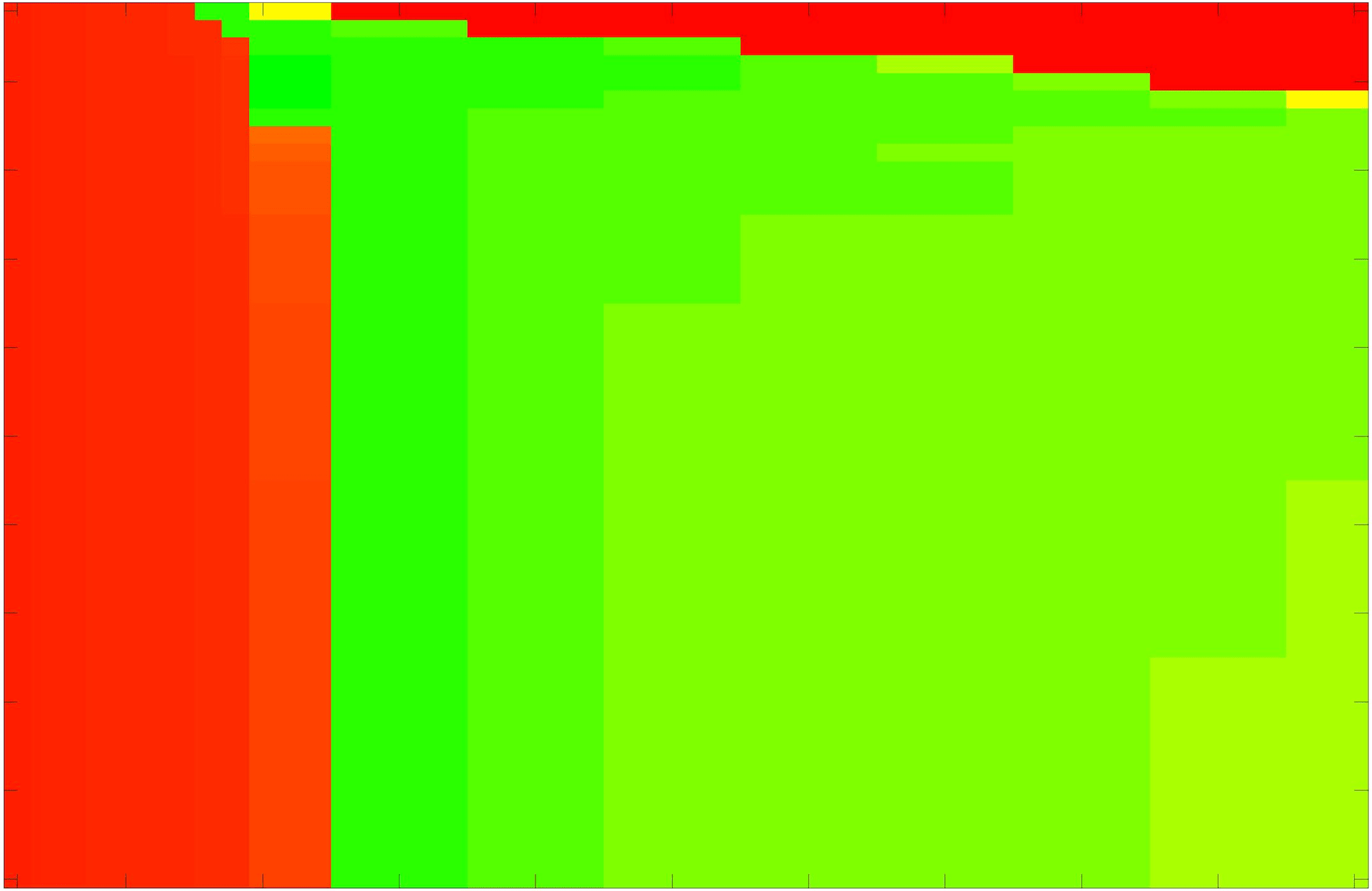}}}
\centering
\floatbox[{\capbeside\thisfloatsetup{capbesideposition={left,center},capbesidewidth=1.5in,font =normalsize}}]{figure}[\FBwidth]
{\caption*{GAV \cite{Ali:17}}}
{\subfloat{\includegraphics[width=1.6in,height=1.1in]{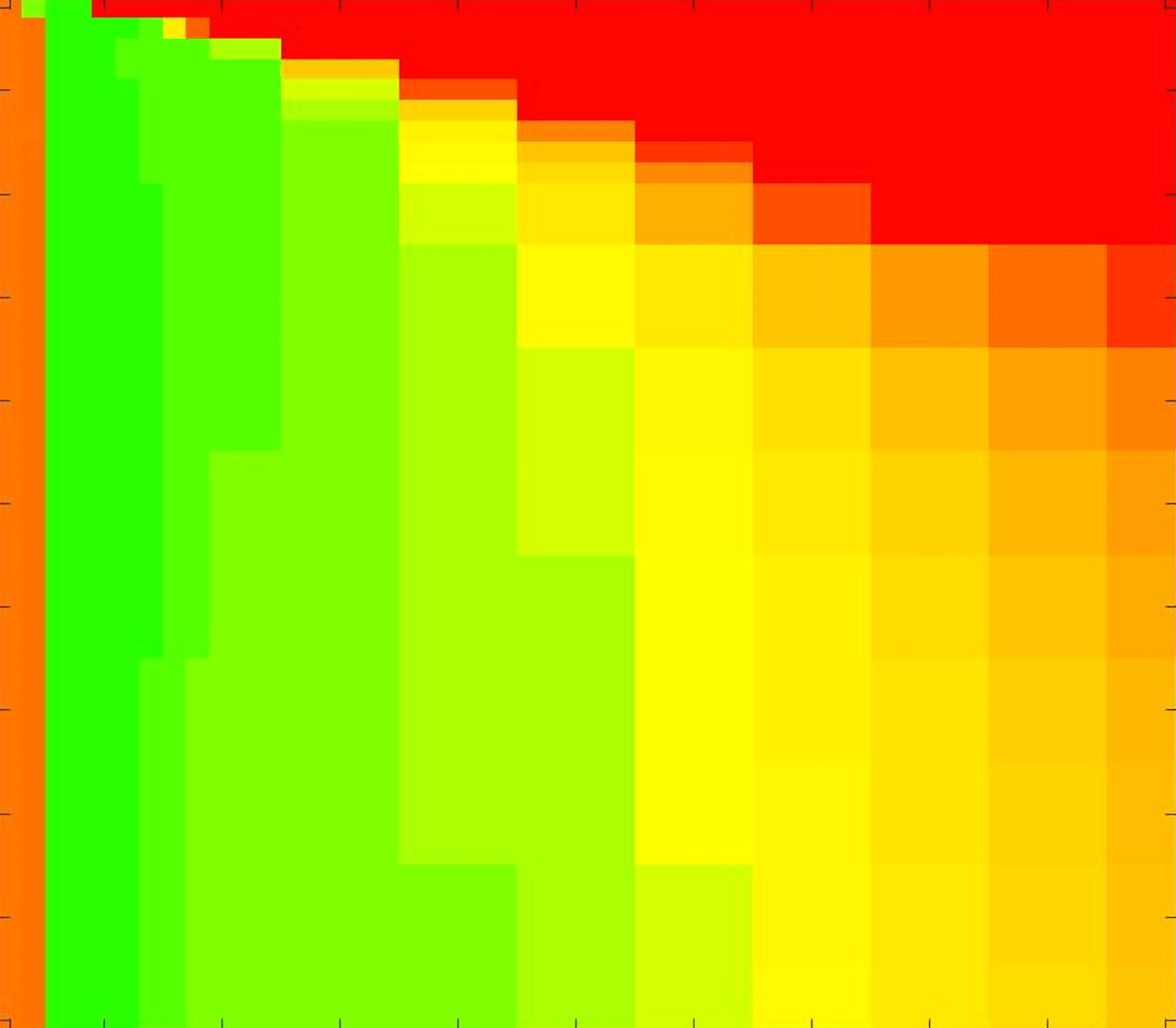}}\quad
\subfloat{\includegraphics[width=1.6in,height=1.1in]{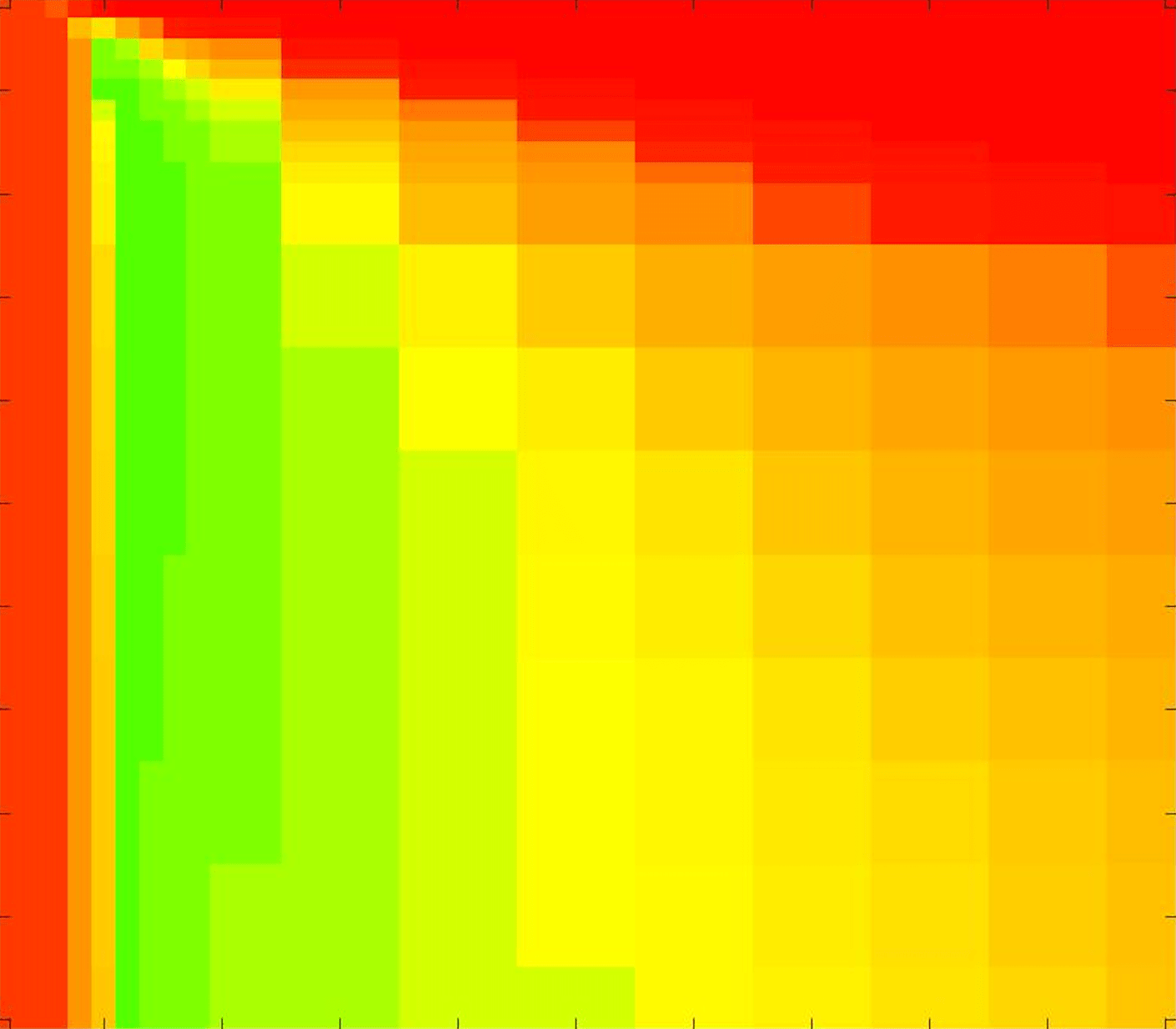}}\quad
\subfloat{\includegraphics[width=1.6in,height=1.1in]{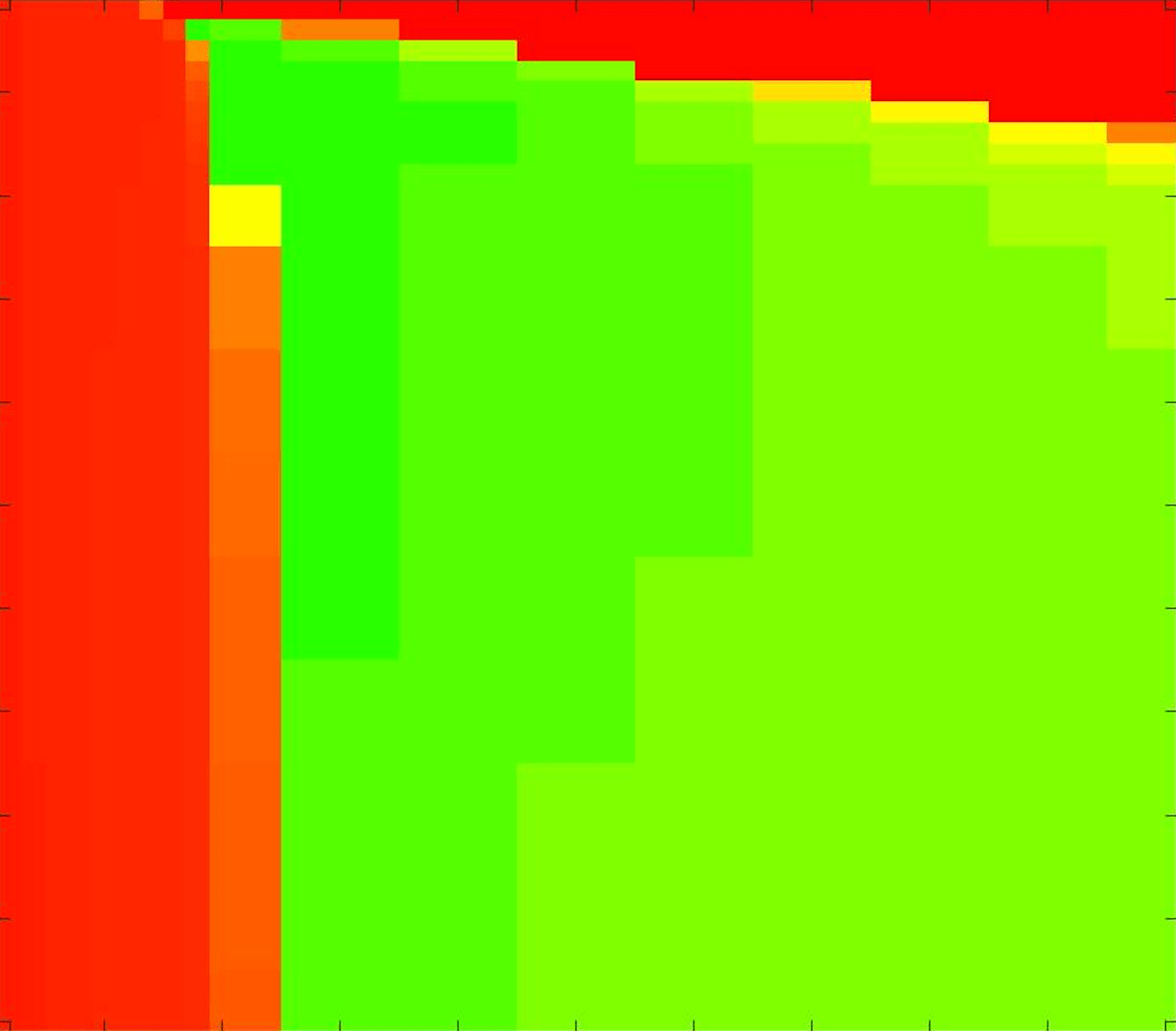}}}
\centering
\floatbox[{\capbeside\thisfloatsetup{capbesideposition={left,center},capbesidewidth=1.5in,font =normalsize}}]{figure}[\FBwidth]
{\caption*{Proposed}}
{\subfloat{\includegraphics[width=1.6in,height=1.1in]{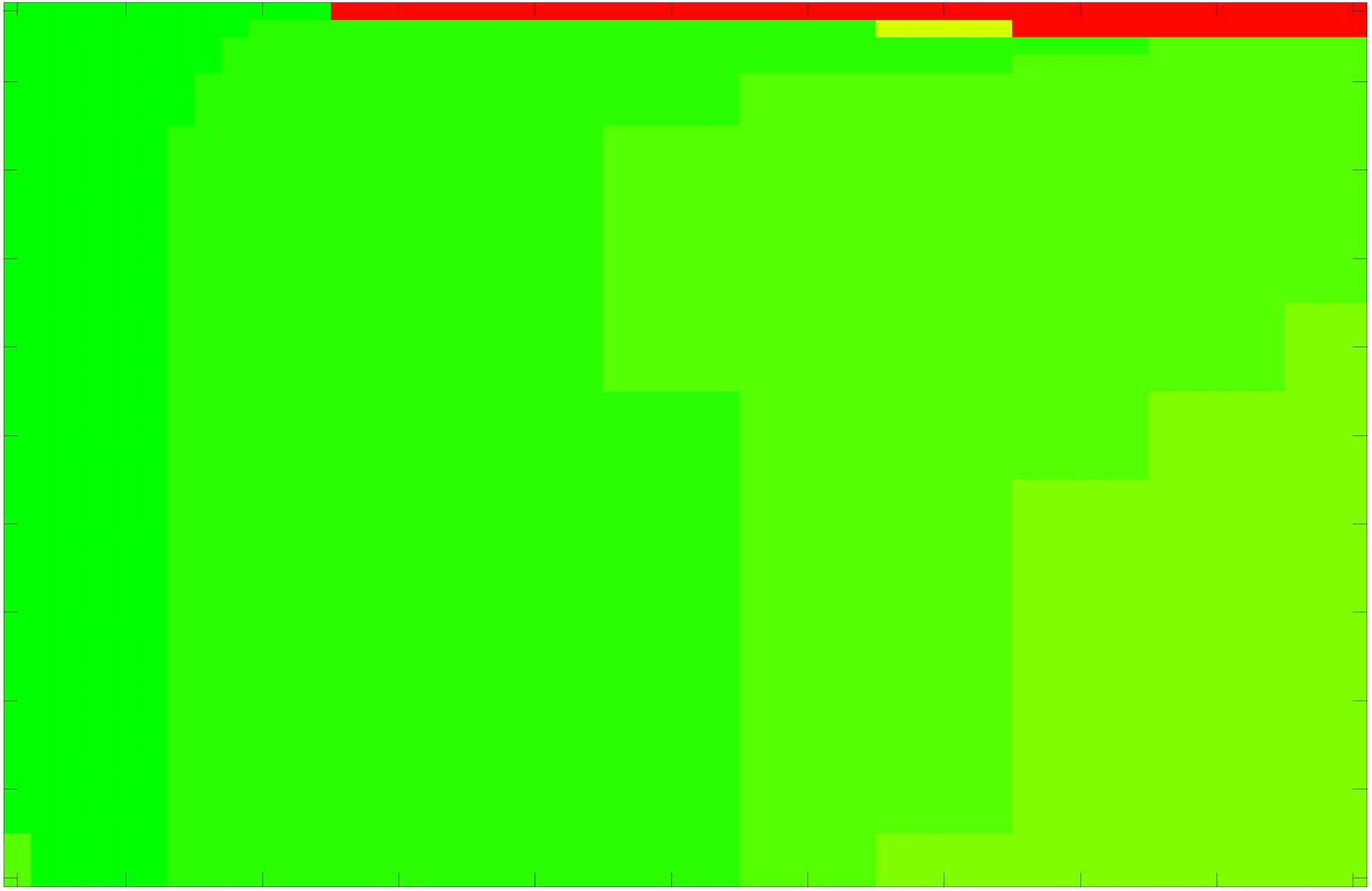}}\quad
\subfloat{\includegraphics[width=1.6in,height=1.1in]{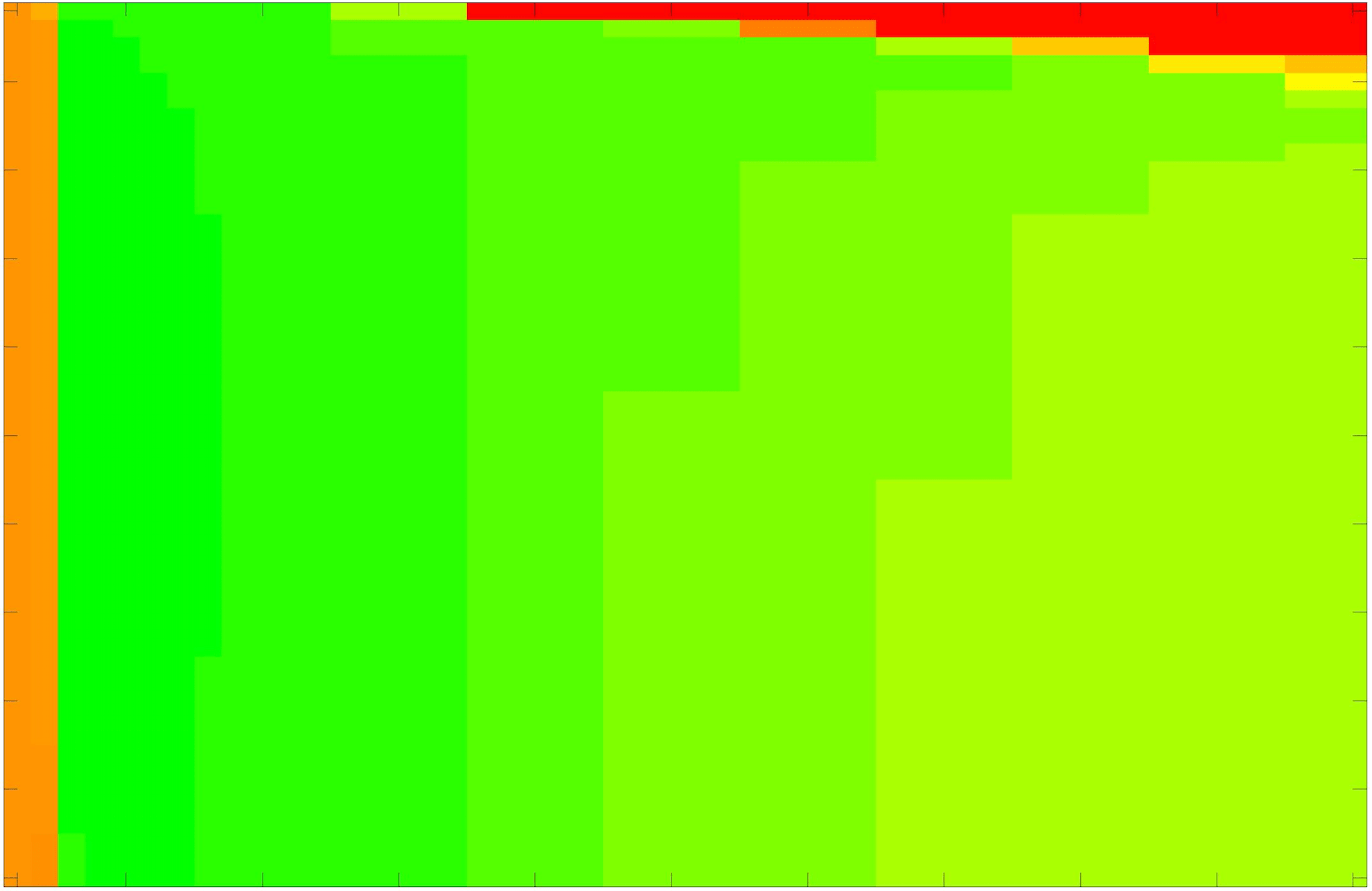}}\quad
\subfloat{\includegraphics[width=1.6in,height=1.1in]{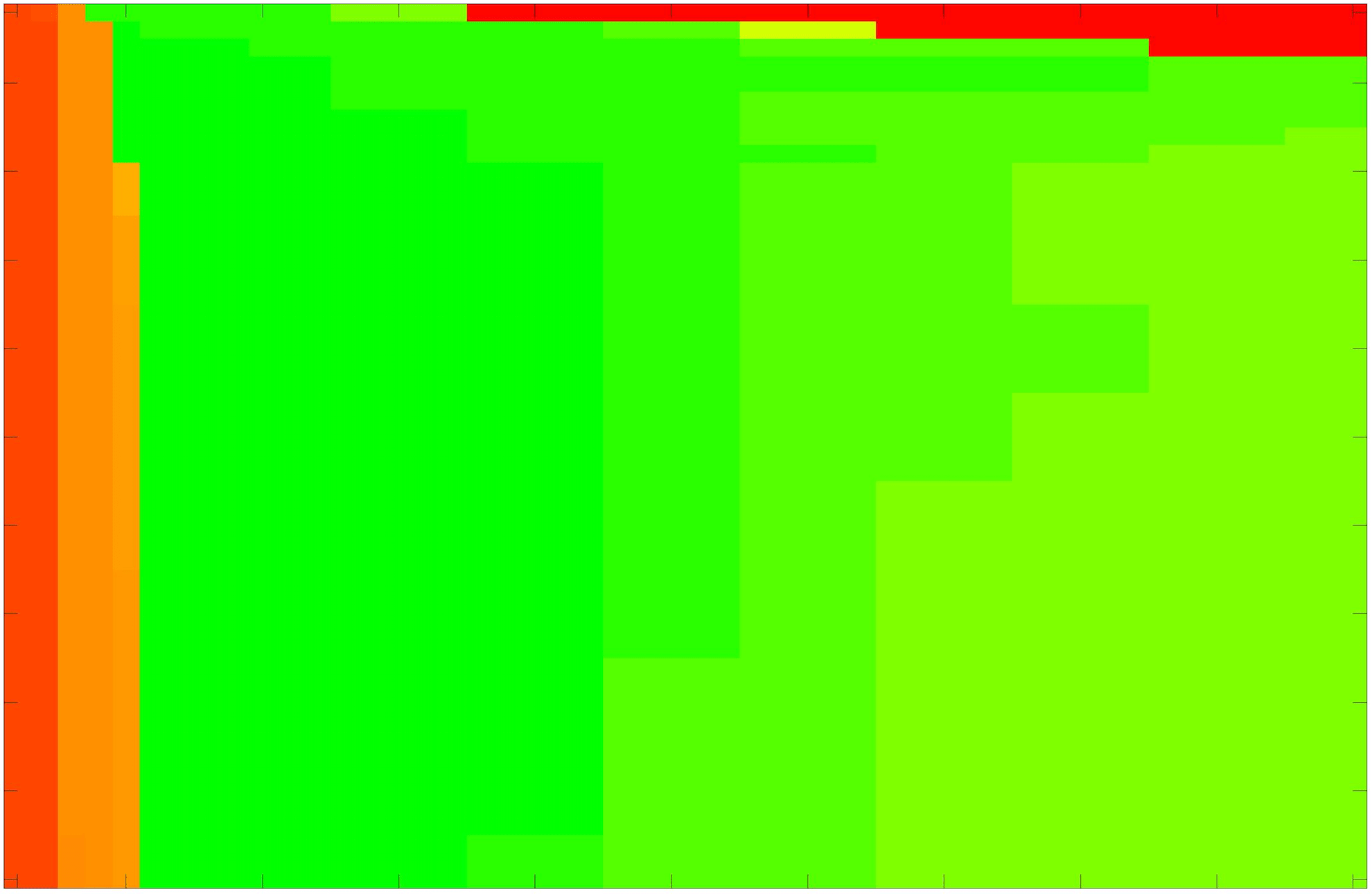}}} 
\end{figure*}

\begin{figure*}
\centering
\floatbox[{\capbeside\thisfloatsetup{capbesideposition={left,top},capbesidewidth=1.5in}}]{figure}[\FBwidth]
{\captionsetup{position=top}\captionsetup[subfigure]{labelformat=empty,font=normalsize} 
\subfloat[Test Image 4]{\includegraphics[width=1.6in,height = 1.4in]{figs/GT_Knee}}\quad
\captionsetup{position=top}\captionsetup[subfigure]{labelformat=empty,font=normalsize} 
\subfloat[Test Image 5]{\includegraphics[width=1.6in,height = 1.4in]{figs/GT_Kidney-min}}\quad
\captionsetup{position=top}\captionsetup[subfigure]{labelformat=empty,font=normalsize} 
\subfloat[Test Image 6]{\includegraphics[width=1.6in,height = 1.4in]{figs/GT_Organ-min}}}
{\caption{Heatmaps of TC values for permutations of $\tilde{\lambda}$ and $\theta$. Each row and column is labelled according to the model used and the image tested. The colour is consistent with the scale in Fig.~\ref{fig:colorbar}. Here, we present Test Images 4 -- 6. \label{fig:heat2}}}\vspace{-0.1in}
\centering
\floatbox[{\capbeside\thisfloatsetup{capbesideposition={left,center},capbesidewidth=1.5in,font =normalsize}}]{figure}[\FBwidth]
{\caption*{CV \cite{ACWE}}}
{\subfloat{\includegraphics[width=1.6in,height=1.1in]{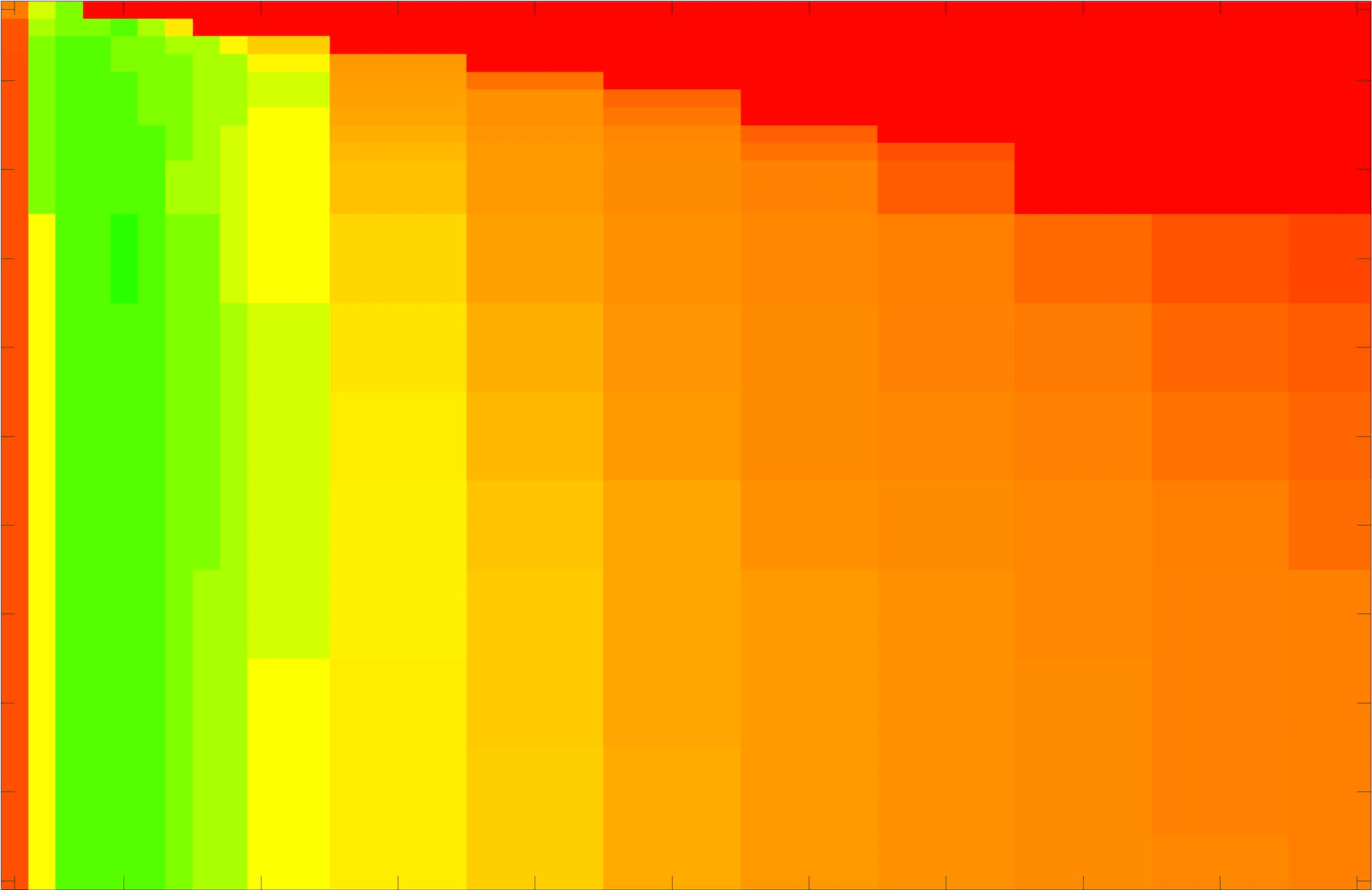}}\quad
\subfloat{\includegraphics[width=1.6in,height=1.1in]{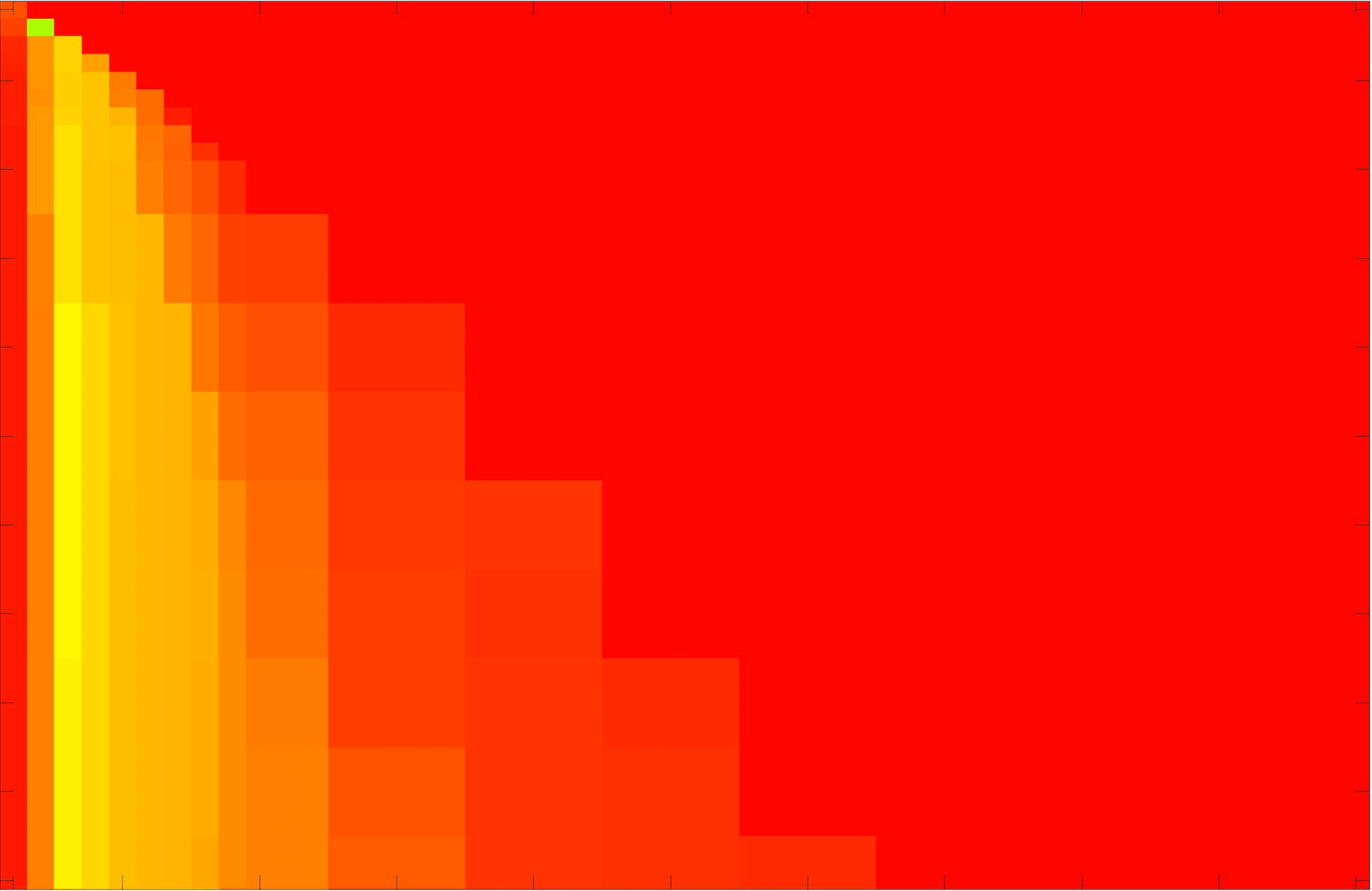}}\quad
\subfloat{\includegraphics[width=1.6in,height=1.1in]{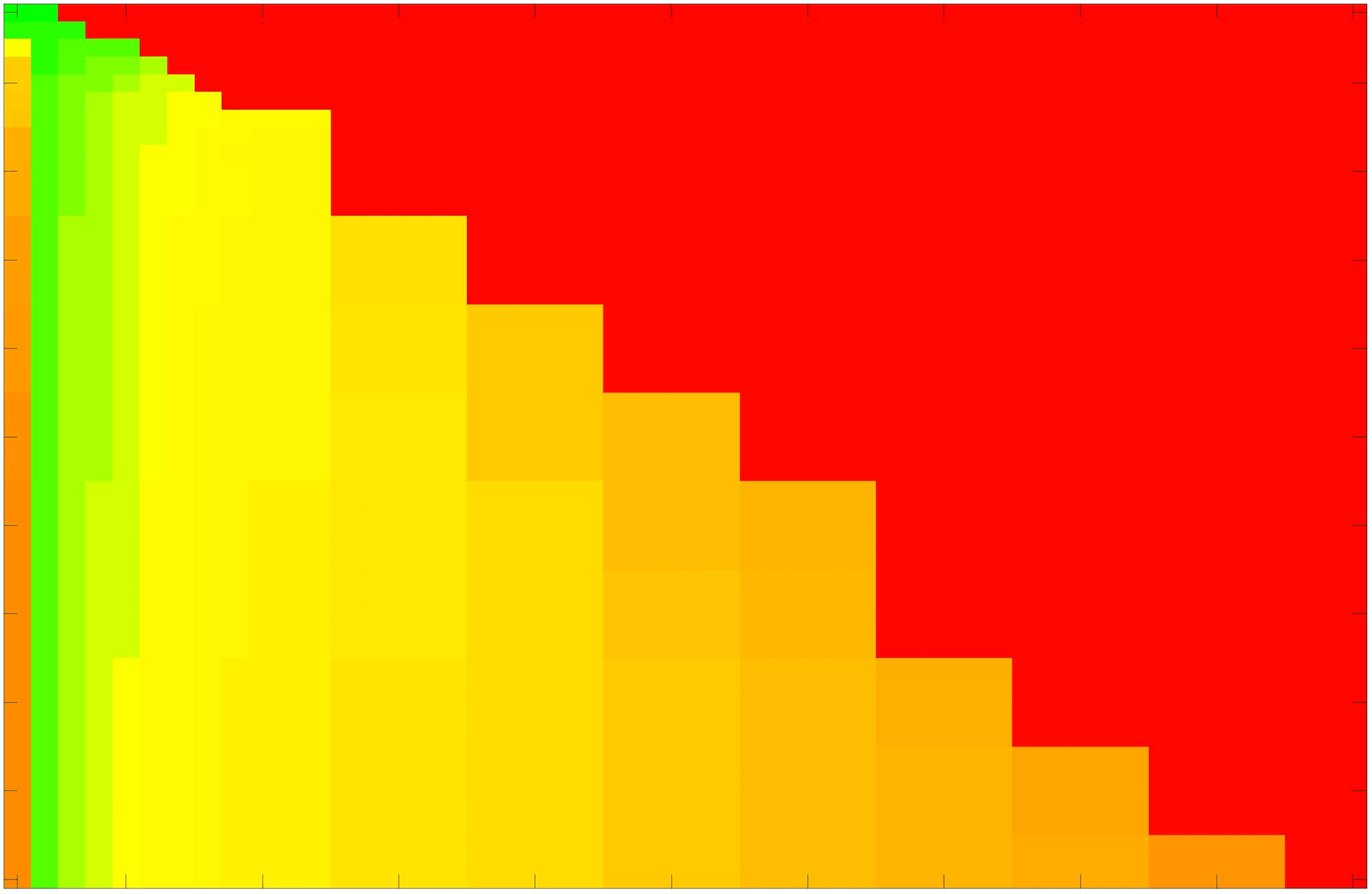}}}
\centering
\floatbox[{\capbeside\thisfloatsetup{capbesideposition={left,center},capbesidewidth=1.5in,font =normalsize}}]{figure}[\FBwidth]
{\caption*{RSF \cite{RSF}}}
{\subfloat{\includegraphics[width=1.6in,height=1.1in]{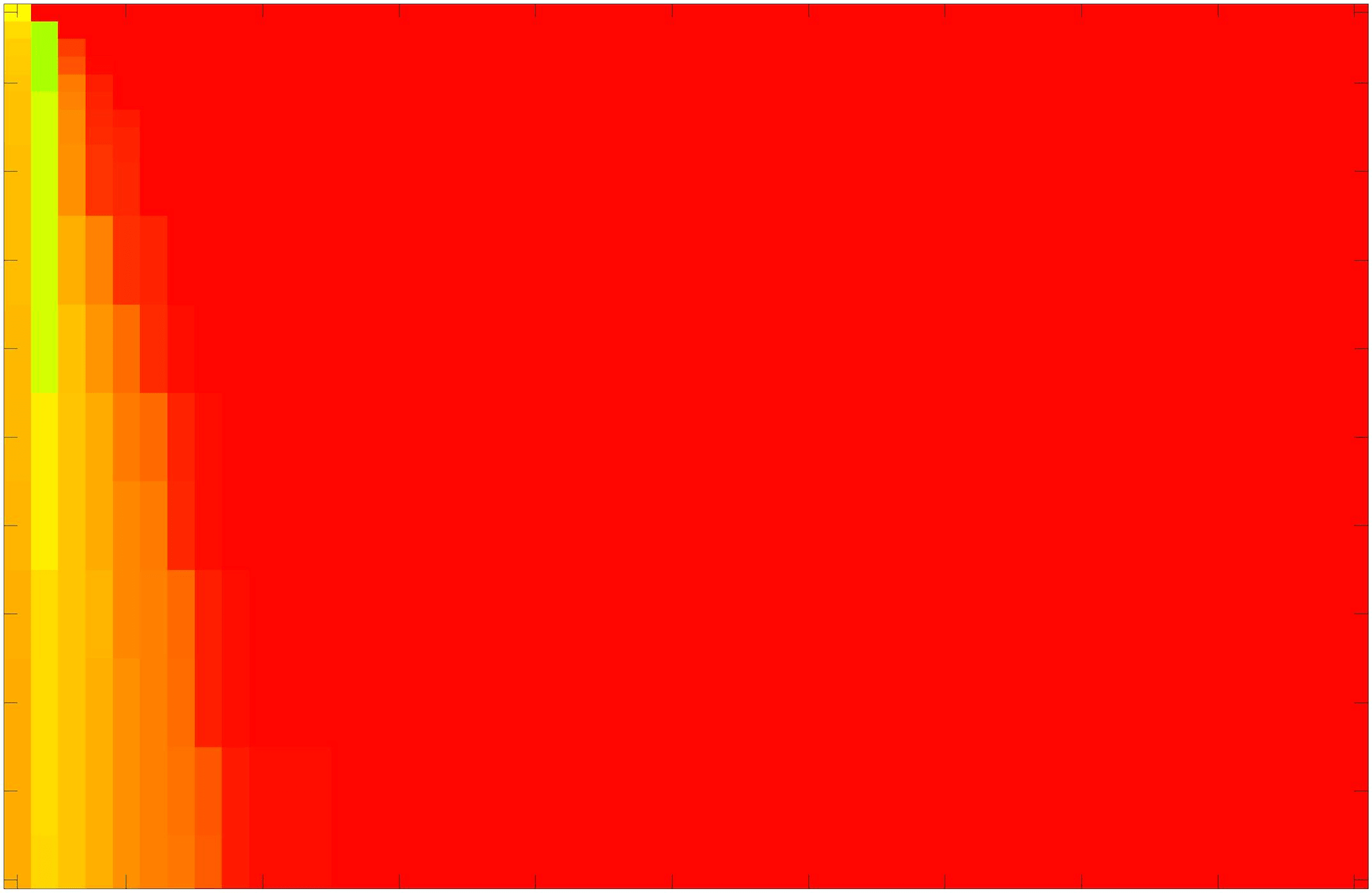}}\quad
\subfloat{\includegraphics[width=1.6in,height=1.1in]{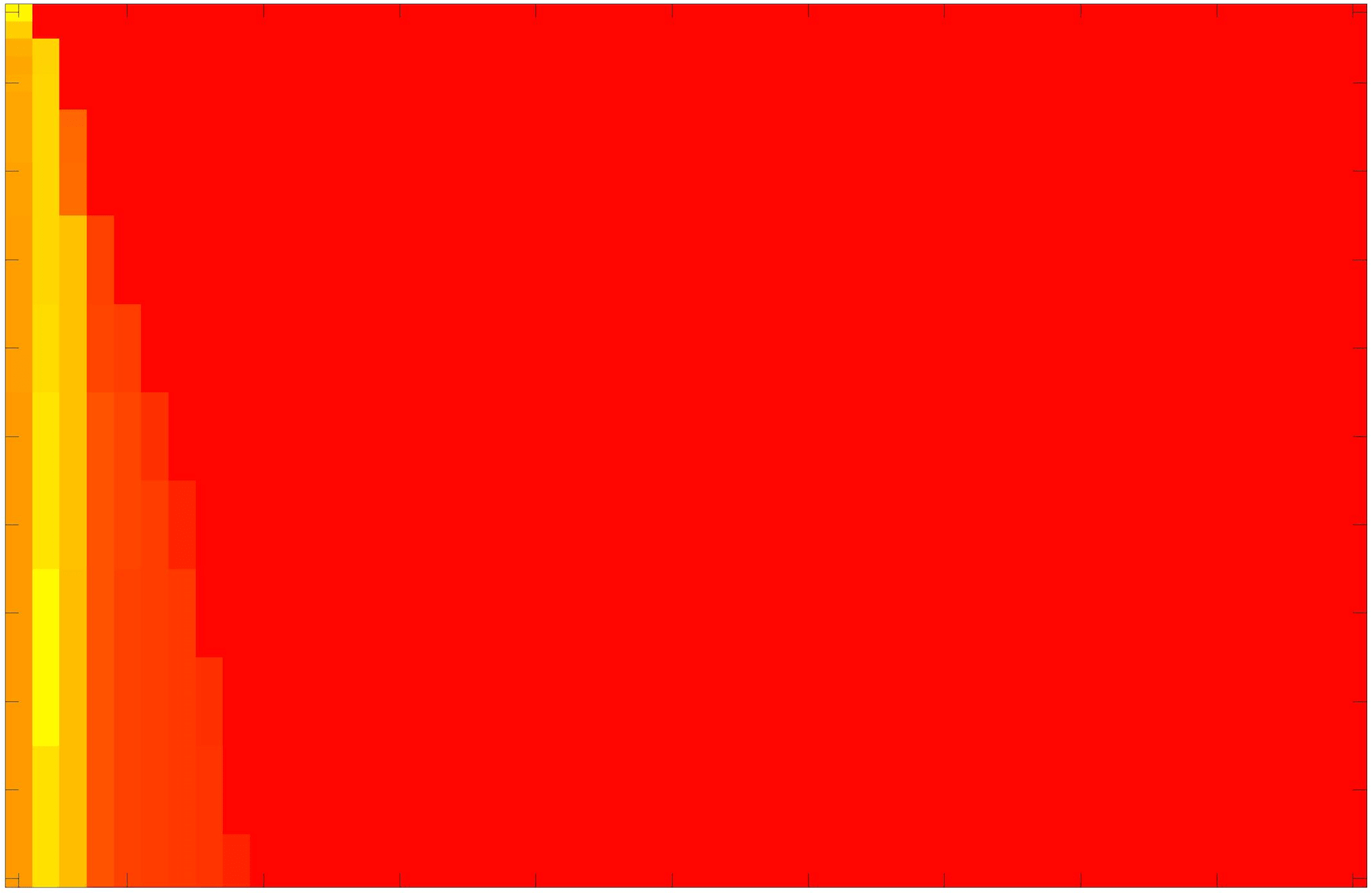}}\quad
\subfloat{\includegraphics[width=1.6in,height=1.1in]{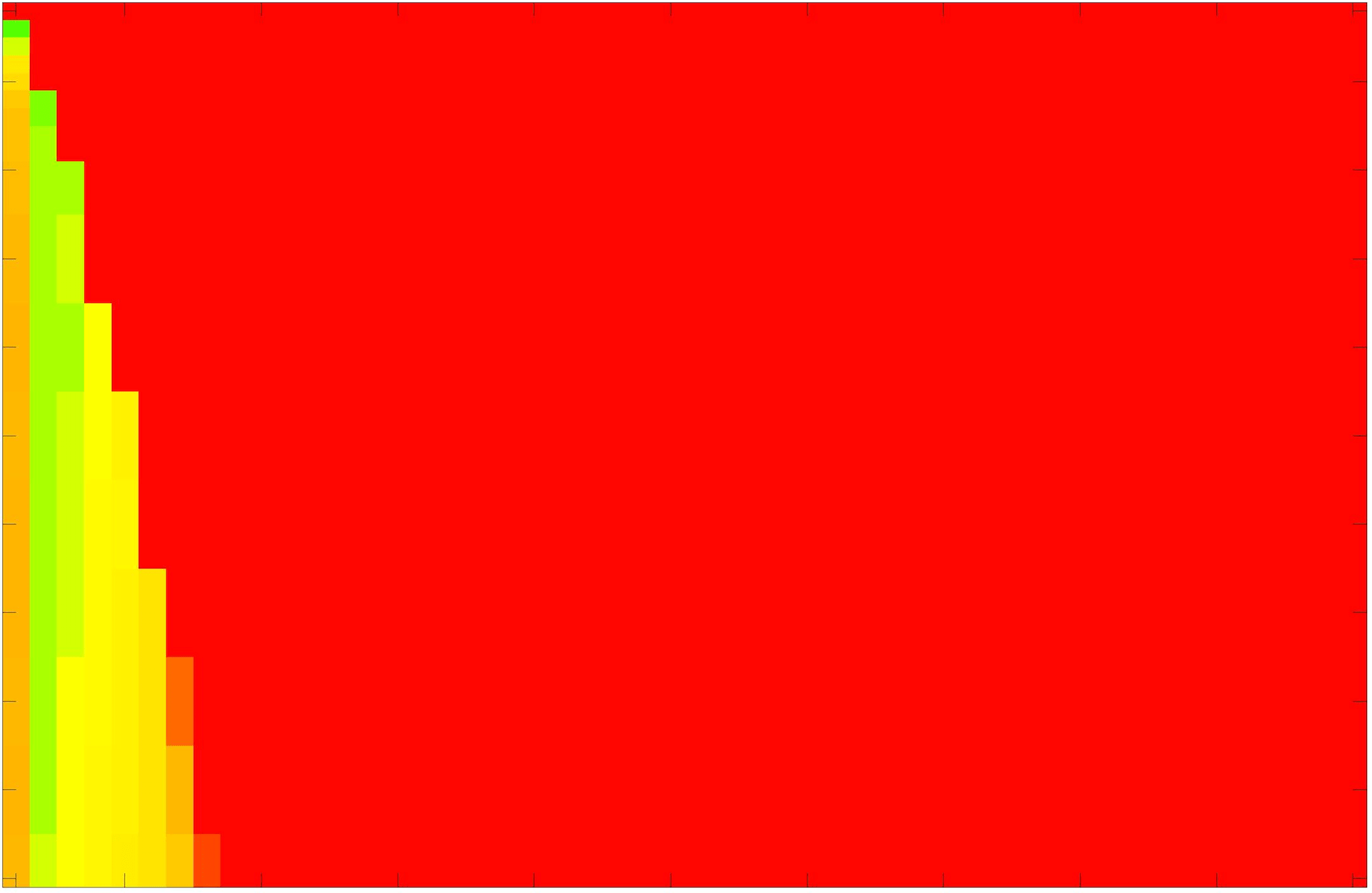}}}
\centering
\floatbox[{\capbeside\thisfloatsetup{capbesideposition={left,center},capbesidewidth=1.5in,font =normalsize}}]{figure}[\FBwidth]
{\caption*{LCV \cite{LCV}}}
{\subfloat{\includegraphics[width=1.6in,height=1.1in]{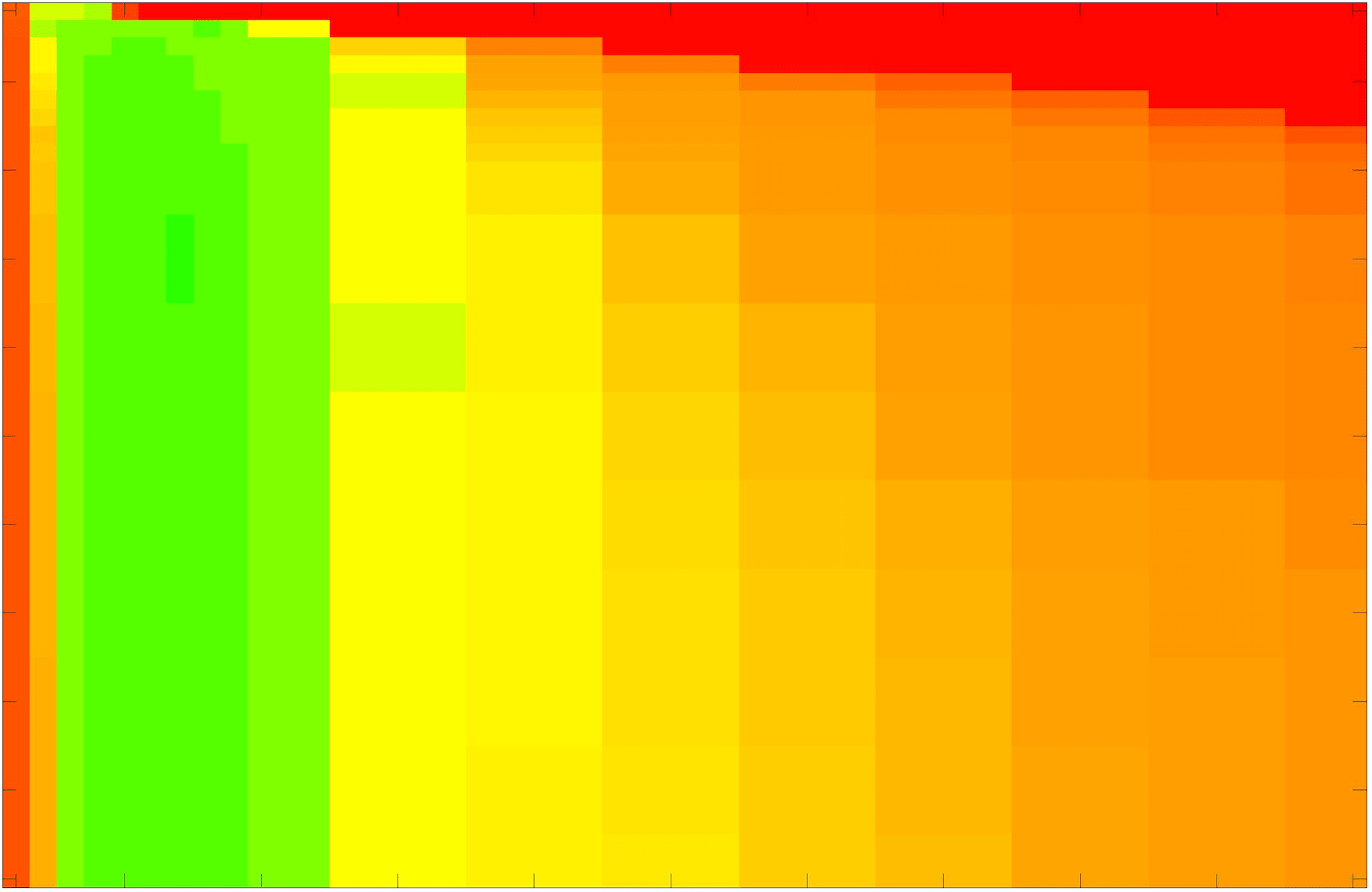}}\quad
\subfloat{\includegraphics[width=1.6in,height=1.1in]{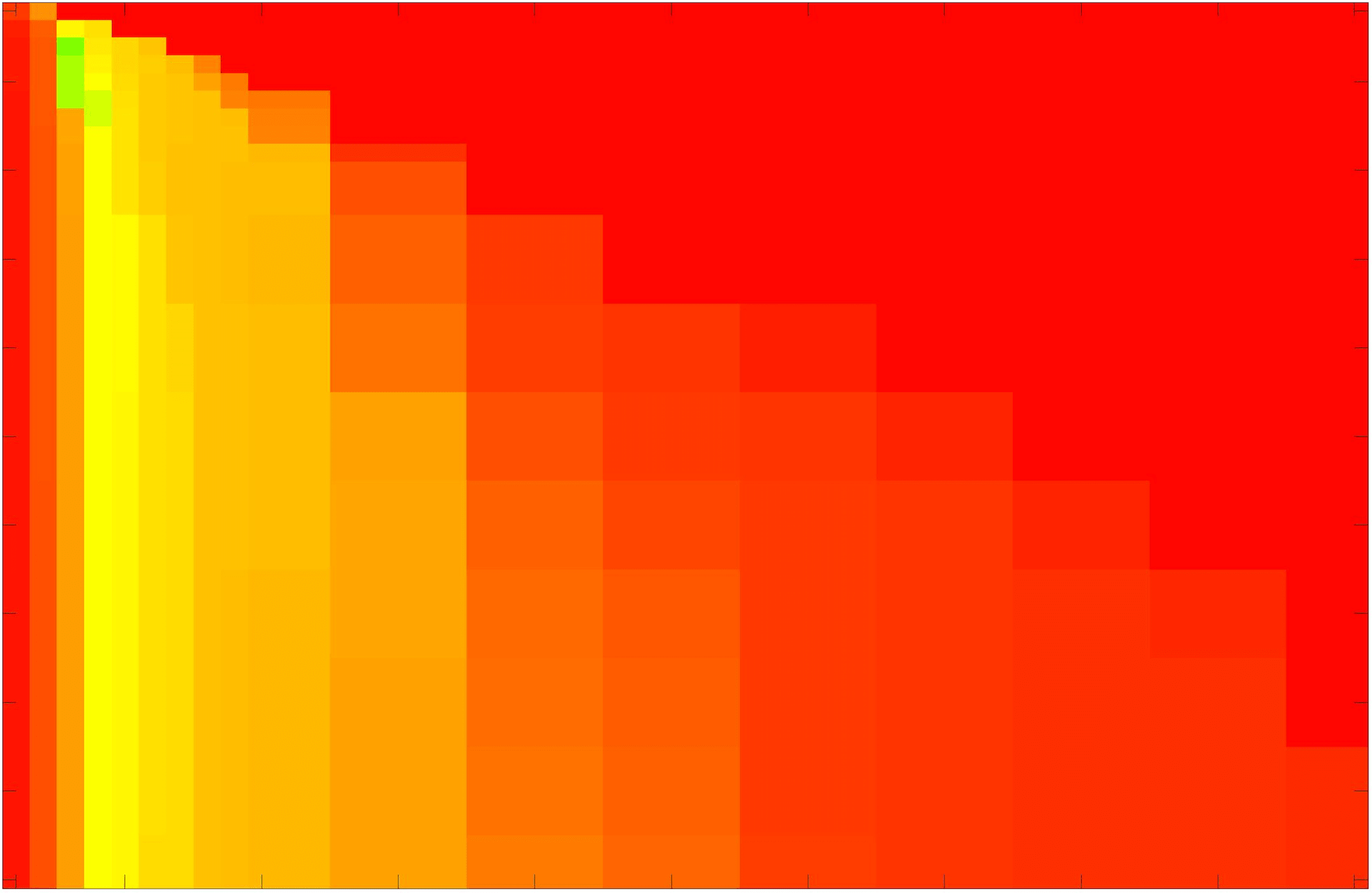}}\quad
\subfloat{\includegraphics[width=1.6in,height=1.1in]{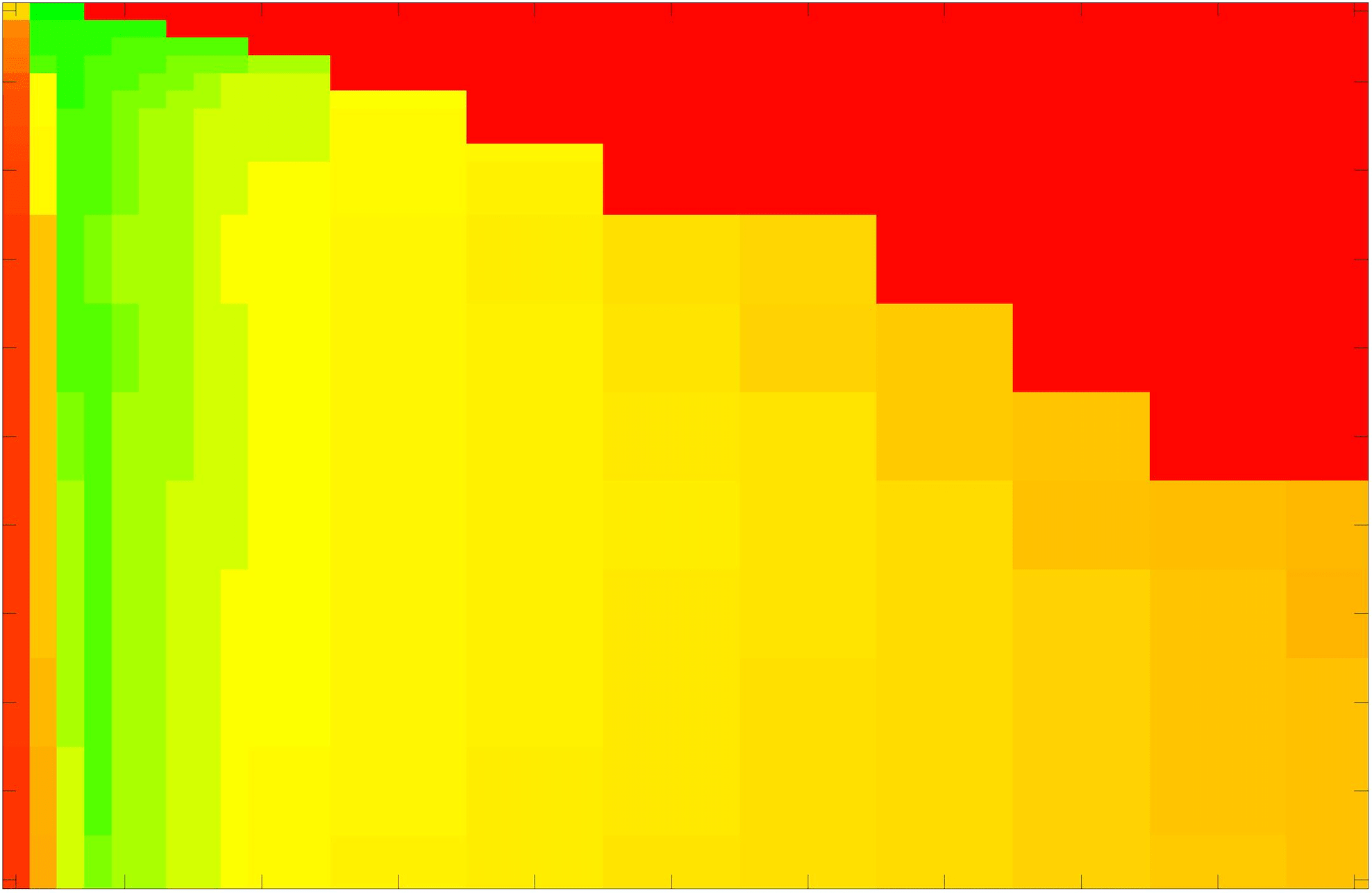}}} 
\centering
\floatbox[{\capbeside\thisfloatsetup{capbesideposition={left,center},capbesidewidth=1.5in,font =normalsize}}]{figure}[\FBwidth]
{\caption*{HYB \cite{Ali:16}}}
{\subfloat{\includegraphics[width=1.6in,height=1.1in]{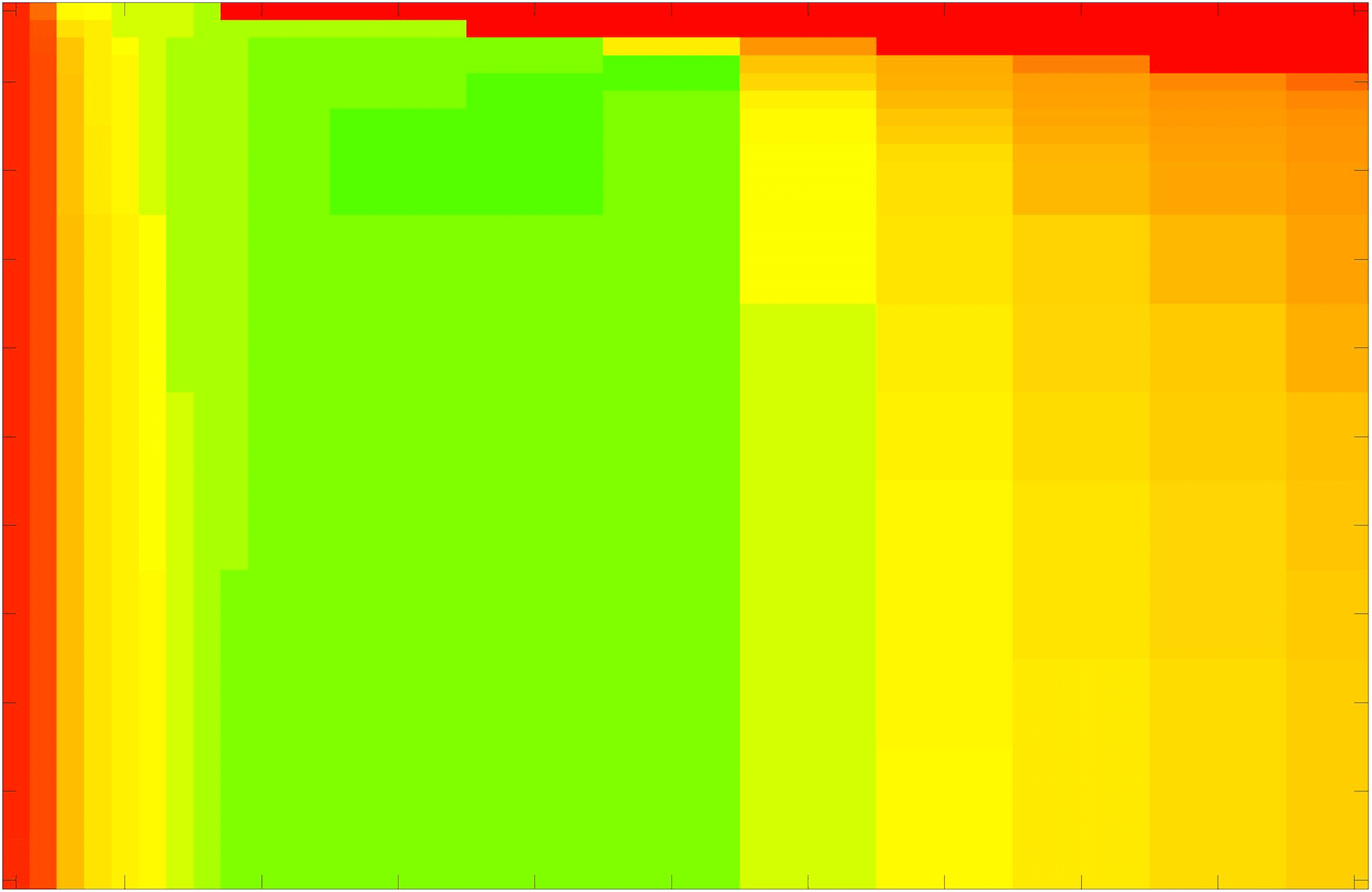}}\quad
\subfloat{\includegraphics[width=1.6in,height=1.1in]{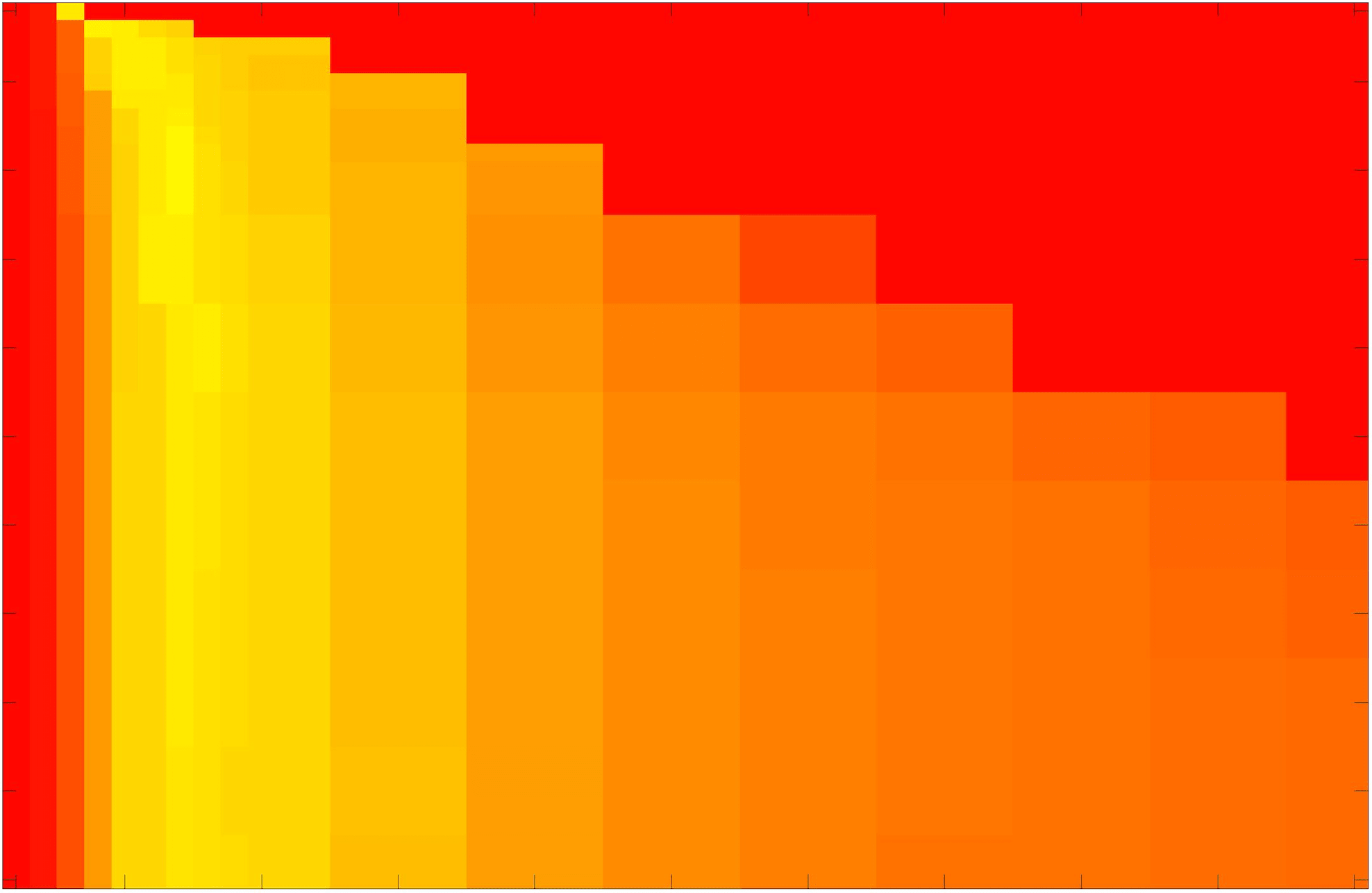}}\quad
\subfloat{\includegraphics[width=1.6in,height=1.1in]{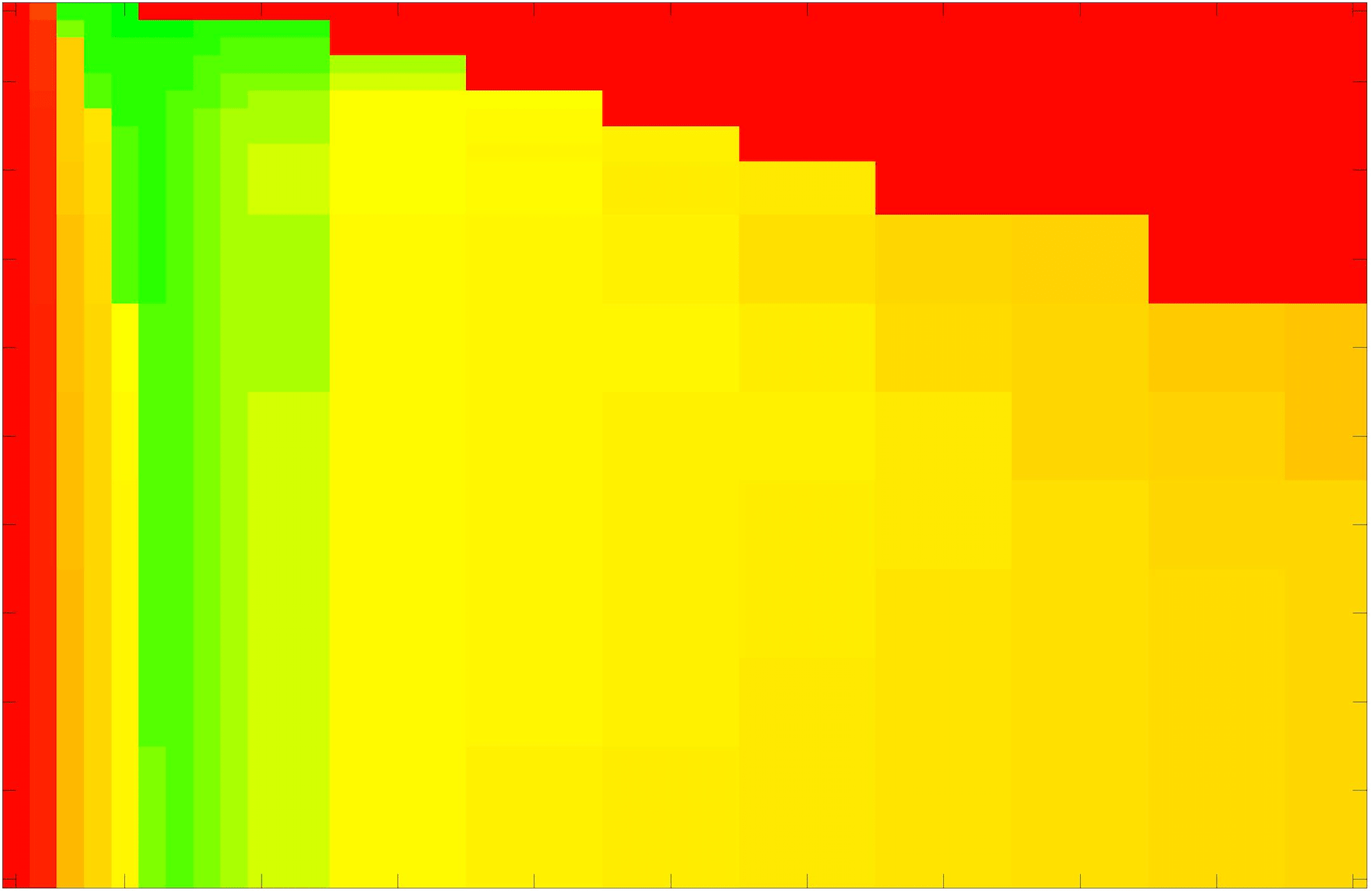}}}
\centering
\floatbox[{\capbeside\thisfloatsetup{capbesideposition={left,center},capbesidewidth=1.5in,font =normalsize}}]{figure}[\FBwidth]
{\caption*{GAV \cite{Ali:17}}}
{\subfloat{\includegraphics[width=1.6in,height=1.1in]{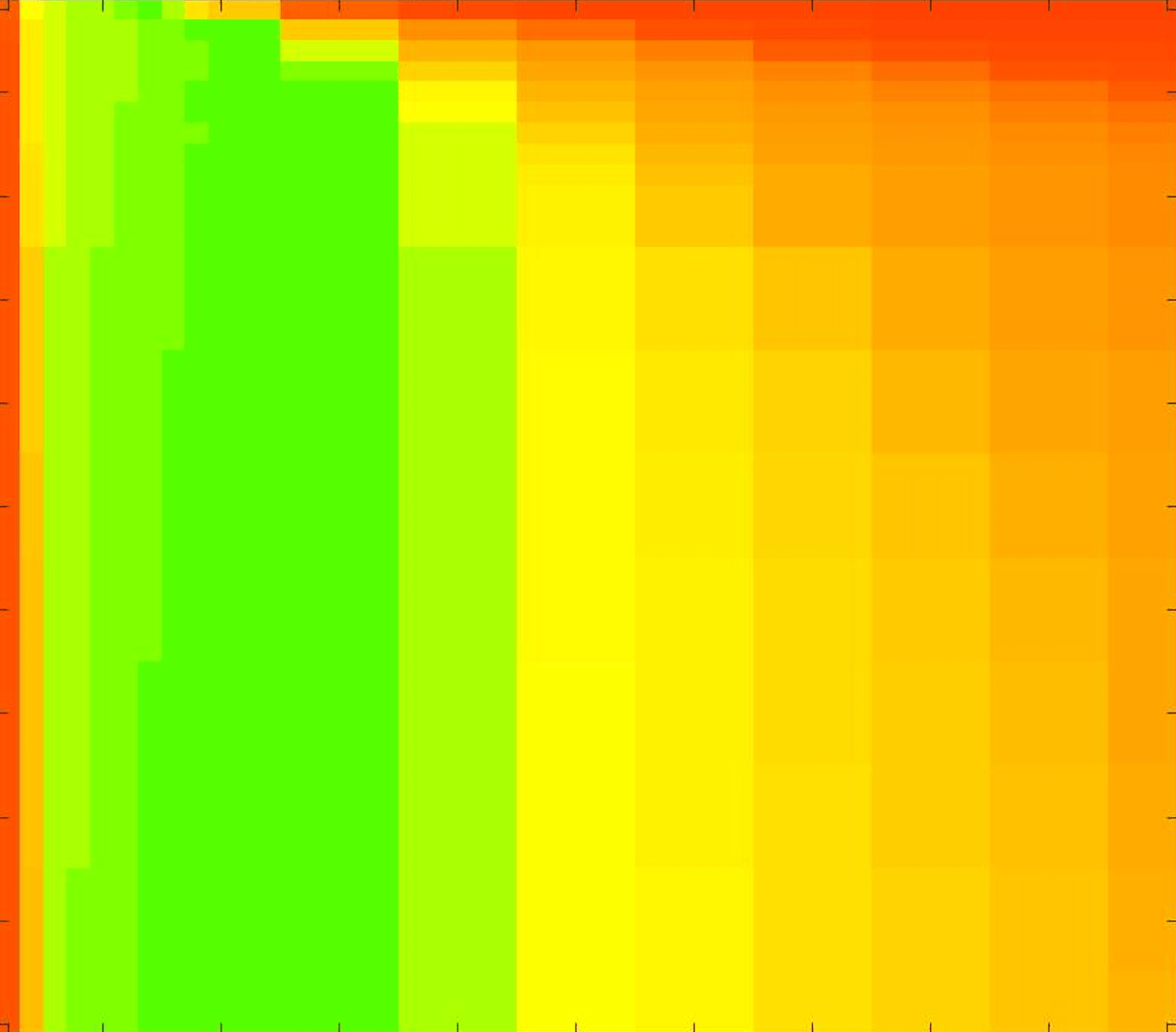}}\quad
\subfloat{\includegraphics[width=1.6in,height=1.1in]{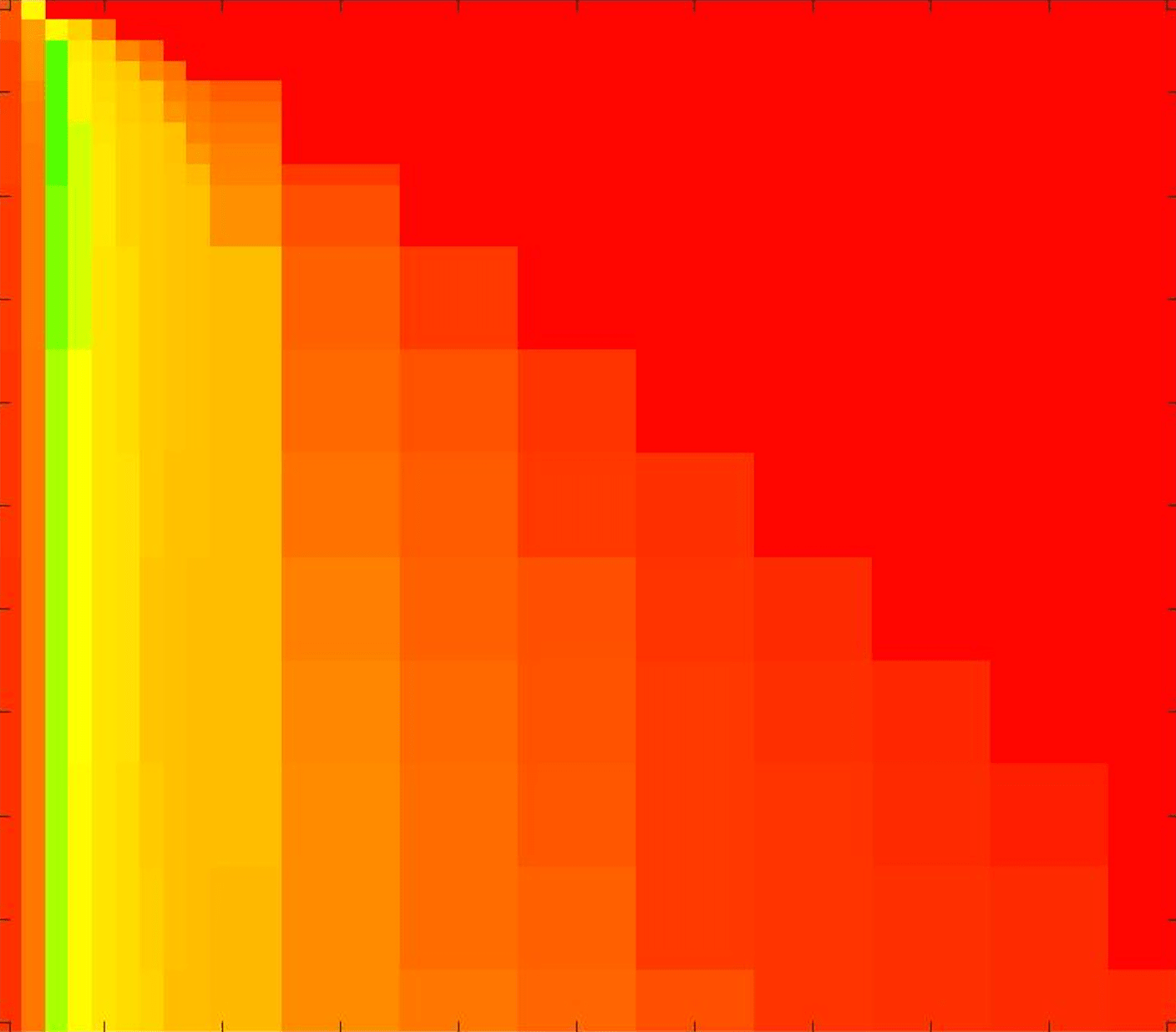}}\quad
\subfloat{\includegraphics[width=1.6in,height=1.1in]{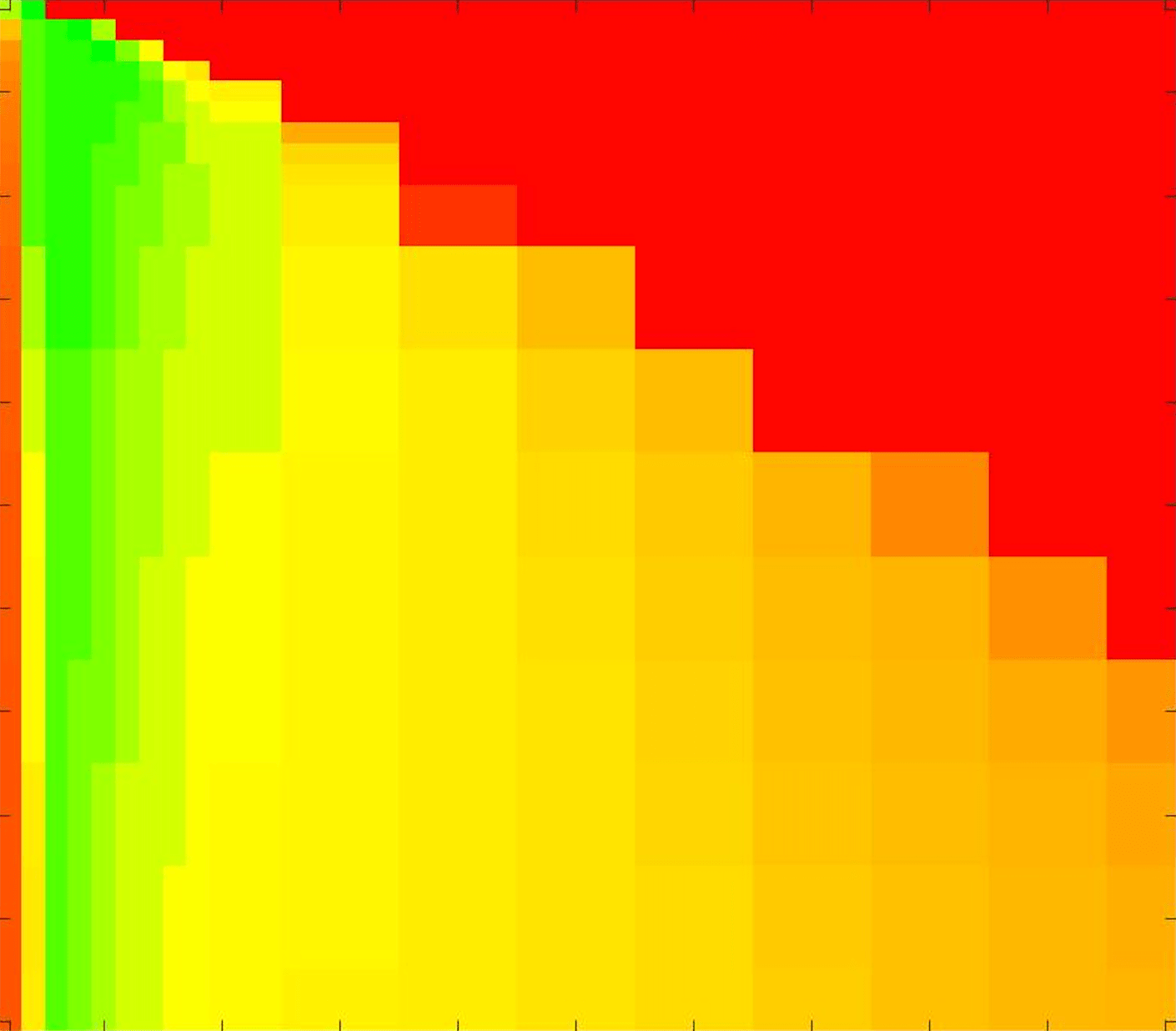}}}
\centering
\floatbox[{\capbeside\thisfloatsetup{capbesideposition={left,center},capbesidewidth=1.5in,font =normalsize}}]{figure}[\FBwidth]
{\caption*{Proposed}}
{\subfloat{\includegraphics[width=1.6in,height=1.1in]{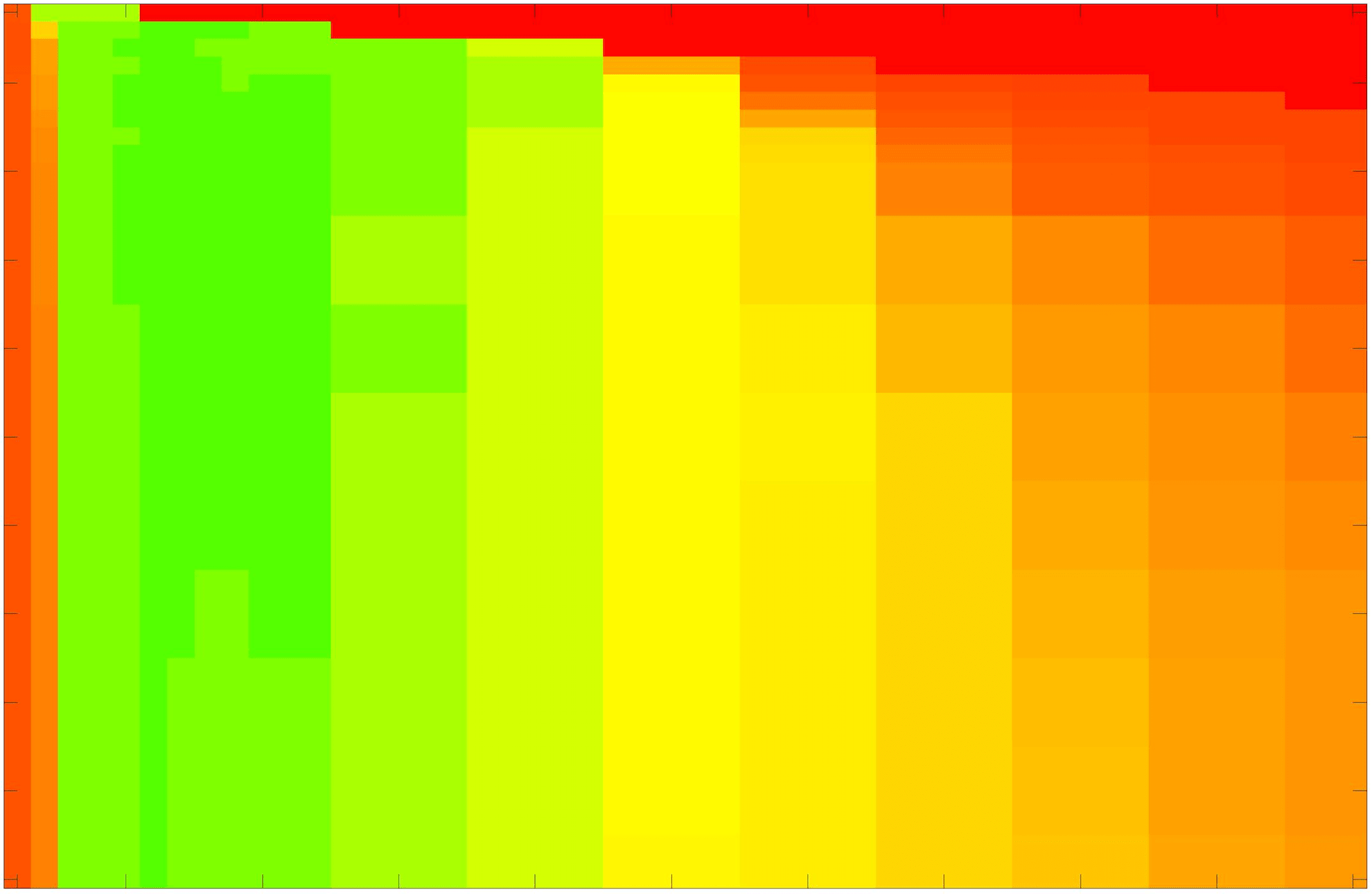}}\quad
\subfloat{\includegraphics[width=1.6in,height=1.1in]{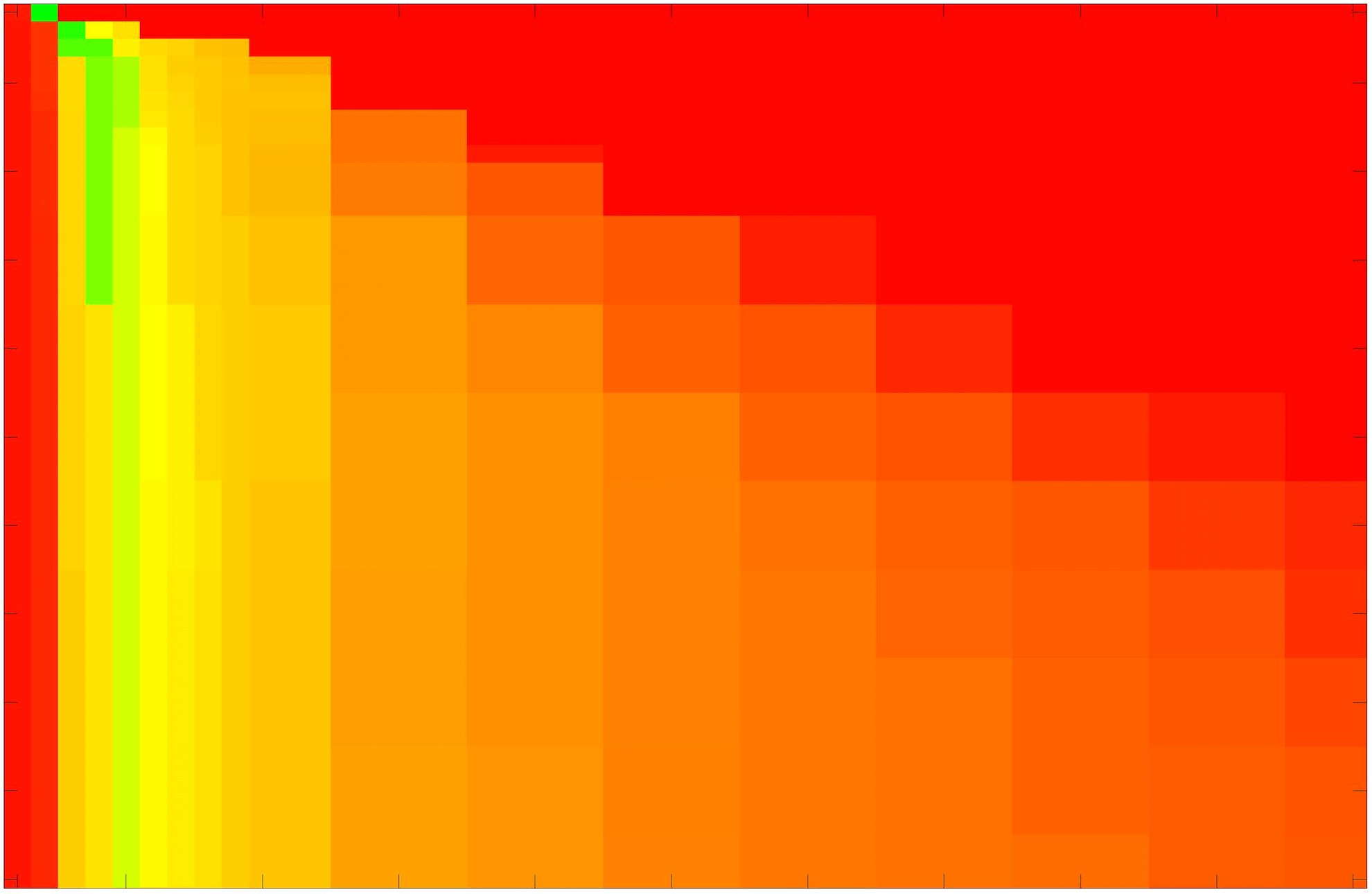}}\quad
\subfloat{\includegraphics[width=1.6in,height=1.1in]{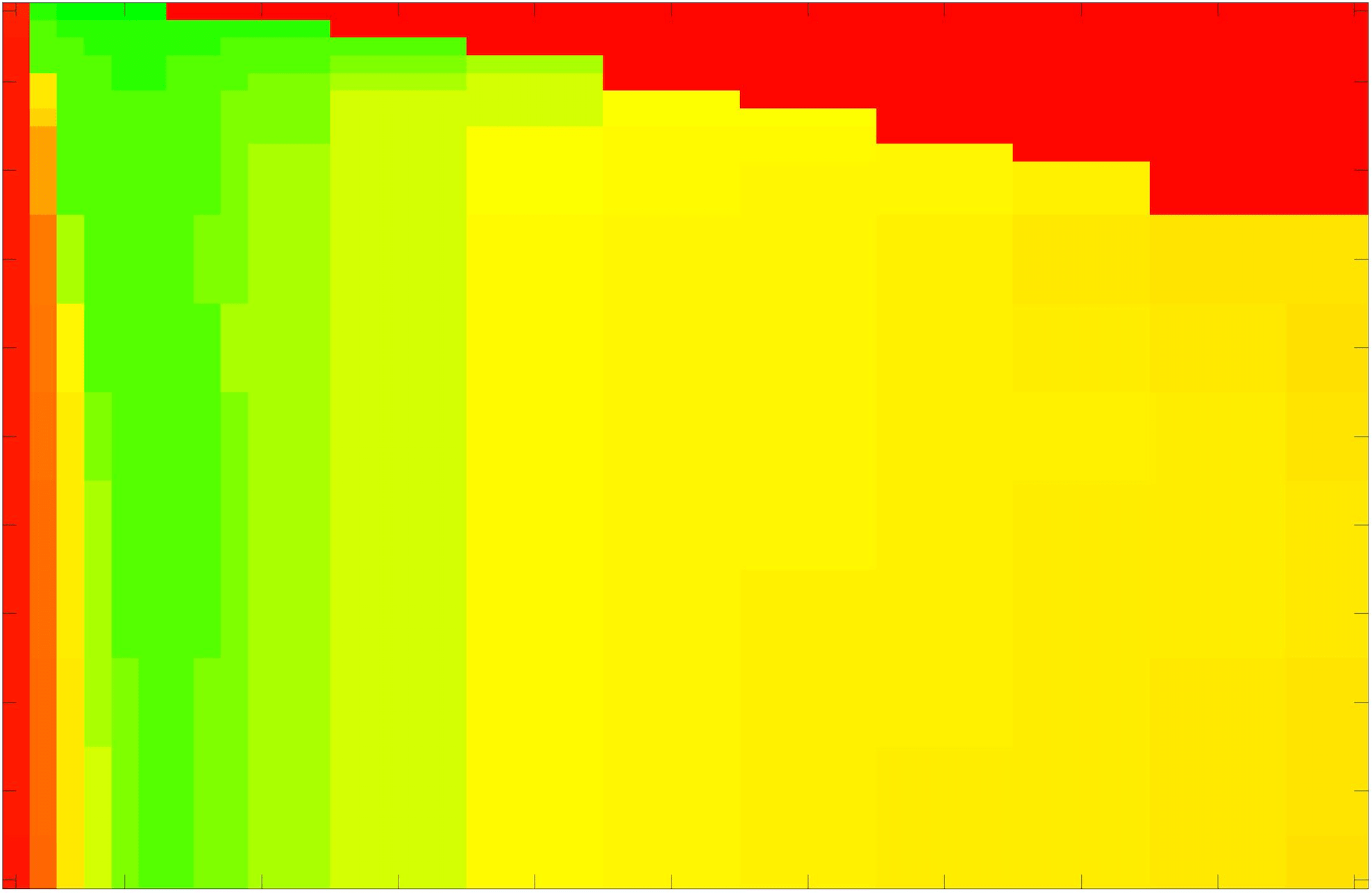}}} 
\end{figure*}

\begin{figure*}
\centering
\floatbox[{\capbeside\thisfloatsetup{capbesideposition={left,top},capbesidewidth=1.5in}}]{figure}[\FBwidth]
{\captionsetup{position=top}\captionsetup[subfigure]{labelformat=empty,font=normalsize} 
\subfloat[Test Image 7]{\includegraphics[width=1.6in,height = 1.4in]{figs/GT_Lung-min}}\quad
\captionsetup{position=top}\captionsetup[subfigure]{labelformat=empty,font=normalsize} 
\subfloat[Test Image 8]{\includegraphics[width=1.6in,height = 1.4in]{figs/GT_Lung33-min}}\quad
\captionsetup{position=top}\captionsetup[subfigure]{labelformat=empty,font=normalsize} 
\subfloat[Test Image 9]{\includegraphics[width=1.6in,height = 1.4in]{figs/GT_Lung31-min}}}
{\caption{Heatmaps of TC values for permutations of $\tilde{\lambda}$ and $\theta$. Each row and column is labelled according to the model used and the image tested. The colour is consistent with the scale in Fig.~\ref{fig:colorbar}. Here, we present Test Images 7 -- 9. \label{fig:heat3}}}\vspace{-0.1in}
\centering
\floatbox[{\capbeside\thisfloatsetup{capbesideposition={left,center},capbesidewidth=1.5in,font =normalsize}}]{figure}[\FBwidth]
{\caption*{CV \cite{ACWE}}}
{\subfloat{\includegraphics[width=1.6in,height=1.1in]{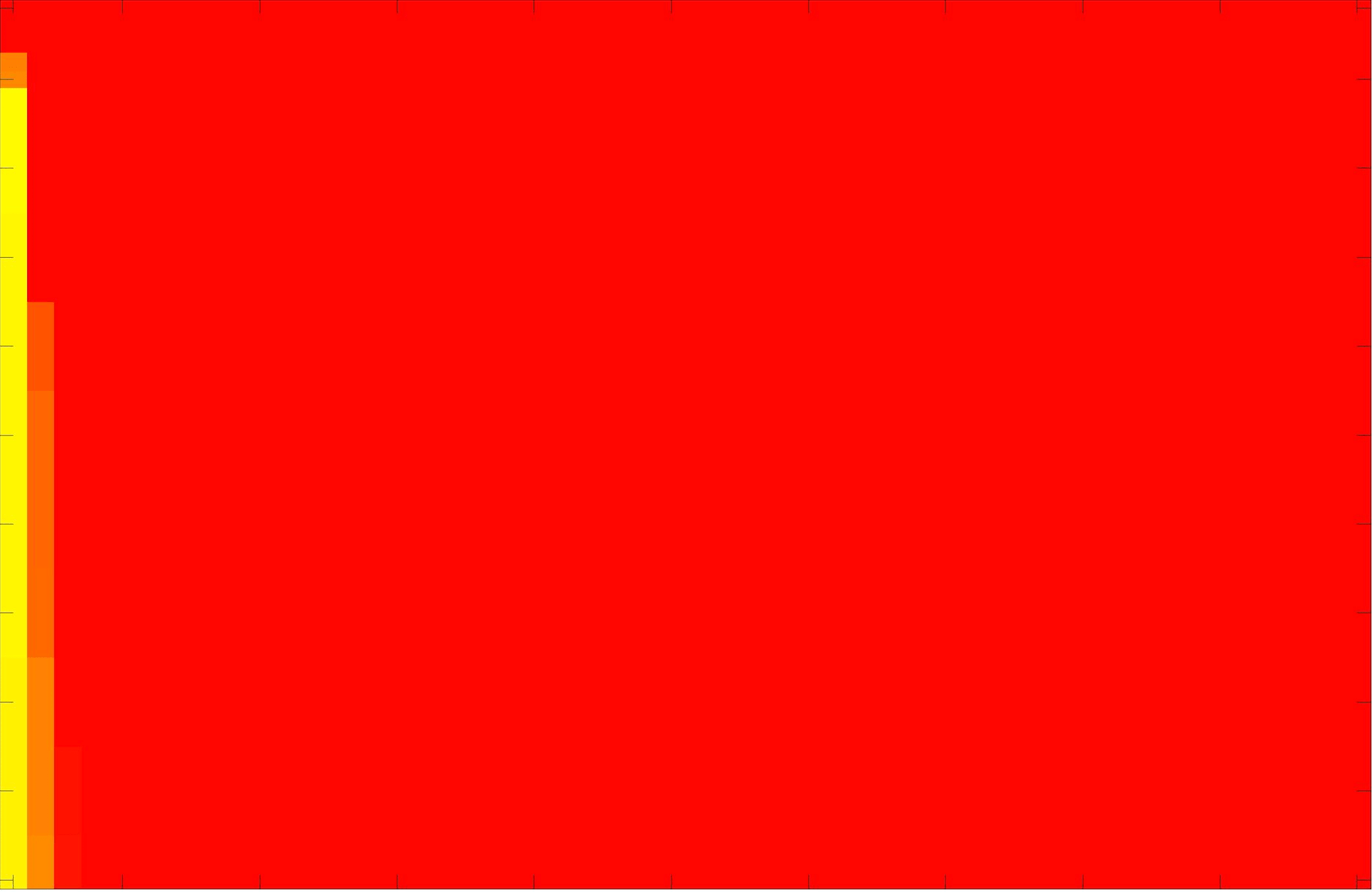}}\quad
\subfloat{\includegraphics[width=1.6in,height=1.1in]{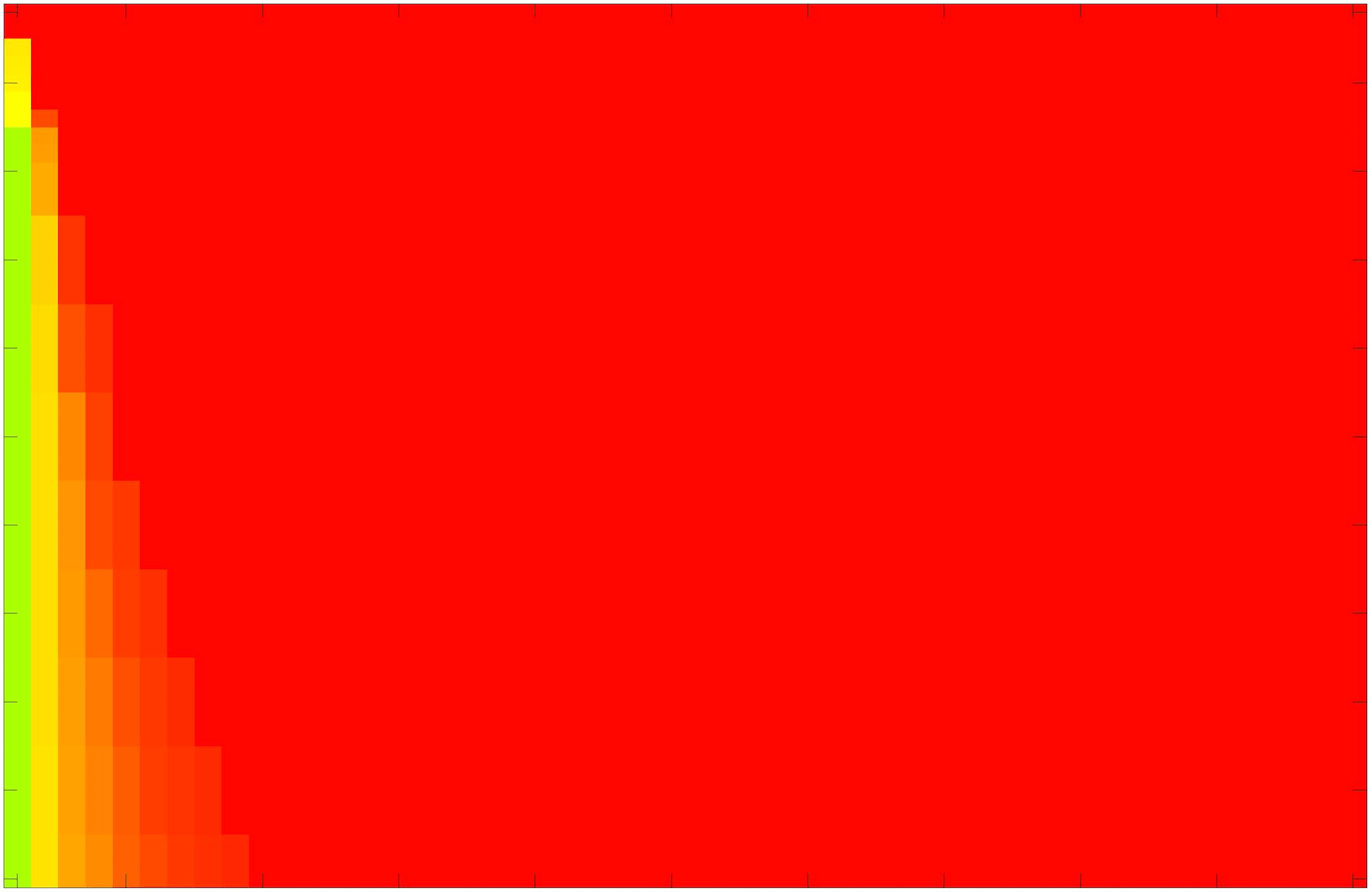}}\quad
\subfloat{\includegraphics[width=1.6in,height=1.1in]{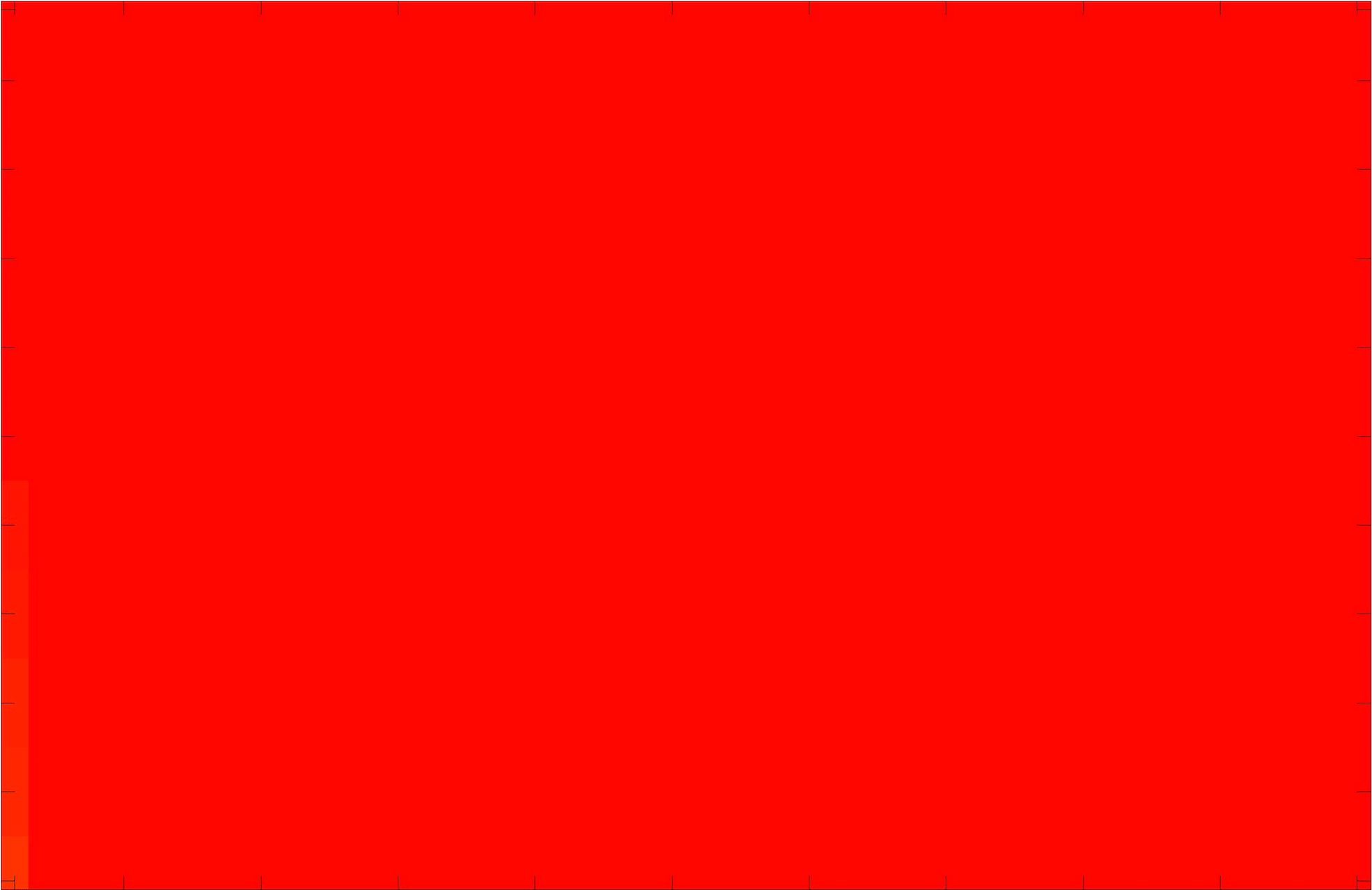}}}
\centering
\floatbox[{\capbeside\thisfloatsetup{capbesideposition={left,center},capbesidewidth=1.5in,font =normalsize}}]{figure}[\FBwidth]
{\caption*{RSF \cite{RSF}}}
{\subfloat{\includegraphics[width=1.6in,height=1.1in]{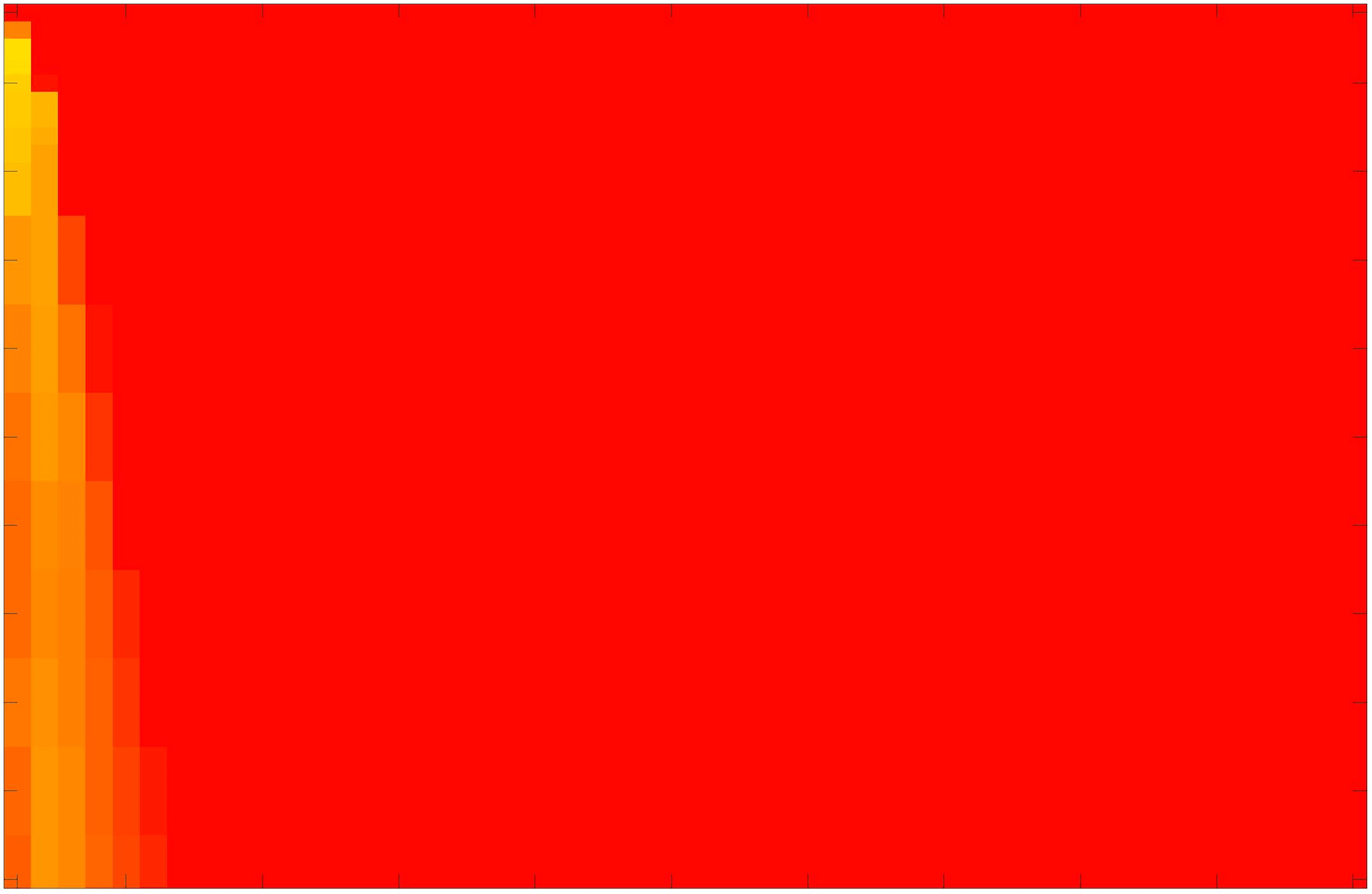}}\quad
\subfloat{\includegraphics[width=1.6in,height=1.1in]{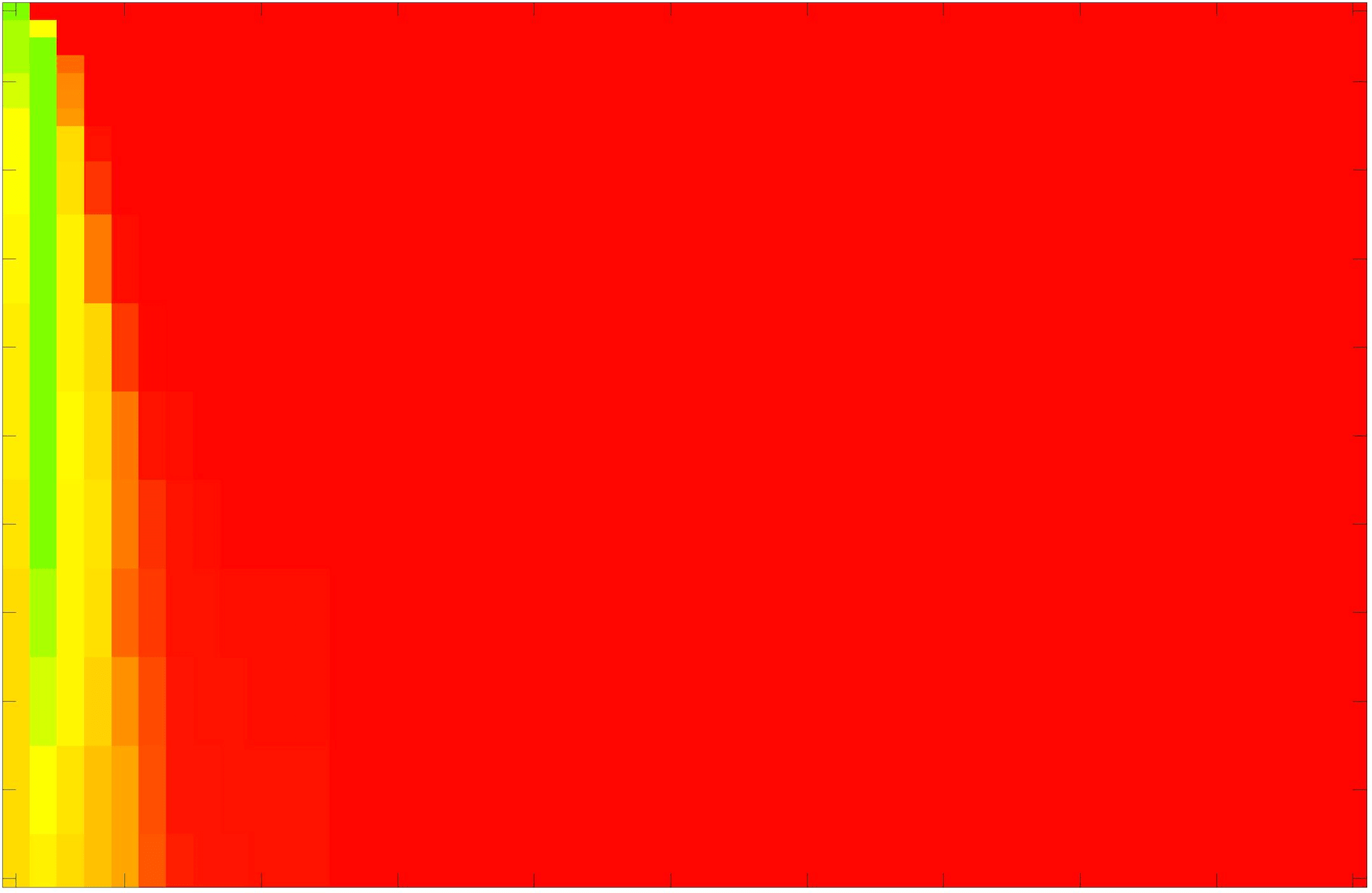}}\quad
\subfloat{\includegraphics[width=1.6in,height=1.1in]{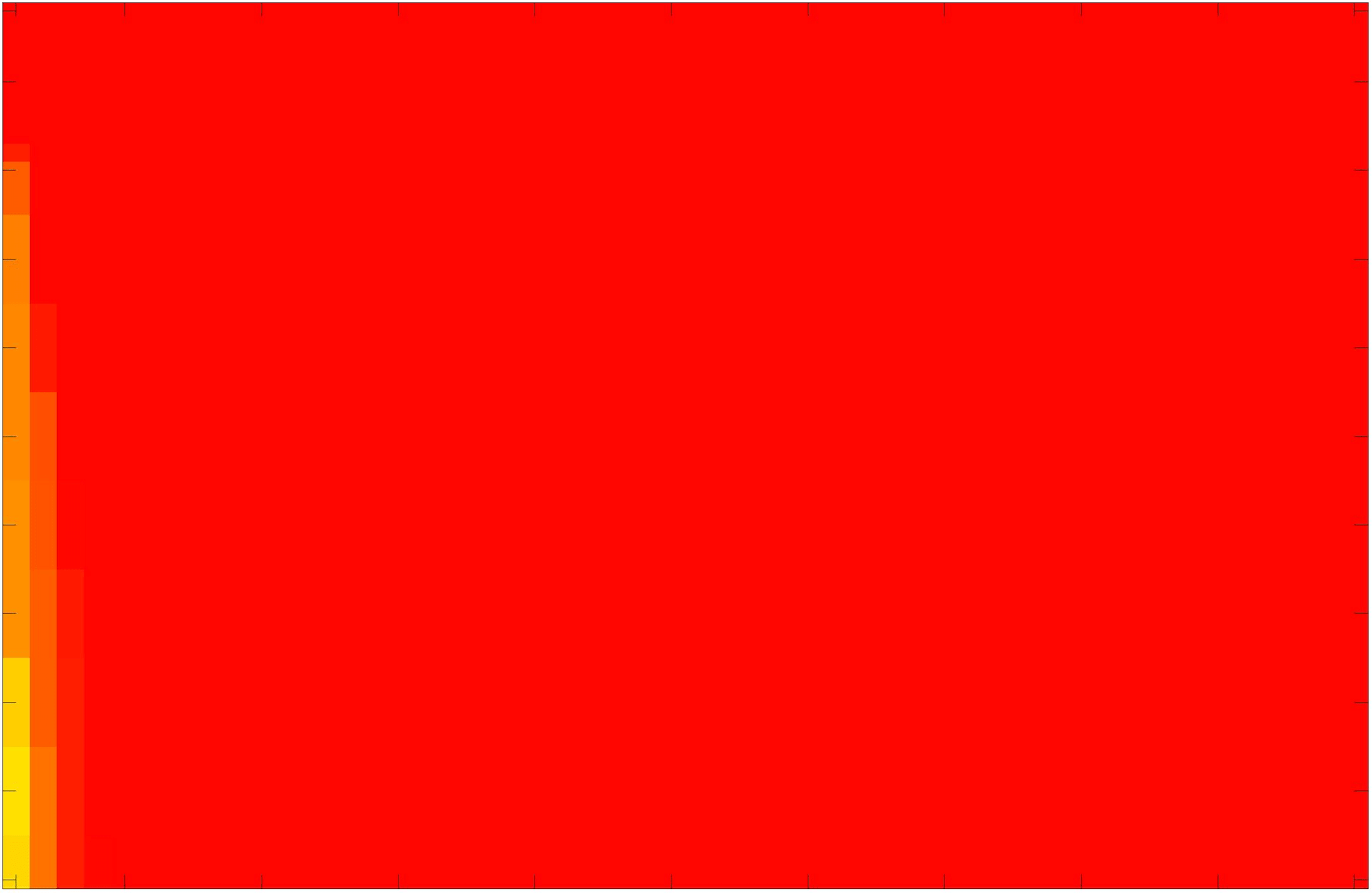}}}
\centering
\floatbox[{\capbeside\thisfloatsetup{capbesideposition={left,center},capbesidewidth=1.5in,font =normalsize}}]{figure}[\FBwidth]
{\caption*{LCV \cite{LCV}}}
{\subfloat{\includegraphics[width=1.6in,height=1.1in]{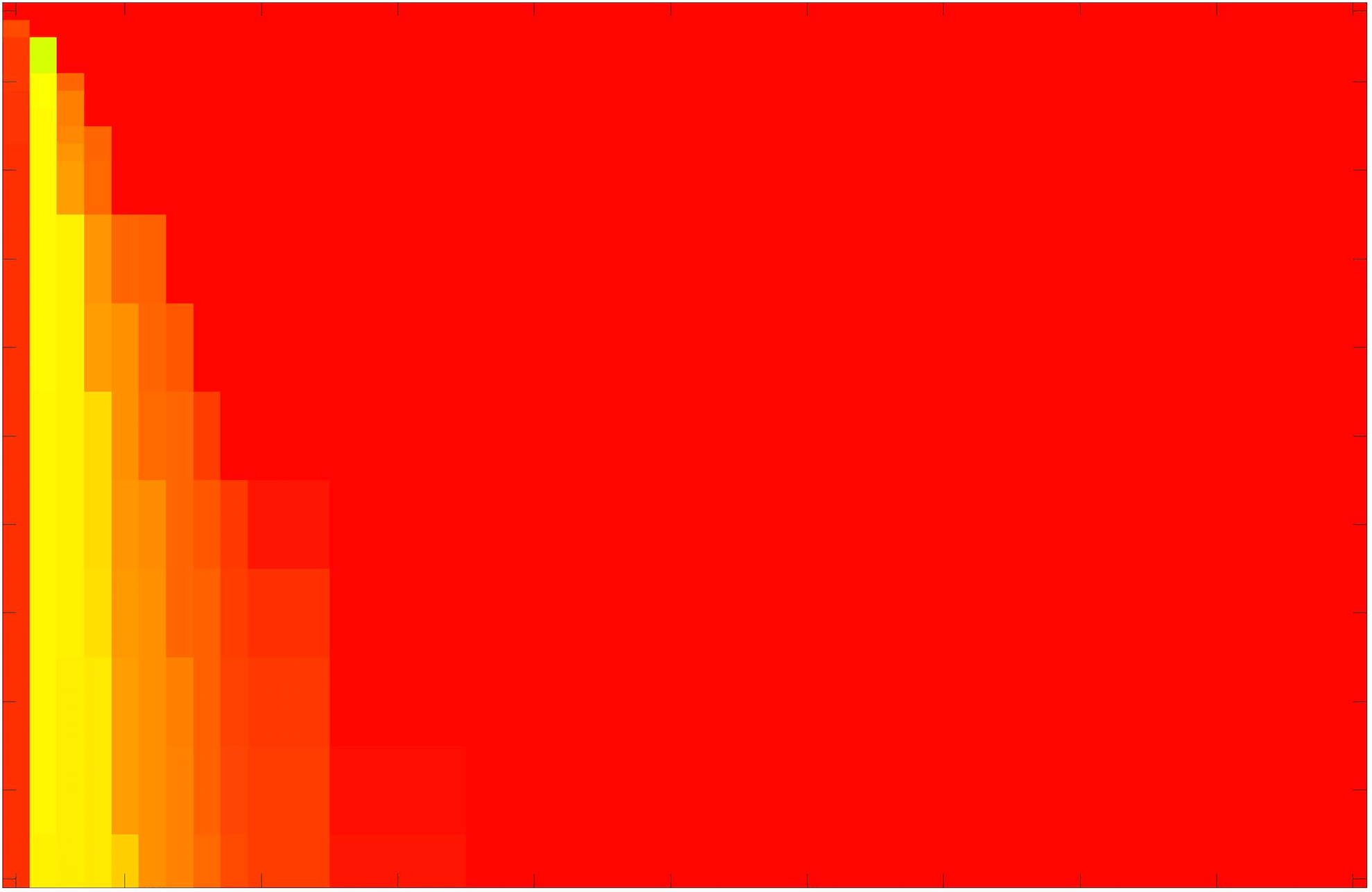}}\quad
\subfloat{\includegraphics[width=1.6in,height=1.1in]{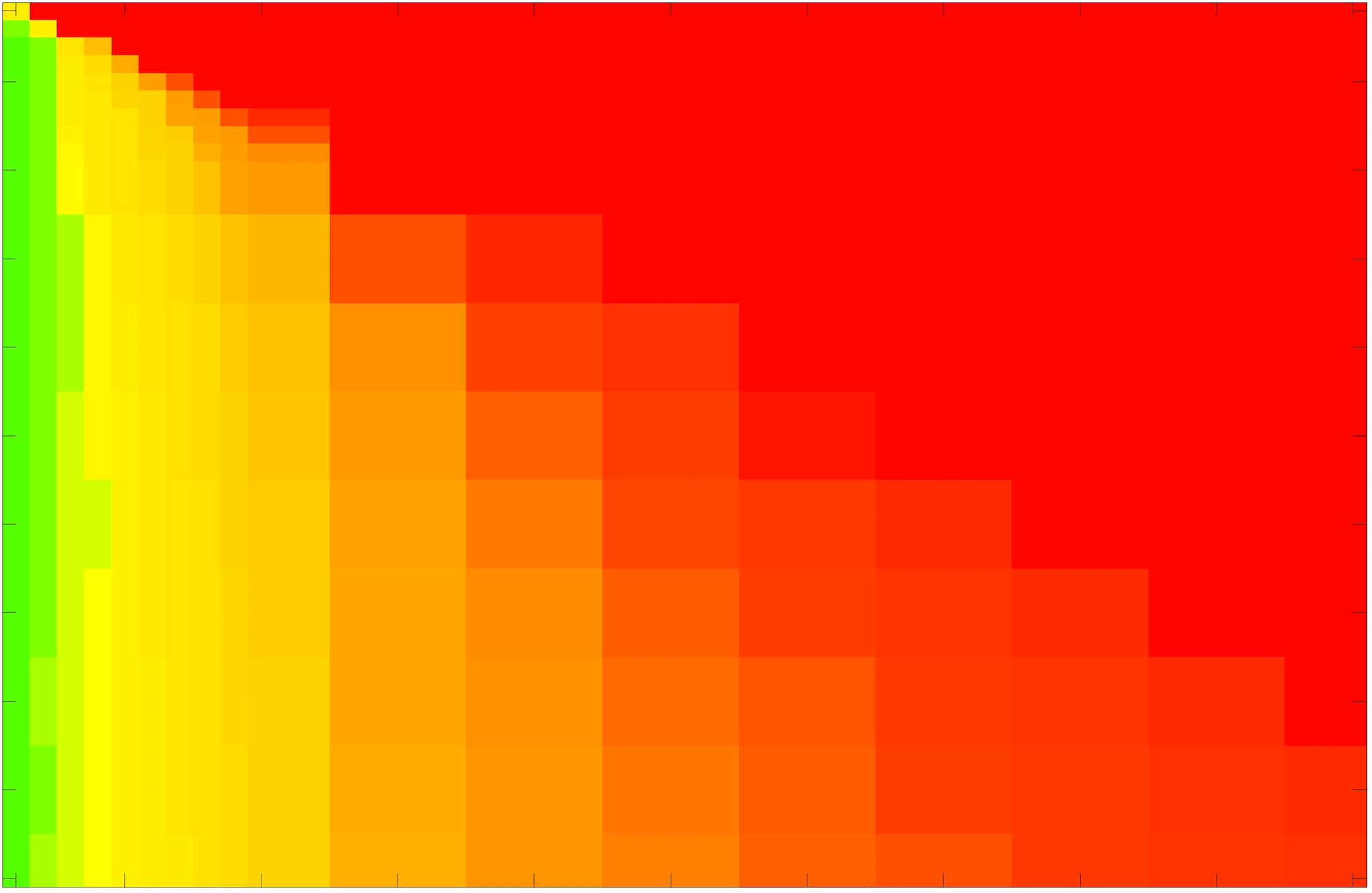}}\quad
\subfloat{\includegraphics[width=1.6in,height=1.1in]{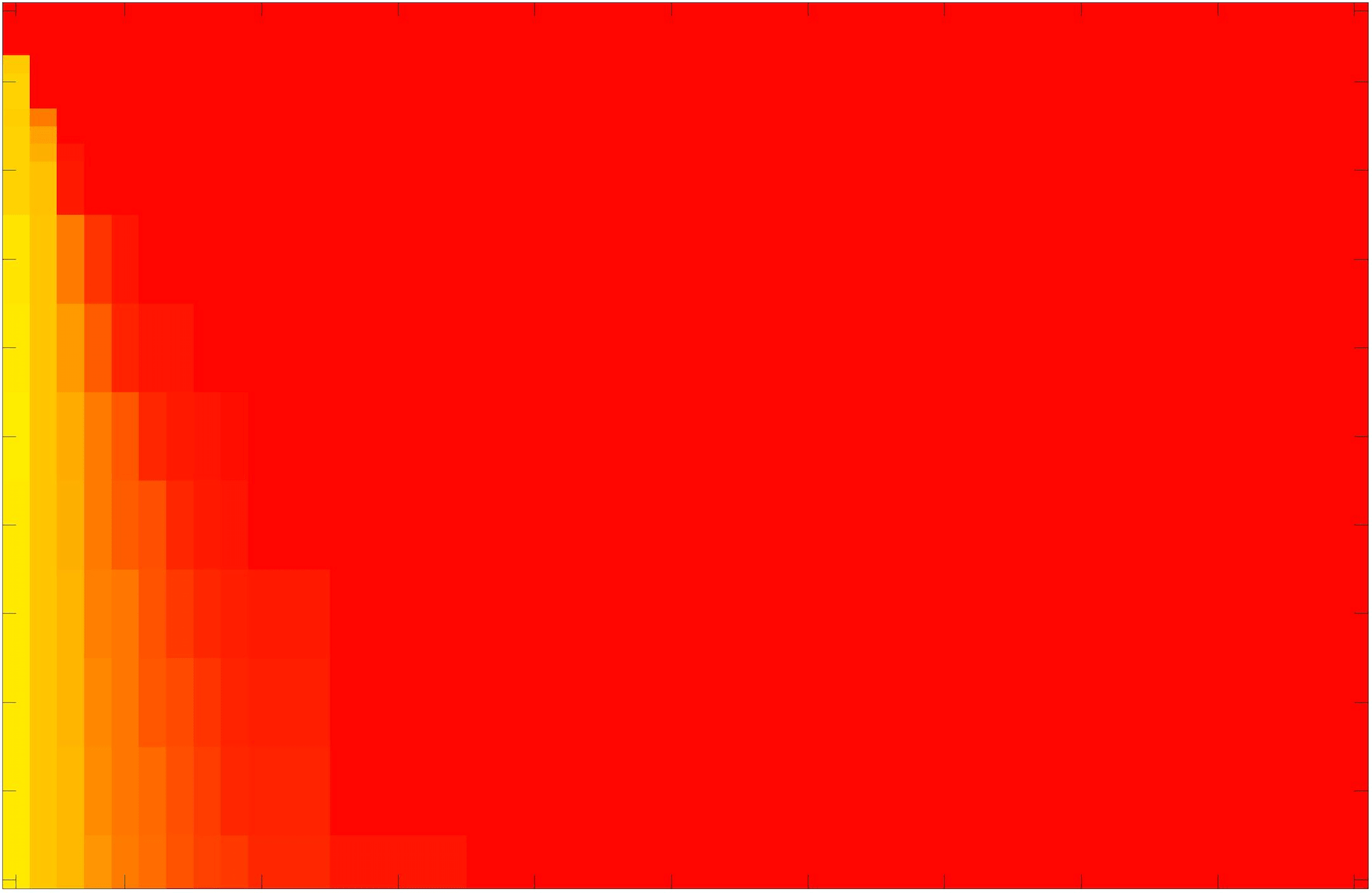}}} 
\centering
\floatbox[{\capbeside\thisfloatsetup{capbesideposition={left,center},capbesidewidth=1.5in,font =normalsize}}]{figure}[\FBwidth]
{\caption*{HYB \cite{Ali:16}}}
{\subfloat{\includegraphics[width=1.6in,height=1.1in]{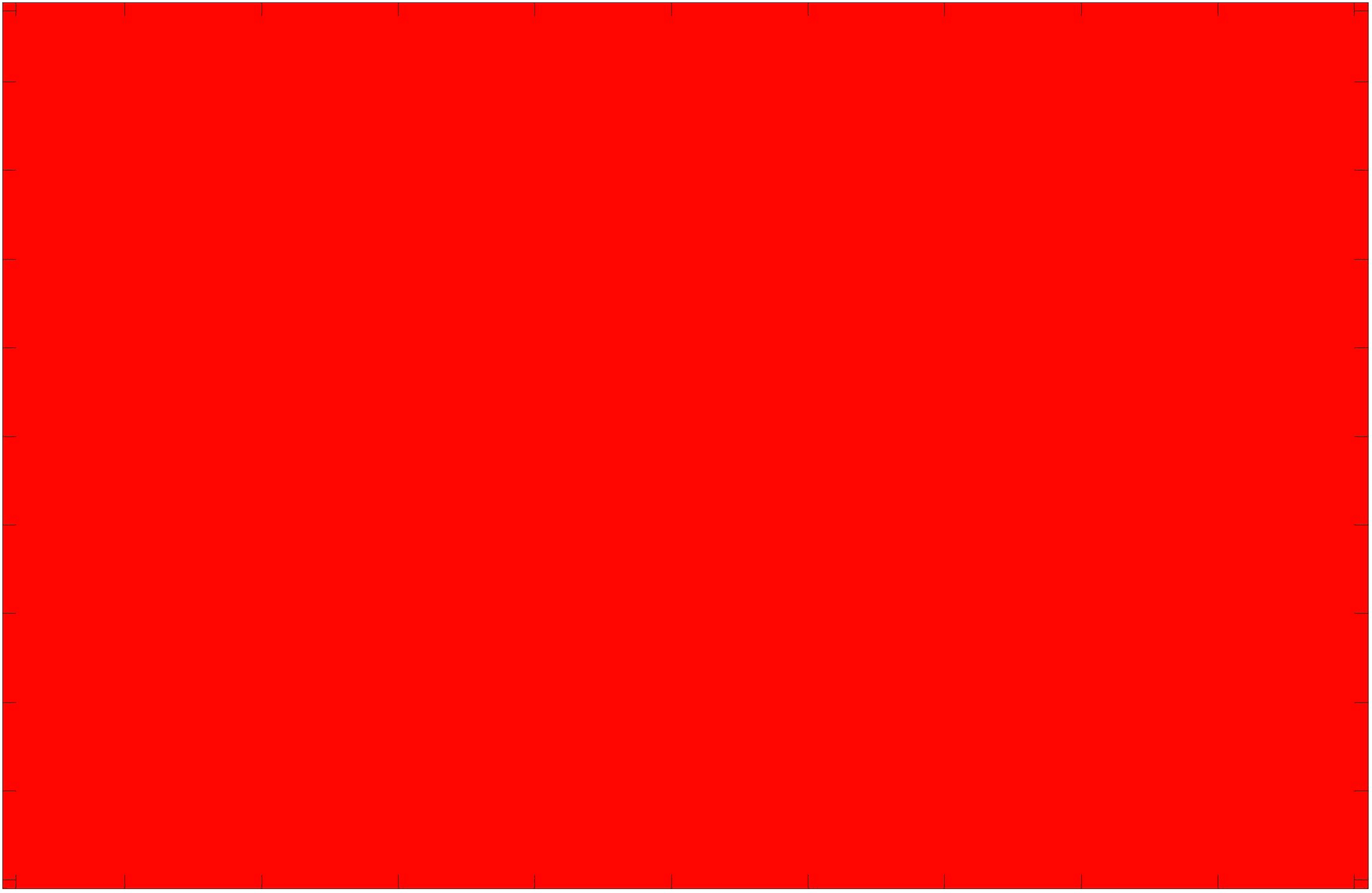}}\quad
\subfloat{\includegraphics[width=1.6in,height=1.1in]{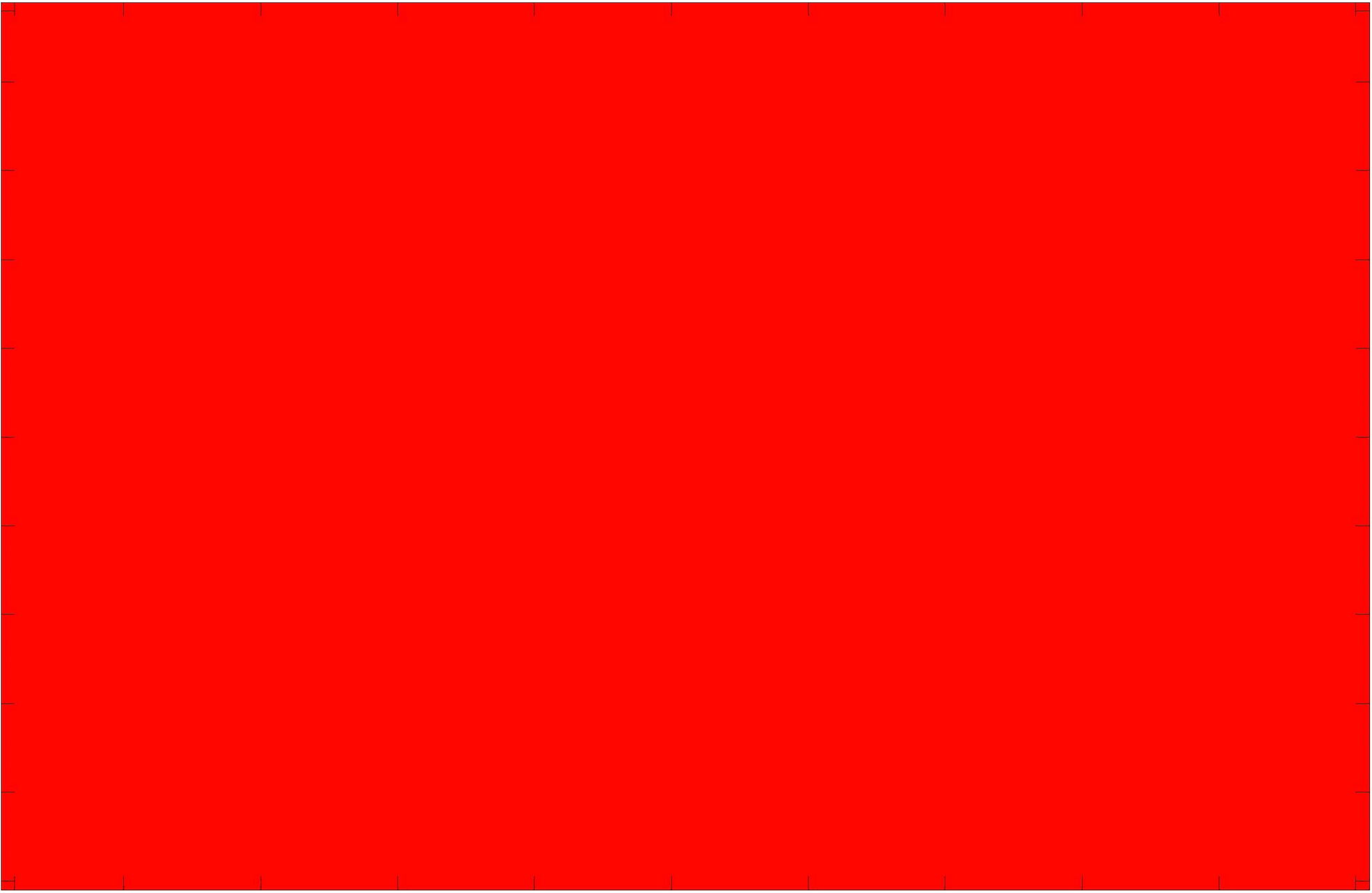}}\quad
\subfloat{\includegraphics[width=1.6in,height=1.1in]{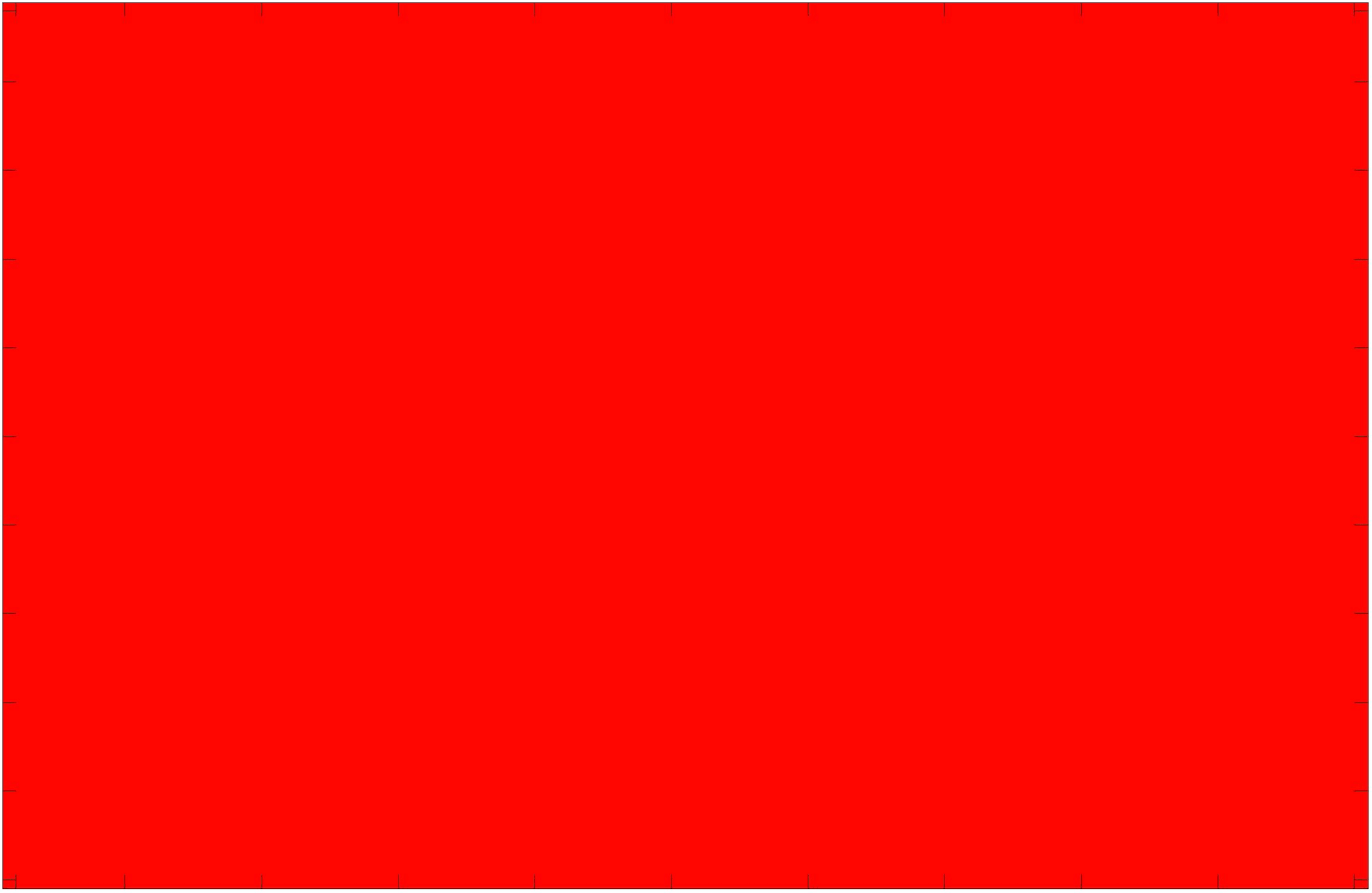}}}
\centering
\floatbox[{\capbeside\thisfloatsetup{capbesideposition={left,center},capbesidewidth=1.5in,font =normalsize}}]{figure}[\FBwidth]
{\caption*{GAV \cite{Ali:17}}}
{\subfloat{\includegraphics[width=1.6in,height=1.1in]{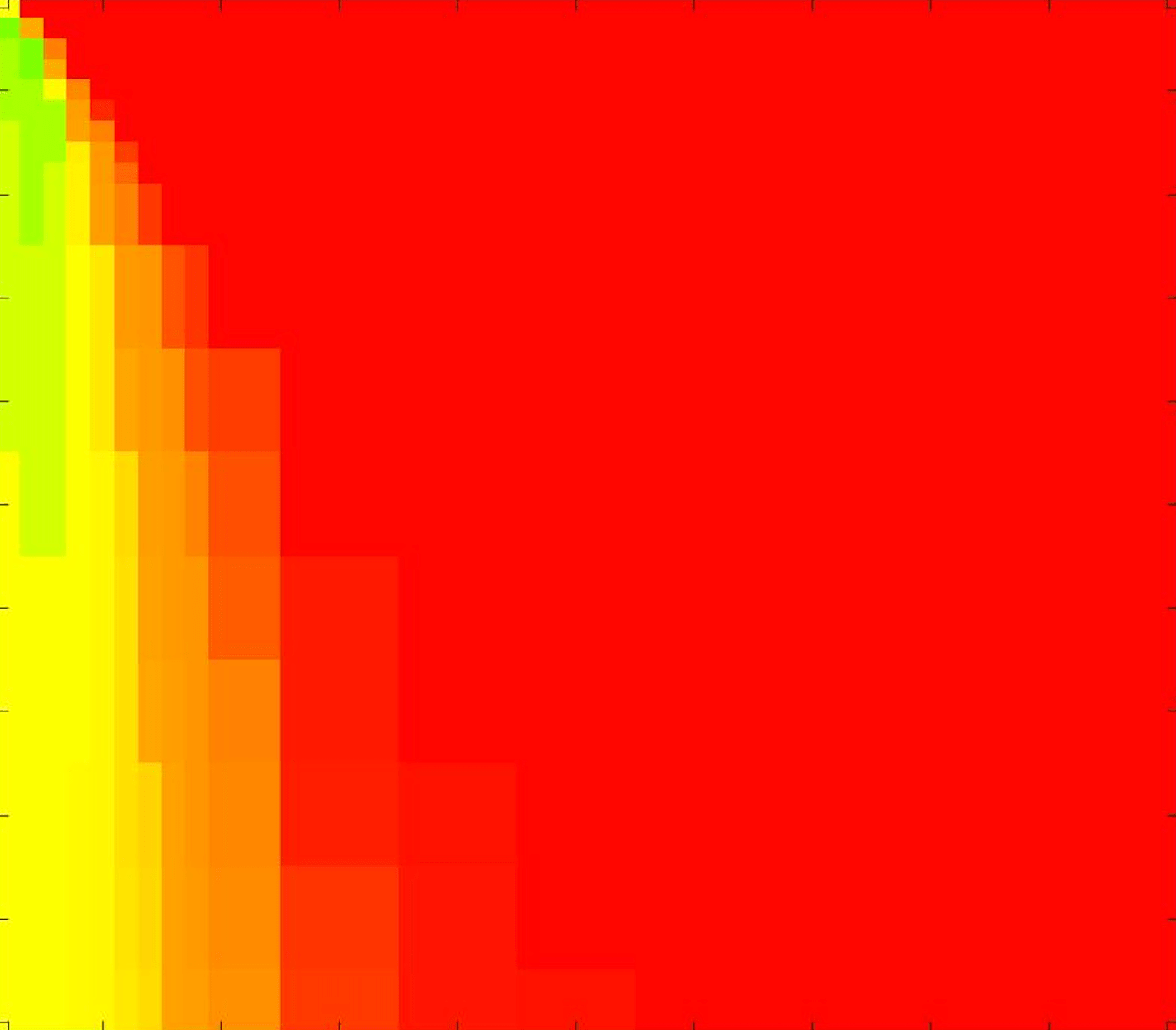}}\quad
\subfloat{\includegraphics[width=1.6in,height=1.1in]{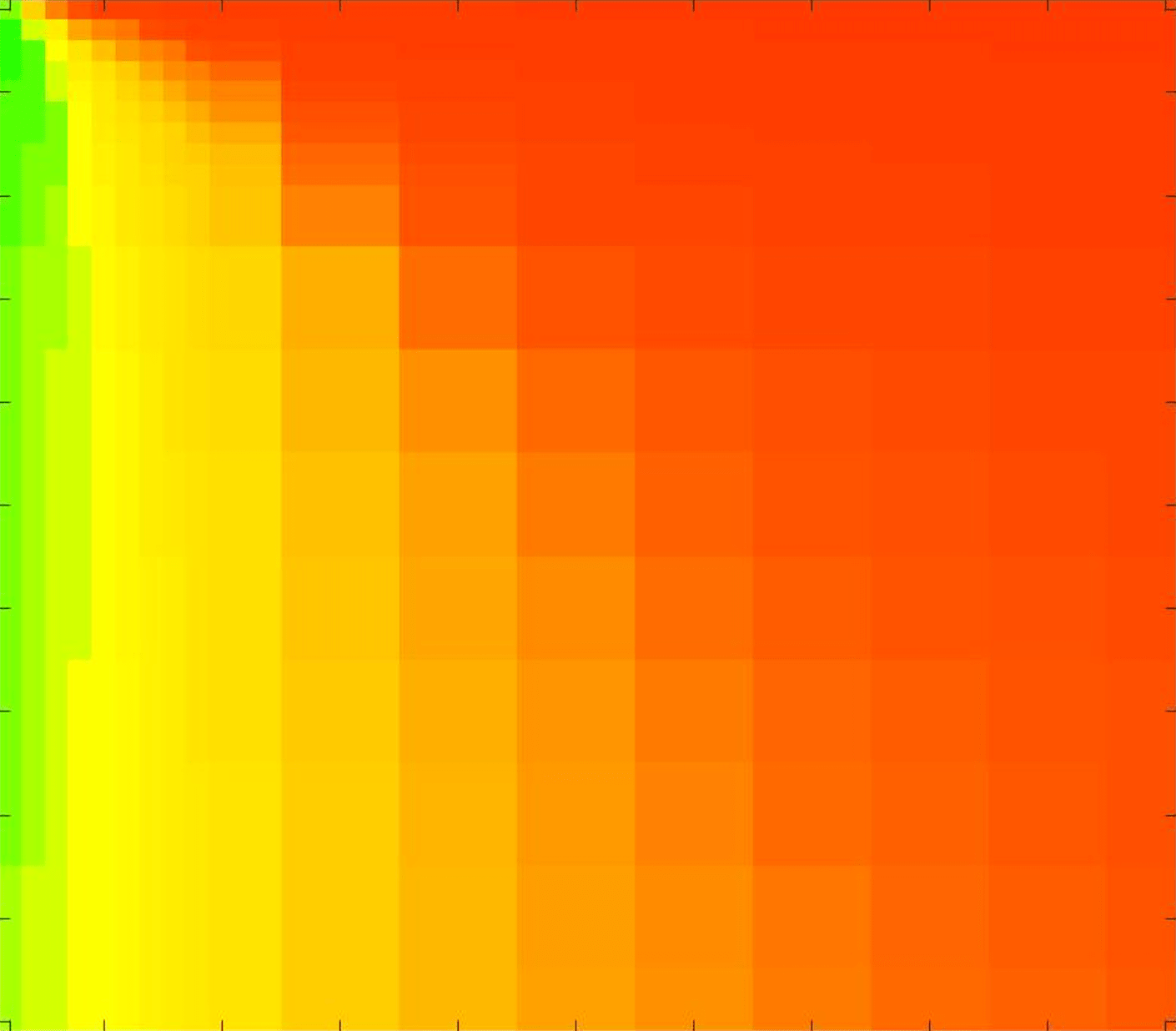}}\quad
\subfloat{\includegraphics[width=1.6in,height=1.1in]{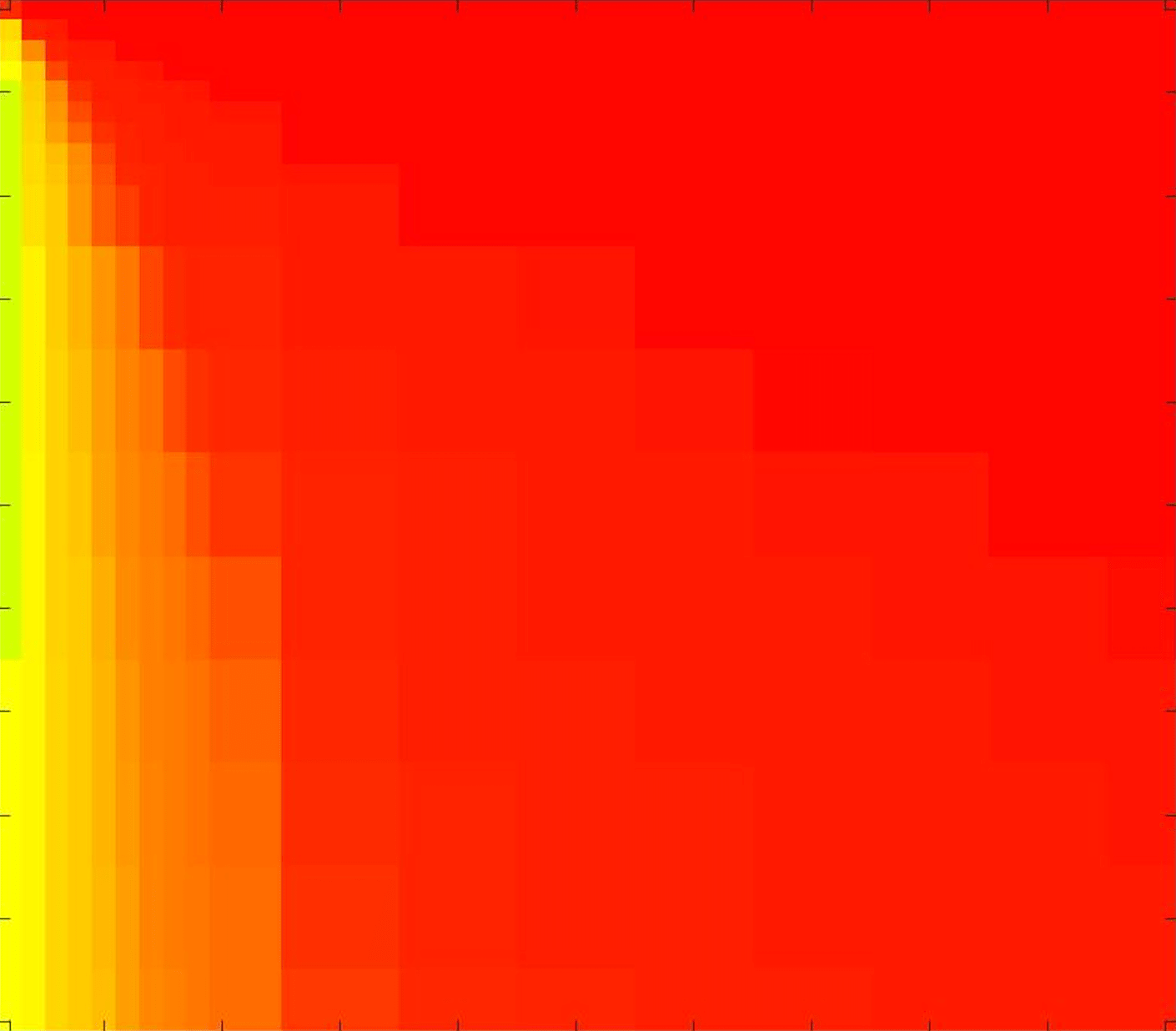}}}
\centering
\floatbox[{\capbeside\thisfloatsetup{capbesideposition={left,center},capbesidewidth=1.5in,font =normalsize}}]{figure}[\FBwidth]
{\caption*{Proposed}}
{\subfloat{\includegraphics[width=1.6in,height=1.1in]{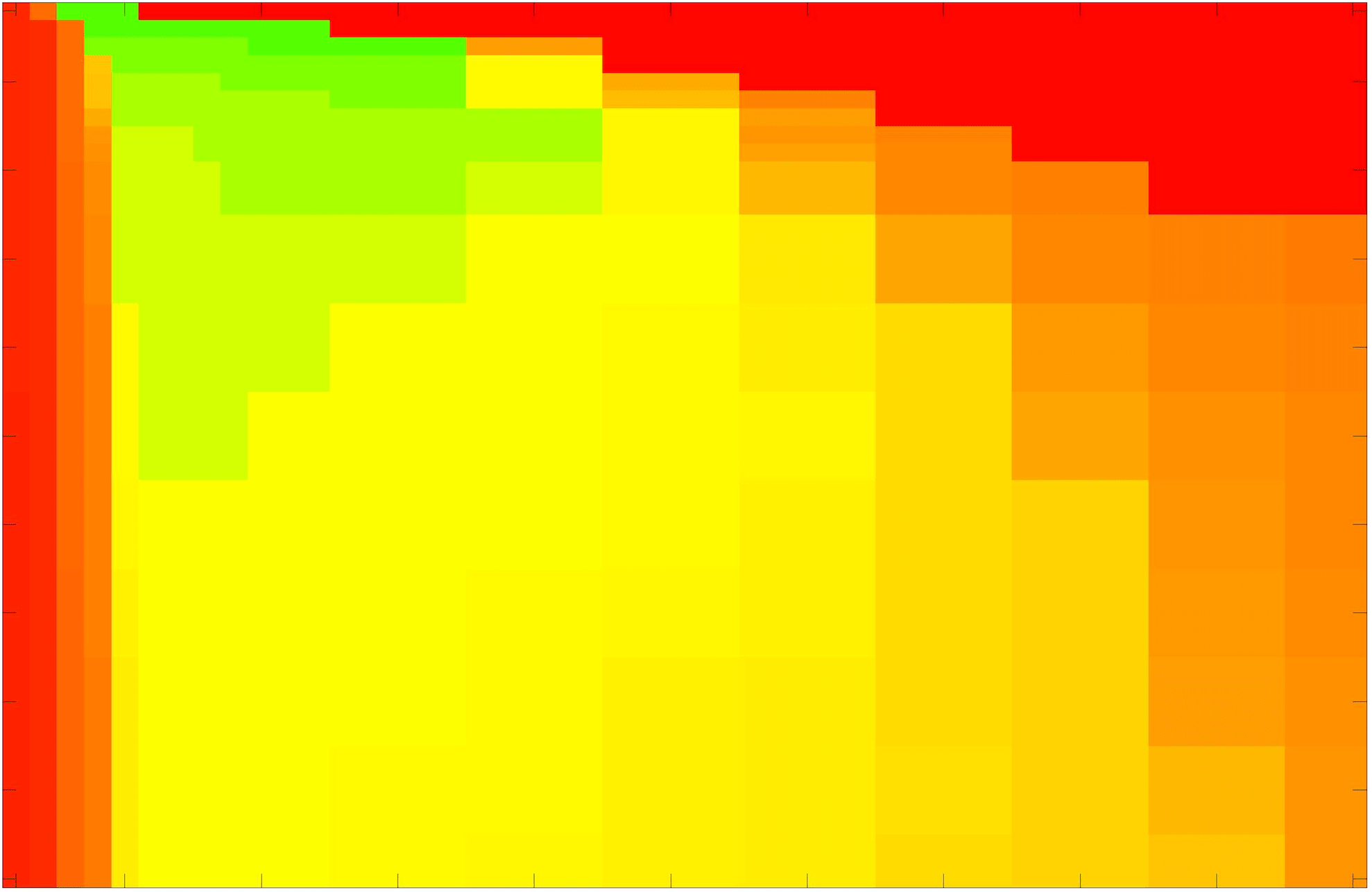}}\quad
\subfloat{\includegraphics[width=1.6in,height=1.1in]{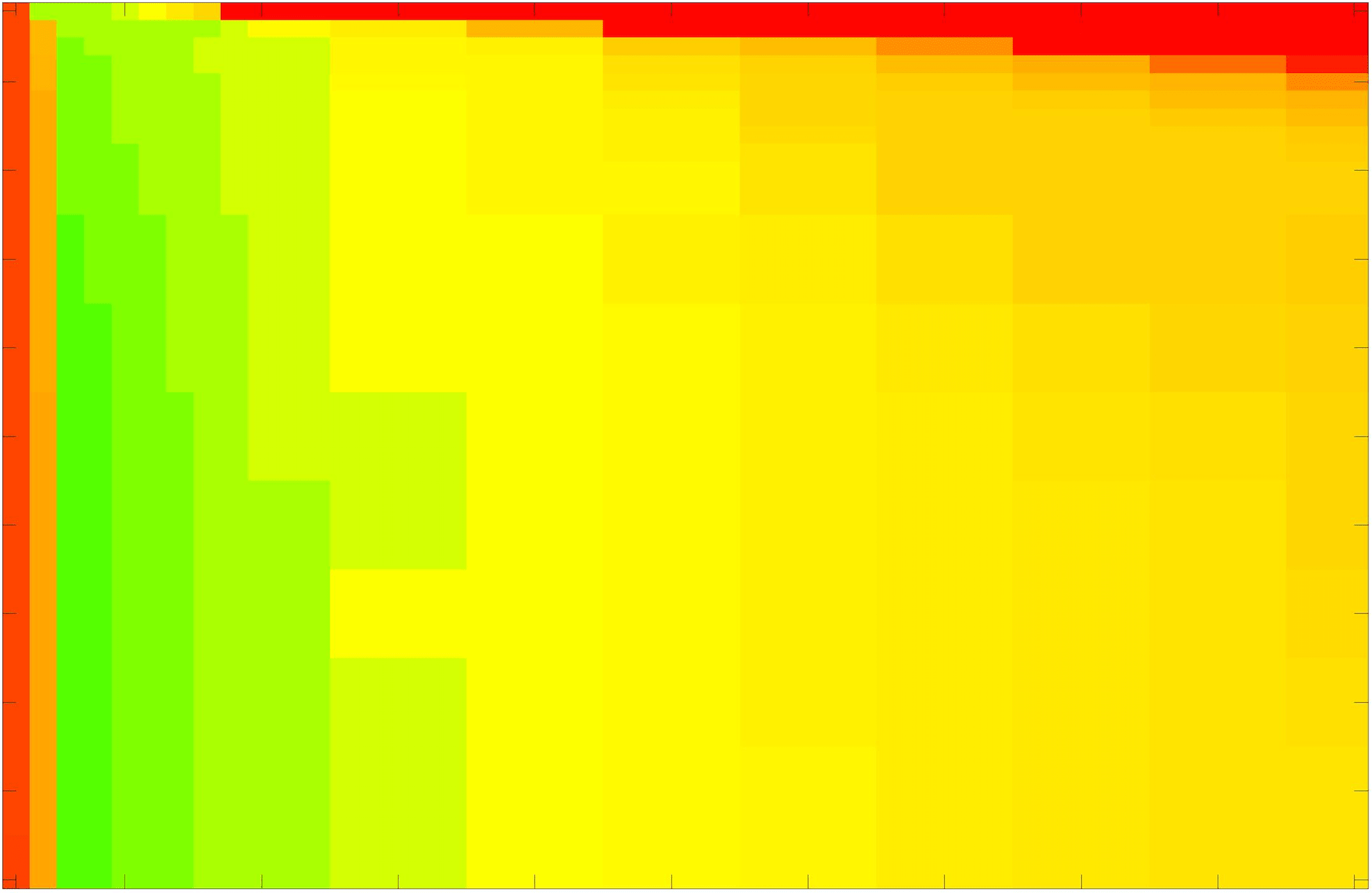}}\quad
\subfloat{\includegraphics[width=1.6in,height=1.1in]{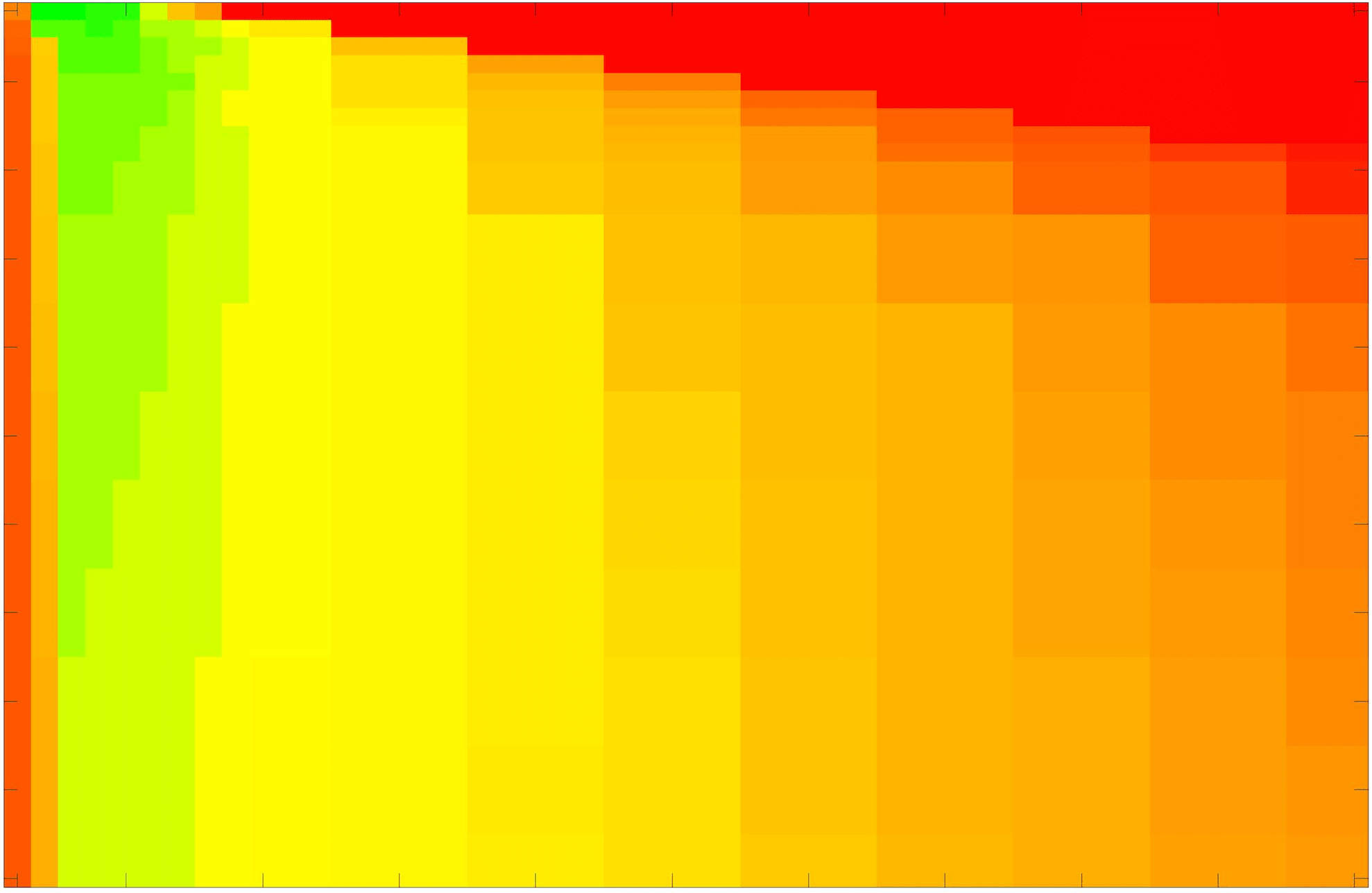}}} 
\end{figure*}

In these tests we aim to demonstrate how sensitive to parameter choices each choice of fitting term is. To accomplish this we perform the segmentations for each of the models discussed (CV, RSF, LCV, HYB, GAV) and the proposed model for a wide range of parameters and compute the TC value. The parameter range used is $\tilde{\lambda},\theta\in[1,50]$. Due to computational constraints, we run for each integer $\tilde{\lambda},\theta$ between 1 and 10, and every fifth from 15 to 50. This aspect of a model's performance is vital when used in practice. The less sensitive to parameter choices a model is the more relevant it is in relation to potential applications. It should be noted that we neglect to test the selective models detailed in \S\ref{sec:selective} with respect to parameter robustness as we are using the authors' implementation of each approach. Instead, we make direct comparisons in the following sections.

\begin{table*}[t]
\centering
\floatbox[{\capbeside\thisfloatsetup{capbesideposition={left,top},capbesidewidth=1.5in}}]{table}[\FBwidth]
{\captionsetup[subfigure]{labelformat=empty,font=normalsize}
\subfloat{
\renewcommand{\arraystretch}{1.5}
\centering
\resizebox{5in}{!}{$
\begin{tabular}{c c c c c c c c c c }
\hline	
\ \ \ \multirow{2}{*}{Model }
& \multicolumn{9}{c}{Test Image}\\	
\cline{2-10}				
&	1&2&3&4&5&6&7&8&9 \\			
\hline																			
CV 	& 0.000 	& 0.000 	& 0.970 	&	0.969	& 0.933 	& 0.988 	& 0.889 	& 0.931 	& 0.180 	\\
RSF 	& {\bf 1.000}  & 0.997 	& 0.993 	&   	0.924	& 0.884 	& 0.956 	& 0.785 	& 0.950 	& 0.782	\\	
LCV 	& 0.313 	& 0.142 	& 0.970  	&   	0.970	& 0.941 	& 0.988 	& 0.911 	& 0.960 	& 0.828 	\\
HYB 	& 0.184 	& 0.091 	& 0.988 	&   	0.960	& 0.870 	& 0.988 	& 0.000 	& 0.000 	& 0.000	\\
GAV 	& 0.984 	& 0.960 	& 0.988 	&   	0.967	& 0.965 	& 0.988 	& 0.950 	& 0.954 	& 0.919	\\
CAC 	& 0.985 	& 0.949 	& 0.946 	&   	0.881	& 0.916 	& 0.961 	& 0.916 	& 0.967 	& 0.952 \\
SRW & {\bf 1.000} & {\bf 1.000} & {\bf 1.000} &   0.761	& 0.724 & 0.708 & 0.917 & {\bf 0.978} 	& 0.957 \\
\hline		
Proposed & {\bf 1.000} & {\bf 1.000 } & {\bf 1.000 }& {\bf 0.973}  & {\bf 0.989} & {\bf 0.990}
& {\bf 0.965}  & 0.961 & {\bf 0.971} \\				
\hline																	
\end{tabular}
$}
}}
{\caption{Optimal TC values for Test Images 1--9, for the models introduced in \S\ref{sec:related} (CV,RSF,LCV,HYB,GAV), \S\ref{sec:selective} (CAC,SRW) and the proposed approach. The best result for each image is given in bold. \label{tab:tableTC}}}
\end{table*}

The TC values for the parameter sets $(\tilde{\lambda},\theta)$ are presented as heatmaps in Figs.~\ref{fig:heat1}--\ref{fig:heat3}. A heatmap is a convenient way to display accuracy results for hundreds of tests concisely. In Fig.~\ref{fig:exampleheat} we give an example heatmap with the same axes used for those in Figs.~\ref{fig:heat1}--\ref{fig:heat3}. For each of the combinations of parameter values $(\tilde{\lambda},\theta)$ we give the TC value of the segmentation result and represent it by the appropriate colour. The corresponding colour scale is shown in Fig.~\ref{fig:colorbar}. Qualitatively, the more green areas of the heatmap the more accurate the model is for a wider set of parameters. Example results for Test Image 5 when varying $\tilde{\lambda}$ (with $\theta=4$) for the proposed model are given in Fig. \ref{fig:kidneyres}. Here it can be seen what each accuracy result corresponds to visually.

{\bf Note.} The axes have been removed from the heatmaps in Figs.~\ref{fig:heat1}--\ref{fig:heat3} for presentational clarity. However, to be explicit, the axes used in all heatmaps are the same as those in Fig.~\ref{fig:exampleheat}.

{\bf Synthetic Images.} These results are presented in Fig. \ref{fig:heat1}. For Test Images 1--2 we see poor parameter robustness from all competing models, except for GAV which performs reasonably well. However, the proposed model has minimal parameter sensitivity for these images, with good results achieved for almost every combination of values tested. For Test Image 3 all models have a reasonable parameter range (except for RSF), however the proposed model gives better quality results for a wider parameter range. The other models achieve reasonable results here as the foreground intensity of the ground truth is greater than the background $(c_1=0.75, c_2=0.49)$, whereas for Test Images 1--2 they are equal $(c_1=c_2=0.50)$. These results highlight the key advantage of the proposed model. 

{\bf Real Images.} In Fig \ref{fig:heat2} we present results for Test Images 4--6. Here, the proposed model performs in a similar way to its competitors because these images are more typical selective segmentation problems in the sense that there is a clear distinction between the foreground and background intensities. In particular, the values in each case are: Test Image 4 $(c_1=0.85, c_2=0.25)$, Test Image 5 $(c_1=0.70, c_2=0.19)$, and Test Image 6 $(c_1=0.73, c_2=0.20)$. It can be seen that the proposed model is competitive compared to previous approaches. The performance is quite poor for Test Image 5, but is arguably still the best for this challenging case. In Fig. \ref{fig:heat3} we present results for Test Images 7--9. Here the proposed model outperforms previous approaches significantly for each image. This is mainly due to the type of image considered. Specifically, the true intensities are: Test Image 7 $(c_1=0.12, c_2=0.24)$, Test Image 8 $(c_1=0.10, c_2=0.23)$, and Test Image 9 $(c_1=0.08, c_2=0.14)$. The proposed model is capable of achieving results where $c_1\approx c_2$, with other models failing completely in these cases.

\subsection{Accuracy Comparisons}
\label{sec:r1}

\begin{figure*}
\centering
\floatbox{figure}[\FBwidth]
{\captionsetup[subfigure]{labelformat=empty,font=normalsize}
\subfloat[{(i) CV \cite{ACWE}, TC = 0.18}]{\includegraphics[width=1.6in,height = 1.6in]{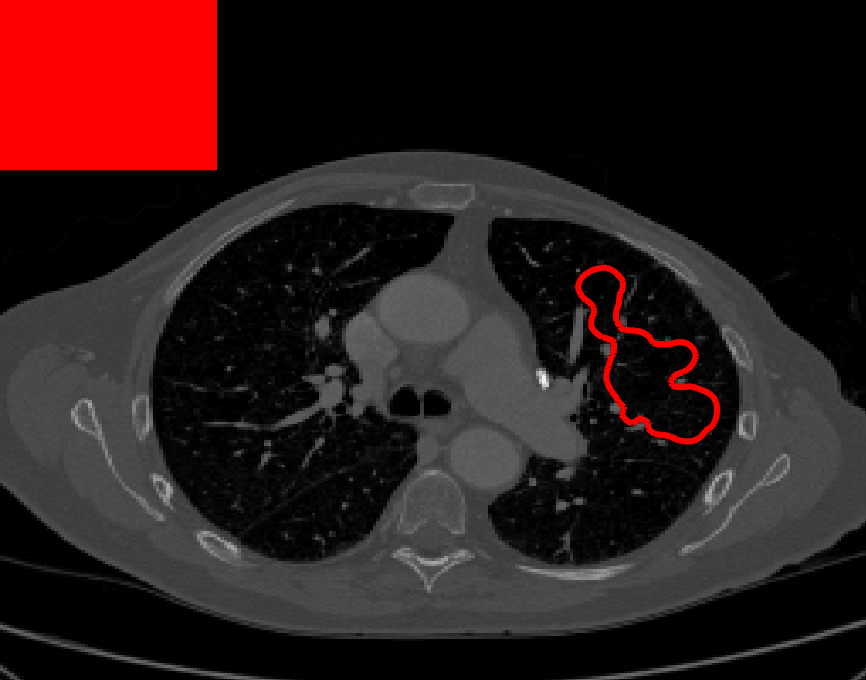}}
\subfloat[{(ii) RSF \cite{RSF}, TC = 0.78}]{\includegraphics[width=1.6in,height = 1.6in]{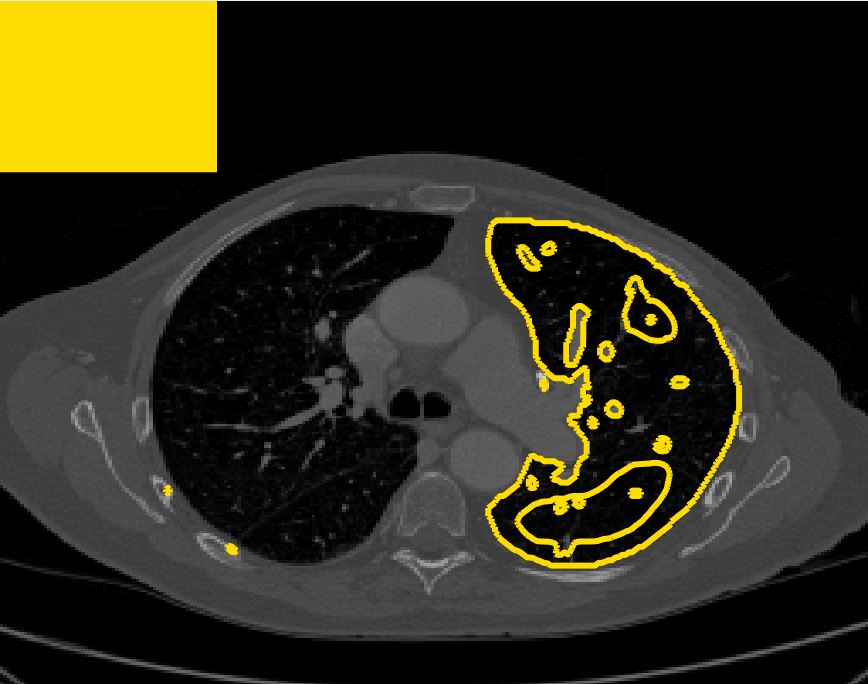}}
\subfloat[{(iii) LCV \cite{LCV}, TC = 0.83}]{\includegraphics[width=1.6in,height = 1.6in]{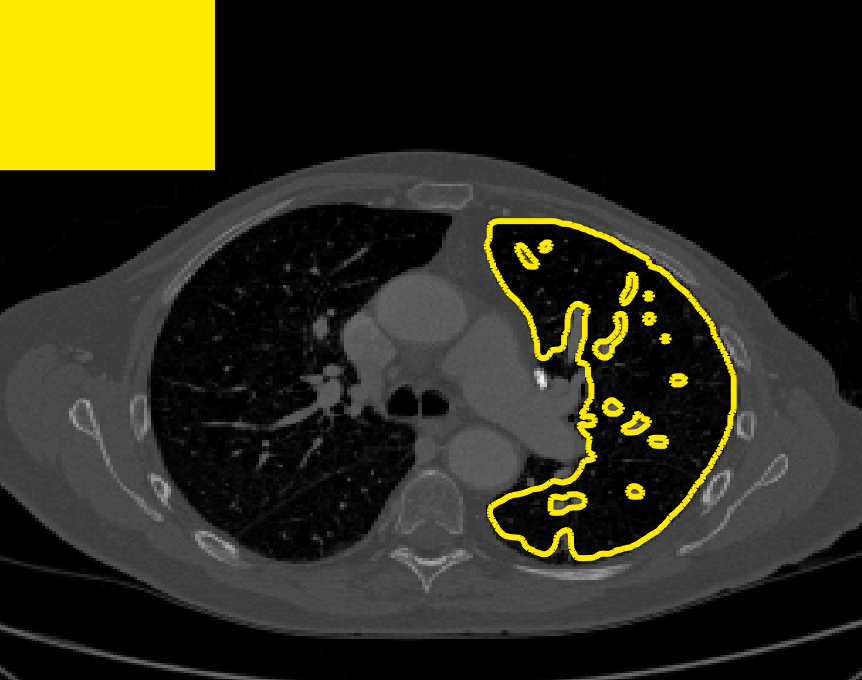}}
\subfloat[{(iv) HYB \cite{Ali:16}, TC = 0.00}]{\includegraphics[width=1.6in,height = 1.6in]{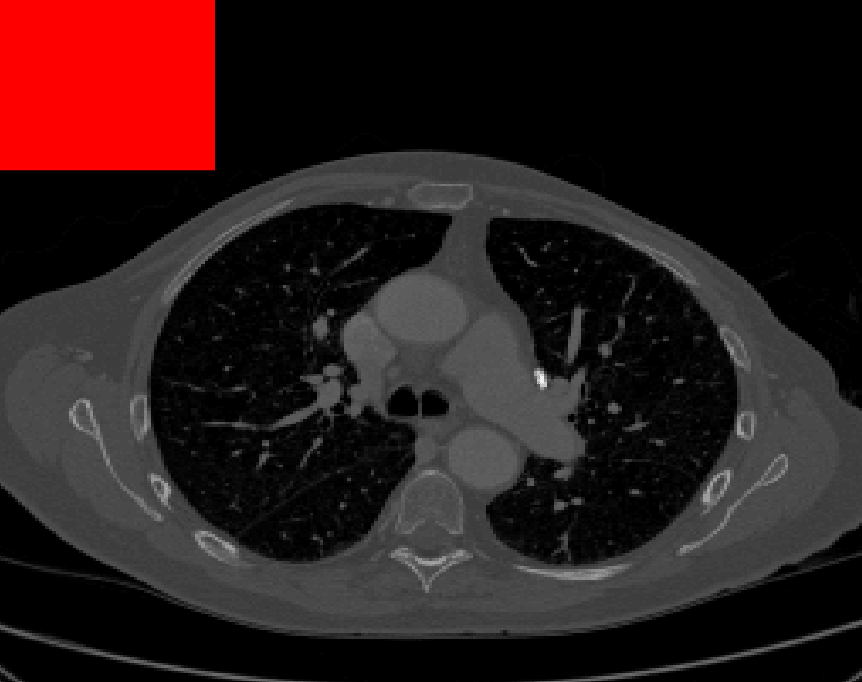}}}
\centering
\floatbox{figure}[\FBwidth]
{\captionsetup[subfigure]{labelformat=empty,font=normalsize}
\subfloat[{(v) GAV \cite{Ali:17}, TC = 0.92}]{\includegraphics[width=1.6in,height = 1.6in]{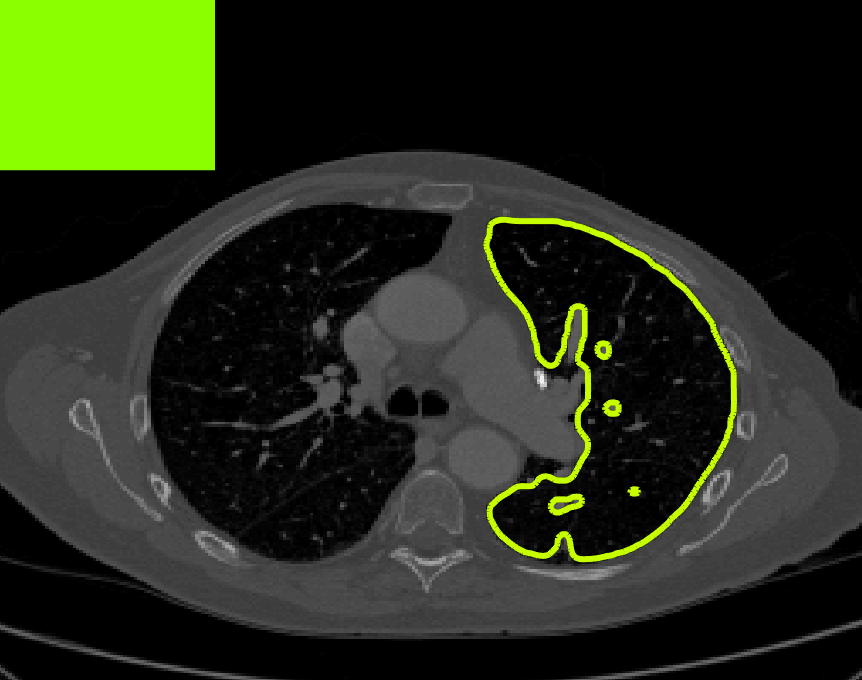}}
\subfloat[{(vi) CAC \cite{Nguyen:12}, TC = 0.95}]{\includegraphics[width=1.6in,height = 1.6in]{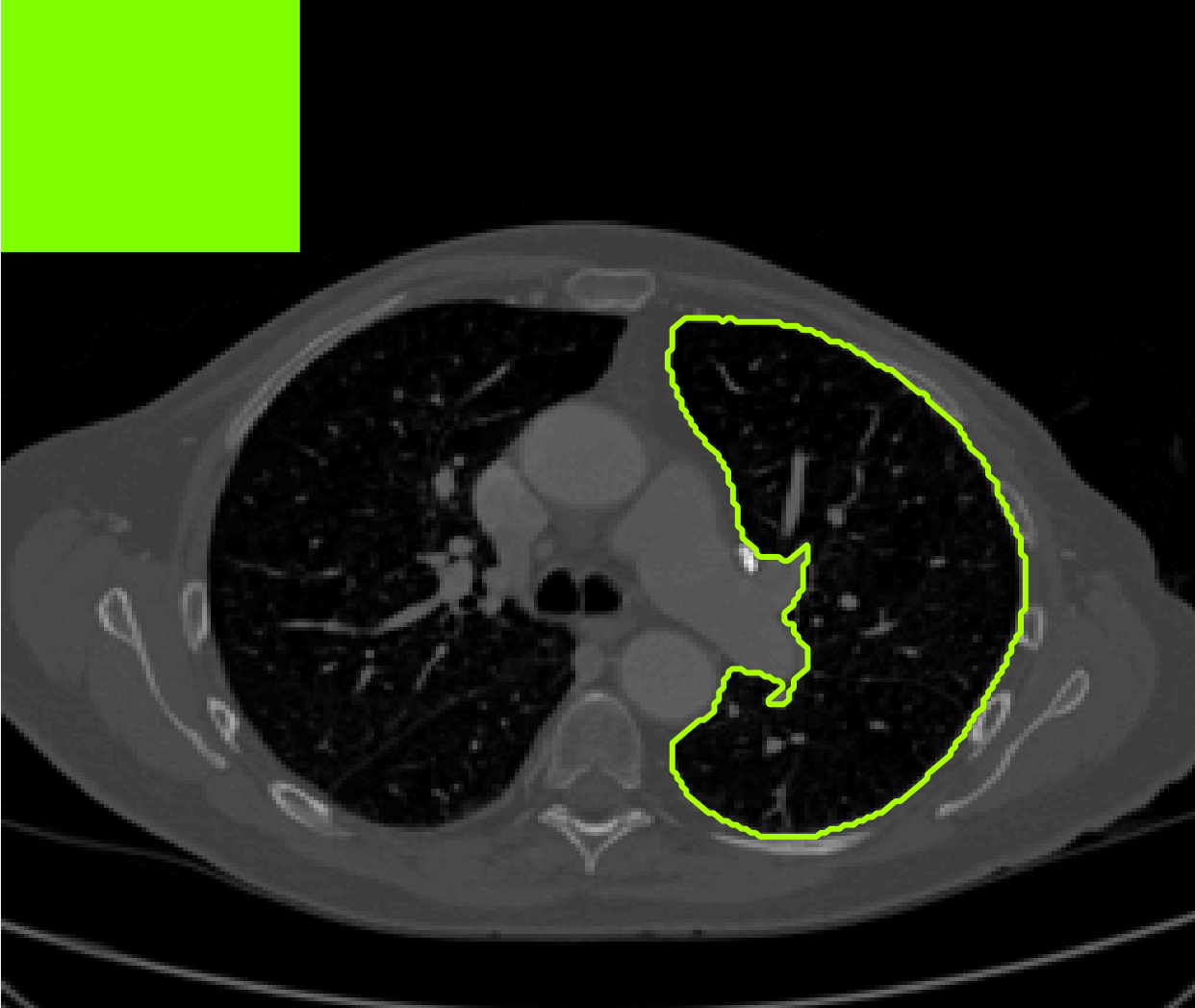}}
\subfloat[{(vii) SRW \cite{SRW}, TC = 0.96}]{\includegraphics[width=1.6in,height = 1.6in]{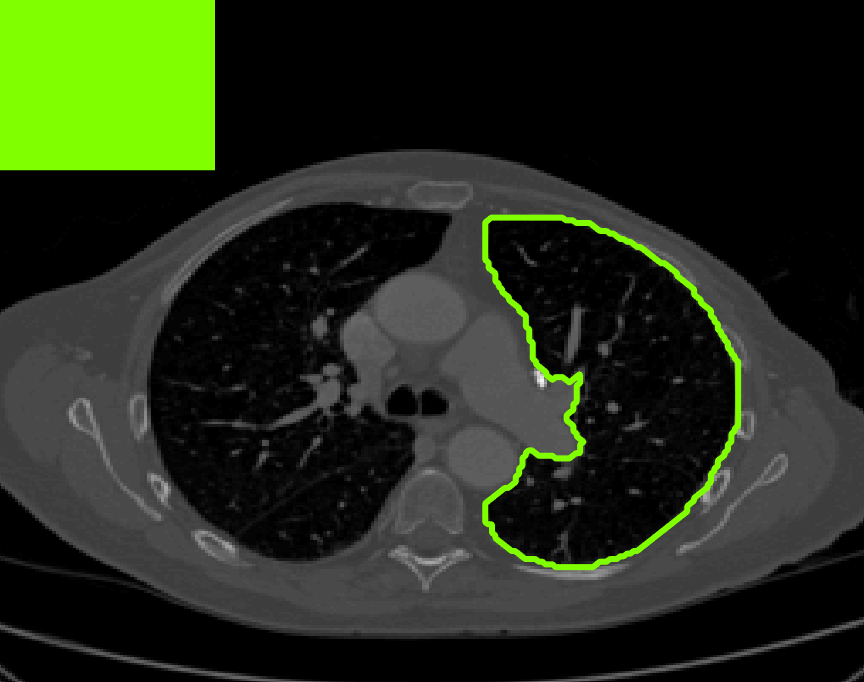}}
\subfloat[{(viii) Proposed, TC = 0.97}]{\includegraphics[width=1.6in,height = 1.6in]{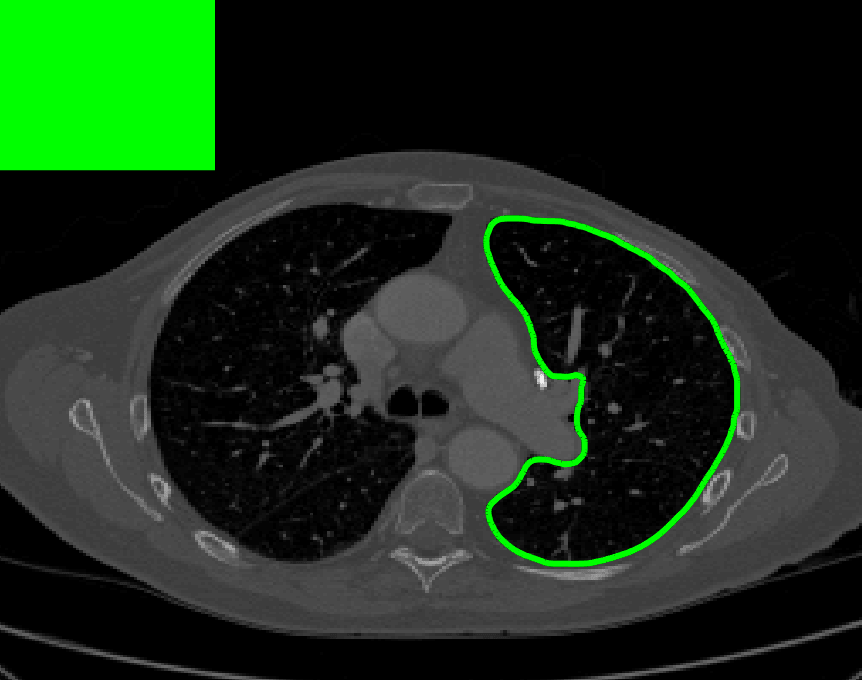}}}
{\caption{We present the optimal results for Test Image 9. The accuracy is represented by colour, consistent with the scale in Fig.~\ref{fig:colorbar}. The proposed model often significantly outperforms previous approaches in this case. \label{fig:new1}}}
\end{figure*}

Here we aim to address the question of whether each model is capable of achieving an accurate result. In other words, assuming that factors such as parameter and user input sensitivity are ignored, how successful is each approach. In Table~\ref{tab:tableTC} we present the optimal TC values for each model found from the tests described in the previous section, with the highest value in bold. We include values for CAC \cite{Nguyen:12} and SRW \cite{SRW}, which we have obtained by iteratively refining the user input and running the algorithm. It is worth mentioning that we are using the authors' implementation of each method. For each image, the results presented in Table \ref{tab:tableTC} are the most accurate we could obtain given a reasonable level of input (comparisons with identical input are discussed in \S\ref{sec:r4}). Immediately we can see that the proposed model consistently outperforms the other models in terms of accuracy for the test images (RSF equals it for Test Image 1, SRW equals it for Test Images 1-3, and beats it for Test Image 8). Below we will discuss some relevant details of the results, again by splitting the test images into synthetic and real.

{\bf Synthetic Images.} We observe that for Test Images 1 and 2 (where $c_{1}=c_{2}$ , CV, LCV, and HYB fail completely. GAV performs well, with the proposed model and RSF being the most accurate with perfect results. For Test Image 3, all models are capable of achieving a good result. It should be noted that in this case $c_1=0.75$ and $c_2=0.49$. This difference enables the other models to perform well, although the proposed model is slightly superior with a perfect result. The alternative selective models also perform well for these images, although CAC has minor errors on the boundaries of the foreground for each image. 

{\bf Real Images.} In Table \ref{tab:tableTC} we can see that the proposed model is the most successful in terms of optimal accuracy. It is worth noting some inconsistency in the other models, with all but GAV having results that fall below TC $= 0.9$ for at least one image. GAV performs well for Test Images 4--9, with the proposed model slightly outperforming it in each case. It is worth reminding the reader that for GAV the parameters $(\beta_{1},\beta_{2})$ have been refined for each example. Fixing this results in more variability in the quality of results. The proposed model has no such parameter optimisation between examples. CAC and SRW perform reasonably well for these images, although are sometimes substandard for Test Images 4-7. This is despite extensive refinement of the user input to achieve an acceptable result. We present the optimal results for Test Image 9 in Fig. \ref{fig:new1}. Here we can see how much variation there is in the quality of results for this lung CT image. CAC and SRW are competitive in this instance. Of the remaining approaches GAV is the most competitive (TC $= 0.919$), but is visually inadequate. Two other models (CV, HYB) fail completely. In this case, the problem looks quite straightforward and yet other fitting terms are insufficient to produce a good result. Again, the proposed model tends to be superior in cases where $c_{1}\approx c_{2}$ and is capable of achieving very good results for all the images considered. This highlight the advantages of the proposed fitting term.

\subsection{User Input Randomisation}
\label{sec:r3}

\begin{figure*}
\floatbox[{\capbeside\thisfloatsetup{capbesideposition={left,top},capbesidewidth=1.5in}}]{figure}[\FBwidth]
{\caption{Boxplots of the TC values for $1000$ random user inputs using the proposed model. We observe that the method is remarkably consistent. Even the worst results, excluding outliers, are competitive with the optimal results of the existing approaches shown in Table \ref{tab:tableTC}. \label{fig:TCboxplots}}}
{\includegraphics[width=5in,height = 3in]{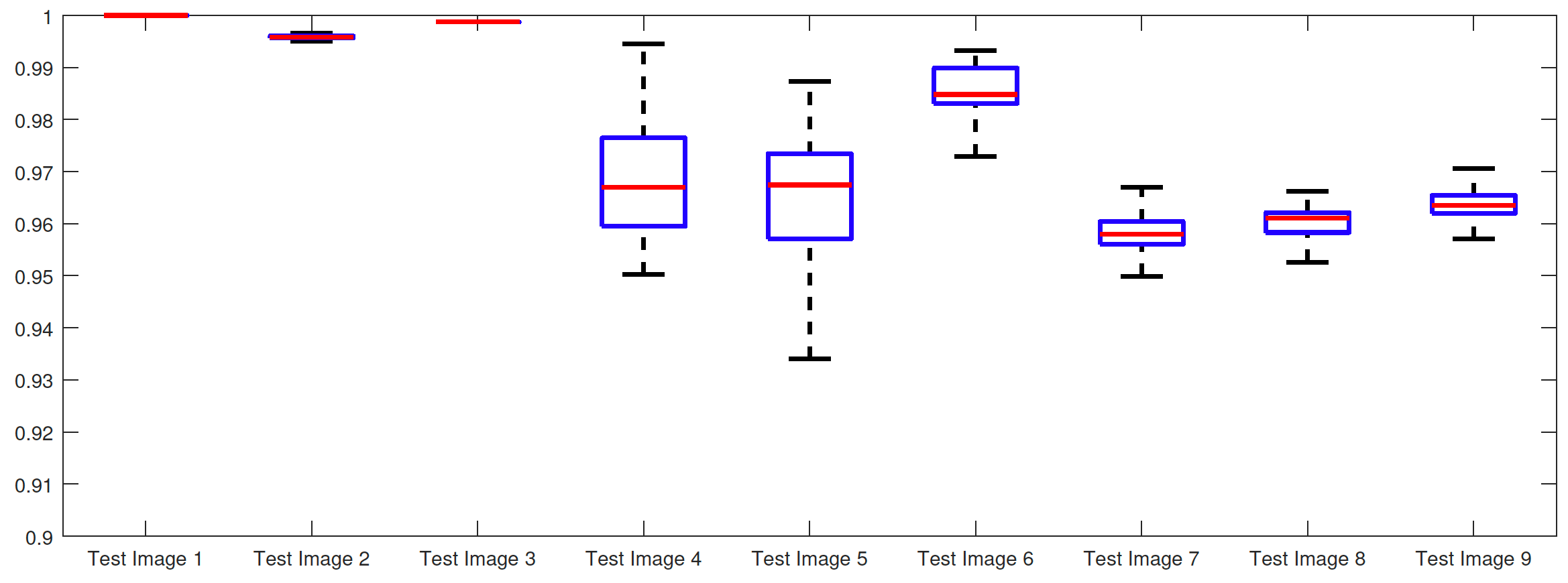}}
\end{figure*}

\begin{figure*}
\centering
\floatbox[{\capbeside\thisfloatsetup{capbesideposition={left,top},capbesidewidth=1.5in}}]{figure}[\FBwidth]
{\captionsetup[subfigure]{labelformat=empty,font=normalsize}
\subfloat[(i) TC = 1.00]{\includegraphics[width=1.5in,height = 1.5in]{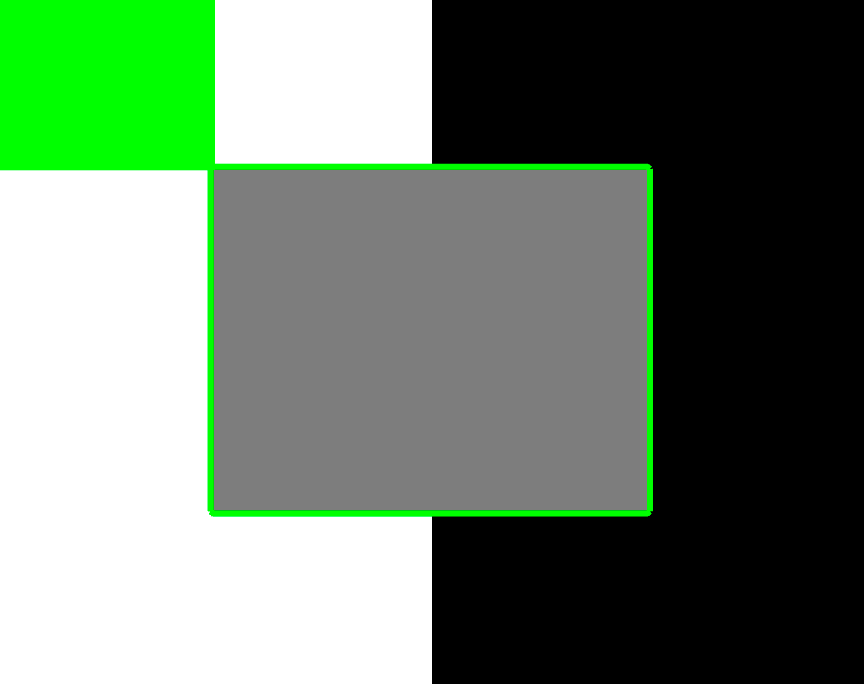}}\hspace{0.25in}
\subfloat[(ii) TC = 0.99]{\includegraphics[width=1.5in,height = 1.5in]{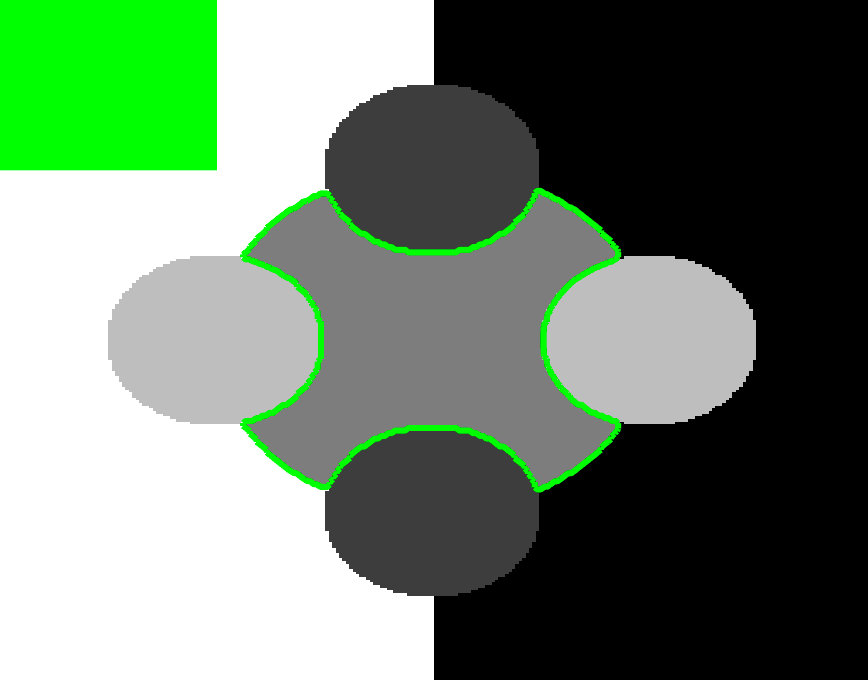}}\hspace{0.25in}
\subfloat[(iii) TC = 1.00]{\includegraphics[width=1.5in,height = 1.5in]{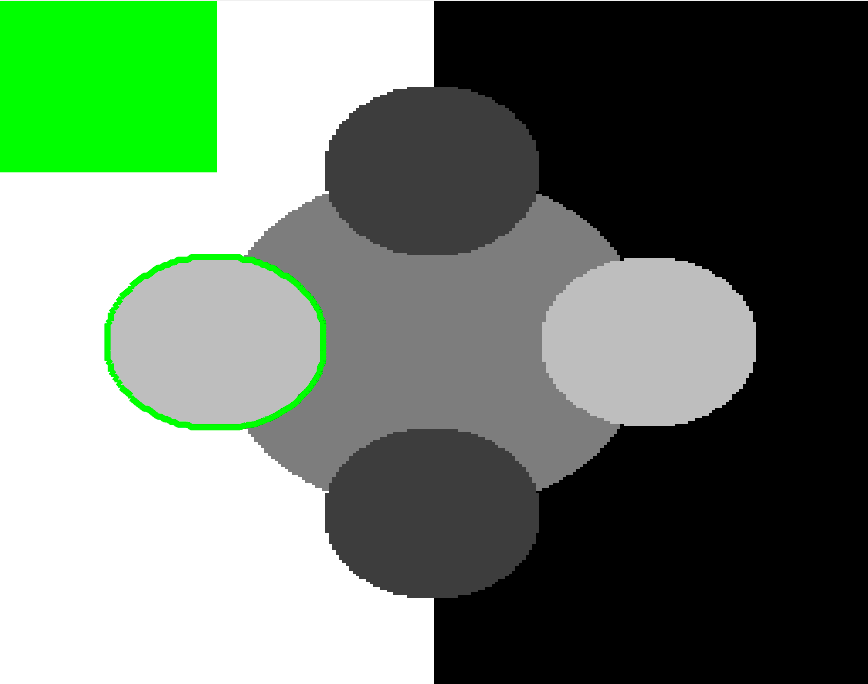}}}
{\caption{Results for the proposed model for each image, including TC values. The worst result, excluding outliers, of $1000$ random user inputs for each example is presented. This demonstrates that the model is robust to user input, with poor results being competitive with the optimal result of competitors. \label{fig:new2}}}
\centering
\vspace{-0.2in}
\floatbox[{\capbeside\thisfloatsetup{capbesideposition={left,center},capbesidewidth=1.5in,font =normalsize}}]{figure}[\FBwidth]
{\caption*{}}
{\captionsetup[subfigure]{labelformat=empty,font=normalsize}
\subfloat[(iv) TC = 0.95]{\includegraphics[width=1.5in,height=1.5in]{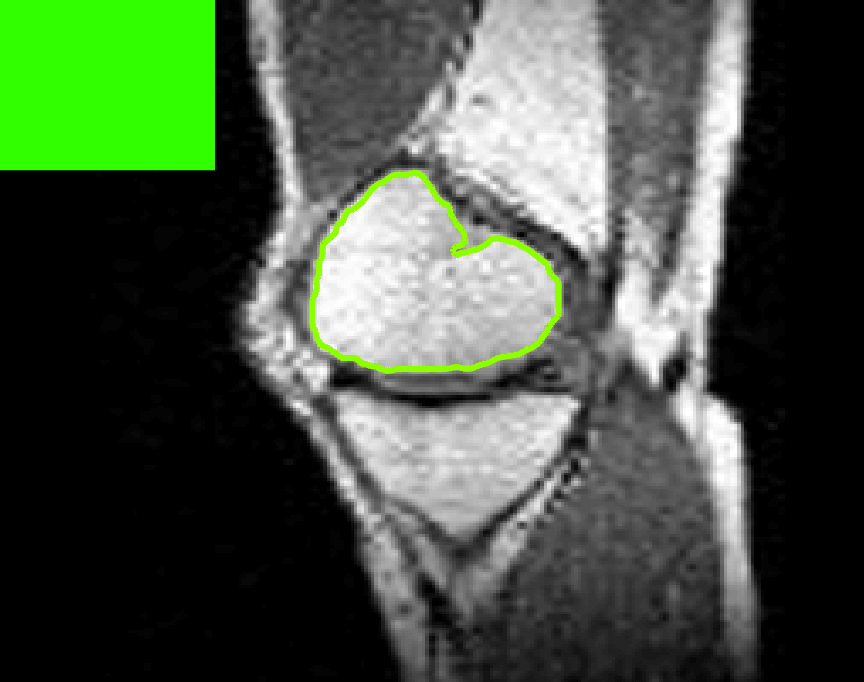}}\hspace{0.25in}
\subfloat[(v) TC = 0.93]{\includegraphics[width=1.5in,height=1.5in]{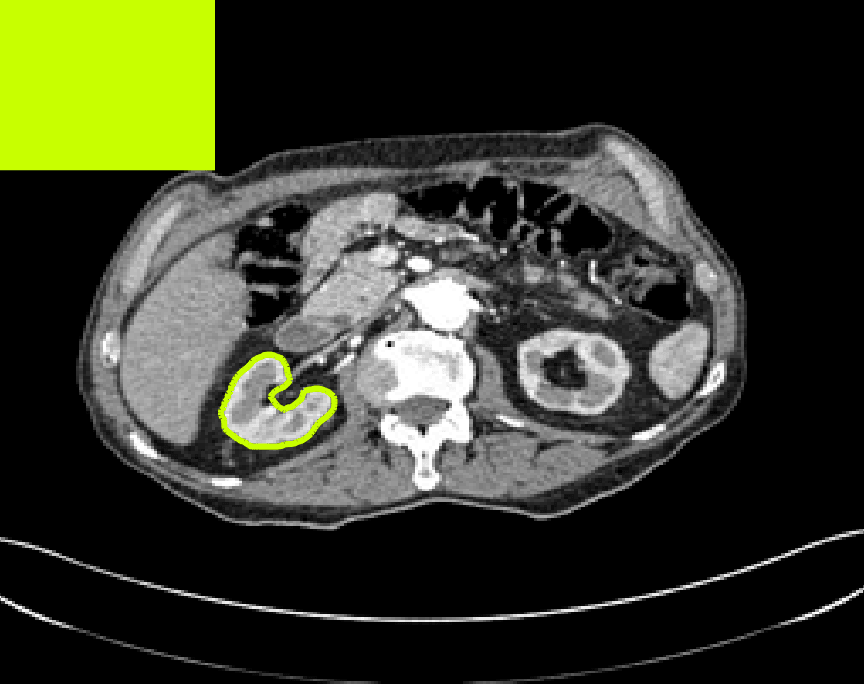}}\hspace{0.25in}
\subfloat[(iv) TC = 0.97]{\includegraphics[width=1.5in,height=1.5in]{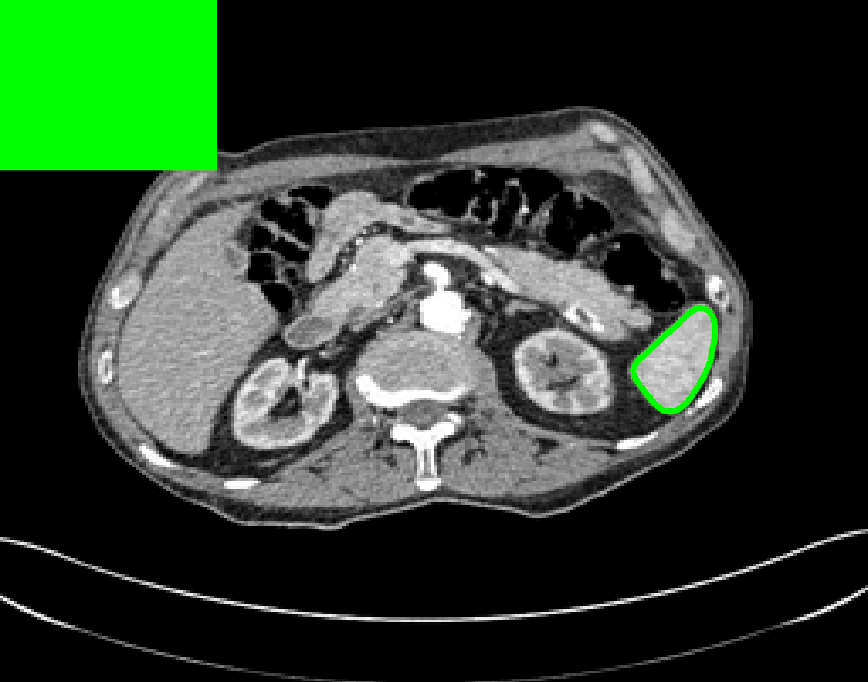}}}
\centering
\floatbox[{\capbeside\thisfloatsetup{capbesideposition={left,center},capbesidewidth=1.5in,font =normalsize}}]{figure}[\FBwidth]
{\caption*{}}
{\captionsetup[subfigure]{labelformat=empty,font=normalsize}
\subfloat[(vii) TC = 0.95]{\includegraphics[width=1.5in,height=1.5in]{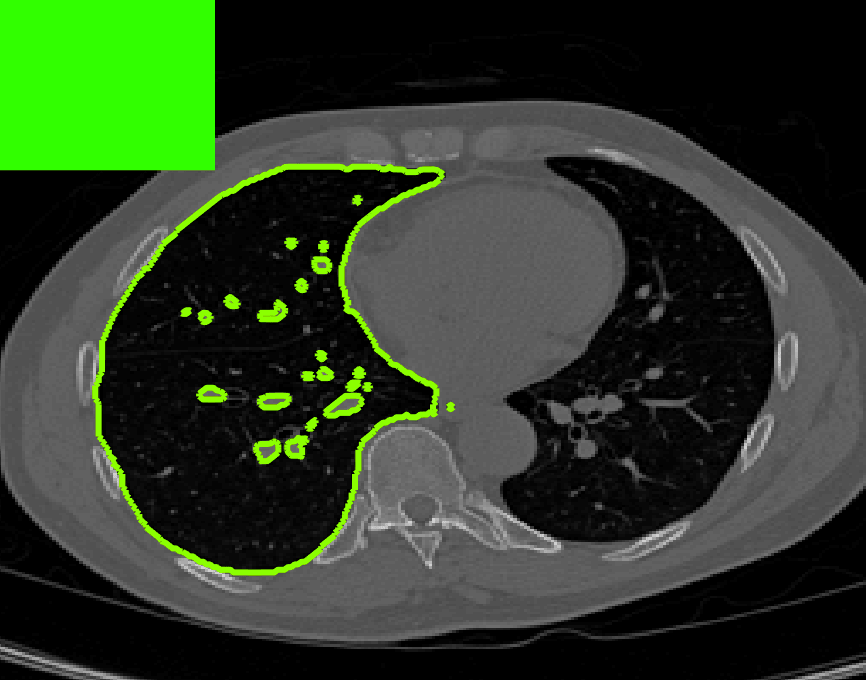}}\hspace{0.25in}
\subfloat[(viii) TC = 0.95]{\includegraphics[width=1.5in,height=1.5in]{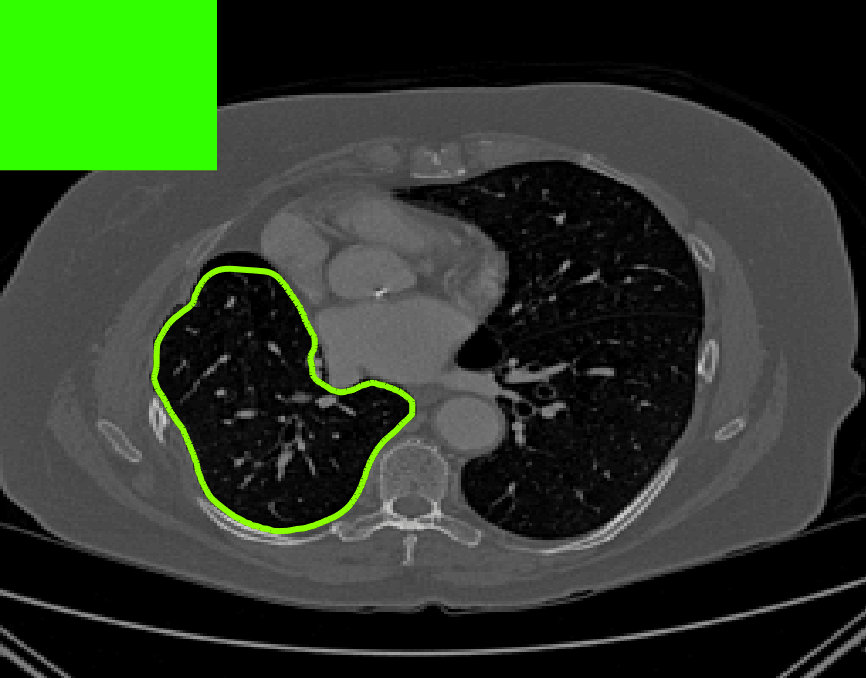}}\hspace{0.25in}
\subfloat[(ix) TC = 0.96]{\includegraphics[width=1.5in,height=1.5in]{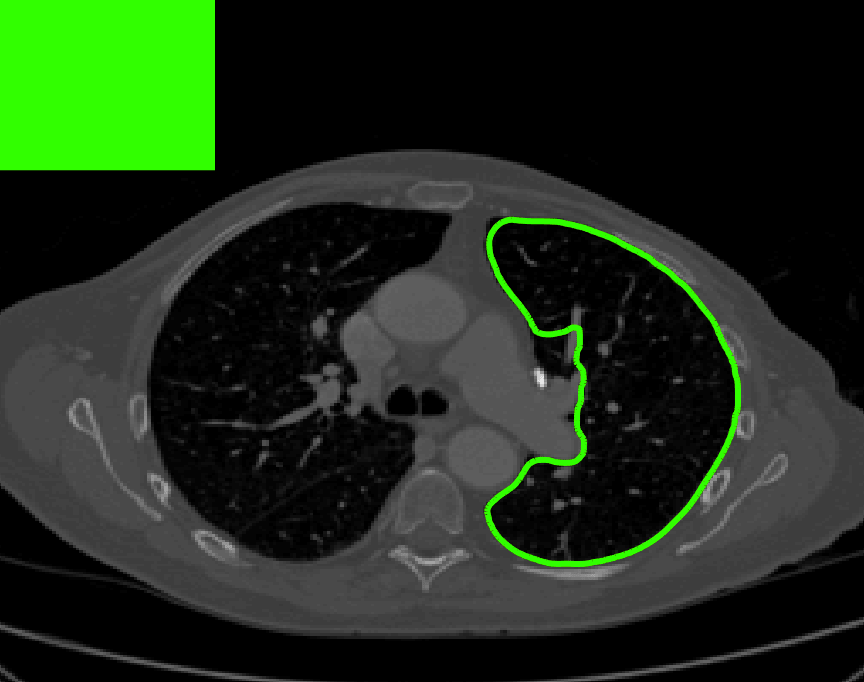}}}
\end{figure*}

One key consideration for the practical use of selective segmentation models is that the result is not too reliant on user input. With intricate user input accurate results are almost guaranteed. However, the benefit of this kind of approach is that accuracy should be attainable with minimal, intuitive user input. One challenge in this setting is how to ascertain to what extent a method is dependent on the user input. In this section we will generalise the user input for the proposed model in order to determine how sensitive it is in this respect. By generalising in this way we will make two assumptions about the markers, $\mathcal{M}$, consistent with the above considerations:
\begin{enumerate}
\item[(i)] All points are within the target object.
\item[(ii)] Only 3 markers are selected.
\end{enumerate}
We regard neither of these assumptions to be too onerous on a user, and are quite consistent with practical use. To perform this test, we randomly choose $1000$ sets of 3 marker points and run each algorithm using them. The parameters $\tilde{\lambda}$ and $\theta$ are fixed at those which gave the optimal TC values in Table~\ref{tab:tableTC}. For each set of marker points we compute the corresponding TC value of applying the proposed model with this input. The results for each image are summarised by boxplots in Fig.~\ref{fig:TCboxplots} with examples of the worst results, excluding outliers, shown in Fig. \ref{fig:new2}. Here, it can be seen that the worst result often outperforms the optimal results of the alternative models considered, which is impressive. Below we discuss the results for the test images, by again splitting them into synthetic and real images. Based on the authors' implementation of CAC and SRW it was not possible to generalise the input in this way. Instead we make direct comparisons of input in the next section.

{\bf Synthetic Images.} For the Test Images 1--3 we achieve near perfect segmentations in all cases, shown by the mean TC being between 0.99 and 1.00 in all cases (for Test Image 1, the mean is precisely 1.00) and a small variance around the mean. Therefore, we can conclude that for images of this type, where the foreground is homogeneous, our method is very robust to user input. Essentially, any reasonable set of markers should produce excellent results. It should be noted that the optimal results from comparable approaches are less than the mean result of $1000$ random tests for our method (except for SRW). This can be observed in Table~\ref{tab:tableTC}. Furthermore, these methods often fail completely. This is a key result highlighting the advantages of our method. In visually simple cases (Test Images 1--3) our new data fitting term is an improvement on existing approaches by modifying the underlying assumptions involved.

{\bf Real Images.} In all cases for Test Images 4--9 the mean values show that the segmentation results are highly accurate. Also, we notice that the variances are very reasonable demonstrating the robustness of varying the user input. This is an important aspect of selective segmentation, and highlights the advantages of the proposed fitting term. For Test Images 4--6 we observe more variability in the accuracy due to minor intensity inhomogeneity in the foreground. This means randomising the user input will be more sensitive. However, we can see that the results are very good with the mean accuracy being competitive with the optimal accuracy of comparable methods. In the case of the lung CT images (Test Images 7--9) the variance in TC values is very small, due to the homogeneity of the foreground. Again, it is important to compare the results of $1000$ random results using our proposed model to the optimal result of comparable methods. For these images all of the methods (except GAV,CAC, and SRW) have at least one TC value below 0.9. However, GAV requires the tuning of additional parameters $(\beta_{1},\beta_{2})$ whilst the proposed model does not. The results for CAC and SRW also rely on extensive requirements of the user input to achieve this accuracy, whereas random input compares favourably here. Compared to GAV, we can see that the mean of our tests is similar to the optimal value of GAV. One exception is for Test Image 9 (shown in Fig. \ref{fig:new1}), where there is a significant gap in favour of our model. Again, from Fig. \ref{fig:new2}, we can see that the worst result of randomising the user input for the proposed model is competitive with the optimal results of the alternatives. This is one of the most encouraging aspects of the tests; the proposed model is remarkably robust to varying user input. This proves that successful results with minimal, intuitive user input is possible for a range of examples.

\subsection{Alternative Selective Methods}
\label{sec:r4}

\begin{figure*}
\floatbox[{\capbeside\thisfloatsetup{capbesideposition={left,top},capbesidewidth=1.5in}}]{figure}[\FBwidth]
{\caption{Boxplots of the TC values comparing our method to CAC \cite{Nguyen:12} and SRW \cite{SRW} for 30 test images. {Ours (i)} refers to using identical user input to CAC and SRW, with a sample shown in Fig. \ref{fig:inputimages}. {Ours (ii)} refers to 1000 random variations of the user input for each image. \label{fig:bp2}}}
{\includegraphics[width=5in,height = 3in]{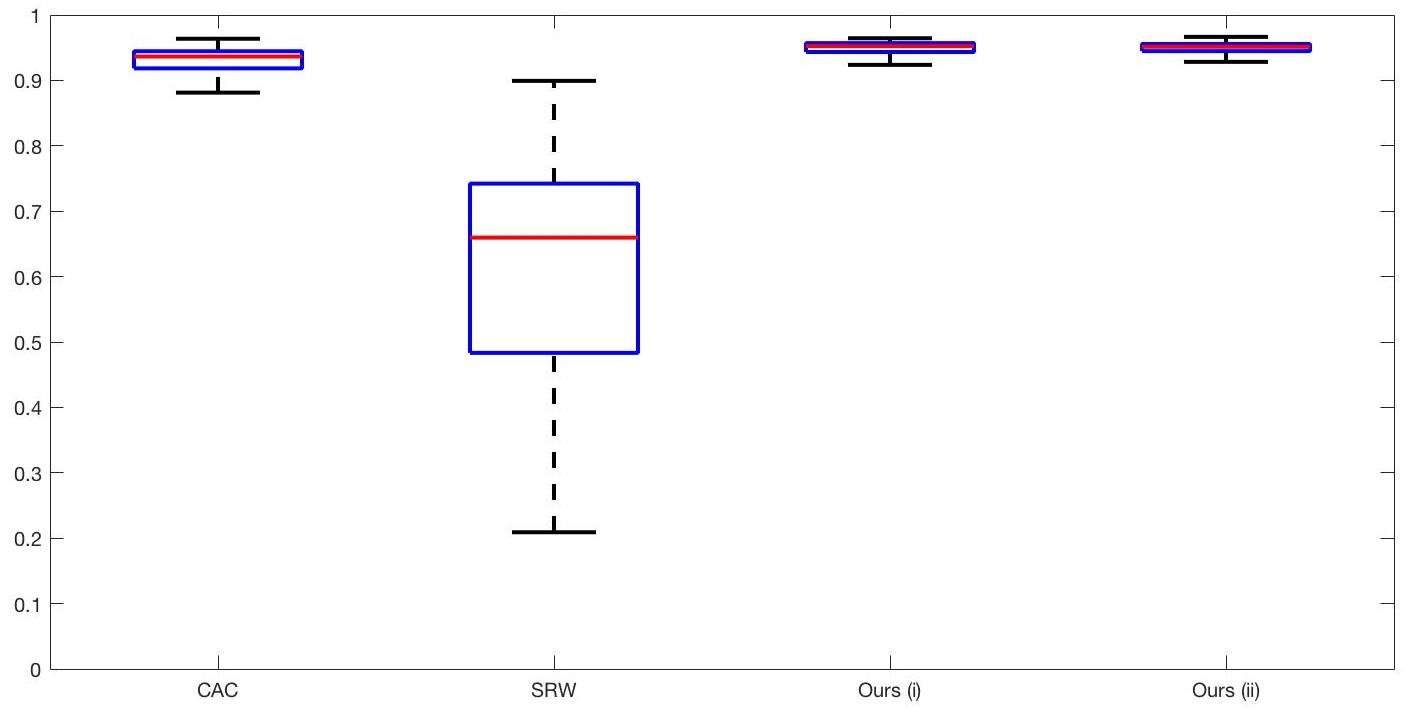}}
\end{figure*}

\begin{figure*}
\centering
\floatbox{figure}[\FBwidth]
{\captionsetup[subfigure]{labelformat=empty,font=normalsize}
\subfloat{\includegraphics[width=1.2in,height = 1.2in]{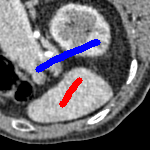}}\quad
\subfloat{\includegraphics[width=1.2in,height = 1.2in]{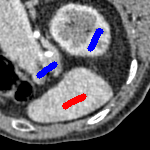}}\quad
\subfloat{\includegraphics[width=1.2in,height = 1.2in]{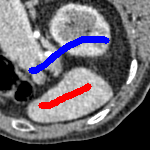}}\quad
\subfloat{\includegraphics[width=1.2in,height = 1.2in]{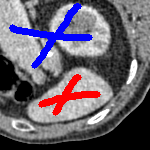}}\quad
\subfloat{\includegraphics[width=1.2in,height = 1.2in]{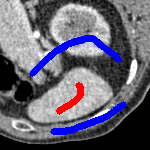}}}
\centering
\floatbox{figure}[\FBwidth]
{\captionsetup[subfigure]{labelformat=empty,font=normalsize}
\subfloat{\includegraphics[width=1.2in,height = 1.2in]{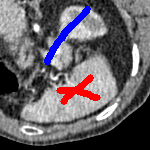}}\quad
\subfloat{\includegraphics[width=1.2in,height = 1.2in]{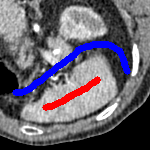}}\quad
\subfloat{\includegraphics[width=1.2in,height = 1.2in]{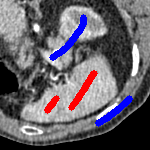}}\quad
\subfloat{\includegraphics[width=1.2in,height = 1.2in]{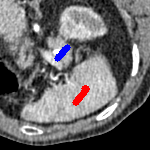}}\quad
\subfloat{\includegraphics[width=1.2in,height = 1.2in]{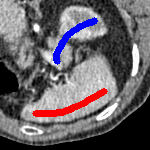}}}
\centering
\floatbox{figure}[\FBwidth]
{\captionsetup[subfigure]{labelformat=empty,font=normalsize}
\subfloat{\includegraphics[width=1.2in,height = 1.2in]{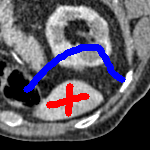}}\quad
\subfloat{\includegraphics[width=1.2in,height = 1.2in]{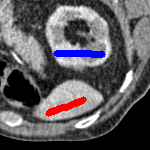}}\quad
\subfloat{\includegraphics[width=1.2in,height = 1.2in]{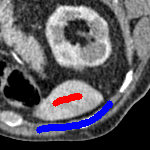}}\quad
\subfloat{\includegraphics[width=1.2in,height = 1.2in]{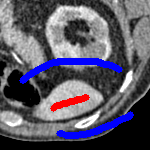}}\quad
\subfloat{\includegraphics[width=1.2in,height = 1.2in]{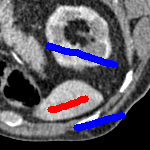}}}
{\caption{Examples of the input used to compare our method to CAC \cite{Nguyen:12} and SRW \cite{SRW}. Each row represents an image in the dataset and we present five variations of the input used in the tests described in \S\ref{sec:r4}. \label{fig:inputimages}}}
\end{figure*}

In order to further establish the robustness of our method, we now introduce the results of testing our approach against competing interactive segmentation methods on a larger data set. The results are presented in Fig. \ref{fig:bp2}, showing a boxplot of accuracy in terms of TC on a set of 30 CT images (excluding outliers). The target structure we consider is the spleen, as this consists of a relatively homogeneous foreground, appropriate for the approach considered. The data has been manually contoured providing ground truth data for the image set. We compare CAC \cite{Nguyen:12} and SRW \cite{SRW} against our method with five variations of user input for each image. It is worth emphasising here that the input used in the tests is identical for each approach and was not refined in any way. It was designed to mimic what a user, unfamiliar with each approach, might select intuitively. A representative example for three images is shown in Fig. \ref{fig:inputimages}. This shows foreground (red) and background (blue) user input regions. For our method, we define the red region as $\mathcal{P}$ as discussed in \S\ref{sec:intro} and enforce hard constraints on the blue region. We refer to the results of the proposed approach using this input as Ours (i). We also include results of randomising the user input in an identical way to \S\ref{sec:r3}. For each image we generate 1000 simulated user input choices, which we present as Ours (ii). It is important to note that the difference between Ours (i) and (ii) is only the definition of $\mathcal{P}$. The method and parameters are fixed between each.

\begin{figure*}
\floatbox[{\capbeside\thisfloatsetup{capbesideposition={left,top},capbesidewidth=1.5in}}]{figure}[\FBwidth]
{\caption{Boxplots of the TC values from Fig. \ref{fig:bp2} for $\text{TC}\in[0.8,1]$. Here, the extent to which the proposed method outperforms CAC \cite{Nguyen:12} is clearer for both types of input. \label{fig:bp3}}}
{\includegraphics[width=5in,height = 3in]{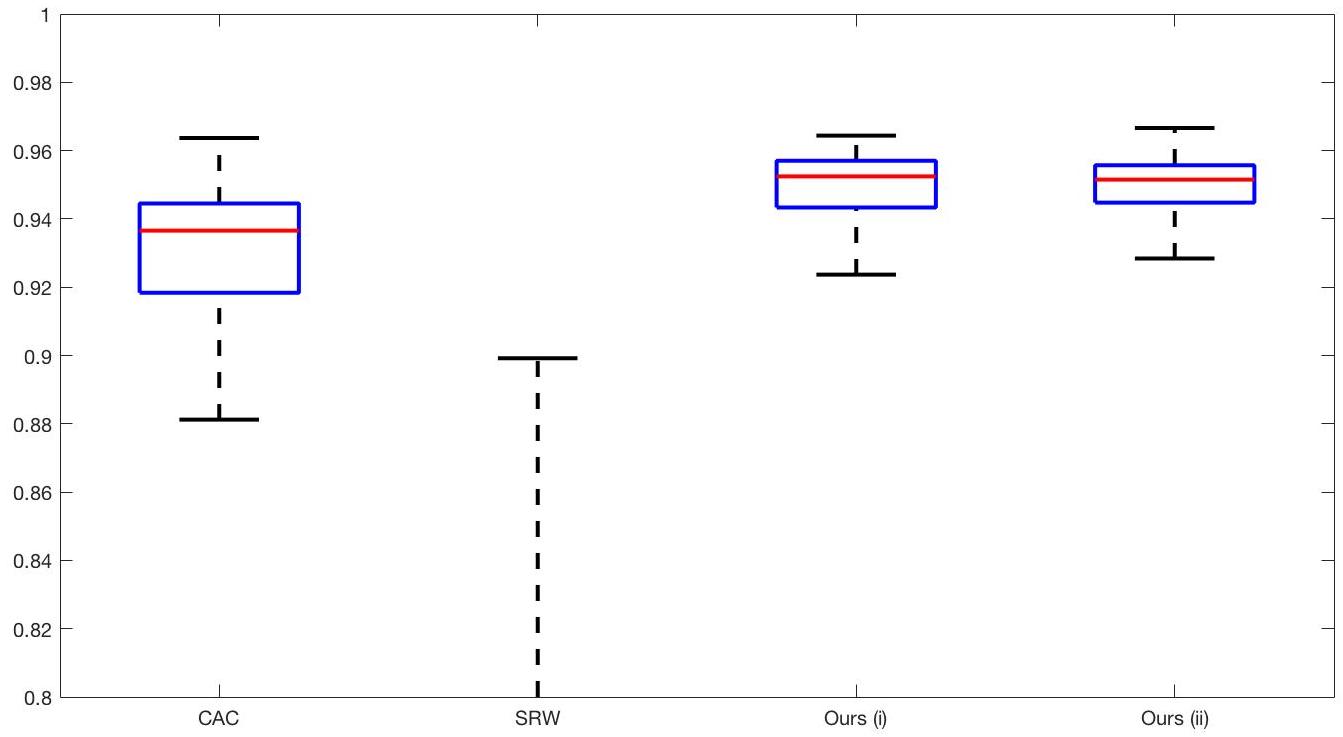}}
\end{figure*}

The performance of CAC \cite{Nguyen:12} is very good, as shown in Fig. \ref{fig:bp2}. We have included an additional figure to highlight the difference between CAC and Ours (i) and (ii) more precisely. This is shown in Fig. \ref{fig:bp3} (this is the same as Fig. \ref{fig:bp2} with TC restricted to [0.8,1]). Here we can see that the proposed approach has a slightly better median (0.96 compared to 0.94) and is generally more consistent than CAC. This is particularly evident when considering the worst TC results of CAC ($0.19$) against ours ($0.87$).

In Fig. \ref{fig:bp2} it can be seen that our method exceeds the performance of SRW by a large margin (0.66 compared to 0.95). One possible reason for this is that the input used, as displayed in Fig. \ref{fig:inputimages}, is restricted to be as intuitive as possible. SRW is capable of achieving improved results with more elaborate foreground/background input. However, it is generally reliant on a trial and error approach which is not ideal in practice. This highlights an important advantage of our method. It is able to achieve a high standard of results with simple user input. This is reinforced by considering Ours (ii), where the results of 30000 random variations of the user input does not cause a drop off in accuracy compared to the 150 manual user input selections. Again, this can be seen more clearly in Fig. \ref{fig:bp3}. In fact, the results for the proposed approach with the random input are slightly better than with the manual input. This underlines the robustness to user input in the model, which is a vital aspect of selective segmentation.

\section{Conclusion}
\label{sec:conclude}

In this paper we have proposed a new intensity fitting term, for use in selective segmentation. We have compared it to fitting terms from comparable approaches (CV, RSF, LCV, HYB, GAV), in order to address an underlying problem in selective segmentation: if the foreground is approximately homogeneous what is the best way to define the intensity fitting term? Previous methods \cite{Rada:13,Geo,CDSS} involve contradictions in the formulation, which we attempt to address. 

We have evaluated the success of the proposed model in four respects: parameter robustness, optimal accuracy, dependence on user input, and comparisons to competing selective models. Our focus is on medical applications, where the target object has approximately homogeneous intensity. In each way, the proposed model performs very well, particularly in cases where the true foreground and background intensities are similar. We have shown that our method is remarkably insensitive to varying user input, highlighting its potential for use in practice, and also outperforms competitive algorithms in the literature.    

\begin{acknowledgements}
The authors would like to thank the Isaac Newton Institute for Mathematical Sciences, Cambridge, for support and hospitality during the programme ``Variational methods and effective algorithms for imaging and vision'' where work on this paper was undertaken. This work was supported by EPSRC grant no EP/K032208/1. The first author wishes to thank the UK EPSRC, the Smith Institute for Industrial Mathematics, and the Liverpool Heart and Chest Hospital for supporting the work through an Industrial CASE award. The second author would like to acknowledge the support of the EPSRC grant EP/N014499/1. This work was generously supported by the Wellcome Trust Institutional Strategic Support Award (204909/Z/16/Z).
\end{acknowledgements}

\bibliographystyle{spmpsci}
\bibliography{CVref}

\end{document}